\documentclass{article}

\usepackage[utf8]{inputenc}
\usepackage[T1]{fontenc}
\usepackage{amsmath,amssymb,amsfonts,amsthm}
\usepackage{graphicx}
\usepackage{booktabs}
\usepackage{algorithm}
\usepackage{algorithmic}
\usepackage{enumitem}
\usepackage{hyperref}
\usepackage{cleveref}
\usepackage{xcolor}
\usepackage{natbib}
\usepackage{tikz}
\usetikzlibrary{positioning,arrows.meta,calc,patterns,decorations.pathreplacing}
\usepackage{subcaption}
\usepackage{wrapfig}
\usepackage{thmtools}
\usepackage{makecell}
\usepackage{xcolor}
\usepackage[table]{xcolor}
\usepackage{array}

\newtheorem{theorem}{Theorem}[section]
\newtheorem{proposition}[theorem]{Proposition}
\newtheorem{lemma}[theorem]{Lemma}

\newcommand{\TC}{\mathrm{TC}}

\definecolor{severered}{RGB}{255,190,190}
\definecolor{warnamber}{RGB}{255,232,190}
\definecolor{lightgray}{gray}{0.85}
\definecolor{midgray}{gray}{0.55}
\definecolor{sectiongray}{gray}{0.92}

\usepackage{booktabs,multirow}

\newcommand{\Ours}[0]{\texttt{MUNI}}
 \usepackage[preprint]{neurips_2026}

\usepackage[utf8]{inputenc} 
\usepackage[T1]{fontenc}    
\usepackage{hyperref}       
\usepackage{url}            
\usepackage{booktabs}       
\usepackage{amsfonts}       
\usepackage{nicefrac}       
\usepackage{microtype}      
\usepackage{xcolor}         
\usepackage{makecell}
\usepackage{tabularx}
\usepackage{array}

\newcommand{\myicona}{%
  \raisebox{-0.15em}{\includegraphics[height=1em]{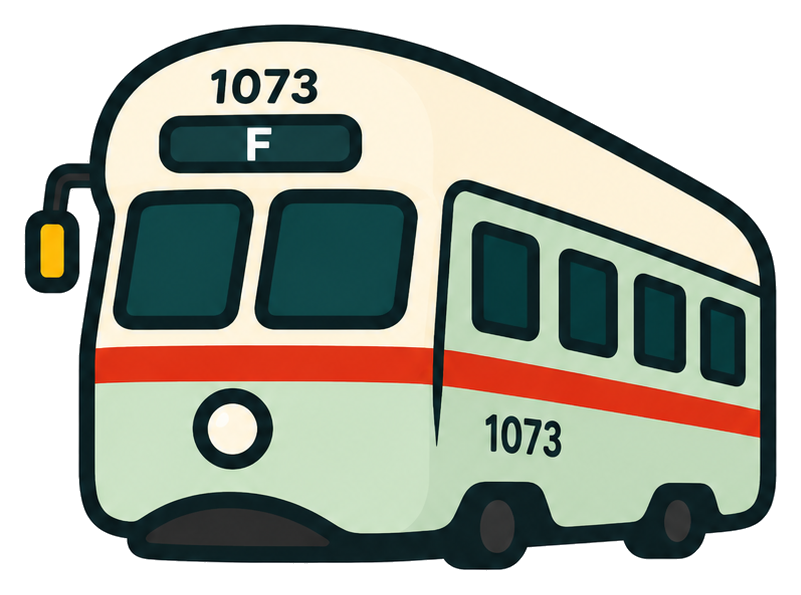}}%
}

\title{\myicona\hspace{0.05em} \underline{MUNI}: \underline{M}ultimodal \underline{Uni}fied Latent Diffusion for Coherent Any-to-Any Generation}
\author{Kyeongmin Yeo\footnotemark[1] \quad Yunhong Min\footnotemark[1] \quad Minhyuk Sung \\[0.2em]
KAIST\\
\texttt{\{aaaaa, dbsghd363, mhsung\}@kaist.ac.kr} 
}

\begin{document}

\maketitle
\begingroup
\renewcommand\thefootnote{}\footnotetext{\textsuperscript{$\ast$}Equal contribution.}
\endgroup

\begin{abstract}
We introduce \Ours{}, an end-to-end multimodal latent diffusion framework for any-to-any generation that unifies subset-conditioned cross-modal generation and unconditional joint sampling through a shared stochastic latent. Existing multimodal generative models are largely LLM-based, which limits leveraging modality-specific generators and requires text-paired data for training. Recent diffusion- and flow-based any-to-any extensions take a different direction but still rely on text-aligned embeddings, fully-paired training, or matched-dimensionality deterministic mappings. \Ours{} rests on two complementary contributions, one architectural and one in the training objective. First, we extend latent diffusion to multimodal any-to-any generation \emph{end-to-end}: instead of the standard two-stage recipe that precomputes a frozen latent space and then fits a prior over it, \Ours{} jointly trains modality-specific encoders, expressive decoders, and a single shared flow-based prior under one objective. Second, we identify that the standard aggregation rules of multimodal variational inference are insufficient once coupled with a learned prior and expressive decoders. A suitable shared latent must simultaneously satisfy \emph{coherence} across generated modalities, \emph{predictive sufficiency} of subset latents, and \emph{minimality} of the latent content.
We propose a routed training objective whose structural choices align the latent with these criteria and admit a minimal-sufficiency characterization in the realizable setting.
Experiments on PolyMNIST-Quadrant-Labels and a large-scale image-text-audio benchmark show \Ours{} matching or exceeding the strongest baselines on conditional generation while opening its largest margins on unconditional coherence. Project page: \href{https://muni-proj.github.io/}{\texttt{muni-proj.github.io}}.

\end{abstract}


\section{Introduction}
\label{sec:introduction}

Diffusion- and flow-based models~\cite{ho2020:DDPM, song2021:scoresde, lipman2023:fm, karras2022:edm} have become the dominant paradigm for generative modeling across many modalities and have set the state of the art in cross-modal conditional generation, including text-to-image~\cite{rombach2022:stablediffusion, flux2024}, text-to-audio~\cite{evans2024:stableaudio, hung2025:tangoflux, liu2023:audioldm}, and image-to-video~\cite{chen2023:videocrafter1, yang2025:cogvideox}. Our key question is how to move beyond these one-to-one settings: how can the capability of diffusion- and flow-based models be effectively extended to \emph{any-subset-conditioned} multimodal generation?

Concretely, given a set of data modalities, we consider a model that takes any subset of those modalities as input and generates the remaining ones. The conditioning subset ranges from the empty set---corresponding to unconditional joint generation of all modalities---to any proper subset of the full collection.

Despite the emergence of large foundational multimodal generative models~\cite{chameleonteam2025:chameleon, lu2023:unifiedio2, zhan2024:anygpt, wu2024:nextgpt}, most such models adopt large language models as their backbone, representing every modality as discrete tokens in a unified sequence. This design makes it difficult to leverage modality-specialized generative architectures and pretrained generators---such as latent diffusion for images~\cite{rombach2022:stablediffusion, esser2024:sd3} and flow matching for audio~\cite{hung2025:tangoflux, evans2024:stableaudio, liu2023:audioldm}---which operate in continuous spaces and currently lead in their respective modalities. It also ties each modality to what can be expressed through language, limiting both the fidelity of per-modality representations and the model's ability to capture direct cross-modal correlations that do not pass through text. Finally, this paradigm requires every training example to be paired with text, which is not the case for many natural multimodal sources, such as uncaptioned audio-visual recordings~\cite{chen2020:vggsound, google2017:audioset} or scientific multimodal measurements~\cite{Parker2024:astroclip, chandrasekaran2023:jump}.

Following this spirit, recent work has explored diffusion- and flow-based models for multimodal generation, demonstrating the promise of this direction~\cite{tang2023:codi, li2025:omniflow, cha2026:flowbind}. However, these approaches still have limitations: relying on text-aligned embedding spaces that may lose modality information not expressed in language~\cite{tang2023:codi}; requiring conditional paths to be learned for every subset configuration and depending on fully-paired data for stable training~\cite{li2025:omniflow}; or imposing matched dimensionality across modalities and learning deterministic source-to-target mappings, despite the non-unique correspondences of multimodal data~\cite{cha2026:flowbind}.

In this work, we propose a novel diffusion-based framework for any-to-any multimodal generation that does not rely on text-aligned embeddings, fully-paired training data, or pre-embedding for dimensionality matching. Our key ideas are twofold: (1) we extend latent diffusion~\cite{rombach2022:stablediffusion} to the multimodal setting by jointly learning the unified prior together with all modality-specific components; and (2) we propose a novel training objective for the underlying variational inference component, addressing the limitations of existing aggregation rules when coupled with a diffusion or flow-based prior and expressive decoders.

Latent diffusion is the de facto standard for image and video generation~\cite{rombach2022:stablediffusion, esser2024:sd3}, but it follows a two-stage recipe in which a latent space is fixed by a pretrained encoder \emph{before} a diffusion prior is fitted over it. Recent work~\cite{heek2026:ul} departs from this separation by training the latent prior \emph{jointly} with the encoder and decoder under a single objective, so that the latent space is shaped by both reconstruction and prior modeling at once.
Our \emph{first contribution} extends this jointly trained framework to multimodal any-to-any generation by introducing modality-specific encoders, modality-specific expressive decoders, and a single unified flow-based prior shared across modalities.

Under this extension, jointly learning the latent embedding and the decoders aligns with multimodal variational inference, a literature that has primarily studied how unimodal posterior experts should be aggregated, including product of experts (PoE)~\cite{wu2018:mvae}, mixture of experts (MoE)~\cite{shi2019:mmvae}, mixture of products of experts (MoPoE)~\cite{sutter2021:mopoe}, and their variants~\cite{vo2026:helvae}. Our \emph{second contribution} identifies that aggregation rules alone are insufficient once coupled with a diffusion or flow-based prior and expressive decoders, and introduces a novel training objective designed around three latent-space criteria: (i) \emph{coherence-sufficiency} for unconditional co-generation, (ii) \emph{predictive sufficiency} for cross-modal generation, and (iii) \emph{minimality} of the latent content.
Existing objectives do not enforce these criteria jointly: latents inferred from partial observations may support conditional generation, but need not preserve dependencies among unobserved modalities required for unconditional co-generation, while self-reconstruction rewards encoding modality-private details that need not belong to the shared latent.
\Ours{} addresses these failures through non-mixture aggregation, target-detached self-reconstruction, and prior learning on leave-one-out latents.

We validate \Ours{} on a controlled PolyMNIST-Quadrant-Labels benchmark and a large-scale image-text-audio any-to-any benchmark. Across both, \Ours{} matches or exceeds the strongest multimodal-VAE and any-to-any generalist baselines on conditional generation while opening its largest margins precisely where our analysis predicts: on unconditional coherence. On image-text-audio, \Ours{} achieves the strongest generalist many-to-one performance, with audio-image alignment gains of up to $+7.14$ AIS over the next-best generalist, and improves unconditional co-generation over the only directly comparable generalist by $+5.57$ CLIP, $+10.60$ CLAP, and $+31.85$ AIS. These results show that \Ours{} scales to large image-text-audio any-to-any generation while retaining strong coherence against frontier generalist baselines.

\vspace{-0.3\baselineskip}
\section{Background and Problem Setup}
\label{sec:background}
\vspace{-0.2\baselineskip}

\subsection{MultiModal Generation with Unified Latents}
\label{sec:background-ul}

\begin{figure}[t]
    \centering
    \includegraphics[width=\linewidth]{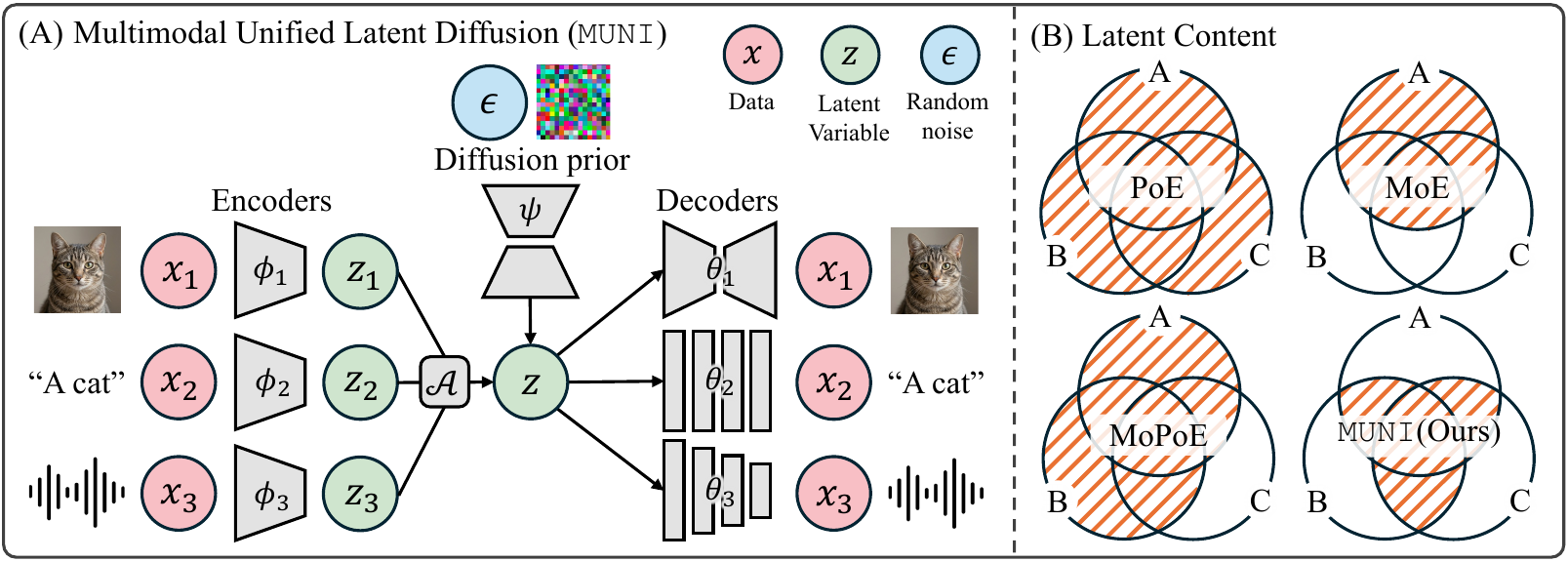}
    \caption{
    \textbf{Overview of \Ours{} and latent content.}
    \textbf{(A)} \Ours{} uses modality-specific encoders and decoders with a shared latent $z$ and a learned prior. 
    \textbf{(B)} Schematic of the modality-shared and modality-specific factors retained in the latent under PoE, MoE, MoPoE, and \Ours{}.
    }
    \label{fig:pairlat_framework}
\end{figure}

Let $X_{1:M}=(X_1,\ldots,X_M)$ denote the random variables associated with $M$ modalities, with joint data distribution $p_{\mathrm{data}}(x_{1:M})$. 
multimodal generation requires a model to represent the family of conditional distributions
\begin{equation}
    p_{\mathrm{data}}(x_{S^c}\mid x_S),
    \qquad S\subseteq[M],
    \label{eq:cross-modal-task}
\end{equation}
where $S$ is the observed subset and $S^c=[M]\setminus S$ is the set of target modalities. 
Given $x_S$, the model should sample $x_{S^c}$ jointly, preserving consistency with the observed modalities. 
Especially, the case $S=\emptyset$ corresponds to unconditional co-generation, in which the model samples coherent tuples from the full joint distribution $p_{\mathrm{data}}(x_{1:M})$.


Recent diffusion- and flow-based any-to-any models extend continuous generative modeling to multimodal generation~\cite{tang2023:codi,li2025:omniflow,cha2026:flowbind}. 
Existing approaches, however, often rely on text-aligned conditioning representations~\cite{tang2023:codi}, require learning conditional paths across all conditioning--target modality subsets rather than inferring a shared latent variable~\cite{li2025:omniflow}, or use invertible flows that require matched dimensionality and impose deterministic one-to-one correspondence between the latent and each modality~\cite{cha2026:flowbind}. 
This motivates a probabilistic shared-latent framework that can be inferred from arbitrary observed subsets, decoded stochastically into each modality, and sampled from a learned prior for unconditional co-generation.

Latent diffusion models use a learned latent space for high-fidelity generation. 
They move diffusion modeling from data space to latent space, where a prior models latent samples and a decoder maps them back to data.
In the standard two-stage pipeline, a variational autoencoder is trained first, and a diffusion prior is then fitted to the frozen latent distribution. 
Unified Latents (UL)~\cite{heek2026:ul} instead jointly trains the encoder $\mathcal E_\phi(z_0\mid x)$, latent prior $q_\psi(z_0)$, and decoder $p_\theta(x\mid z_0)$:
\begin{equation}
    \mathcal L_{\mathrm{UL}}(\phi,\psi,\theta)
    =
    \mathbb E_{x\sim p_{\mathrm{data}}}
    \mathbb E_{z_0\sim\mathcal E_\phi(\cdot\mid x)}
    \left[
    \mathcal L_x(\theta;x,z_0)
    +
    \beta \mathcal L_z(\psi;z_0)
    \right],
    \label{eq:ul-objective}
\end{equation}
where $\mathcal L_z$ and $\mathcal L_x$ are the diffusion evidence lower bound (ELBO) terms for the latent prior and conditional decoder, respectively. 
By optimizing reconstruction and prior modeling in the same latent space, UL avoids treating the autoencoder representation as a fixed target for the prior. 
The latent is therefore shaped not only for accurate decoding, but also to remain tractable for the prior to model.

We extend this idea to multimodal generation by introducing subset encoders and modality-specific expressive decoders. 
Specifically, for each non-empty subset $A\subseteq[M]$, we introduce a subset encoder $q_{\phi,A}(z \mid x_A)$, a shared learned prior $q_\psi(z)$, and modality-specific decoders
\vspace{-0.1\baselineskip}
\begin{equation}
    p_\theta(x_{1:M}\mid z)
    =
    \prod_{m=1}^M p_{\theta_m}(x_m\mid z).
    \label{eq:factorized-decoder}
\end{equation}
At inference time, conditional generation from $S$ is represented by
\begin{equation}
    p_{\theta,\phi}(x_{S^c}\mid x_S)
    =
    \int q_{\phi,S}(z\mid x_S)
    \prod_{m\in S^c} p_{\theta_m}(x_m\mid z)\,dz,
    \label{eq:conditional-generation}
\end{equation}
while unconditional co-generation is represented by
\begin{equation}
    p_{\theta,\psi}(x_{1:M})
    =
    \int q_\psi(z)
    \prod_{m=1}^M p_{\theta_m}(x_m\mid z)\,dz.
    \label{eq:unconditional-generation}
\end{equation}
The same latent variable therefore mediates both subset-conditioned generation and unconditional joint sampling, while expressive decoders can model modality-specific variation.
Fig.~\ref{fig:pairlat_framework} (A) illustrates this construction.

A direct multimodal UL objective trains every subset latent to decode all modalities and uses the same latents to supervise the prior:
\begin{equation}
    \mathcal L_{\mathrm{MMUL}}
    =
    \mathbb E_{x_{1:M}\sim p_{\mathrm{data}}}
    \sum_{A\in\mathcal P^*([M])}
    w_A
    \mathbb E_{z_A\sim q_{\phi,A}(\cdot\mid x_A)}
    \left[
    \sum_{m=1}^M
    \mathcal L_x^m(\theta_m;x_m,z_A)
    +
    \beta\mathcal L_z(\psi;z_A)
    \right],
    \label{eq:cmul-naive}
\end{equation}
where $\mathcal P^*([M])=\{A:A\subseteq[M],A\ne\emptyset\}$ and $w_A$ is a subset sampling weight. 
This objective is a natural multimodal analogue of UL. 
However, applying the prior loss to every subset fits the prior to a mixture of subset latents, including latents inferred from insufficient evidence to determine dependencies among the missing modalities. 
Such prior samples need not yield coherent unconditional co-generation. 
In addition, self-reconstruction terms for $m\in A$ can encourage the shared latent to store modality-specific details that are unnecessary for cross-modal generation.


\subsection{Connection to Multimodal Variational Inference}
\label{sec:background-mvae}

This multimodal factorization based on unified latents closely resembles the latent-variable structure of multimodal VAEs~\cite{wu2018:mvae, shi2019:mmvae, sutter2021:mopoe, palumbo2023:mmvaeplus, vo2026:helvae, cui2025:shala}. 
Both families introduce subset posteriors $q_{\phi,A}(z\mid x_A)$ for partial observations and decode a shared latent through modality-specific likelihood factors. 
Most multimodal VAE formulations, however, rely on a fixed Gaussian prior and relatively simple likelihood decoders, which limits generation fidelity. 
Multimodal UL instead retains the learned latent prior and expressive conditional decoders of the latent-diffusion setting.

In multimodal variational inference, subset posteriors are parameterized by aggregating unimodal posterior experts $q_{\phi_m}(z\mid x_m)$. 
Tab.~\ref{tab:posterior-aggregation} summarizes representative aggregation rules. 
The next section analyzes how these choices shape the induced subset latents, which properties are required for coherent multimodal generation, and why aggregation alone does not specify which latents should serve as targets for prior learning.

\begin{table*}[t]
\centering
\caption{Posterior aggregation rules and their alignment with the latent-space criteria in Sec.~\ref{sec:analysis}. 
Coh., Pred., and Min. denote coherence, predictivity, and minimality, respectively.
Here, $q_{\mathrm{PoE}}(z\mid x_A)=\prod_{m\in A}q_{\phi_m}(z\mid x_m)$.}
\label{tab:posterior-aggregation}
\label{tab:posterior-aggregation}
\small
\setlength{\tabcolsep}{4pt}
\renewcommand{\arraystretch}{1.18}
\newcolumntype{Y}[1]{>{\hsize=#1\hsize\linewidth=\hsize\centering\arraybackslash}X}
\newcolumntype{C}{>{\centering\arraybackslash}p{2.3em}}
\begin{tabularx}{\textwidth}{@{}lY{1.15}Y{0.85}CCC@{}}

\toprule
Method
& Aggregation (Training)
& Aggregation (Inference)
& Coh.
& Pred.
& Min. \\
\midrule
MVAE (PoE)~\cite{wu2018:mvae}
&
$\prod_{m\in [M]} q_{\phi_m}(z\mid x_m)$
&
$\prod_{m\in S} q_{\phi_m}(z\mid x_m)$
&
$\checkmark$
&
$\triangle$
&
$\times$ \\
MMVAE (MoE)~\cite{shi2019:mmvae}
&
$\frac{1}{M}\sum_{m\in [M]} q_{\phi_m}(z\mid x_m)$
&
$\frac{1}{|S|}\sum_{m\in S} q_{\phi_m}(z\mid x_m)$
&
$\times$
&
$\times$
&
$\times$ \\
MoPoE~\cite{sutter2021:mopoe}
&
$\frac{1}{2^{M}-1}\sum_{A \in \mathcal{P}^*([M])} q_{\mathrm{PoE}}(z\mid x_A)$
&
$\prod_{m\in S} q_{\phi_m}(z\mid x_m)$
&
$\times$
&
$\checkmark$
&
$\times$ \\
HELVAE~\cite{vo2026:helvae}
&
$\left(\sum_{m\in [M]}\sqrt{q_{\phi_m}(z\mid x_m)}\right)^2$
&
$\left(\sum_{m\in S}\sqrt{q_{\phi_m}(z\mid x_m)}\right)^2$
&
$\checkmark$
&
$\triangle$
&
$\times$ \\
\bottomrule
\end{tabularx}
\end{table*}

\vspace{-0.2\baselineskip}
\section{Latent-Space Criteria for Coherent Multimodal Generation}
\label{sec:analysis}

The direct objective in Eq.~\ref{eq:cmul-naive} and the aggregation rules in Tab.~\ref{tab:posterior-aggregation} both induce subset latents, but do not determine what information those latents should carry. 
Under the factorized decoder in Eq.~\ref{eq:factorized-decoder}, modalities are decoded independently conditioned on $z$, so cross-modal dependence must be represented in the latent. 
This yields three latent-space criteria: prior targets must support coherent joint sampling, subset latents must predict missing modalities while explaining dependencies within the observed subset, and the shared latent should avoid modality-private variation that expressive decoders can model.

\vspace{-0.2\baselineskip}
\paragraph{Coherence-sufficient latents.}
For unconditional co-generation, factorized decoding preserves coherence only when the latent explains the dependencies among modalities. 
We measure the residual dependence by conditional total correlation and require
\begin{equation}
    \TC(X_{1:M}\mid Z)=0.
    \label{eq:global-tc}
\end{equation}
We call such a latent \emph{coherence-sufficient}: conditioned on it, factorized decoding does not discard cross-modal dependence.

\paragraph{Predictively sufficient subset latents.}
For conditional generation from $X_A$, the subset latent $Z_A$ must retain the information in $X_A$ needed to predict each missing modality, while also explaining dependencies within the observed subset:
\begin{align}
    \TC(X_A\mid Z_A)&=0,
    \label{eq:subset-observed-tc}
    \\
    I(X_A;X_m\mid Z_A)&=0
    \qquad \forall m\in A^c.
    \label{eq:subset-predictive-mi}
\end{align}
The mutual-information condition states that $Z_A$ is a sufficient statistic of $X_A$ for each target modality $X_m$. 
The total-correlation condition requires the latent to capture dependencies within the observed subset, which is necessary when such latents are used as prior targets for coherent joint sampling.

\vspace{-0.3\baselineskip}
\paragraph{Minimal latent content.}
The preceding conditions specify the information that the latent must preserve for coherence and conditional alignment. 
Among the latents that satisfy these conditions, however, minimizing the information retained from $X_A$ yields a simpler latent marginal for the prior to model and prevents the shared latent from absorbing modality-specific variation. 
The desired subset latent is therefore
\vspace{-0.1\baselineskip}
\begin{equation}
    Z_A^\star
    =
    \arg\min_{\substack{
    Z_A:\,\TC(X_A\mid Z_A)=0\\
    I(X_A;X_m\mid Z_A)=0\ \forall m\in A^c
    }}
    I(X_A;Z_A).
    \label{eq:minimal-subset-latent}
\end{equation}
This criterion imposes a separation of roles: the shared latent captures the information needed to relate modalities, while modality-specific variation is modeled by $p_{\theta_m}(x_m\mid z)$.

Fig.~\ref{fig:pairlat_framework} (B) visualizes the shared and modality-specific factors retained by PoE, MoE, MoPoE, and \Ours{}, while Tab.~\ref{tab:posterior-aggregation} summarizes their alignment with the three criteria.
PoE and HELVAE aggregate all observed experts into a single posterior, making them structurally compatible with the coherence criterion. 
Their predictivity is only partial, however, because the standard objective does not explicitly train each subset latent to predict held-out modalities. 
In contrast, mixture-based aggregation, as in MMVAE, can draw the latent from a single expert, so the resulting representation may not contain enough information for either conditional alignment or coherent joint decoding. 
MoPoE differs from MMVAE at inference time: although it is trained with a mixture over subset products, it uses the full product posterior for conditional generation, which satisfies the predictivity requirement. 
Its prior, however, is still trained on mixture components that include smaller-subset latents, which need not be valid targets for co-generation. 
Finally, none of these aggregation rules enforces minimality, since self-reconstruction encourages the shared latent to store modality-specific details. 
Thus, aggregation specifies how evidence is combined, but not which subset latents are valid targets for prior learning or how modality-private information should be excluded from the shared latent.


These limitations motivate the objective introduced in Sec.~\ref{sec:method}, while Apps.~\ref{app:additional-analysis} and~\ref{app:main-proof} provide further analysis of the prior-learning targets and the corresponding latent-space constraints.

\vspace{-0.2\baselineskip}
\section{Multimodal Unified Latent Diffusion (\Ours{})}
\label{sec:method}
\vspace{-0.2\baselineskip}

\Ours{} modifies the direct multimodal UL objective in Eq.~\ref{eq:cmul-naive} according to the latent-space criteria in Sec.~\ref{sec:analysis}. 
For each modality $m$, we use a unimodal expert encoder $q_{\phi_m}(z\mid x_m)$ and an expressive decoder $p_{\theta_m}(x_m\mid z)$. 
For a non-empty subset $A$, the subset posterior is obtained by applying a non-mixture aggregation rule to the available unimodal experts:
$q_{\phi,A}(z\mid x_A)=\mathcal A(\{q_{\phi_j}(z\mid x_j)\}_{j\in A})$, 
where $\mathcal A_A$ is either product or Hellinger aggregation.
Unlike mixture-based aggregation, this forms a single posterior from all observed modalities.

For a reconstruction target $m$, we use a target-dependent subset posterior. 
If $m\in A$, the target expert is detached before aggregation; otherwise the usual subset posterior is used:
\begin{equation}
    \bar q_{\phi_j}^{(m)}(z\mid x_j)
    =
    \begin{cases}
    \operatorname{sg} \left[q_{\phi_j}(z\mid x_j)\right], & j=m,\\
    q_{\phi_j}(z\mid x_j), & j\ne m,
    \end{cases}
    \qquad
    q_{\phi,A}^{(m)}(z\mid x_A)
    =
    \mathcal A \left(\{\bar q_{\phi_j}^{(m)}(z\mid x_j)\}_{j\in A}\right),
    \label{eq:self-detached-posterior}
\end{equation}
where $\operatorname{sg}[\cdot]$ is the stop-gradient operator. 
Thus, self-reconstruction still trains the decoder, but does not reward the encoder for transmitting modality-specific details.

The training objective is
\begin{equation}
\begin{aligned}
    \mathcal L_{\mathrm{ours}}
    =
    \mathbb E_{x_{1:M}\sim p_{\mathrm{data}}}
    \sum_{A \in \mathcal P^*([M])}
    w_A
    \Bigg[
    &\sum_{m=1}^M
    \mathbb E_{z\sim q_{\phi,A}^{(m)}(\cdot\mid x_A)}
    \mathcal L_x^m(\theta_m;x_m,z)
    \\
    &+
    \beta\,\mathbf 1[|A|\ge M-1]
    \mathbb E_{z\sim q_{\phi,A}(\cdot\mid x_A)}
    \mathcal L_z(\psi;z)
    \Bigg].
\end{aligned}
\label{eq:ours-objective}
\end{equation}
\vspace{-0.2\baselineskip}
The reconstruction term covers both held-out prediction and target-detached self-reconstruction, while the prior term is applied only to full-modality and leave-one-out subset latents. 
In practice, Eq.~\ref{eq:ours-objective} is optimized by Monte Carlo sampling subsets and reconstruction targets.

\begin{proposition}[(Informal) minimal sufficient latent]
\label{prop:main-proof-informal}
Under realizability and optimization assumptions, Eq.~\ref{eq:ours-objective} selects subset latents that satisfy predictive sufficiency while approaching the minimality criterion in Eq.~\ref{eq:minimal-subset-latent}. 
On the subset routes used for prior learning, these latents are coherence-sufficient and therefore provide valid targets for the learned prior.
\end{proposition}

Apps.~\ref{app:additional-analysis} and~\ref{app:main-proof} detail the objective design and provide a formal version and proof of Proposition~\ref{prop:main-proof-informal}. 
App.~\ref{app:loss-implementation} gives the concrete prior loss $\mathcal L_z$ and the modality-specific reconstruction losses $\mathcal L_x^m$ used in our experiments.

At inference time, the same encoders and decoders are used for every conditioning subset. 
Given an observed subset $S$, we sample a latent vector from the subset posterior $z\sim q_{\phi,S}(z\mid x_S)$ and generate each target modality from $p_{\theta_m}(x_m\mid z)$ for $m\in S^c$. 
When $S=\emptyset$, we instead sample $z\sim q_\psi(z)$ and decode all modalities.
\vspace{-0.2\baselineskip}
\section{Related Work}
\label{sec:related}
\vspace{-0.2\baselineskip}

\vspace{-0.2\baselineskip}
\paragraph{Any-to-any generation.}
Recent diffusion- and flow-based frameworks support generation beyond fixed modality pairs.
CoDi~\citep{tang2023:codi} composes modality-specific diffusion models through a shared text embedding space, enabling flexible input-output combinations but relying on a representation that may lose information not captured by text; its separately trained encoders and decoders can also weaken input-output alignment. 
OmniFlow~\citep{li2025:omniflow} formulates any-to-any generation with rectified flows over conditioning--target modality subsets, but learning paths across all subset configurations increases data and training demands. 
FlowBind~\citep{cha2026:flowbind} maps each modality to a shared space through invertible flows; this requires all modalities to have the same dimensionality, and generation is deterministic given the source modality rather than sampled from a conditional distribution.
\Ours{} instead learns a shared latent variable: subset posteriors are used for conditional generation, modality decoders model target variation, and a prior model is used for unconditional co-generation.

\vspace{-0.2\baselineskip}
\paragraph{Multimodal variational inference.}
Multimodal VAEs model multiple modalities through a shared latent variable and modality-specific decoders, with the main inference question being how to infer the latent from arbitrary observed subsets. 
MVAE~\citep{wu2018:mvae} uses product-of-experts aggregation, MMVAE~\citep{shi2019:mmvae} uses mixture-of-experts aggregation, and MoPoE~\citep{sutter2021:mopoe} mixes products over modality subsets. 
Prior analyses show that these choices induce trade-offs among joint generation, cross-modal prediction, and generation quality~\citep{daunhawer2022:limitationsmvae}. 
We use this literature as a reference for posterior aggregation, but build on the unified-latent formulation with a learned prior and expressive decoders, rather than a standard multimodal VAE with a fixed Gaussian prior and simple likelihood decoder.
Moreover, prior multimodal VAE studies have largely focused on controlled or relatively simple benchmarks; to our knowledge, \Ours{} is the first to scale this probabilistic framework to practical image-text-audio any-to-any generation.

\vspace{-0.2\baselineskip}
\paragraph{Private latents and expressive models.}
Another line of work separates shared and modality-specific information with private latent variables, as in DMVAE~\citep{daunhawer2020:dmvae}, MMVAE+~\citep{palumbo2023:mmvaeplus}, MMVAE++~\citep{martens2024:mmvaeplusplus}, and IDMVAE~\citep{zhang2025:idmvae}. 
These private latents are natural choices when decoders cannot represent target-modality variation without encoding it explicitly. 
Other methods improve multimodal VAEs with stronger generative components, such as a second-stage diffusion prior in ShaLa~\citep{cui2025:shala}, diffusion decoders in Diff-MVAE~\citep{wesego2025:diffmvae}, an energy-based prior in MVEBM~\citep{yuan2024:ebmvae}, or mixture-structured latent spaces in CMVAE~\citep{palumbo2024:cmvae}. 
These extensions improve capacity, but do not specify which subset latents should train the prior or how self-reconstruction should affect the encoder and the corresponding latent. 
\Ours{} keeps a single shared latent, assigns modality-specific variation to expressive decoders, and uses the criteria in Sec.~\ref{sec:analysis} to determine prior learning and self-reconstruction routes.
\vspace{-0.2\baselineskip}
\section{Experiments}
\label{sec:experiments}


In this section, we evaluate \Ours{} on two main benchmarks. 
We first describe the baselines and shared experimental setup in \cref{sec:exp-common}. 
We then evaluate on a standard PolyMNIST-based benchmark with mixed image and label modalities in \cref{sec:exp-polymnist}, followed by a large-scale image-text-audio benchmark in Sec.~\ref{sec:exp-ita}. 
In both settings, we interpret the results through the latent-space criteria in Sec.~\ref{sec:analysis}, testing whether the observed patterns support our analysis and objective design. 
App.~\ref{app:gaussian_mixture} further provides a controlled Gaussian-mixture experiment with known shared-private structure, making the predicted baseline failure modes and the role of our proposed design more explicit.

\vspace{-0.2\baselineskip}
\subsection{Common Setups}
\label{sec:exp-common}
\vspace{-0.2\baselineskip}

\paragraph{Baselines.}
Across all benchmarks, we compare against four representative multimodal VAE baselines that span the dominant choices of posterior aggregation:
MVAE~\cite{wu2018:mvae} (product of experts, PoE),
MMVAE~\cite{shi2019:mmvae} (mixture of experts, MoE),
MoPoE~\cite{sutter2021:mopoe} (mixture of products of experts, generalizing PoE and MoE),
and HELVAE~\cite{vo2026:helvae} (Hellinger aggregation, a symmetric interpolation between PoE and MoE).
For MVAE, we also train partially observed PoE posteriors with subset-level terms, matching the role of subset ELBOs in the original formulation.
On the image-text-audio benchmark in Sec.~\ref{sec:exp-ita}, we additionally compare against three recent diffusion- and flow-based any-to-any generalists: CoDi~\cite{tang2023:codi}, OmniFlow~\cite{li2025:omniflow}, and FlowBind~\cite{cha2026:flowbind}.

\paragraph{Implementation.}
The original multimodal VAE baselines are typically implemented with a fixed Gaussian prior and deterministic likelihood decoders. For a fair comparison, we instantiate all methods under the same architecture and generative framework as \Ours{}. 
Specifically, all methods use a learned flow-based prior and expressive modality decoders, so that performance differences primarily reflect aggregation and prior-routing choices rather than model capacity. 
Full details about training, evaluation, and architecture are provided in App.~\ref{app:imple_details}.

\begin{figure}[t]
\centering
\setlength{\tabcolsep}{4pt}
\renewcommand{\arraystretch}{1.0}
\begin{tabular}{ccccc}
\footnotesize MVAE &
\footnotesize MMVAE &
\footnotesize MoPoE &
\footnotesize HELVAE &
\footnotesize \textbf{\Ours{}} \\
\includegraphics[width=0.18\textwidth]{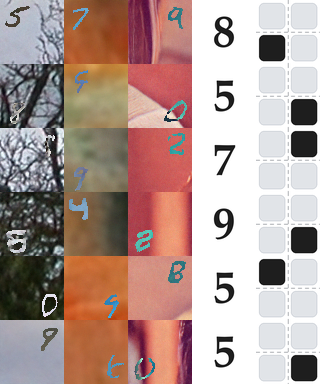} &
\includegraphics[width=0.18\textwidth]{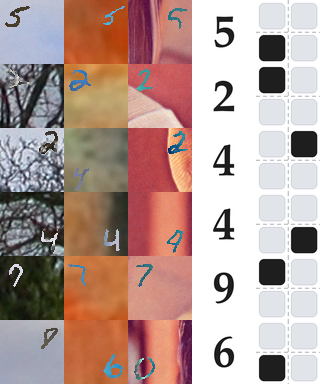} &
\includegraphics[width=0.18\textwidth]{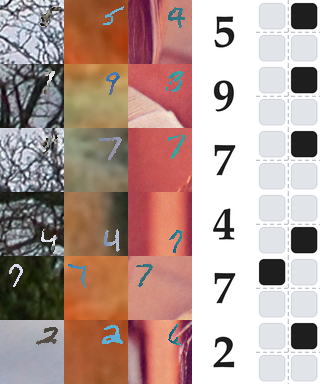} &
\includegraphics[width=0.18\textwidth]{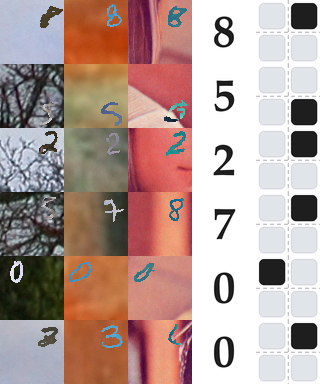} &
\includegraphics[width=0.18\textwidth]{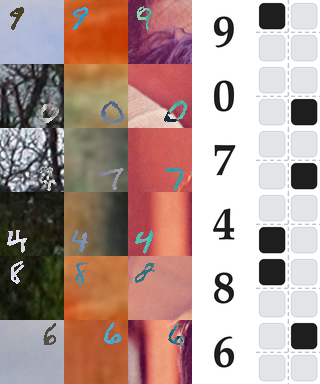} \\
\end{tabular}
\caption{\textbf{Unconditional co-generation on PolyMNIST-Quadrant-Labels.} 
Each column shows one tuple $(M_1,M_2,M_3,\text{digit},\text{quadrant})$ jointly sampled from the prior of a given method. 
Coherent samples preserve digit identity and quadrant position across the generated image and label modalities.}
\vspace{-0.8\baselineskip}
\label{fig:polymnist-qualitative}
\end{figure}

\vspace{-0.2\baselineskip}
\subsection{PolyMNIST-Quadrant-Labels}
\label{sec:exp-polymnist}

\paragraph{Setup.}
We construct PolyMNIST-Quadrant-Labels from PolyMNIST-Quadrant~\cite{zhang2025:idmvae} by aligning the quadrant position across image modalities and adding digit and quadrant labels as discrete modalities. 
Each example contains three $64{\times}64$ RGB image modalities $M_1,M_2,M_3$ with distinct backgrounds, together with a digit label and a quadrant label. 
The image modalities share digit identity and quadrant position but differ in background, so coherent generation requires preserving the shared labels while leaving background variation to the modality decoders. 
The dataset contains $220$k training and $10$k held-out evaluation samples, and \Ours{} uses product-based aggregation.

\paragraph{Metrics.}
We evaluate three tasks: \emph{single-label$\to$image}, \emph{multi-label$\to$image}, and \emph{unconditional} co-generation. 
The first conditions on either the digit or quadrant label, the second conditions on both labels, and the third samples all five modalities jointly from the prior. 
For conditional generation, we report \emph{alignment}: the fraction of generated images whose digit or quadrant is correctly recovered by a held-out verifier, averaged over image modalities $M_1,M_2,M_3$. 
For multi-label$\to$image, a sample is counted as correct only when both labels are recovered. 
For unconditional co-generation, we report \emph{coherence}, which requires verifier predictions on the three generated images to agree with one another and with the generated digit and quadrant labels.

\begin{wraptable}{r}{0.51\textwidth}
\vspace{-1.0em}
\centering
\caption{\textbf{PolyMNIST-Quadrant-Labels.}
Single-L $\to$ I and Multi-L $\to$ I report verifier accuracy on generated images conditioned on a single label (Digit or Quadrant) or on both labels jointly. \textbf{Bold} and \underline{underline} mark the best and second-best results.}
\label{tab:polymnist}
\footnotesize
\setlength{\tabcolsep}{2pt}
\renewcommand{\arraystretch}{1.1}
\begin{tabular}{lcccc}
\toprule
& \multicolumn{2}{c}{Single-L $\to$ I}
& \multicolumn{1}{c}{Multi-L $\to$ I}
& \multicolumn{1}{c}{Uncond.} \\
\cmidrule(lr){2-3}\cmidrule(lr){4-4}\cmidrule(lr){5-5}
Method & Digit $\uparrow$ & Quadrant $\uparrow$ & Both $\uparrow$ & Coherence $\uparrow$ \\
\midrule
MVAE~\cite{wu2018:mvae}     & 0.4137 & \underline{0.6803} & 0.7053 & 0.0079 \\
MMVAE~\cite{shi2019:mmvae}    & \underline{0.9279} & \textbf{0.9999} & 0.1630 & 0.3167 \\
MoPoE~\cite{sutter2021:mopoe}    & 0.9199 & \textbf{0.9999} & \underline{0.9283} & 0.3943 \\
HELVAE~\cite{vo2026:helvae}   & \textbf{0.9297} & \textbf{0.9999} & 0.7654 & \underline{0.4246} \\
\textbf{\Ours{}} \textbf{(Ours)} & 0.9131 & \textbf{0.9999} & \textbf{0.9346} & \textbf{0.4841} \\
\bottomrule
\end{tabular}
\vspace{-0.5em}
\end{wraptable}


\paragraph{Results.}
Tab.~\ref{tab:polymnist} summarizes verifier accuracy across the four metrics. 
On Single-L\,$\to$\,I, \Ours{} remains competitive with the strongest baselines, and on Multi-L\,$\to$\,I it achieves the best joint alignment. 
The largest gap appears in unconditional co-generation, where \Ours{} achieves the best coherence by a noticeable margin. 
This pattern is consistent with the criteria in Sec.~\ref{sec:analysis}. 
MMVAE samples from a mixture of single-modality experts, so the latent can ignore part of the observed subset and fail the predictivity criterion in Eqs.~\ref{eq:subset-observed-tc}--\ref{eq:subset-predictive-mi}, explaining its weak Multi-L\,$\to$\,I alignment. 
MoPoE uses the full product posterior at inference, which explains its strong conditional alignment; however, its prior is trained on mixture components that include smaller-subset latents, violating the coherence criterion in Eq.~\ref{eq:global-tc}. 
MVAE and HELVAE aggregate all observed experts, but their standard objectives do not enforce cross-reconstruction or the minimality criterion in Eq.~\ref{eq:minimal-subset-latent}, limiting overall performance. 
\Ours{} aligns more directly with all three criteria, yielding strong conditional alignment together with the strongest unconditional coherence. 
Fig.~\ref{fig:polymnist-qualitative} shows the same trend, with \Ours{} producing more consistent digit identity and quadrant position across jointly generated modalities.

\vspace{-0.3\baselineskip}
\subsection{Image-Text-Audio Any-to-Any Generation}
\label{sec:exp-ita}

\paragraph{Setup.}
We evaluate \Ours{} on a large-scale image-text-audio any-to-any benchmark, following the data setup of FlowBind~\cite{cha2026:flowbind}. 
Training uses three pairwise datasets across text, image, and audio: text-image (242K from LAION-COCO~\cite{laion2025relaioncoco} and 30K from Flickr-30k~\cite{plummer2015:flickr30k}), text-audio (91K from AudioCaps v2~\cite{kim2019audiocaps}), and image-audio (184K from VGGSound~\cite{chen2020:vggsound}). 
For many-to-one and unconditional co-generation, where no standard text-image-audio triplet benchmark exists, we follow the synthetic triplet protocol of FlowBind by generating images with FLUX.1-dev~\cite{flux2024} from text-audio pairs in the AudioCaps v2~\cite{kim2019audiocaps} test set to obtain (text, image, audio) triplets.

We report two variants of \Ours{} with aggregation rules compatible with the predictivity criterion: \Ours{} uses product aggregation, while \Ours{}$^\dagger$ uses Hellinger aggregation, following HELVAE~\cite{vo2026:helvae}.
Our main comparisons are the generalist diffusion- and flow-based baselines from Sec.~\ref{sec:exp-common}: CoDi, OmniFlow, and FlowBind. 
To isolate aggregation and prior-learning effects from model capacity, we also include MMVAE and MoPoE under the same backbone and compute as \Ours{}. 
The MVAE and HELVAE baselines are excluded because their standard training requires fully observed triplets, which are unavailable in this training setup. 
We further report LLM-based generalists and task-specific specialists as reference rows in App.~\ref{app:imple_details}.


\vspace{-0.2\baselineskip}
\paragraph{Metrics.}
We focus the main paper on two settings where aggregation and prior-routing effects are most visible. 
(i) \emph{Many-to-one alignment} conditions on two source modalities, generates the third, and reports cross-modal alignment to each source separately, using CLIPScore~\cite{hessel2021:clipscore} for text-image, CLAPScore~\cite{elizalde2023:clap} for text-audio, and Audio-Image Similarity (AIS)~\cite{wu2022:wav2clip} for image-audio. 
(ii) \emph{Unconditional coherence} jointly generates all three modalities and reports pairwise alignment between generated modalities using the same scores. 
Due to space constraints, full \emph{one-to-one fidelity} and \emph{alignment} results are provided in App.~\ref{app:imple_details}.

\begin{table}[t]
\centering
\caption{\textbf{Many-to-one cross-modal alignment on image-text-audio.}
\Ours{} uses PoE aggregation and \Ours{}$^\dagger$ uses Hellinger aggregation.
\textbf{Bold} and \underline{underline} mark the best and second-best generalist results.}
\vspace{0.2\baselineskip}
\label{tab:ita-many-to-one}
\small
\begin{tabular}{>{\centering\arraybackslash}p{1.6cm}!{\color{lightgray}\vrule}l!{\color{midgray}\vrule}cccccc}
\toprule
\multirow{2}{*}{Category} & \multicolumn{1}{c!{\color{midgray}\vrule}}{\multirow{2}{*}{Method}}
& \multicolumn{2}{c}{(I+A)$\to$T}
& \multicolumn{2}{c}{(T+A)$\to$I}
& \multicolumn{2}{c}{(T+I)$\to$A}\\
\cmidrule(lr){3-4}\cmidrule(lr){5-6}\cmidrule(lr){7-8}
& \multicolumn{1}{c!{\color{midgray}\vrule}}{}
& CLIP $\uparrow$ & CLAP $\uparrow$ & CLIP $\uparrow$ & AIS $\uparrow$ & CLAP $\uparrow$ & AIS $\uparrow$\\
\midrule
\multirow{5}{*}{\emph{Generalists}}
& CoDi~\cite{tang2023:codi}      & 24.05            & 33.72            & 24.98            & 85.52            & 11.06            & 65.31 \\
& OmniFlow~\cite{li2025:omniflow}  & 24.73            & 36.26            & \underline{26.41} & 81.51            & 13.50            & 63.55 \\
& FlowBind~\cite{cha2026:flowbind}  & \underline{27.54}   & \underline{39.56}   & 25.23            & 86.44 & 26.83 & 80.15 \\
& \textbf{\Ours{} (Ours)}   & 25.20 & \textbf{39.75}& 25.44   & \textbf{93.42}   & \underline{26.88}   & \textbf{87.29} \\
& \textbf{\Ours{}$^\dagger$ (Ours)}   & \textbf{27.60} & 36.61 & \textbf{26.43}   & \underline{90.94}   & \textbf{30.16}   & \underline{86.89} \\
\midrule
\multirow{2}{*}{\makecell{\emph{Multimodal} \\ \emph{VAEs}}}
& MMVAE~\cite{shi2019:mmvae}     & 26.57 & 35.24 & 25.70 & 88.22 & 25.33 & 82.49 \\
& MoPoE~\cite{sutter2021:mopoe}     & 27.81 & 36.02 & 26.25 & 86.25 & 30.96 & 84.07 \\
\bottomrule
\end{tabular}
\end{table}

\begin{figure}[t]
\centering
\setlength{\tabcolsep}{2pt}
\renewcommand{\arraystretch}{1.0}
\begin{tabular}{ccc@{\hspace{1.5em}}ccc}
\includegraphics[width=0.145\textwidth]{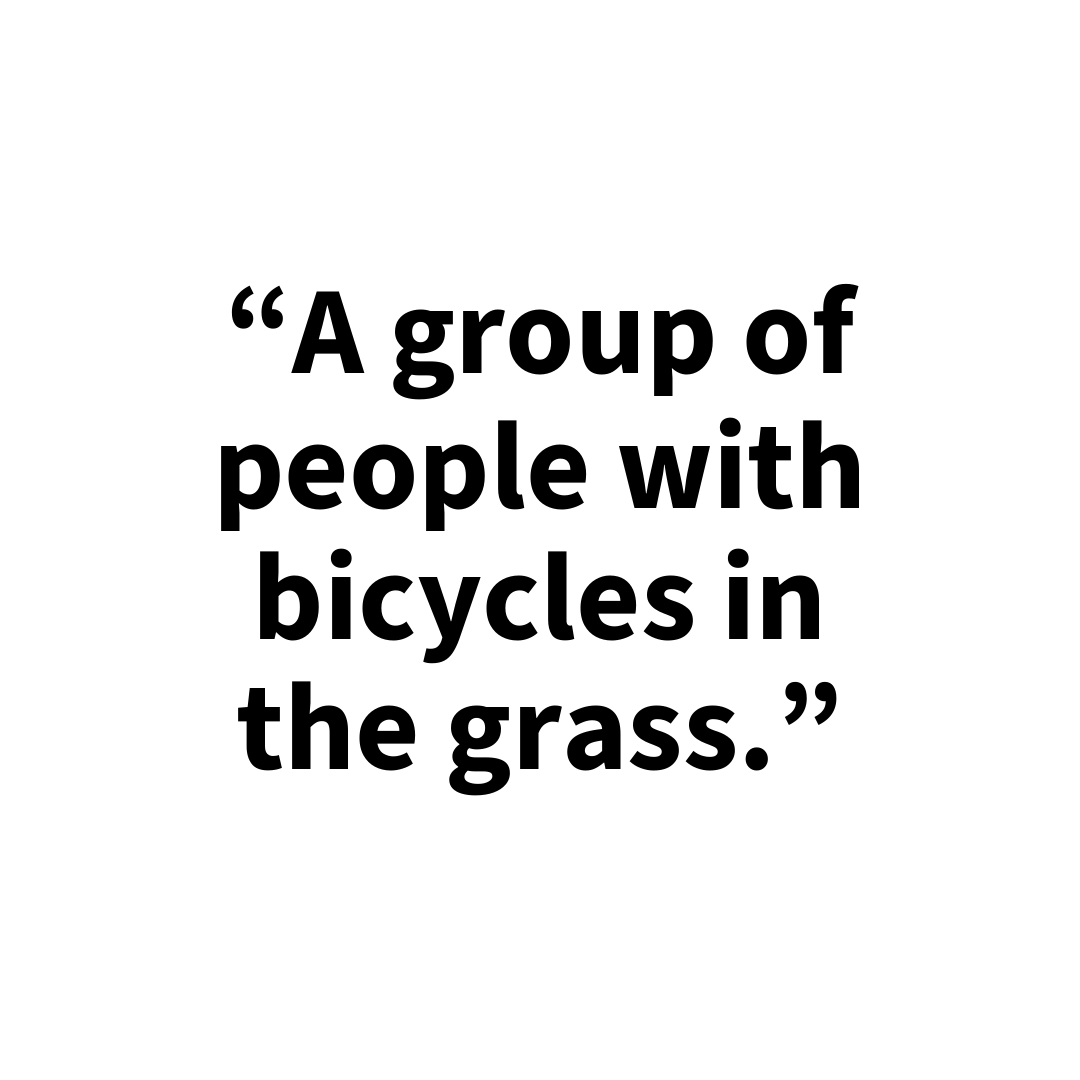} &
\includegraphics[width=0.145\textwidth]{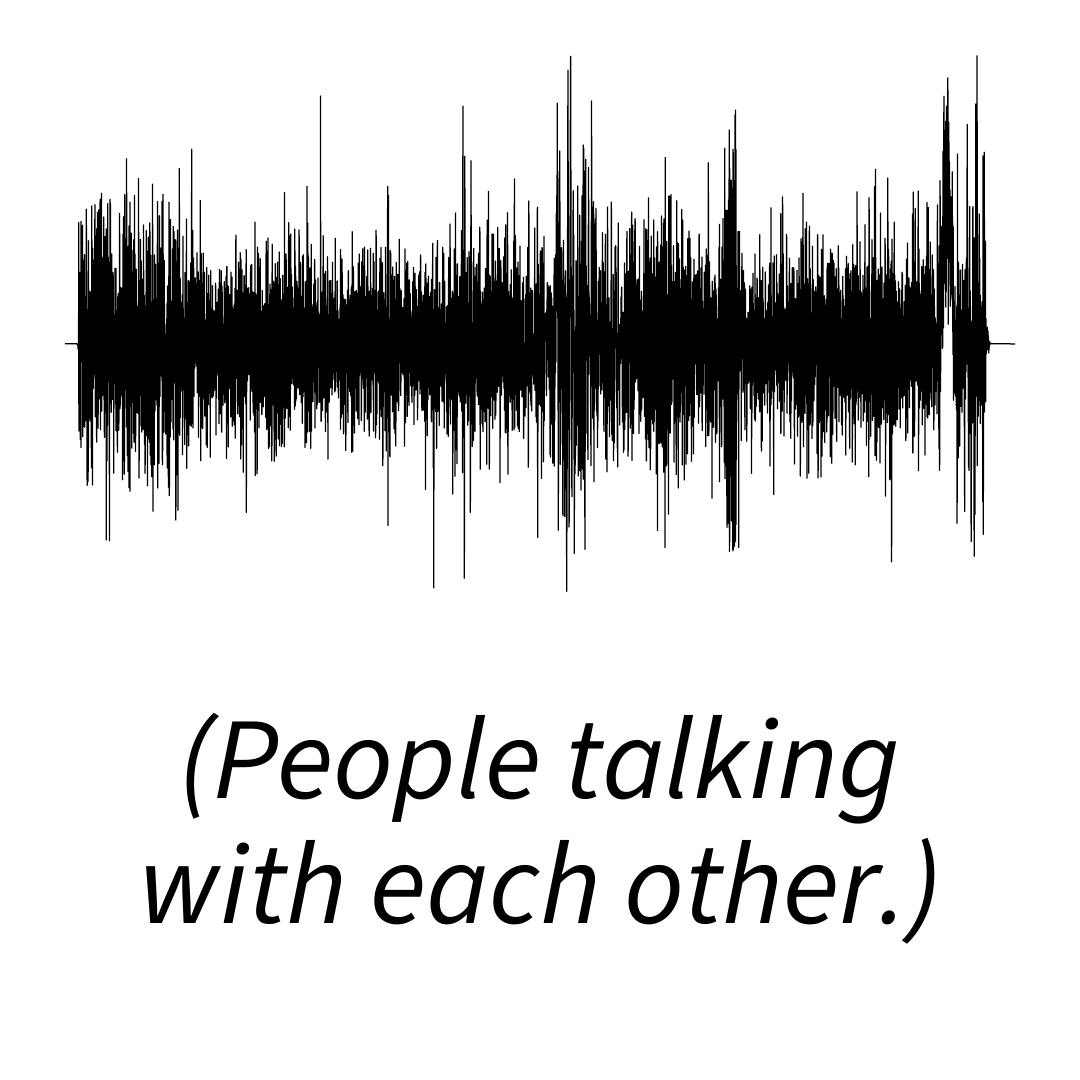} &
\includegraphics[width=0.145\textwidth]{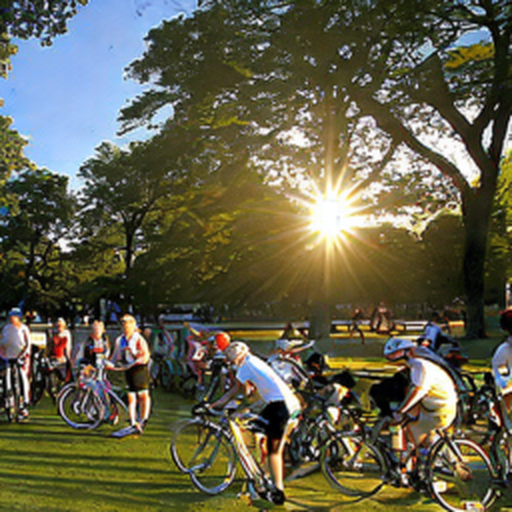} &
\includegraphics[width=0.145\textwidth]{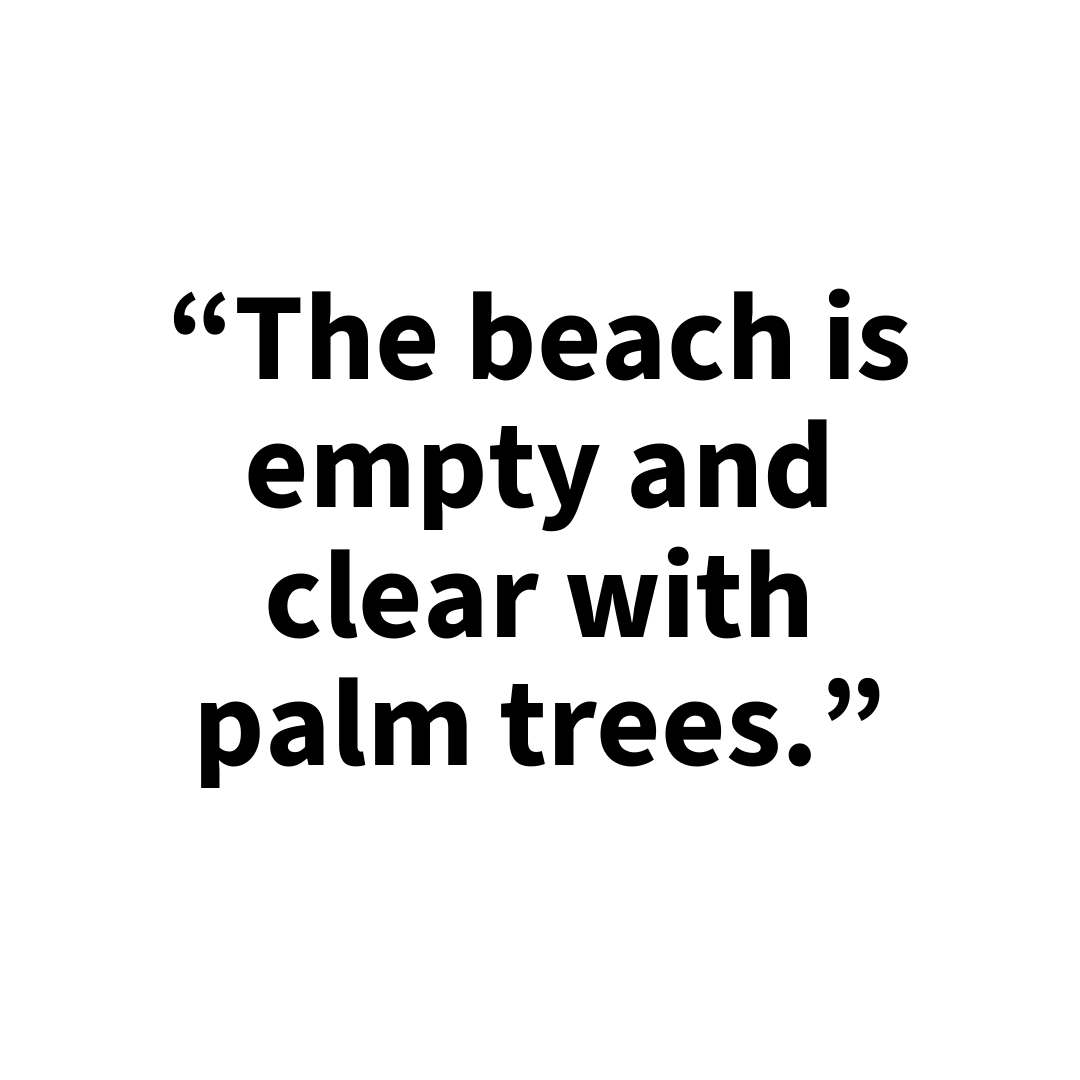} &
\includegraphics[width=0.145\textwidth]{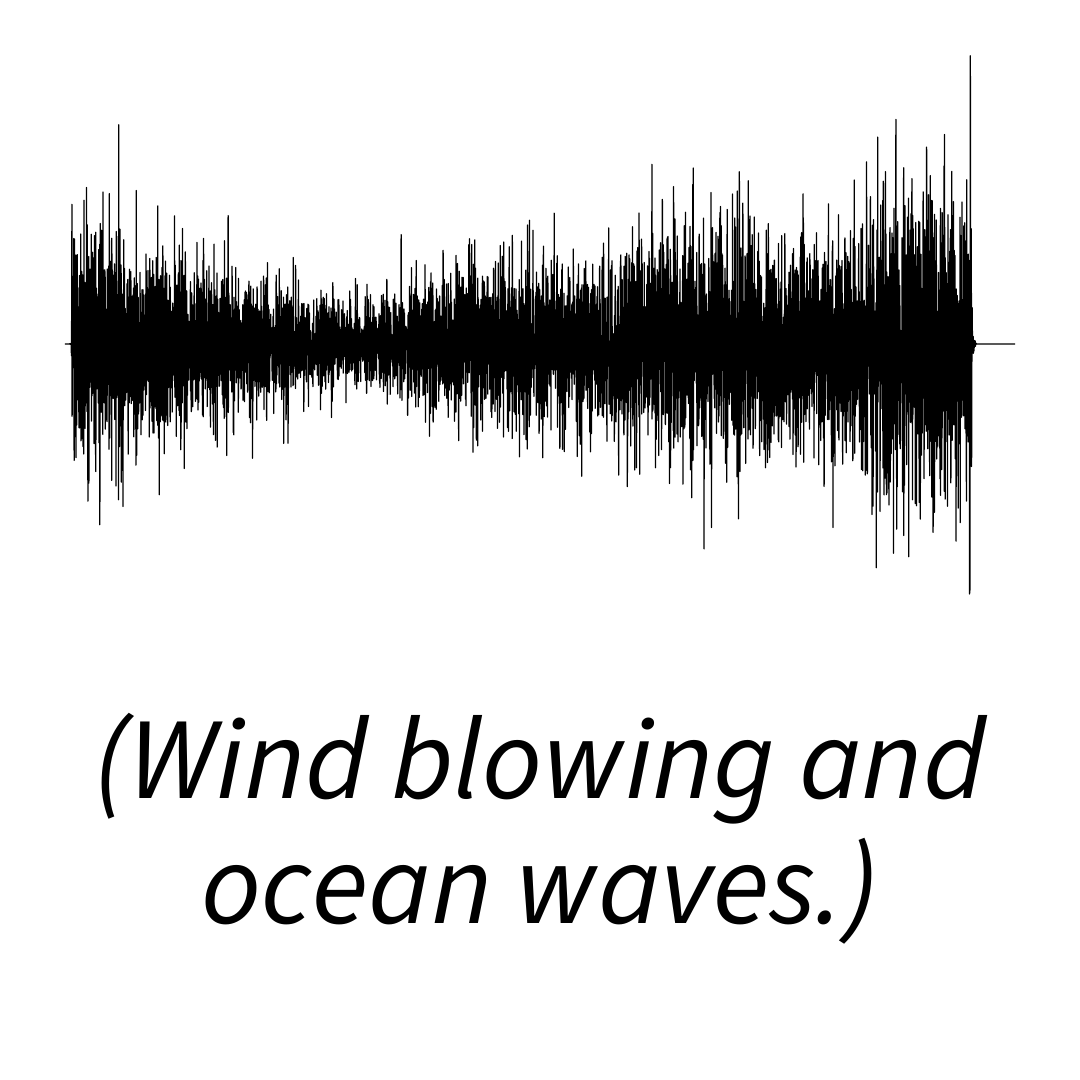} &
\includegraphics[width=0.145\textwidth]{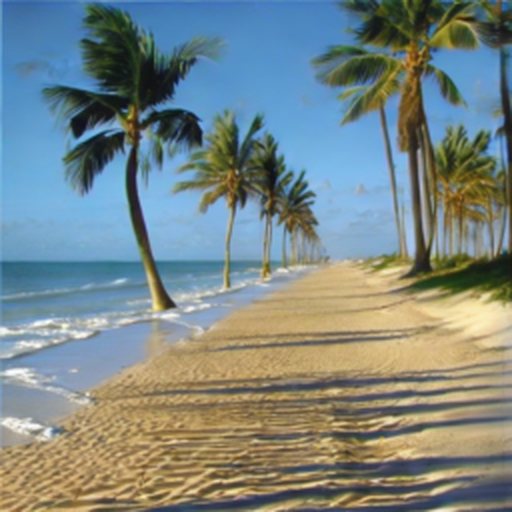} \\
\includegraphics[width=0.145\textwidth]{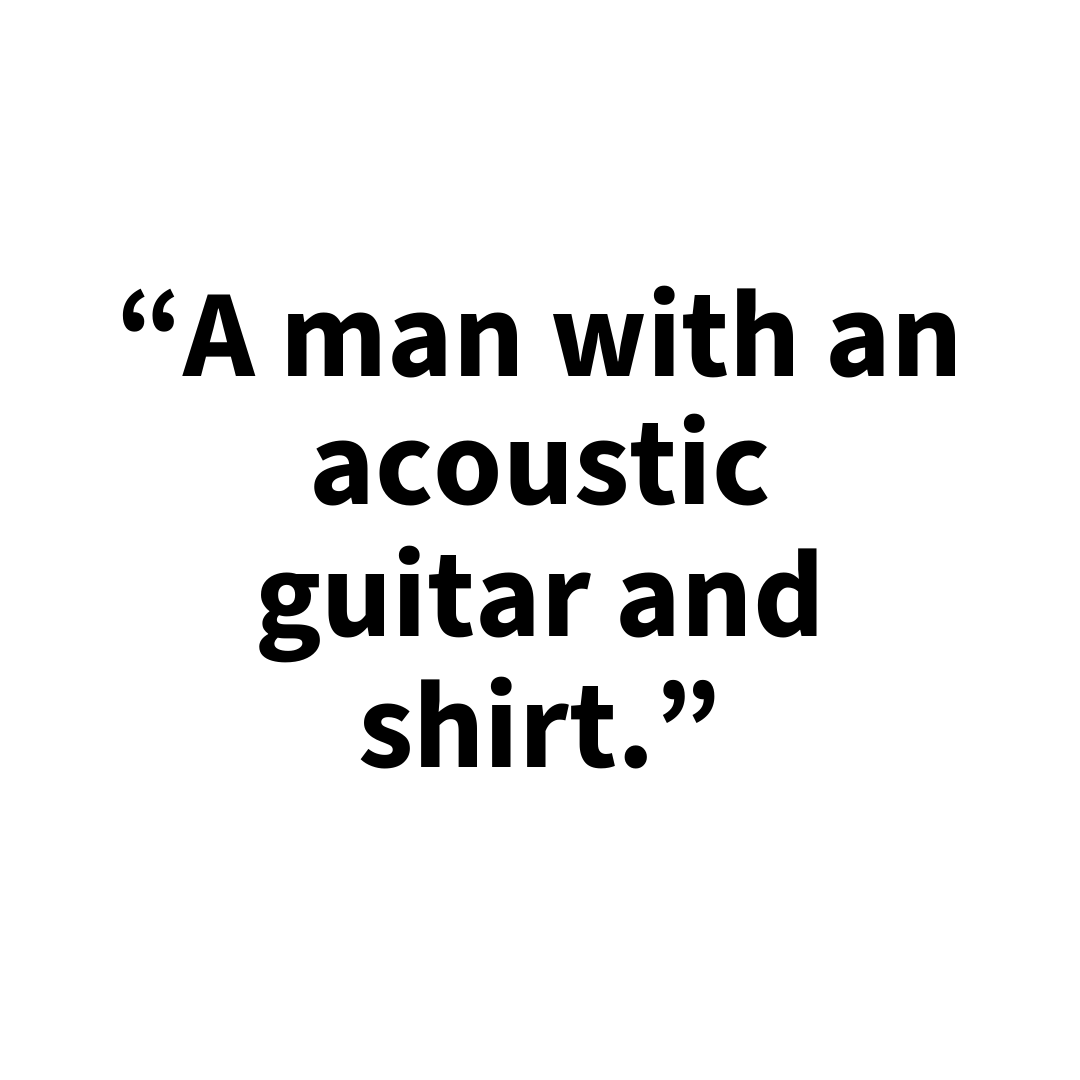} &
\includegraphics[width=0.145\textwidth]{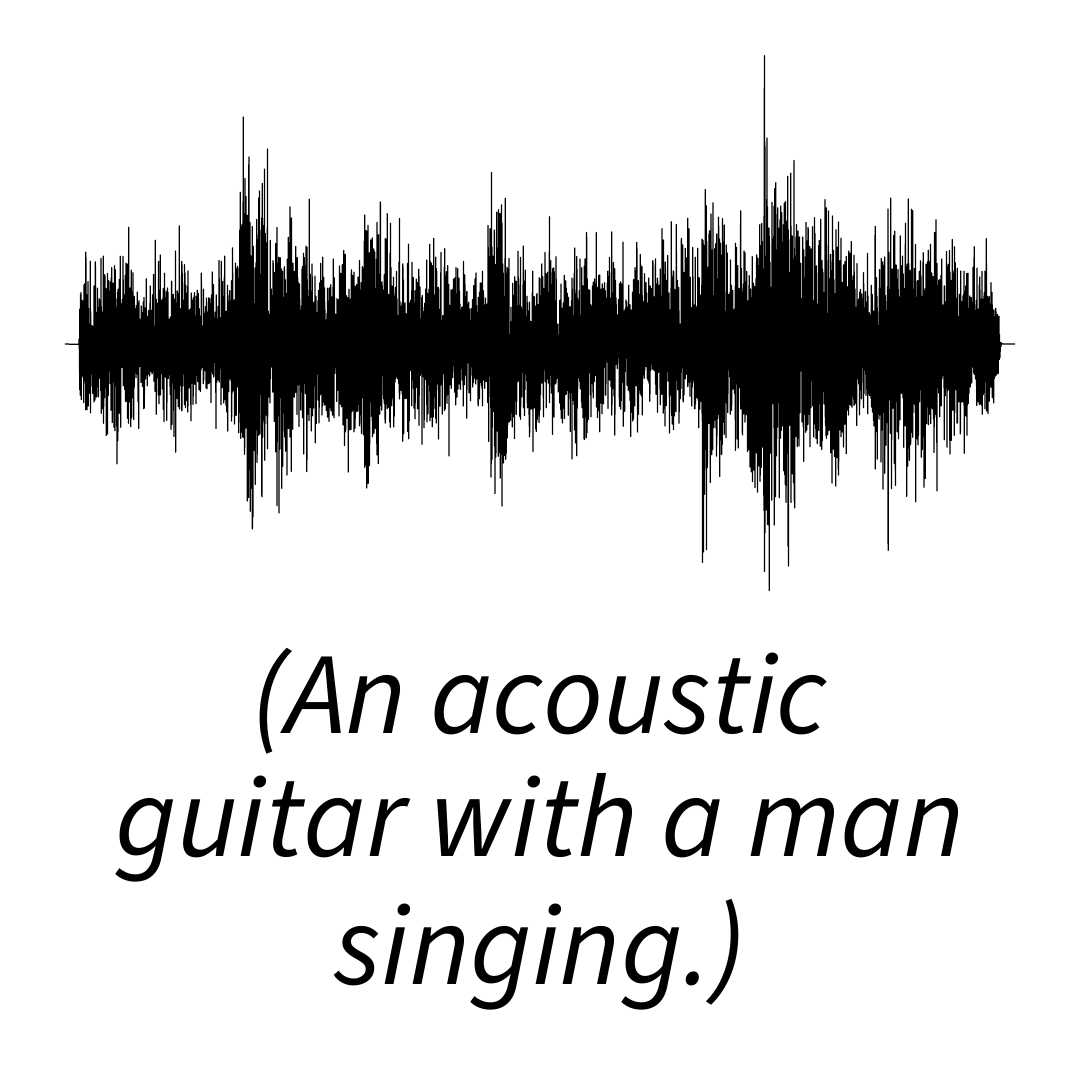} &
\includegraphics[width=0.145\textwidth]{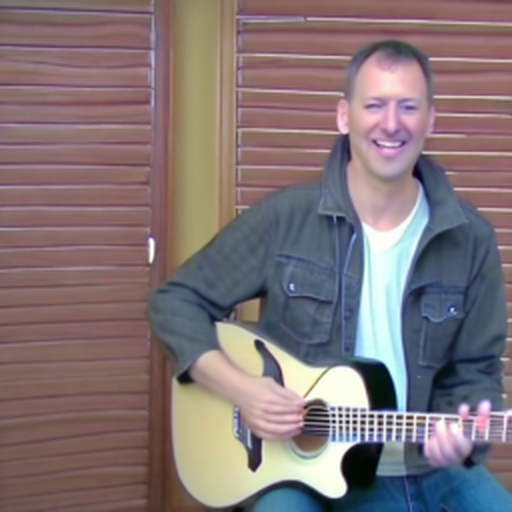} &
\includegraphics[width=0.145\textwidth]{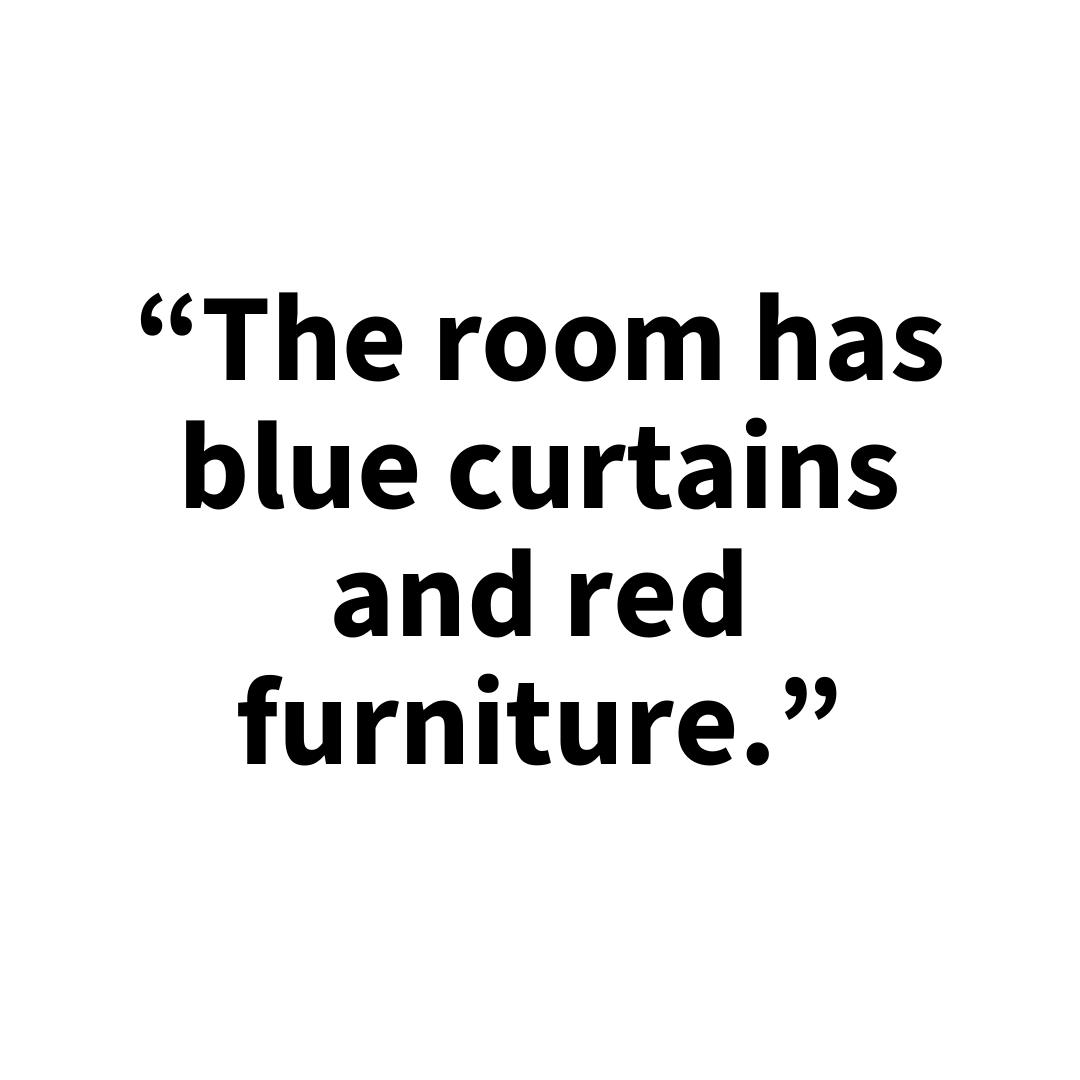} &
\includegraphics[width=0.145\textwidth]{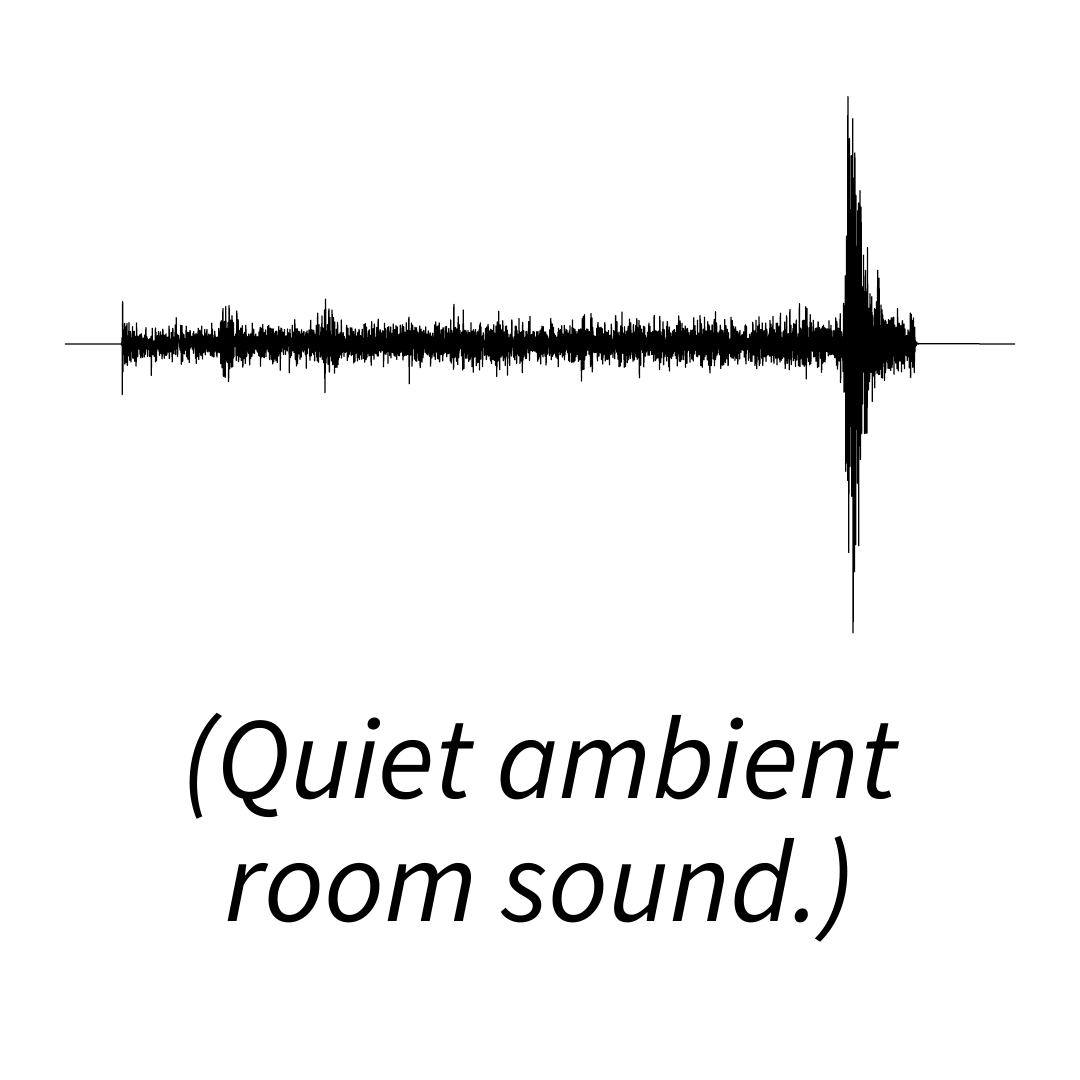} &
\includegraphics[width=0.145\textwidth]{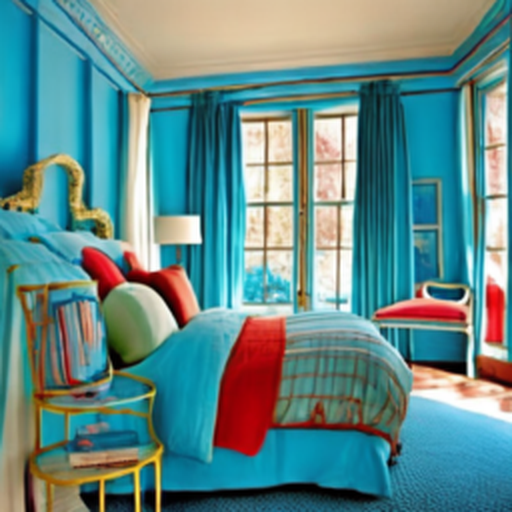} \\
\end{tabular}
\caption{\textbf{Qualitative results of \Ours{}$^\dagger$ in unconditional co-generation of image-text-audio.} Each example shows a (text, audio, image) triple jointly sampled from the prior.}
\vspace{-1.3\baselineskip}
\label{fig:ita-qualitative}
\end{figure}

\begin{wraptable}[14]{R}{0.46\linewidth}
\vspace{-0.1em}
\centering
\caption{\textbf{Unconditional coherence on image-text-audio.} CoDi~\cite{tang2023:codi} and FlowBind~\cite{cha2026:flowbind} are excluded since they do not support unconditional joint sampling.
\textbf{Bold} and \underline{underline} mark the best and second-best generalist results.}
\label{tab:ita-unconditional}
\small
\setlength{\tabcolsep}{6pt}
\renewcommand{\arraystretch}{1.1}
\begin{tabular}{lccc}
\toprule
\multirow{2}{*}{Method} & T--I & T--A & A--I \\
& CLIP $\uparrow$ & CLAP $\uparrow$ & AIS $\uparrow$ \\
\midrule
OmniFlow~\cite{li2025:omniflow}                       & 21.17             & 14.23             & 50.95 \\
\textbf{\Ours{} (Ours)}        & \textbf{26.76}    & \underline{23.73} & \underline{81.24} \\
\textbf{\Ours{}$^\dagger$ (Ours)} & \underline{26.74} & \textbf{24.83}    & \textbf{82.80} \\
\midrule
MMVAE~\cite{shi2019:mmvae}     & 25.52 & 20.06 & 76.68 \\
MoPoE~\cite{sutter2021:mopoe}     & 25.83 & 21.47 & 74.37 \\
\bottomrule
\end{tabular}
\vspace{-0.5em}
\end{wraptable}

\vspace{-0.2\baselineskip}
\paragraph{Results.}
On many-to-one generation (Tab.~\ref{tab:ita-many-to-one}), \Ours{} and \Ours{}$^\dagger$ achieve the best generalist result on every metric across the three input compositions. 
The gains are especially clear on audio--image alignment: \Ours{} improves AIS from $86.44$ to $93.42$ for (T+A)$\to$I and from $80.15$ to $87.29$ for (T+I)$\to$A over the next-best generalist. 
This pattern is consistent with the predictivity criterion in Eq.~\ref{eq:subset-predictive-mi}: CoDi's text-aligned conditioning can miss information not expressed in text, OmniFlow must learn conditional paths separately across subset configurations, making reliable multi-source prediction harder, and FlowBind maps a source to a deterministic target rather than sampling from $p(x_{S^c}\mid x_S)$, limiting its ability to represent the true conditional distribution. 
Among the multimodal VAE baselines, MMVAE remains weaker due to mixture-based single-expert sampling, while MoPoE is competitive conditionally because it uses the full product posterior at inference.

Unconditional co-generation (Tab.~\ref{tab:ita-unconditional}) shows a larger separation. 
Among the generalist baselines, only OmniFlow supports unconditional co-generation, but it is trained primarily through conditional generation paths and yields weak co-generation: \Ours{}$^\dagger$ improves over it by $+5.57$ CLIP on T--I, $+10.60$ CLAP on T--A, and $+31.85$ AIS on A--I.
Both \Ours{} variants also outperform MMVAE and MoPoE on all three pairwise alignments, matching the PolyMNIST trend and supporting the coherence criterion in Eq.~\ref{eq:global-tc}.
Qualitative unconditional samples from \Ours{} are shown in Fig.~\ref{fig:ita-qualitative}. 
Additionally, App.~\ref{app:imple_details} reports the full one-to-one results and extended comparisons with task-specific specialists and LLM-based generalists.

\vspace{-0.5\baselineskip}
\section{Conclusion}
\label{sec:conclusion}
\vspace{-0.2\baselineskip}


We introduced \Ours{}, a multimodal unified-latent framework for any-to-any generation. 
\Ours{} extends Unified Latents to arbitrary subset conditioning, providing a single probabilistic interface for both conditional generation and unconditional co-generation. 
By identifying predictivity, coherence, and minimality as criteria for the shared latent, \Ours{} addresses the failure modes of existing multimodal VAE methods through three objective choices: non-mixture aggregation, target-detached self-reconstruction, and prior learning on leave-one-out latents. Experiments on PolyMNIST-Quadrant-Labels and image-text-audio generation show that \Ours{} preserves competitive conditional generation while substantially improving unconditional coherence. 
Our results further show that the multimodal variational-inference perspective can scale beyond controlled benchmarks when paired with expressive generative components. 
Overall, the results support the main message of this work: coherent any-to-any generation depends not only on how modality evidence is aggregated, but also on what information the shared latent is trained to contain.



\bibliographystyle{abbrv}
\bibliography{ref}

\clearpage
\newpage

\appendix
\definecolor{softgray}{gray}{0.55}

\section*{Societal Impacts}
Our work enables any-to-any multimodal generation from partially observed data, reducing the need for expensive fully observed multimodal tuples. 
However, stronger image-text-audio generation may also increase risks of synthetic-media misuse, including misinformation, impersonation, and privacy or copyright violations.

\section*{Limitations and Future Work}
\Ours{} uses a general objective across all conditioning settings and does not explicitly exploit known structure among modalities.
In practice, cross-modal relations can be highly asymmetric: for example, text-to-speech generation contains substantial variation, whereas speech-to-text is nearly deterministic. 
Incorporating such modality-specific structure into route sampling, loss design, or architecture choices could make the framework more effective for particular multimodal domains. 
Extending \Ours{} to exploit these known relationships is an important direction for future work.

\section{Additional Analysis}
\label{app:additional-analysis}

\subsection{Objective Components}
\label{app:objective-components}

Relative to the direct cross-modal UL objective, Eq.~\ref{eq:ours-objective} modifies three components: how subset posteriors are aggregated, which posterior is used for each reconstruction term, and which subset latents are used for prior learning.

\paragraph{Aggregation.}
Given a modality subset $A$, the subset posterior is constructed as
\begin{equation}
    q_{\phi,A}(z\mid x_A)
    =
    \mathcal A\left(\{q_{\phi_j}(z\mid x_j)\}_{j\in A}\right),
    \label{eq:app-aggregation}
\end{equation}
where $\mathcal A$ is instantiated as product aggregation or Hellinger aggregation.
Both choices form a single posterior from all observed experts, unlike mixture rules that may sample from only part of the observed subset. 
This makes the aggregation rule compatible with the predictivity criterion in Eq.~\ref{eq:subset-predictive-mi}, although aggregation alone does not enforce the criterion.

\paragraph{Target-routed reconstruction posterior.}
For reconstruction target $m$, Eq.~\ref{eq:ours-objective} uses
\begin{equation}
    \bar q_{\phi_j}^{(m)}(z\mid x_j)
    =
    \begin{cases}
    \operatorname{sg}\!\left[q_{\phi_j}(z\mid x_j)\right], & j=m,\\
    q_{\phi_j}(z\mid x_j), & j\ne m,
    \end{cases}
    \qquad
    q_{\phi,A}^{(m)}(z\mid x_A)
    =
    \mathcal A\left(\{\bar q_{\phi_j}^{(m)}(z\mid x_j)\}_{j\in A}\right).
    \label{eq:app-target-dependent-posterior}
\end{equation}
If $m\notin A$, no target expert is present and $q^{(m)}_{\phi,A}$ reduces to the usual subset posterior $q_{\phi,A}$. 
If $m\in A$, the forward pass still uses the target expert, but this expert is detached in the backward pass. 
Consequently, each encoder expert is updated only by reconstruction losses whose target is a different modality. 
The encoder is therefore trained to preserve information useful for cross-modal prediction, rather than details useful only for reconstructing its own modality. 
This removes the self-copying pressure that would otherwise place modality-specific information in the shared latent, supporting the minimality criterion in Eq.~\ref{eq:minimal-subset-latent}.

\paragraph{Latents for prior learning.}
The prior term in Eq.~\ref{eq:ours-objective} is gated as
\begin{equation}
    \mathbf 1[|A|\ge M-1]\,
    \mathbb E_{z\sim q_{\phi,A}(\cdot\mid x_A)}
    \mathcal L_z(\psi;z).
    \label{eq:app-prior-gating}
\end{equation}

Thus, prior learning is applied only to full-modality and leave-one-out subset latents, since smaller subset latents can under-specify the joint structure needed for coherent unconditional co-generation.
For any non-empty proper subset $A\subsetneq[M]$,
\begin{equation}
    \TC(X_{1:M}\mid Z_A)
    =
    \TC(X_A\mid Z_A)
    +
    \TC(X_{A^c}\mid Z_A)
    +
    I(X_A;X_{A^c}\mid Z_A).
    \label{eq:app-subset-decomposition}
\end{equation}

The second term captures dependence among the modalities absent from the conditioning set, which cannot in general be determined by a subset latent $Z_A$ inferred only from $X_A$.
For a leave-one-out subset $A_i=[M]\setminus{i}$, this term vanishes identically because $A_i^c$ contains a single modality.
Hence, if $Z_{A_i}$ makes the observed modalities conditionally independent and is sufficient for the held-out modality, then all terms in Eq.~\eqref{eq:app-subset-decomposition} vanish, giving $\TC(X_{1:M}\mid Z_{A_i})=0$. This motivates applying the prior loss to leave-one-out and full-modality latents.

\section{Proof of Proposition~\ref{prop:main-proof-informal}}
\label{app:main-proof}

We formalize the route-wise population argument underlying Proposition~\ref{prop:main-proof-informal}. 
The stop-gradient operator leaves the forward latent unchanged, but determines which encoder experts receive reconstruction gradients on each route. 
The proof therefore characterizes the sufficient latent selected by this routing rule.

Fix a non-empty subset $A\subseteq[M]$ and let $Z_A$ be a candidate latent inferred from $X_A$. For a target modality $m$, define the evidence that can train the encoder on that route $(A, m)$ as
\begin{equation}
    Y_{A,m}
    =
    \begin{cases}
    X_A, & m\notin A,\\
    X_{A\setminus\{m\}}, & m\in A.
    \end{cases}
    \label{eq:app-route-evidence}
\end{equation}
For missing targets, all observed modalities in $A$ receive prediction gradients. For self targets, the target expert is detached, so only the non-target experts in $A\setminus\{m\}$ can be trained by the reconstruction of $X_m$.

We define the conditional predictive information
\begin{equation}
    \rho_{A,m}(Z_A)
    =
    I(Y_{A,m};X_m\mid Z_A).
    \label{eq:app-route-residual}
\end{equation}
Thus, for $m\notin A$,
\begin{equation}
    \rho_{A,m}(Z_A)
    =
    I(X_A;X_m\mid Z_A),
    \label{eq:app-missing-residual}
\end{equation}
while for $m\in A$,
\begin{equation}
    \rho_{A,m}(Z_A)
    =
    I(X_{A\setminus\{m\}};X_m\mid Z_A).
    \label{eq:app-self-residual}
\end{equation}

We define the set of predictively sufficient latents for subset $A$ as
\begin{equation}
    \mathcal S_A
    =
    \left\{
    Z_A:
    \rho_{A,m}(Z_A)=0
    \ \text{for all } m\in[M]
    \right\}.
    \label{eq:app-route-sufficient-set}
\end{equation}
For missing targets, the condition $\rho_{A,m}(Z_A)=0$ is exactly $I(X_A;X_m\mid Z_A)=0$. For self targets, the conditions $\rho_{A,m}(Z_A)=0$ for all $m\in A$ are equivalent to $\TC(X_A\mid Z_A)=0$, which is proved by the following lemma:

\begin{lemma}[Conditional sufficiency implies conditional factorization]
\label{lem:app-self-residual-tc}
Suppose that, for every modality $m\in A$,
\begin{equation}
    I(X_{A\setminus\{m\}};X_m\mid Z_A)=0.
    \label{eq:app-all-self-residual-zero}
\end{equation}
Then
\begin{equation}
    \TC(X_A\mid Z_A)=0.
    \label{eq:app-observed-tc-zero}
\end{equation}
\end{lemma}
\emph{Proof.}
Write $A=\{a_1,\ldots,a_K\}$. By the chain rule for conditional total correlation,
\begin{equation}
    \TC(X_A\mid Z_A)
    =
    \sum_{k=2}^K
    I(X_{a_k};X_{a_1},\ldots,X_{a_{k-1}}\mid Z_A).
\end{equation}
For each $k$, the variables $(X_{a_1},\ldots,X_{a_{k-1}})$ are a subset of $X_{A\setminus\{a_k\}}$. Therefore,
\begin{equation}
    I(X_{a_k};X_{a_1},\ldots,X_{a_{k-1}}\mid Z_A)
    \le
    I(X_{a_k};X_{A\setminus\{a_k\}}\mid Z_A)
    =
    0,
\end{equation}
where the last equality follows from the assumption. Hence every term in the chain-rule expansion is zero, and $\TC(X_A\mid Z_A)=0$. \hfill$\square$

Therefore, the sufficient set can equivalently be written as
\begin{equation}
    \mathcal S_A
    =
    \left\{
    Z_A:
    \TC(X_A\mid Z_A)=0,\;
    I(X_A;X_m\mid Z_A)=0
    \ \text{for all } m\in A^c
    \right\}.
    \label{eq:app-sufficient-set}
\end{equation}


\paragraph{Assumptions.}
We make the following idealized assumptions. Throughout this proof, all losses are population risks, i.e., expectations over the data distribution.


\emph{Model capacity and optimization.}
We assume that the encoder and decoder families are sufficiently expressive and that the population objectives are optimized ideally. For any route $(A,m)$, if
\begin{equation}
\rho_{A,m}(Z_A)>0,
\label{eq:app-positive-residual}
\end{equation}
then the encoder experts receiving gradients on that route, together with the decoder for modality $m$, admit an update that decreases the corresponding expected target loss. Conversely, when $\rho_{A,m}(Z_A)=0$ and the decoder is optimal given $Z_A$, the route is target-wise optimal. We also assume that the ideal latent $Z_A^\star$ defined in Eq.~\ref{eq:minimal-subset-latent} is representable by the encoder family.


\emph{Prior ordering.}
For any latent $Z_A$, define the optimal prior cost for its marginal distribution as
\begin{equation}
\mathcal C_A(Z_A)
=
\inf_\psi
\mathbb E,\mathcal L_z(\psi;Z_A).
\label{eq:app-prior-cost}
\end{equation}
We assume that, within the sufficient set $\mathcal S_A$, larger retained information from $X_A$ implies larger optimal prior cost:
\begin{equation}
I(X_A;Z)<I(X_A;Z')
\quad\Longrightarrow\quad
\mathcal C_A(Z)<\mathcal C_A(Z'),
\qquad
Z,Z'\in\mathcal S_A.
\label{eq:app-prior-ordering}
\end{equation}
A simple case where this ordering holds is maximum-likelihood training with a prior family that can realize the marginal density of $Z_A$. Then
\begin{equation}
\inf_\psi \mathbb E[-\log q_\psi(Z_A)]
=
H(Z_A)
=
I(X_A;Z_A)+H(Z_A\mid X_A).
\end{equation}
Thus, when $H(Z_A\mid X_A)$ is fixed across the latents being compared, minimizing the optimal prior cost over $\mathcal S_A$ is equivalent to minimizing $I(X_A;Z_A)$.
When the prior is trained with diffusion or flow-matching losses, we use Eq.~\eqref{eq:app-prior-ordering} as the corresponding idealized ordering assumption for the prior loss.

\emph{Small prior weight.}
We assume that $\beta$ is sufficiently small that the prior loss cannot compensate for a reducible target loss. Consequently, solutions are first restricted to target-wise optimal, where no target loss can be further decreased by the corresponding feasible update. The prior cost then selects among these latents. Under this assumption, the target losses enforce sufficiency, while the prior term selects among sufficient latents according to $\mathcal C_A$.

\begin{proposition}[Minimal sufficient latent]
\label{prop:app-prior-selected-minimality}
Let $\mathcal B_A$ be the set of latents for which no target loss can be further reduced by a feasible update of the encoder experts receiving gradients on the corresponding route, together with the corresponding decoder.
Under the assumptions above, let $\widehat Z_A$ be a minimizer of the prior cost over $\mathcal B_A$:
\begin{equation}
    \widehat Z_A
    =
    \operatorname*{argmin}_{Z_A\in\mathcal B_A}
    \mathcal C_A(Z_A)
    \label{eq:app-prior-selected}
\end{equation}
Then $\widehat Z_A$ solves the minimal sufficient problem
\begin{equation}
    \widehat Z_A
    =
    \operatorname*{argmin}_{Z_A\in\mathcal S_A}
    I(X_A;Z_A).
    \label{eq:app-prior-selected-minimal}
\end{equation}
Moreover, if $|A|\ge M-1$, then $\widehat Z_A$ is coherence-sufficient:
\begin{equation}
\TC(X_{1:M}\mid \widehat Z_A)=0.
\label{eq:app-prior-route-coherence}
\end{equation}
\end{proposition}

\emph{Proof.}
We first show that every element of $\mathcal B_A$ is sufficient. Suppose $Z_A\in\mathcal B_A$ but $\rho_{A,m}(Z_A)>0$ for some target modality $m$. By the model capacity and optimization assumption, the encoder experts receiving gradients on that route, together with the decoder for modality $m$, admit a feasible update that decreases the corresponding expected target loss. This contradicts the definition of $\mathcal B_A$. Hence
\begin{equation}
\rho_{A,m}(Z_A)=0
\quad
\text{for all } m\in[M],
\end{equation}
and therefore
\begin{equation}
\mathcal B_A\subseteq\mathcal S_A.
\label{eq:app-B-subset-S}
\end{equation}

Since $Z_A^\star$ is assumed to be representable by the encoder family, it is feasible under the model class. By definition, $Z_A^\star\in\mathcal S_A$, so $\rho_{A,m}(Z_A^\star)=0$ for all $m\in[M]$. Under the idealized optimization assumption, no corresponding target loss can be further decreased by the feasible updates. Hence $Z_A^\star\in\mathcal B_A$.

Let $\widehat Z_A$ be a solution of Eq.~\ref{eq:app-prior-selected}. Since $Z_A^\star\in\mathcal B_A$, prior selection gives
\begin{equation}
\mathcal C_A(\widehat Z_A)
\le
\mathcal C_A(Z_A^\star).
\label{eq:app-prior-selection-ineq}
\end{equation}
Since $\widehat Z_A,Z_A^\star\in\mathcal S_A$, the contrapositive of Eq.~\ref{eq:app-prior-ordering} gives
\begin{equation}
I(X_A;\widehat Z_A)
\le
I(X_A;Z_A^\star).
\end{equation}

On the other hand, $Z_A^\star$ minimizes $I(X_A;Z_A)$ over $\mathcal S_A$, and $\widehat Z_A\in\mathcal S_A$. Therefore the two inequalities are tight, and $\widehat Z_A$ also solves Eq.~\ref{eq:app-prior-selected-minimal}.

It remains to show coherence-sufficiency for the subset latents used as prior targets. If $A=[M]$, then $\widehat Z_A\in\mathcal S_A$ directly gives
\begin{equation}
\TC(X_{1:M}\mid \widehat Z_A)=0.
\end{equation}
Now consider the leave-one-out case $A=A_i=[M]\setminus{i}$. Since $A_i^c={i}$, sufficiency gives
\begin{equation}
I(X_{A_i};X_i\mid \widehat Z_{A_i})=0.
\label{eq:app-loo-mi-zero}
\end{equation}
The self-target conditions inside $A_i$ imply, by Lemma~\ref{lem:app-self-residual-tc},
\begin{equation}
\TC(X_{A_i}\mid \widehat Z_{A_i})=0.
\label{eq:app-loo-observed-tc-zero}
\end{equation}
Using the subset decomposition in Eq.~\ref{eq:app-subset-decomposition},
\begin{equation}
\TC(X_{1:M}\mid \widehat Z_{A_i})
=
\TC(X_{A_i}\mid \widehat Z_{A_i})
+
\TC(X_i\mid \widehat Z_{A_i})
+
I(X_{A_i};X_i\mid \widehat Z_{A_i}).
\end{equation}
Since $X_i$ is a single modality, $\TC(X_i\mid \widehat Z_{A_i})=0$. Combining this with Eqs.~\ref{eq:app-loo-mi-zero} and~\ref{eq:app-loo-observed-tc-zero} in the decomposition above yields $\TC(X_{1:M}\mid \widehat Z_{A_i})=0$, as desired.

Thus full-modality and leave-one-out latents are coherence-sufficient for joint generation. \hfill$\square$

\section{Details of Reconstruction and Prior Losses}
\label{app:loss-implementation}

Eq.~\ref{eq:ours-objective} writes the reconstruction and prior terms abstractly as $\mathcal L_x^m$ and $\mathcal L_z$. 
This section gives the concrete losses used in our experiments. 
The prior term is the analogue of the KL term in a multimodal VAE:
\begin{equation}
    D_{\mathrm{KL}}\left(q_{\phi,A}(z\mid x_A)\,\|\,q_\psi(z)\right)
    =
    \mathbb E_{z\sim q_{\phi,A}}
    \left[-\log q_\psi(z)\right]
    -
    H\left(q_{\phi,A}(\cdot\mid x_A)\right).
    \label{eq:app-prior-kl-decomposition}
\end{equation}
As in Unified Latents~\cite{heek2026:ul}, the negative log-density under the learned prior is represented by a diffusion/flow objective. 
However, since directly optimizing the diffusion ELBO is ineffective for training the prior flow model~\cite{kingma2021:vdm}, we train the prior with the standard conditional flow-matching loss and apply the ELBO-correct gradient only to the encoder.

\paragraph{Flow prior.}
Let $q_{\phi,A}$ be a diagonal Gaussian posterior and sample
\begin{equation}
    z\sim\mathcal N(\mu_{\phi,A}(x_A),\sigma_{\phi,A}^2(x_A)I).
\end{equation}
The flow prior uses the linear path
\begin{equation}
    z_u = u z + (1-u)\epsilon,
    \qquad
    \epsilon\sim\mathcal N(0,I),
    \qquad
    u\sim p(u).
    \label{eq:app-prior-linear-path}
\end{equation}
The prior network is trained with the standard conditional flow-matching loss
\begin{equation}
    \ell_z(\psi;z,u,\epsilon)
    =
    \left\|
    v_\psi(z_u,u) - (z-\epsilon)
    \right\|_2^2.
    \label{eq:app-prior-cfm-loss}
\end{equation}
For the encoder, however, we use the ELBO-corrected gradient for the prior likelihood, following the derivation of Unified Latents~\cite{heek2026:ul}.
For the linear path in Eq.~\ref{eq:app-prior-linear-path},
\begin{equation}
    \mathrm{SNR}(u)
    =
    \frac{u^2}{(1-u)^2},
    \qquad
    \frac{1}{2}\frac{d}{du}\mathrm{SNR}(u)
    =
    \frac{u}{(1-u)^3}.
\end{equation}
Thus, the model-dependent part of the continuous-time ELBO weights the clean-latent prediction error as
\begin{equation}
    \mathcal J_{\mathrm{ELBO}}(\psi;z)
    =
    \int_0^1
    \frac{u}{(1-u)^3}
    \mathbb E_{\epsilon}
    \left[
    \|\hat z_0(z_u,u)-z\|_2^2
    \right]du,
    \label{eq:app-vdm-bound}
\end{equation}
where
\begin{equation}
    \hat z_0(z_u,u)
    =
    z_u+(1-u)v_\psi(z_u,u).
\end{equation}
Since
\begin{equation}
    \|\hat z_0-z\|_2^2
    =
    (1-u)^2
    \|v_\psi(z_u,u)-(z-\epsilon)\|_2^2,
\end{equation}
sampling $u\sim p(u)$ gives the ELBO-correct importance factor for the velocity loss:
\begin{equation}
    \omega_{\mathrm{ELBO}}(u)
    =
    \frac{u}{(1-u)p(u)}.
    \label{eq:app-elbo-v-weight}
\end{equation}
The factor $1/p(u)$ corrects for non-uniform timestep sampling.

In implementation, we apply this weight only to the encoder-side gradient using a gradient-scaling operator $\operatorname{gs}_{\lambda}$. 
It is the identity in the forward pass and scales the backward gradient by $\lambda$:
\begin{equation}
    \operatorname{gs}_{\lambda}(z)=z
    \quad\text{in the forward pass},
    \qquad
    \frac{\partial\,\operatorname{gs}_{\lambda}(z)}{\partial z}
    =
    \lambda I
    \quad\text{in the backward pass}.
    \label{eq:app-scaled-stopgrad}
\end{equation}
For the prior route, we set
\begin{equation}
    \lambda_z(u)
    =
    \gamma_z\,\omega_{\mathrm{ELBO}}(u),
    \label{eq:app-prior-grad-scale}
\end{equation}
where $\gamma_z$ is the configured \texttt{encoder\_grad\_scale}. 
This scalar controls the encoder-side strength of the prior term relative to reconstruction, and its values are reported in Table~\ref{tab:train-config-all}.

The entropy term of the KL in Eq.~\ref{eq:app-prior-kl-decomposition} is analytic for the diagonal Gaussian posterior. 
Let $\lambda_z(u)=\gamma_z\omega_{\mathrm{ELBO}}(u)$, 
$\tilde z=\operatorname{gs}_{\lambda_z(u)}(z)$, and 
$\tilde z_u=u\tilde z+(1-u)\epsilon$. 
We use
\begin{equation}
    \mathcal L_z(\psi,\phi;A)
    =
    \mathbb E_{\substack{
    z\sim q_{\phi,A}(\cdot\mid x_A)\\
    \epsilon\sim\mathcal N(0,I),\,u\sim p(u)
    }}
    \left[
    \left\|
    v_\psi(\tilde z_u,u)-(\tilde z-\epsilon)
    \right\|_2^2
    \right]
    -
    \gamma_z
    H\left(q_{\phi,A}(\cdot\mid x_A)\right).
    \label{eq:app-prior-loss-final}
\end{equation}
Since $\operatorname{gs}_{\lambda}$ is the identity in the forward pass, the prior network is trained with the standard conditional flow-matching loss, while the encoder receives the ELBO-correct gradient scale $\lambda_z(u)$ and the analytic entropy term.

\paragraph{Decoder losses.}
The reconstruction loss $\mathcal L_x^m$ is chosen according to the target modality. 
For a flow decoder with continuous target $x_m$, we use the linear path
\begin{equation}
    x_{m,u}
    =
    u x_m+(1-u)\epsilon,
    \qquad
    \epsilon\sim\mathcal N(0,I),
    \qquad
    u\sim p_m(u),
    \label{eq:app-decoder-linear-path}
\end{equation}
and train with the conditional flow-matching loss
\begin{equation}
    \mathcal L_x^m(\theta_m;x_m,z)
    =
    \mathbb E_{u,\epsilon}
    \left[
    \left\|
    v_{\theta_m}(x_{m,u},u;z)
    -
    (x_m-\epsilon)
    \right\|_2^2
    \right].
    \label{eq:app-flow-decoder-loss}
\end{equation}
For categorical label modalities, we use cross-entropy,
\begin{equation}
    \mathcal L_x^m(\theta_m;y_m,z)
    =
    -\log p_{\theta_m}(y_m\mid z),
    \label{eq:app-classifier-loss}
\end{equation}
and for autoregressive text decoders, we use teacher forcing,
\begin{equation}
    \mathcal L_x^m(\theta_m;x_{m,1:T},z)
    =
    -\sum_{t=1}^T
    \log p_{\theta_m}(x_{m,t}\mid x_{m,<t},z).
    \label{eq:app-ar-loss}
\end{equation}

\section{Controlled Gaussian-Mixture Experiment}
\label{app:gaussian_mixture}

We additionally conduct a controlled synthetic experiment with a known shared-private structure. Unlike the main benchmarks, this setting gives direct control over which factors are shared by all modalities, which factors are shared by only a subset of modalities, and which factors are private to a single modality. This allows us to directly examine how each method handles shared and private factors in conditional and unconditional generation.

\paragraph{Setup.}
We construct a three-modality benchmark with modalities $M_1,M_2,M_3$. 
The data are generated from seven independent ground-truth factors: a $10$-dimensional factor shared by all modalities $z_{123}$, three $10$-dimensional pair-shared factors $z_{12},z_{13},z_{23}$, and three $20$-dimensional modality-private factors $z_1,z_2,z_3$. 
Each factor is sampled from a randomly generated Gaussian mixture with 64 components. 
Each observation $x_i\in\mathbb R^{50}$ is the concatenation of the factors involving modality $M_i$:
\[
\begin{aligned}
    x_1 &= (z_{123}, z_{12}, z_{13}, z_1),\\
    x_2 &= (z_{123}, z_{12}, z_{23}, z_2),\\
    x_3 &= (z_{123}, z_{13}, z_{23}, z_3).
\end{aligned}
\]

\paragraph{Additional baselines.}
We include MMVAE+~\cite{palumbo2023:mmvaeplus} to test whether explicit private latents remain useful with expressive decoders, and ShaLa~\cite{cui2025:shala} as an existing baseline with an expressive prior. 
All methods use the same encoder and decoder architectures For ShaLa, we replace the original Gaussian decoder with the same flow-matching decoder used by the other baselines. 
Since ShaLa does not support multi-conditioned generation, we do not evaluate it on the $2\to1$ setting.

\paragraph{Metrics.}
We evaluate three generation tasks: \emph{one-to-one}, \emph{many-to-one}, and \emph{unconditional}. For the conditional tasks, we report \emph{alignment}, the mean squared error (MSE) on the shared coordinates between each conditioning modality and the generated target. For unconditional co-generation, we report \emph{fidelity} as the sliced Wasserstein distance (SWD) between generated samples and the ground-truth marginal of each generated modality, and \emph{coherence} as the shared-coordinate MSE across pairs of jointly sampled modalities. All metrics are averaged across supported source-target configurations, and lower values are better.

\paragraph{Implementation details.}
We train on $10^6$ procedurally generated training examples and evaluate on a held-out test split. All reported toy results are computed with $N=20{,}000$ samples per task. SWD is computed with 128 random projections.

All methods use the same backbone family: modality-specific MLP encoders, an MLP flow-matching prior, and modality-specific MLP flow-matching decoders. The latent has dimension 48 for MVAE~\cite{wu2018:mvae}, MMVAE~\cite{shi2019:mmvae}, MoPoE~\cite{sutter2021:mopoe}, HELVAE~\cite{vo2026:helvae}, and \Ours{}; MMVAE+~\cite{palumbo2023:mmvaeplus} uses the same 48-dimensional shared latent together with an additional 24-dimensional private latent per modality. Apart from method-specific components such as MMVAE+'s private latents and ShaLa's two-stage conditional prior, the baselines share the same MLP encoder, flow prior, and flow decoder backbones. Flow timesteps are sampled from a logit-normal distribution during training. At evaluation, we use a fixed 50-step sampler budget for both the learned prior and the modality decoders, with the same sampler settings used across methods. Other details are summarized in Tab.~\ref{tab:train-config-all}.

\begin{table}[t]
\small
\centering
\caption{\textbf{Controlled Gaussian-mixture benchmark.}
Alignment is the shared-coordinate MSE between conditioning and target modalities for conditional generation. For unconditional co-generation, fidelity is measured by SWD to the ground-truth modality marginal, and coherence is the shared-coordinate MSE across jointly sampled modalities. Lower is better. \textbf{Bold} marks the best result and \underline{underline} marks the second-best. A dash indicates that the corresponding setting is not evaluated for that method.}
\label{tab:gmm_full}

  \begin{tabular}{lcccc}
  \toprule
  & \multicolumn{1}{c}{1$\to$1}
  & \multicolumn{1}{c}{2$\to$1}
  & \multicolumn{2}{c}{Uncond.} \\
  \cmidrule(lr){2-2}\cmidrule(lr){3-3}\cmidrule(lr){4-5}
  Method & Align. $\downarrow$ & Align. $\downarrow$ & Fid. $\downarrow$ & Coh. $\downarrow$ \\
  \midrule
  MVAE~\cite{wu2018:mvae}       & 0.2265              & 0.3233              & 1.5996              & \underline{0.2801} \\
  MMVAE~\cite{shi2019:mmvae}    & \textbf{0.0267}     & 5.0647              & \underline{0.2080}  & 2.0496 \\
  MoPoE~\cite{sutter2021:mopoe} & 0.0558              & \underline{0.0604}  & 0.2720              & 0.6297 \\
  HELVAE~\cite{vo2026:helvae}   & 0.6325              & 2.1096              & 0.4345              & 4.3642 \\
  MMVAE+~\cite{palumbo2023:mmvaeplus} & \underline{0.0276} & 5.0413        & 2.8070              & 2.0432 \\
  ShaLa~\cite{cui2025:shala}    & 7.2122              & -                   & 0.8075              & 2.5984 \\
  \textbf{\Ours{} (Ours)}       & 0.0508              & \textbf{0.0414}     & \textbf{0.1300}     & \textbf{0.0394} \\
  \bottomrule
  \end{tabular}

\end{table}

\paragraph{Results.}
Tab.~\ref{tab:gmm_full} reports the quantitative results. MoE-based methods, including MMVAE and MMVAE+, achieve strong $1\to1$ alignment, but degrade sharply in the $2\to1$ setting and in unconditional coherence. This suggests that mixture-based aggregation does not reliably fuse evidence from multiple observed modalities or support coherent joint sampling. MoPoE improves conditional alignment through product-based aggregation at inference, but its unconditional coherence remains limited as expected.

MVAE and HELVAE are less competitive overall.
Their shared latent is pressured to carry both shared content and modality-specific variation, which hurts performance under limited latent capacity. 
MMVAE+ introduces private latents, but does not meaningfully improve over MMVAE in this benchmark, indicating that explicit private latents provide limited benefit once the decoders are expressive. 
ShaLa uses an expressive prior but does not support multi-conditioned generation; in the evaluated settings, it remains weak on $1\to1$ alignment and unconditional coherence under the same training budget.

\Ours{} achieves the best $2\to1$ alignment and the strongest unconditional performance with the lowest fidelity and coherence errors.
These results support the latent-space criteria in Sec.~\ref{sec:analysis}: combining all observed evidence supports predictivity in many-to-one generation, while restricting prior learning to leave-one-out latents improves coherence in unconditional co-generation.

\section{More Experimental Details and Results}
\label{app:imple_details}

We provide additional implementation details and extended results for Sec.~\ref{sec:experiments}.

\subsection{PolyMNIST-Quadrant-Labels}
\label{app:polymnist_details}

\paragraph{Architecture.}
All compatible methods use the same backbone family. 
The shared latent is 32-dimensional, while MMVAE+ uses an additional 32-dimensional private latent for each modality.
The image modalities use a shared modality-conditioned ResNet encoder with modality-specific input and posterior heads, and a shared NCSN++ flow-matching decoder whose target modality is specified by a modality label.
The digit and quadrant modalities use MLP encoders and classifier decoders. 
Please refer to Tab.~\ref{tab:train-config-all} for additional architectural details.

\paragraph{Training.}
Image flow decoders use logit-normal timestep sampling following SD3~\cite{esser2024:sd3}, while latent priors use uniform timestep sampling. See Tab.~\ref{tab:train-config-all} for more training details.

\paragraph{Evaluation.}
All metrics are computed on the full 10k held-out split using the same definitions as in Sec.~\ref{sec:exp-polymnist}. 
A frozen verifier predicts digit and quadrant labels from generated images. 
At evaluation time, we use 50 prior sampling steps and 25 image-decoder sampling steps.

\begin{table}[t]
\centering
\small
\caption{\textbf{PolyMNIST-Quadrant-Labels.}
Alignment is verifier accuracy on generated images for conditional label-to-image tasks. 
Coherence is strict agreement between verifier predictions on $M_1,M_2,M_3$ and the generated labels under unconditional sampling. 
Higher is better. \textbf{Bold} marks the best result and \underline{underline} marks the second-best. 
A dash indicates an unsupported setting.}
\label{tab:polymnist_full}
\begin{tabular}{lcccc}
\toprule
& \multicolumn{2}{c}{Single-L $\to$ I}
& \multicolumn{1}{c}{Multi-L $\to$ I}
& \multicolumn{1}{c}{Uncond.} \\
\cmidrule(lr){2-3}\cmidrule(lr){4-4}\cmidrule(lr){5-5}
Method & Digit $\uparrow$ & Quad. $\uparrow$ & Align. $\uparrow$ & Coh. $\uparrow$ \\
\midrule
MVAE~\cite{wu2018:mvae}     & 0.4137 & 0.6803 & 0.7053 & 0.0079 \\
MMVAE~\cite{shi2019:mmvae}    & 0.9279 & \textbf{0.9999} & 0.1630 & 0.3167 \\
MoPoE~\cite{sutter2021:mopoe}    & 0.9199 & \textbf{0.9999} & \underline{0.9283} & 0.3943 \\
HELVAE~\cite{vo2026:helvae}   & \underline{0.9297} & \textbf{0.9999} & 0.7654 & 0.4246 \\
MMVAE+~\cite{palumbo2023:mmvaeplus}   & \textbf{0.9368} & \textbf{0.9999} & 0.1668 & 0.2783 \\
ShaLa~\cite{cui2025:shala}    & 0.8976 & \underline{0.9915} & - & \underline{0.4539} \\
\textbf{\Ours{} (Ours)} & 0.9131 & \textbf{0.9999} & \textbf{0.9346} & \textbf{0.4841} \\
\bottomrule
\end{tabular}
\end{table}

\paragraph{Results.}
Tab.~\ref{tab:polymnist_full} extends the main comparison with MMVAE+~\cite{palumbo2023:mmvaeplus} and ShaLa~\cite{cui2025:shala}. 
MMVAE+ improves single-label digit alignment, but remains weak under multi-label conditioning and unconditional co-generation, suggesting that explicit private latents provide limited additional benefit once the modality decoders are expressive. 
ShaLa achieves the second-best unconditional coherence, but does not support multi-label conditioning and is less competitive under single-label conditioning.
\Ours{} achieves the best multi-label alignment and the best unconditional coherence while remaining competitive under single-label conditioning.

\subsection{Image-Text-Audio Any-to-Any Generation}
\label{app:ita-details}

\paragraph{Additional baselines.}
In addition to the generalist any-to-any baselines reported in the main paper, we include task-specific specialists for all six one-to-one generation directions:
\begin{itemize}[leftmargin=1.5em]
    \item \textbf{T$\rightarrow$I:} FLUX.1-dev~\cite{flux2024}, a 12B rectified-flow transformer for prompt-conditioned image synthesis.
    \item \textbf{I$\rightarrow$T:} LLaVA-NeXT~\cite{liu2024llavanext}, a visual instruction-tuned large multimodal model with improved high-resolution image understanding, OCR, and reasoning capabilities.
    \item \textbf{T$\rightarrow$A:} TangoFlux~\cite{hung2025:tangoflux}, an efficient flow-matching text-to-audio model trained with CLAP-ranked preference optimization for improved text-audio alignment.
    \item \textbf{A$\rightarrow$T:} Qwen2-Audio~\cite{chu2024:qwen2audio}, a large audio-language model that accepts speech, music, and environmental audio as input and produces natural-language responses.
    \item \textbf{I$\rightarrow$A:} Seeing and Hearing~\cite{xing2024:seeinghearing}, a visual-guided audio generation method that aligns pretrained visual and audio diffusion models through a shared latent space.
    \item \textbf{A$\rightarrow$I:} Sound2Vision~\cite{sungbin2024:sound2vision}, an audio-to-image method that maps audio representations into a visual latent space and uses a pretrained image generator to synthesize plausible scenes from sound.
\end{itemize}
These specialist models are optimized for individual translation directions and are therefore reported as reference rows rather than direct comparison to generalist any-to-any systems. We also include UnifiedIO2-L~\cite{lu2023:unifiedio2} as an LLM-based generalist baseline. UnifiedIO2-L tokenizes images, text, audio, and other modalities into a shared autoregressive sequence space, but it does not support audio-image translation.

\paragraph{Dataset.}
We adopt the pairwise training data setup of FlowBind~\cite{cha2026:flowbind}, using paired text, image, and audio data without any fully observed triplets. 
For text--image training, we use 242K LAION-COCO~\cite{laion2025relaioncoco} pairs, filtered to examples with aesthetic scores above 5.0, together with 30K Flickr-30k~\cite{plummer2015:flickr30k} image--text pairs. 
For text--audio training, we use 91K AudioCaps v2~\cite{kim2019audiocaps} examples, which provide natural-language captions for YouTube audio clips. 
For audio--image training, we use 184K VGGSound~\cite{chen2020:vggsound} examples, a large-scale audio-visual dataset collected from YouTube videos.

For one-to-one evaluation, text-to-image and image-to-text are evaluated on the MS-COCO~\cite{lin2015:coco} evaluation split, while text-audio, audio-text, image-audio, and audio-image tasks are evaluated on held-out test splits of the corresponding pairwise datasets. 

\paragraph{Implementation Details.}
We follow FlowBind~\cite{cha2026:flowbind} to define multimodal latent learning in the aligned representation spaces of pretrained modality-specific encoders and decoders. 
For images, we use CLIP~\cite{radford2021:clip} as the encoder and Stable-UnCLIP~\cite{huggingface2025stableunclip} as the pretrained image generation backbone. 
For audio, we use CLAP~\cite{elizalde2023:clap} as the encoder and AudioLDM~\citep{liu2023:audioldm} as the pretrained audio generation backbone. 
For text, we use EmbeddingGemma~\cite{vera2025:embeddinggemma} as the encoder and Gemma3-1B~\cite{gemmateam2025:gemma3technicalreport} as the autoregressive text decoder. 
This design allows us to train over compact semantic representations.

Unlike FlowBind~\cite{cha2026:flowbind}, which maps all modalities into a fixed-dimensional shared latent space, our framework allows modality-specific decoder heads with different output dimensions. 
We use this flexibility to fine-tune the autoregressive text decoder for stronger text generation. 
Specifically, we fine-tune the Gemma3-1B text decoder using LoRA~\cite{hu2022:lora} adapters with rank $r=16$, scaling factor $\alpha=32$, and dropout rate $0.05$.
All other encoders, decoders, and the prior network are implemented as MLPs with residual connections.

For encoder aggregation, we instantiate two variants: \Ours{} uses product-of-experts aggregation, while \Ours{}$^\dagger$ uses Hellinger aggregation. 
For fair comparisons, we reproduce the multimodal VAE baselines~\cite{shi2019:mmvae,sutter2021:mopoe} using the same pretrained encoders and decoders, optimization settings, and compute budget as \Ours{}. Other details are summarized in Tab.~\ref{tab:train-config-all}.

\paragraph{Training.}
We use modality-specific cross-reconstruction losses. 
For text, we train the autoregressive decoder with cross-entropy loss on the first 64 examples of each effective batch of 1024, weighted by 0.05, and add a representation alignment loss between the final 768-dimensional output of the projection MLP and the target EmbeddingGemma~\cite{vera2025:embeddinggemma} vector. 
For image and audio, we use flow matching and MSE losses, respectively. In Monte Carlo training, we sample both cross-reconstruction and prior-learning routes with a 3:2 sampling ratio favoring cross-reconstruction, which empirically improves conditional generation while preserving prior coherence.

\paragraph{Sampling.} For image generation, we use Euler sampling with CFG scale 4.0 and 25 inference steps. For text generation, we use greedy sampling. 

\paragraph{Evaluation.}
We measure fidelity with FID~\cite{heusel2017:fid} for images, FAD~\cite{kilgour2019:fad} for audio, and CIDEr~\cite{vedantam2015:cider} for text. 
For cross-modal alignment, we use CLIPScore~\cite{hessel2021:clipscore} for text--image pairs, CLAPScore~\cite{elizalde2023:clap} for text--audio pairs, and AIS~\cite{wu2022:wav2clip} for audio--image pairs.

\paragraph{One-to-one results.}
Tabs.~\ref{tab:ita-fidelity} and~\ref{tab:ita-alignment} report the full one-to-one fidelity and alignment results. 
In fidelity, \Ours{} and \Ours{}$^\dagger$ achieve the best or second-best generalist performance on most directions, with particularly large gains in text-to-image generation (14.40 FID vs.\ 20.13 for FlowBind~\cite{cha2026:flowbind}) and strong results on image-to-audio and audio-to-image generation (1.90 FAD and 22.97 FID). 
For image-to-text, \Ours{}$^\dagger$ obtains competitive CIDEr scores, while UnifiedIO2-L~\cite{lu2023:unifiedio2} and FlowBind achieve the best scores on image-to-text and audio-to-text, respectively. 
The high CIDEr score of UnifiedIO2-L is expected, as it relies on a LLM-based decoder specialized for text generation.

For alignment, \Ours{}$^\dagger$ is strongest on audio-related directions, achieving the best generalist CLAP score for text-to-audio (31.81) and the best AIS scores for image-to-audio and audio-to-image (90.11 and 95.61). It also remains competitive on text-image alignment.  
Overall, these results show that our framework remains competitive on standard one-to-one generation while also providing stronger many-to-one aggregation and unconditional co-generation, as shown in the main text.

\begin{table}[h]
\centering
\caption{\textbf{Training configuration across experiments.}
Hyperparameters are shared between \Ours{} and the multimodal VAE baselines under matched-compute comparisons. Reported runtimes are measured on a single RTX PRO 6000 GPU.}
\label{tab:train-config-all}
\footnotesize
\setlength{\tabcolsep}{6pt}
\renewcommand{\arraystretch}{1.15}
\begin{tabular}{@{}l ccc@{}}
\toprule
& \textbf{Gaussian-mixture} & \textbf{PolyMNIST-Quadrant-Labels} & \textbf{Image-Text-Audio} \\
\midrule
\rowcolor{sectiongray}
\multicolumn{4}{@{}l}{\textbf{\emph{Optimization}}} \\
Optimizer            & & AdamW ($\beta_1{=}0.9$, $\beta_2{=}0.95$) & \\
Learning rate        & $4{\times}10^{-3}$ & $6{\times}10^{-4}$ & $4{\times}10^{-4}$ \\
LR schedule          & & constant, 5-epoch linear warmup & \\
Weight decay         & 0.0     & 0.0     & 0.0 \\
Grad clip ($\ell_2$) & 1.0     & 1.0     & 1.0 \\
Epochs               & 200 & 300 & $200$ \\
Batch size           & $4000$ & $1024$ & $1024$ \\
Total Runtime        & $\sim$10 min & $\sim$3 h & $\sim$3 h \\
EMA decay            & $0.999$ & $0.999$ & $0.999$ \\
\midrule
\rowcolor{sectiongray}
\multicolumn{4}{@{}l}{\textbf{\emph{Encoders}}} \\
Architecture         & MLP & ResNet (images), MLP (labels) & MLP \\
Aggregation          & PoE     & PoE     & \{PoE, Hellinger\} \\
Total params         & 4.6\,M & 8.8\,M & 90\,M \\
\midrule
\rowcolor{sectiongray}
\multicolumn{4}{@{}l}{\textbf{\emph{Prior}}} \\
Architecture         & MLP & MLP & MLP \\
Encoder grad scale & $0.1$; $0.01$ for ShaLa~\cite{cui2025:shala} & $0.1$ & $0.01$ \\
Total params         & 3.2\,M & 12.6\,M & 30\,M \\
\midrule
\rowcolor{sectiongray}
\multicolumn{4}{@{}l}{\textbf{\emph{Decoders}}} \\
Architecture         & MLP & NCSN++ (images), MLP (labels) & MLP (image, audio), Transformer (text) \\
Total params         & 14.0\,M & 10.7\,M & 140\,M \\
\bottomrule
\end{tabular}
\end{table}

\section{More Qualitative Results}

\begin{figure}[t]
\centering
\setlength{\tabcolsep}{2pt}
\renewcommand{\arraystretch}{1.0}
\begin{tabular}{c|cccc}
\footnotesize Source (T) &
\footnotesize CoDi~\cite{tang2023:codi} &
\footnotesize OmniFlow~\cite{li2025:omniflow} &
\footnotesize FlowBind~\cite{cha2026:flowbind} &
\footnotesize \textbf{\Ours{}$^\dagger$ (Ours)} \\
\midrule
\includegraphics[width=0.18\textwidth]{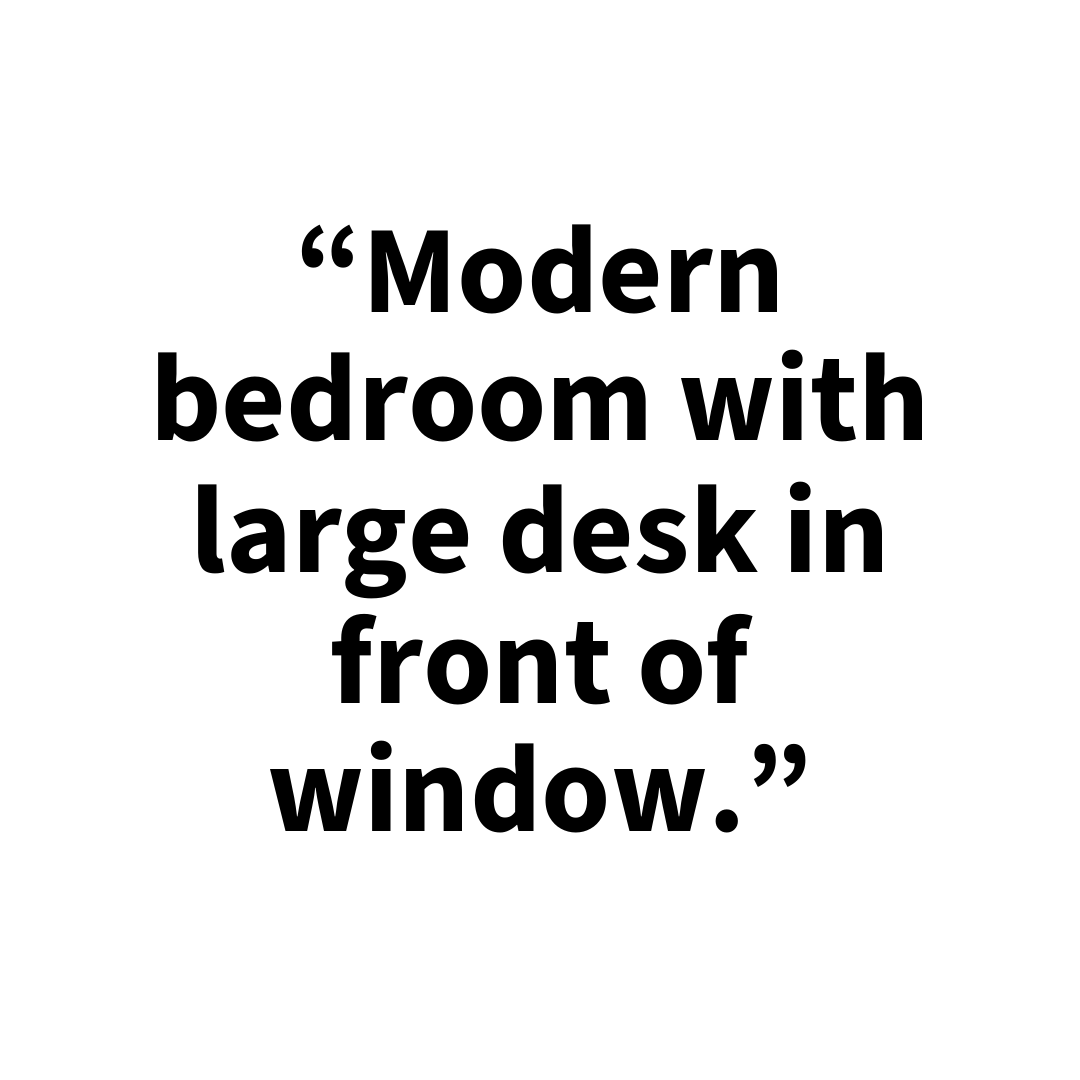} &
\includegraphics[width=0.18\textwidth]{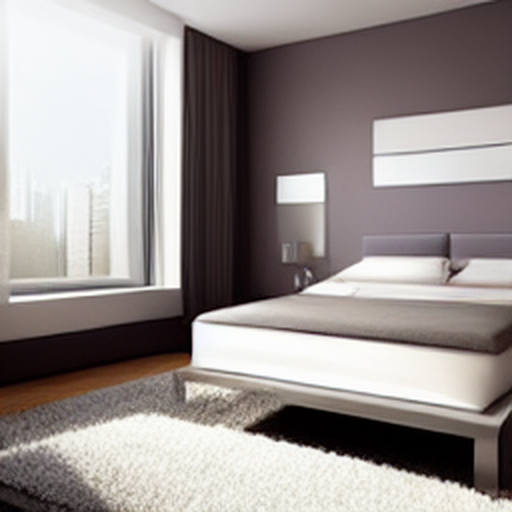} &
\includegraphics[width=0.18\textwidth]{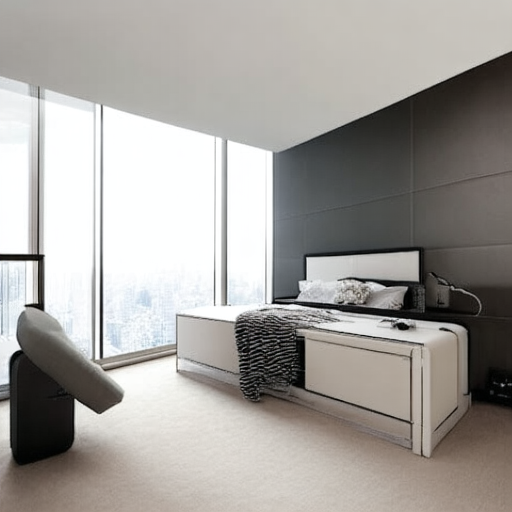} &
\includegraphics[width=0.18\textwidth]{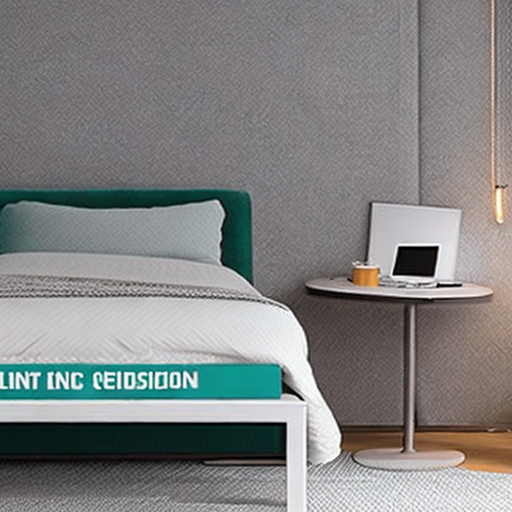} &
\includegraphics[width=0.18\textwidth]{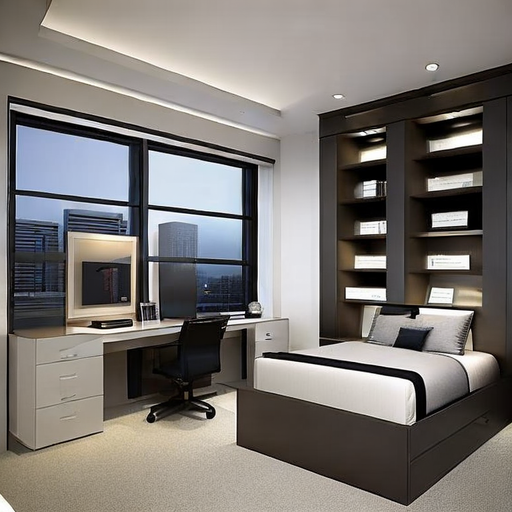} \\

\includegraphics[width=0.18\textwidth]{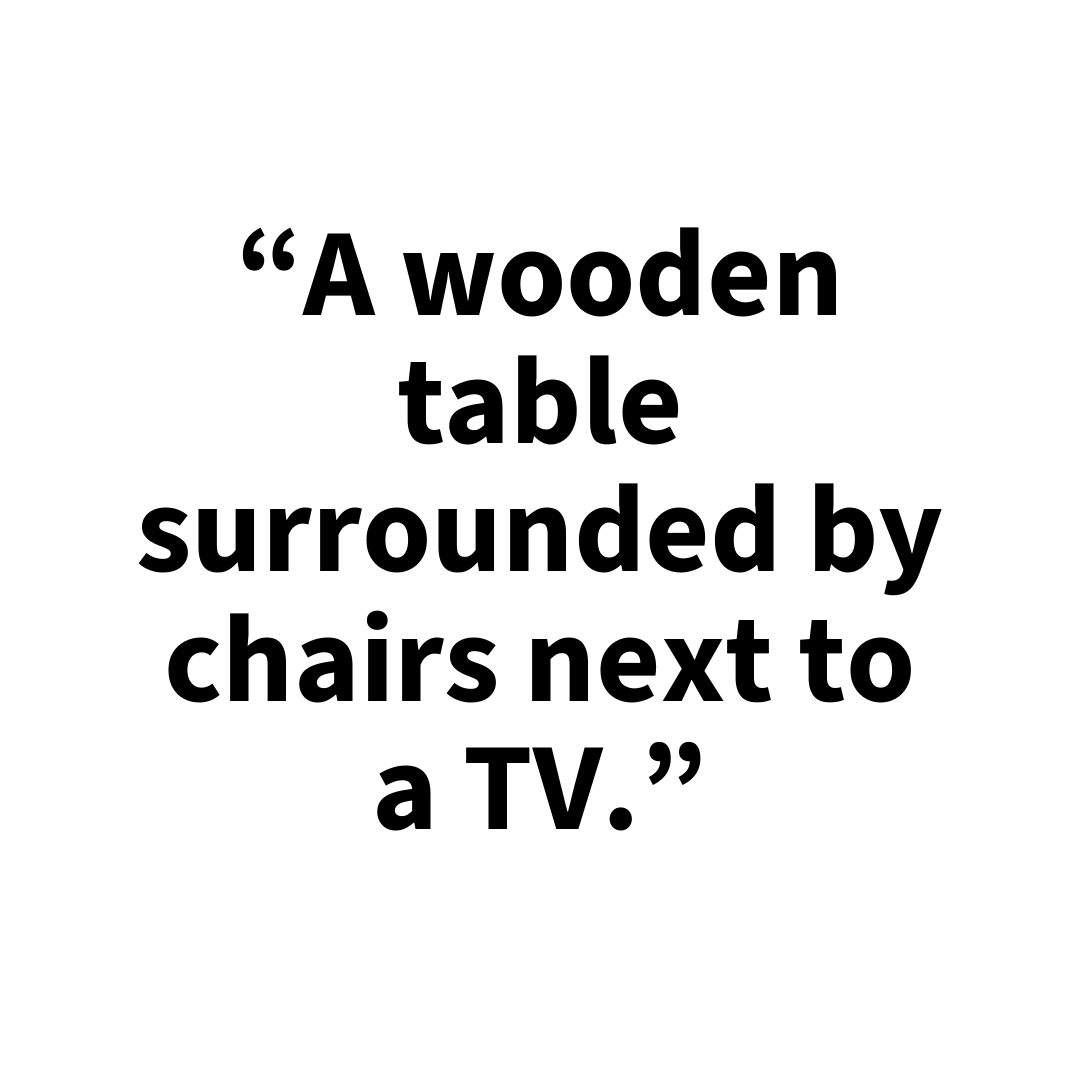} &
\includegraphics[width=0.18\textwidth]{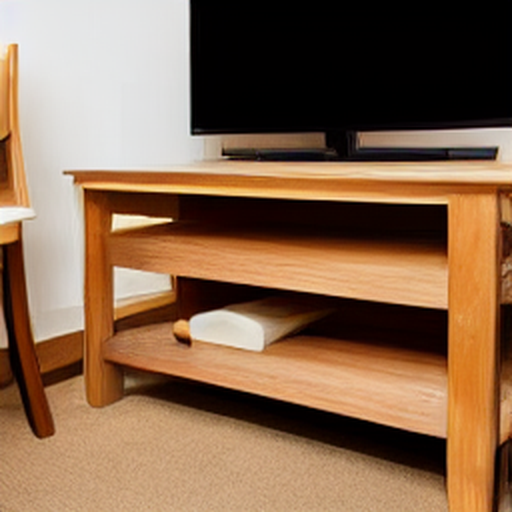} &
\includegraphics[width=0.18\textwidth]{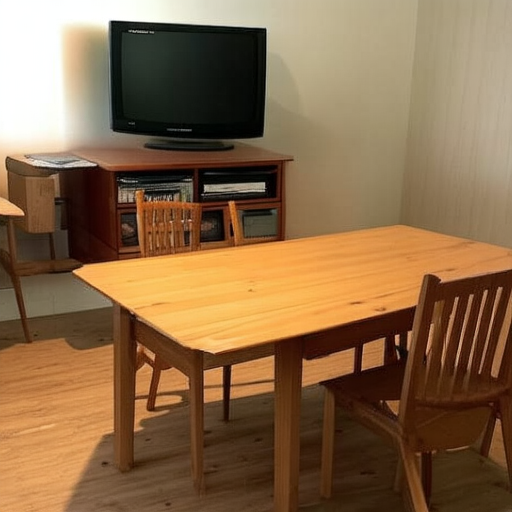} &
\includegraphics[width=0.18\textwidth]{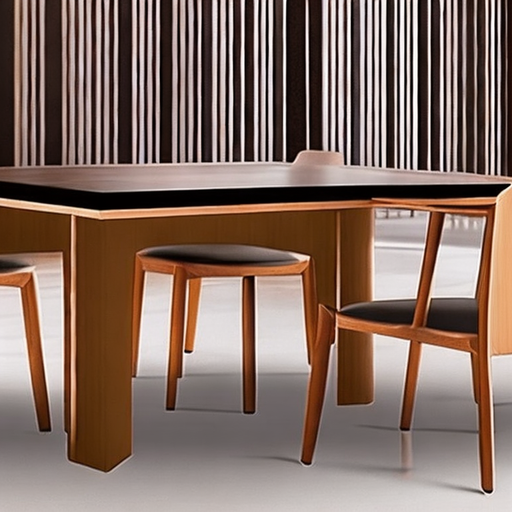} &
\includegraphics[width=0.18\textwidth]{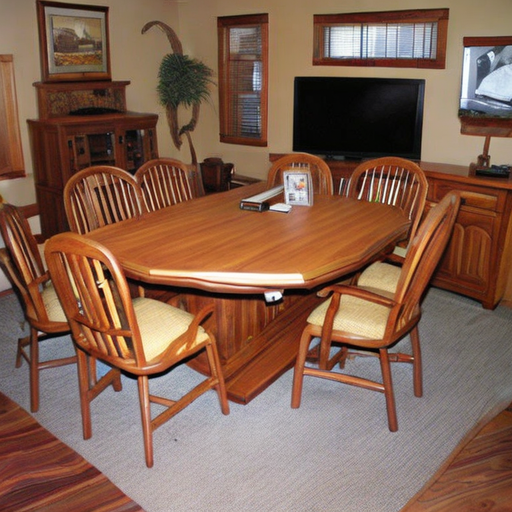} \\

\includegraphics[width=0.18\textwidth]{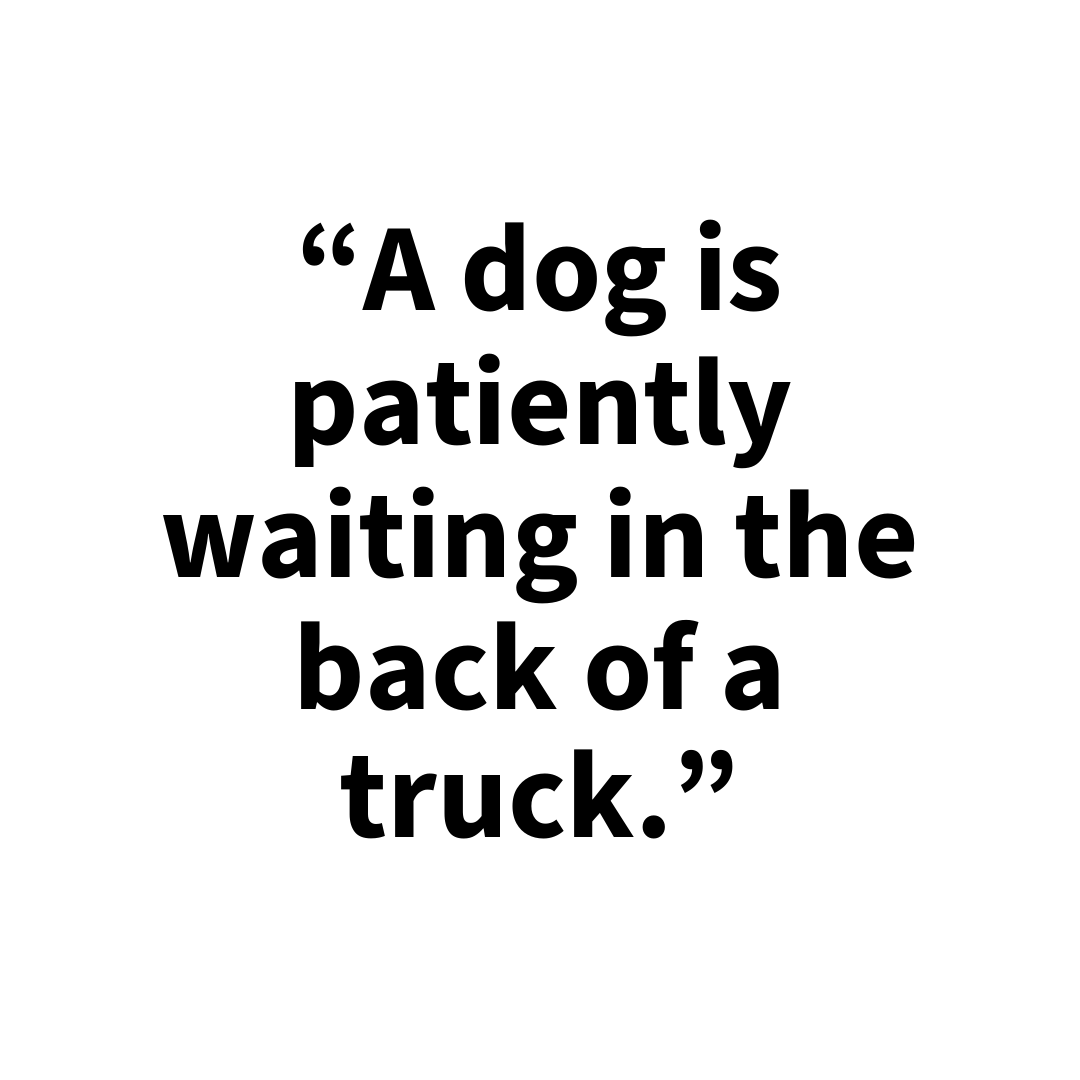} &
\includegraphics[width=0.18\textwidth]{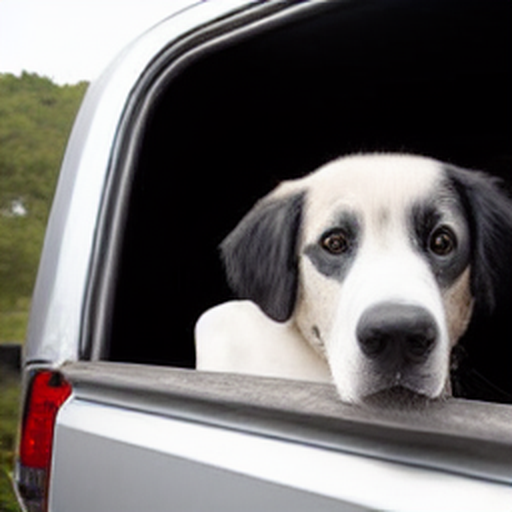} &
\includegraphics[width=0.18\textwidth]{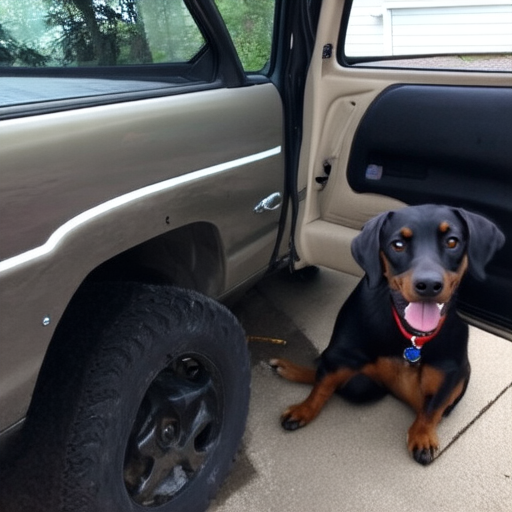} &
\includegraphics[width=0.18\textwidth]{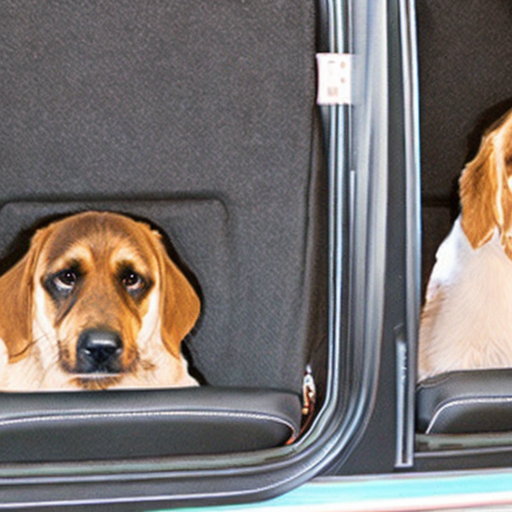} &
\includegraphics[width=0.18\textwidth]{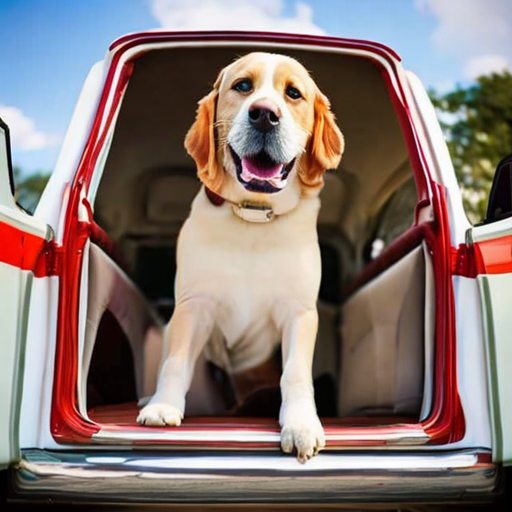} \\

\includegraphics[width=0.18\textwidth]{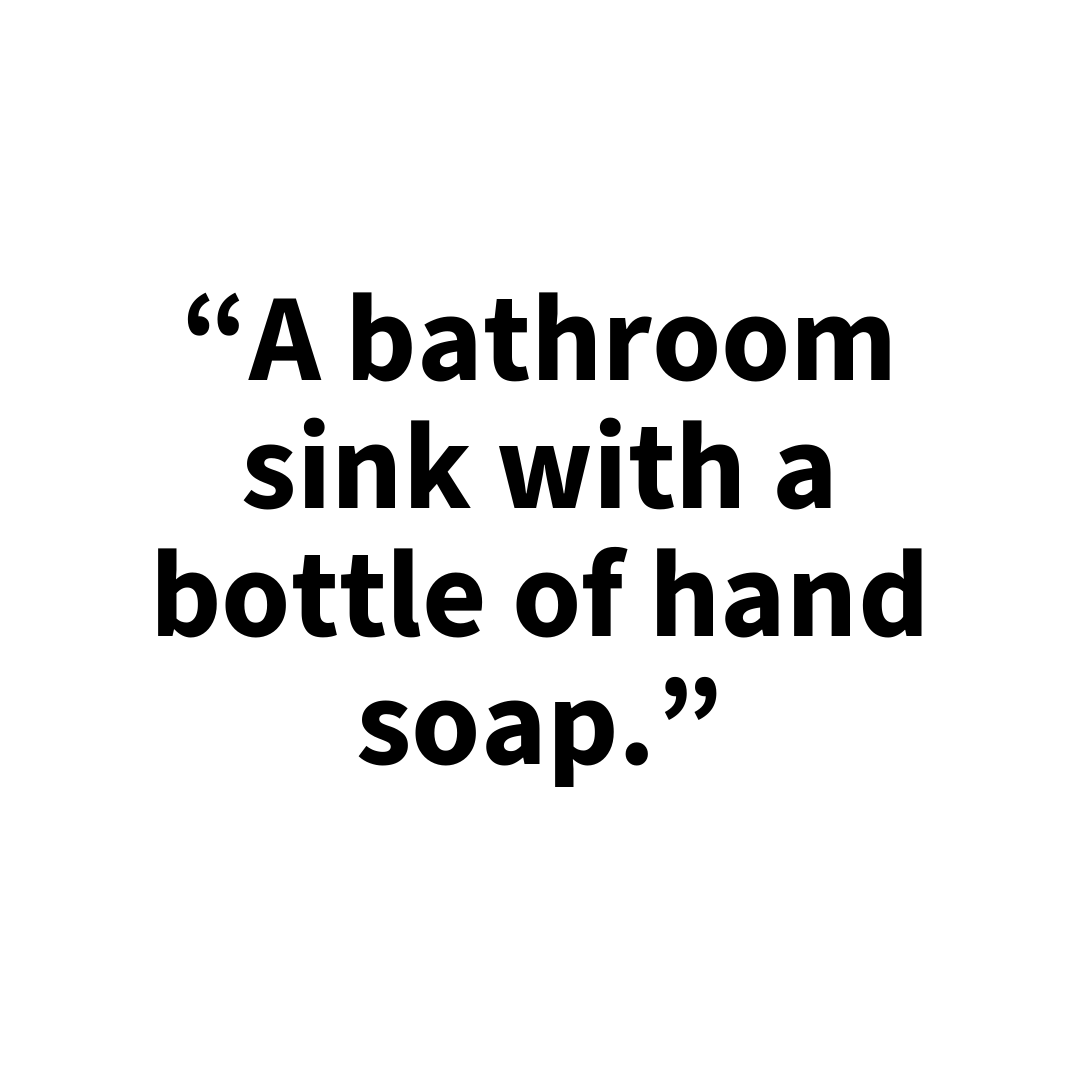} &
\includegraphics[width=0.18\textwidth]{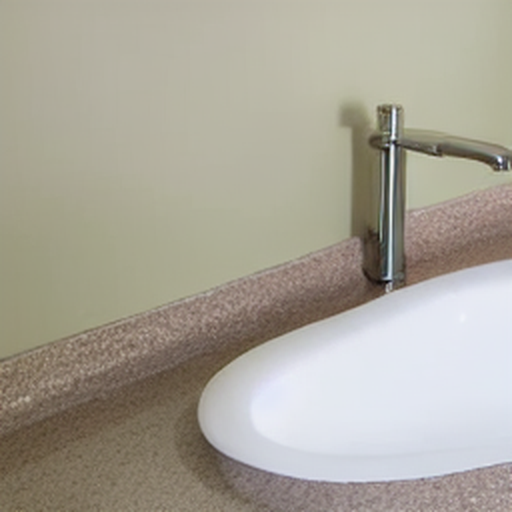} &
\includegraphics[width=0.18\textwidth]{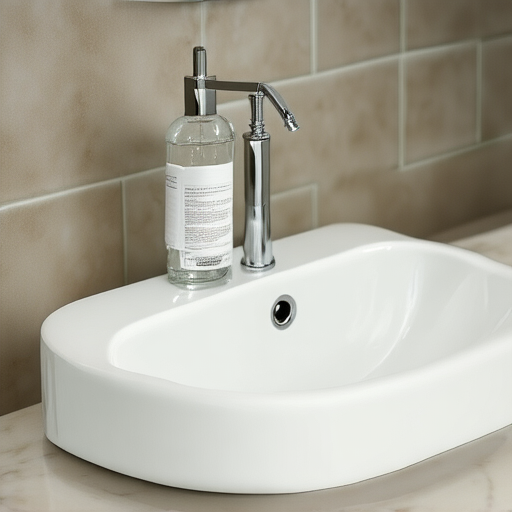} &
\includegraphics[width=0.18\textwidth]{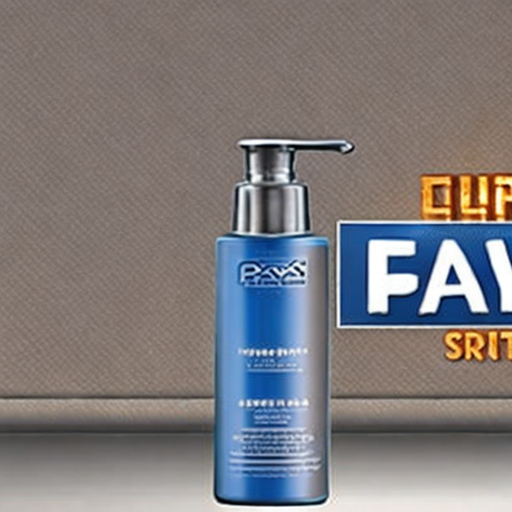} &
\includegraphics[width=0.18\textwidth]{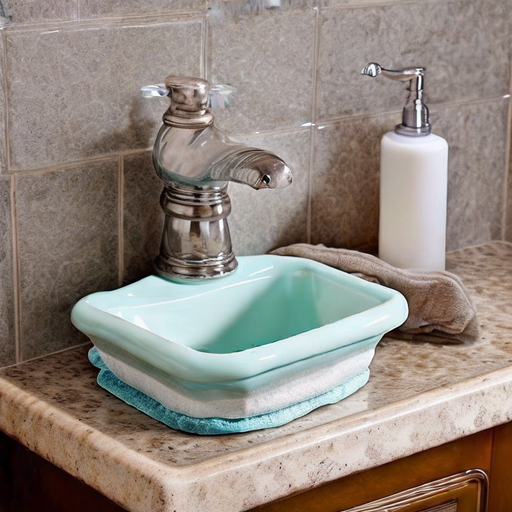} \\

\includegraphics[width=0.18\textwidth]{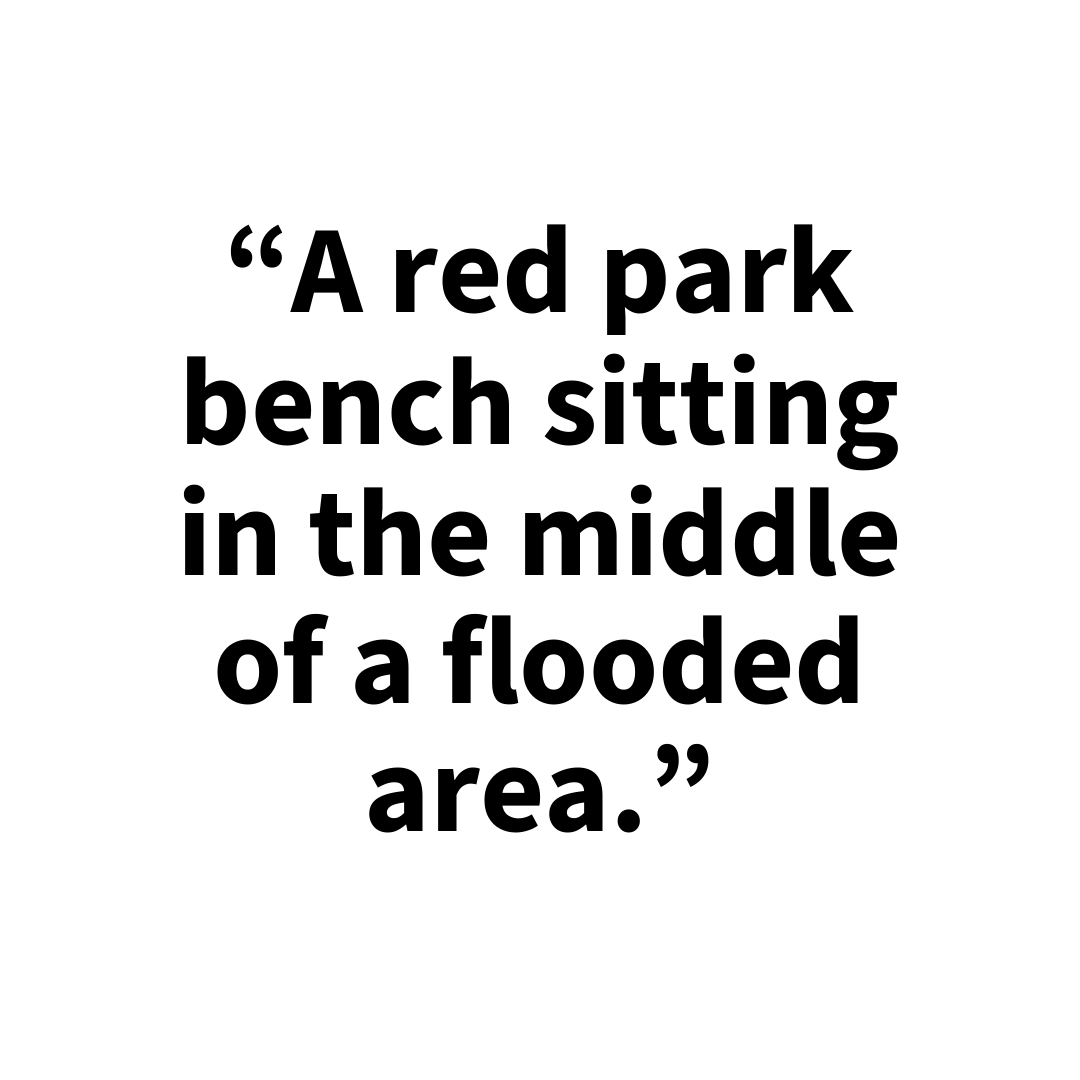} &
\includegraphics[width=0.18\textwidth]{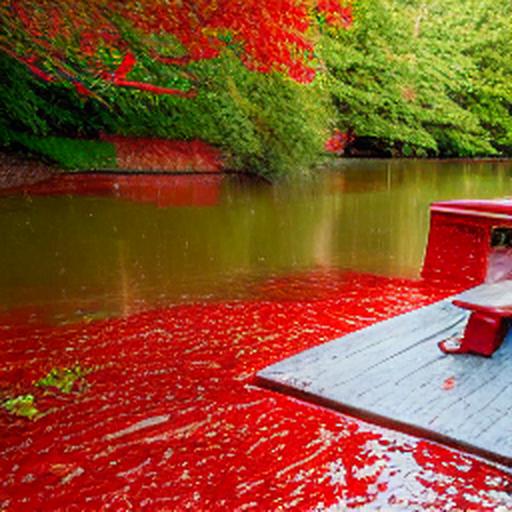} &
\includegraphics[width=0.18\textwidth]{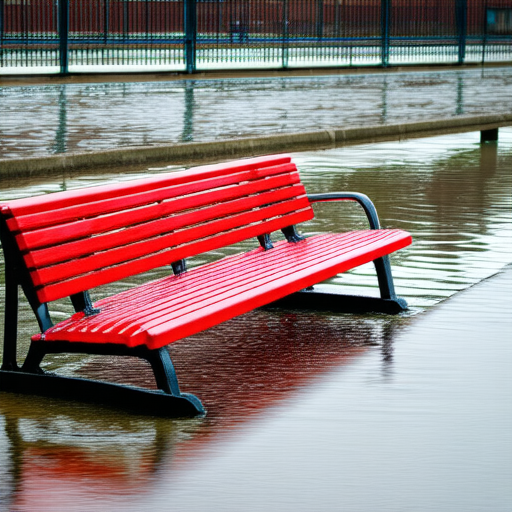} &
\includegraphics[width=0.18\textwidth]{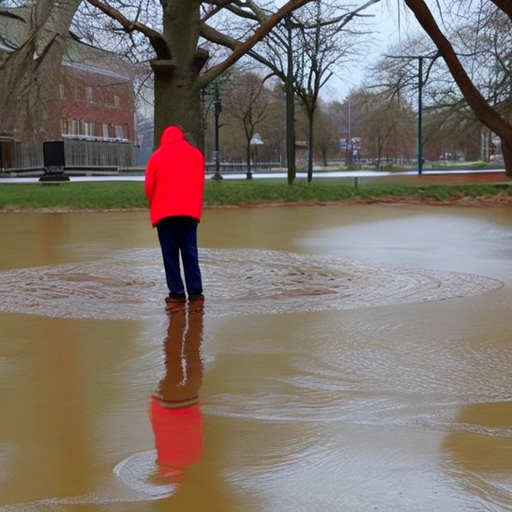} &
\includegraphics[width=0.18\textwidth]{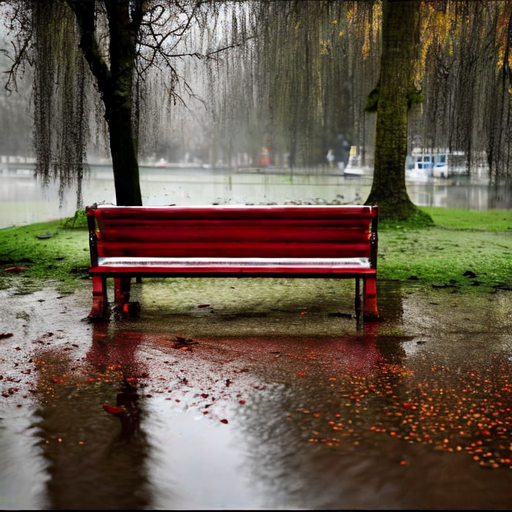} \\

\includegraphics[width=0.18\textwidth]{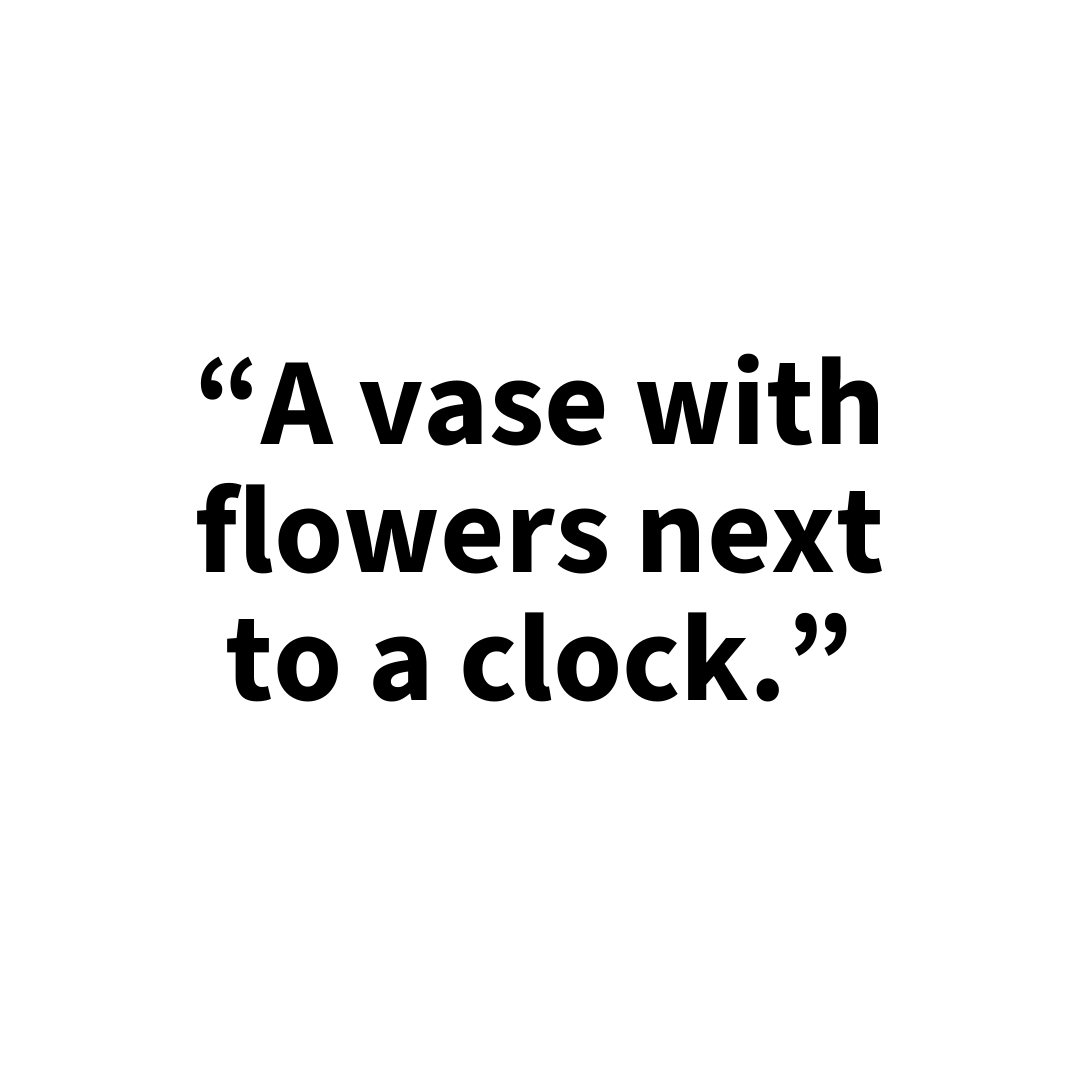} &
\includegraphics[width=0.18\textwidth]{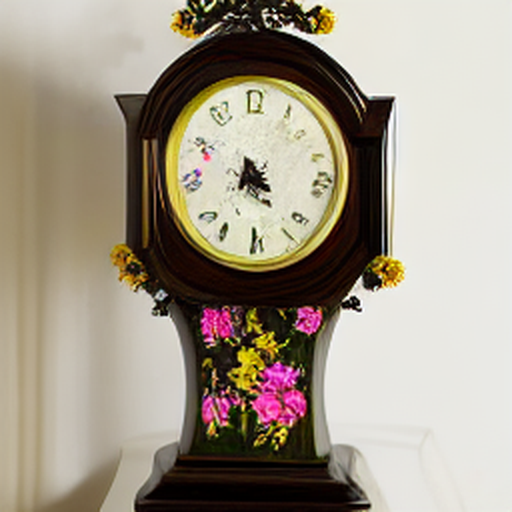} &
\includegraphics[width=0.18\textwidth]{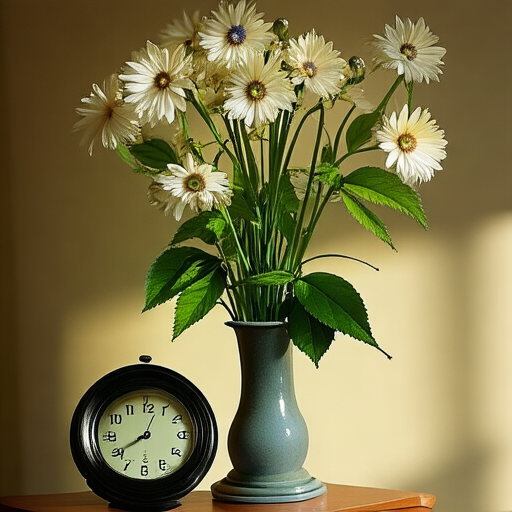} &
\includegraphics[width=0.18\textwidth]{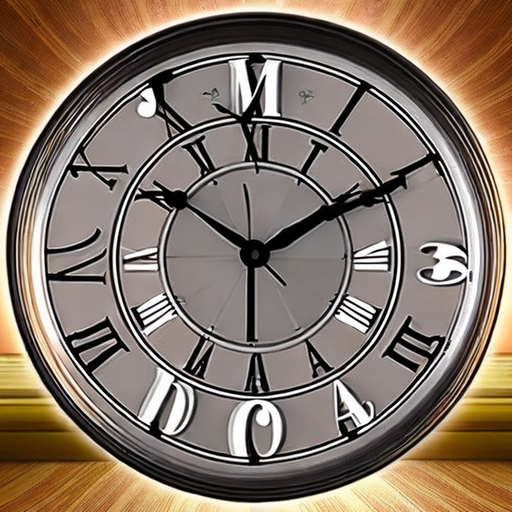} &
\includegraphics[width=0.18\textwidth]{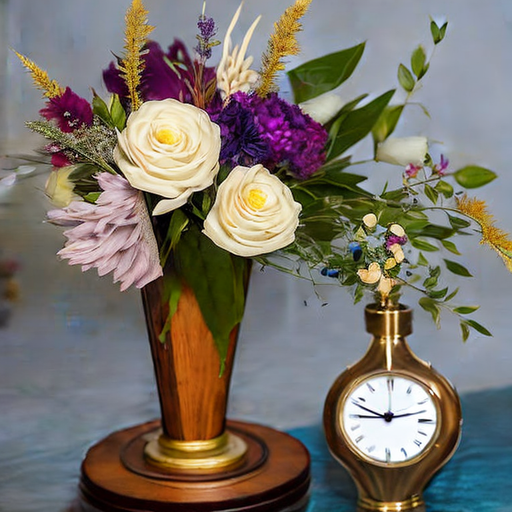} \\
\end{tabular}
\caption{\textbf{Text-to-image generation.} Given a source text (left), each method generates an image. \Ours{}$^\dagger$ produces an image semantically aligned with the source.}
\label{fig:ita-qual-t2i}
\end{figure}

\begin{figure}[t]
\centering
\setlength{\tabcolsep}{2pt}
\renewcommand{\arraystretch}{1.0}
\begin{tabular}{c|cccc}
\footnotesize Source (I) &
\footnotesize CoDi~\cite{tang2023:codi} &
\footnotesize OmniFlow~\cite{li2025:omniflow} &
\footnotesize FlowBind~\cite{cha2026:flowbind} &
\footnotesize \textbf{\Ours{}$^\dagger$ (Ours)} \\
\midrule
\includegraphics[width=0.18\textwidth]{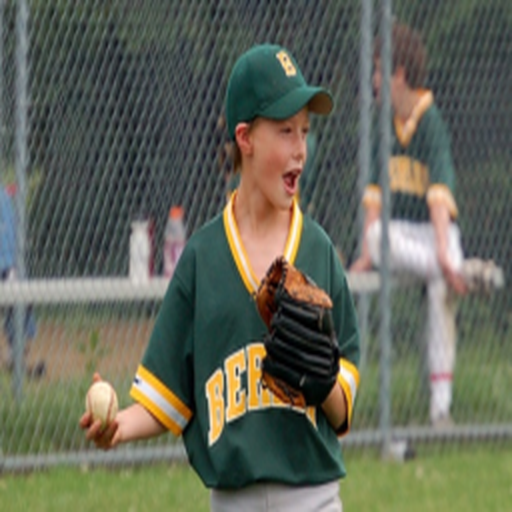} &
\includegraphics[width=0.18\textwidth]{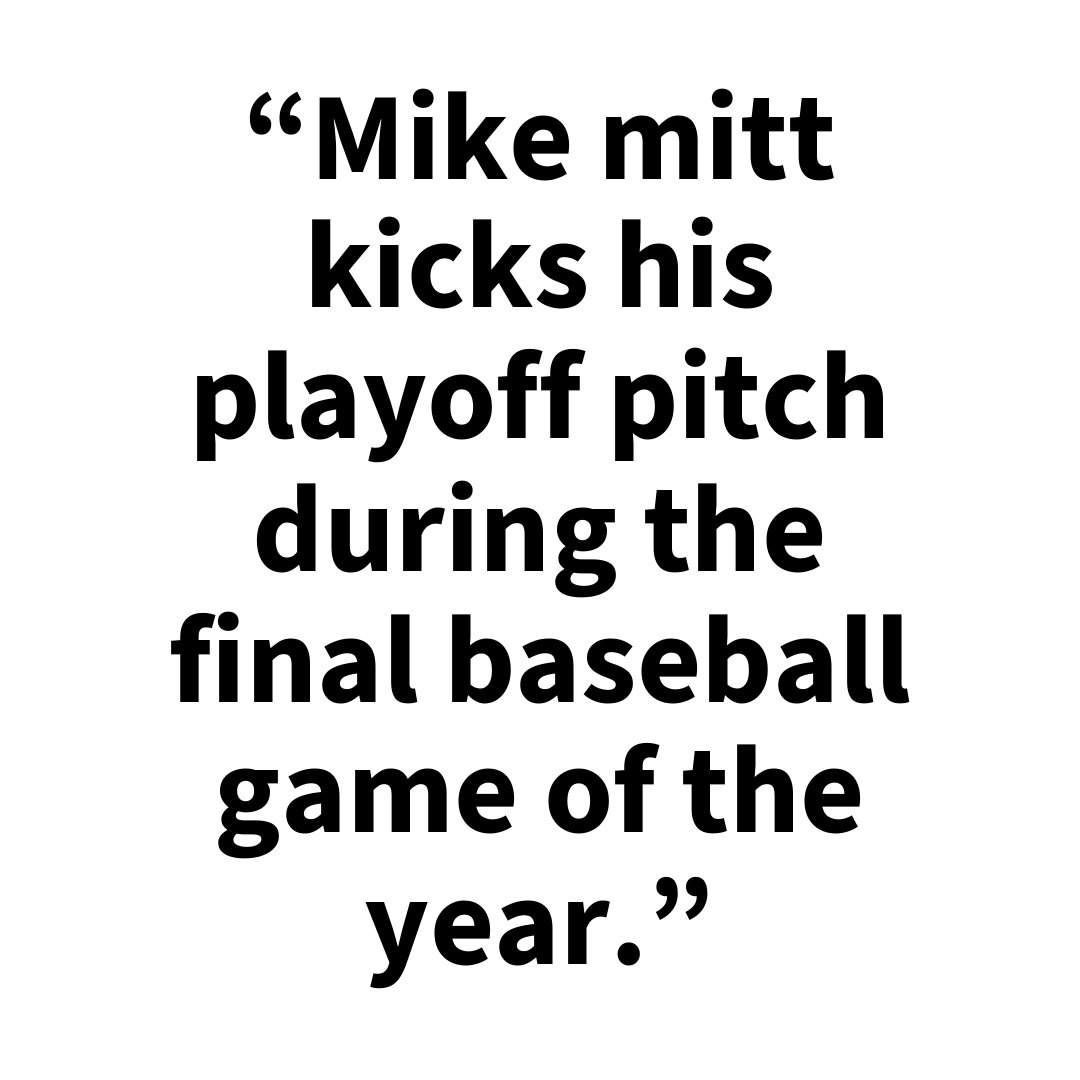} &
\includegraphics[width=0.18\textwidth]{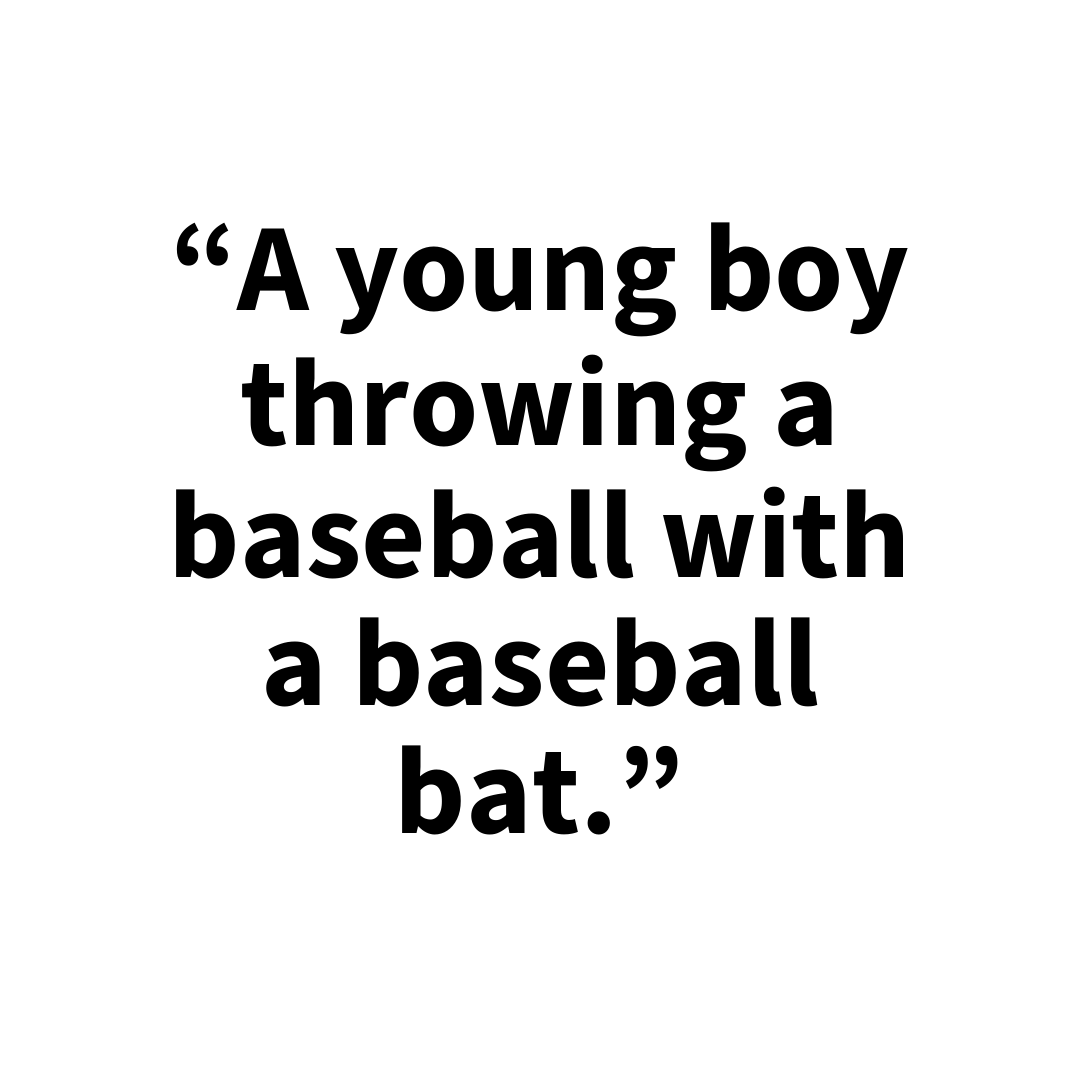} &
\includegraphics[width=0.18\textwidth]{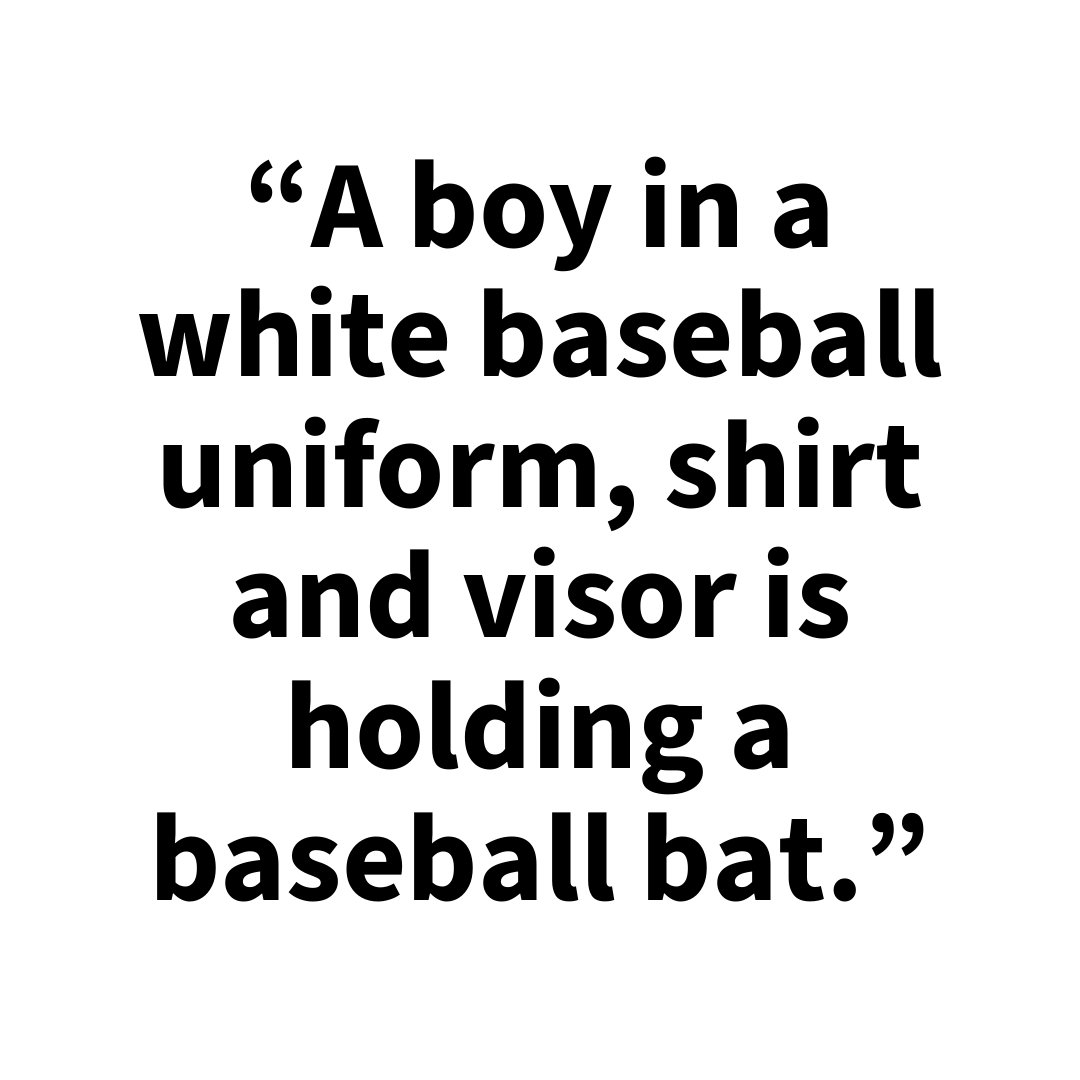} &
\includegraphics[width=0.18\textwidth]{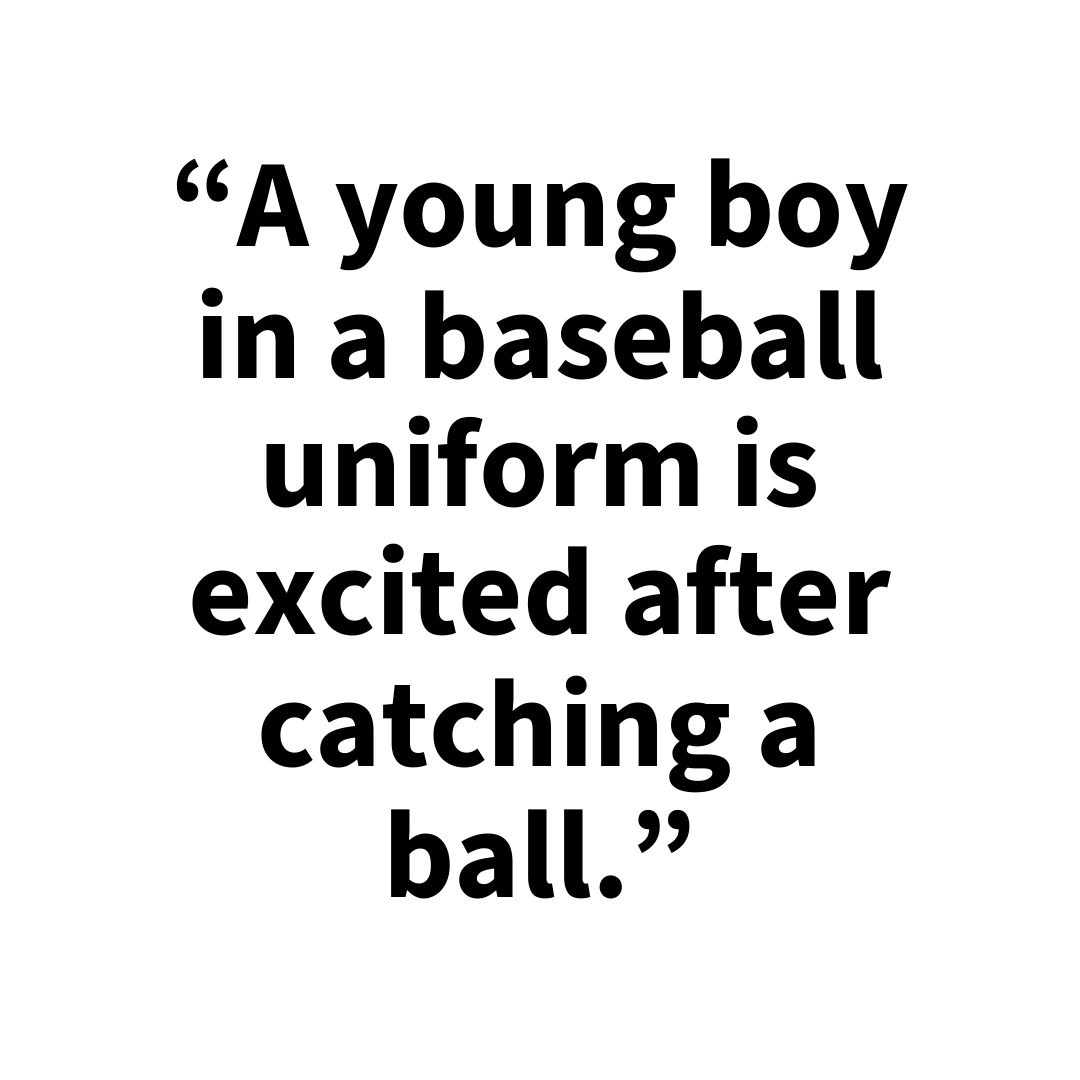} \\

\includegraphics[width=0.18\textwidth]{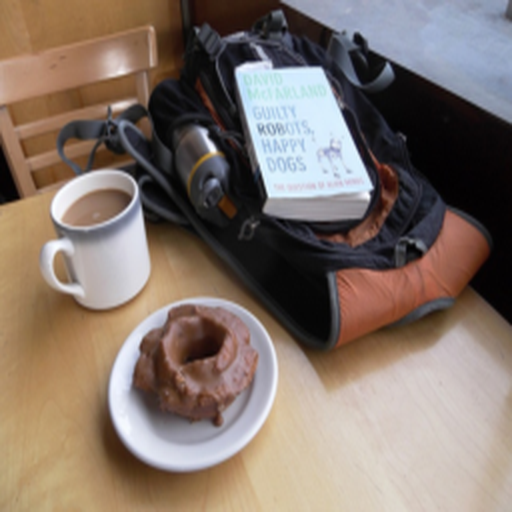} &
\includegraphics[width=0.18\textwidth]{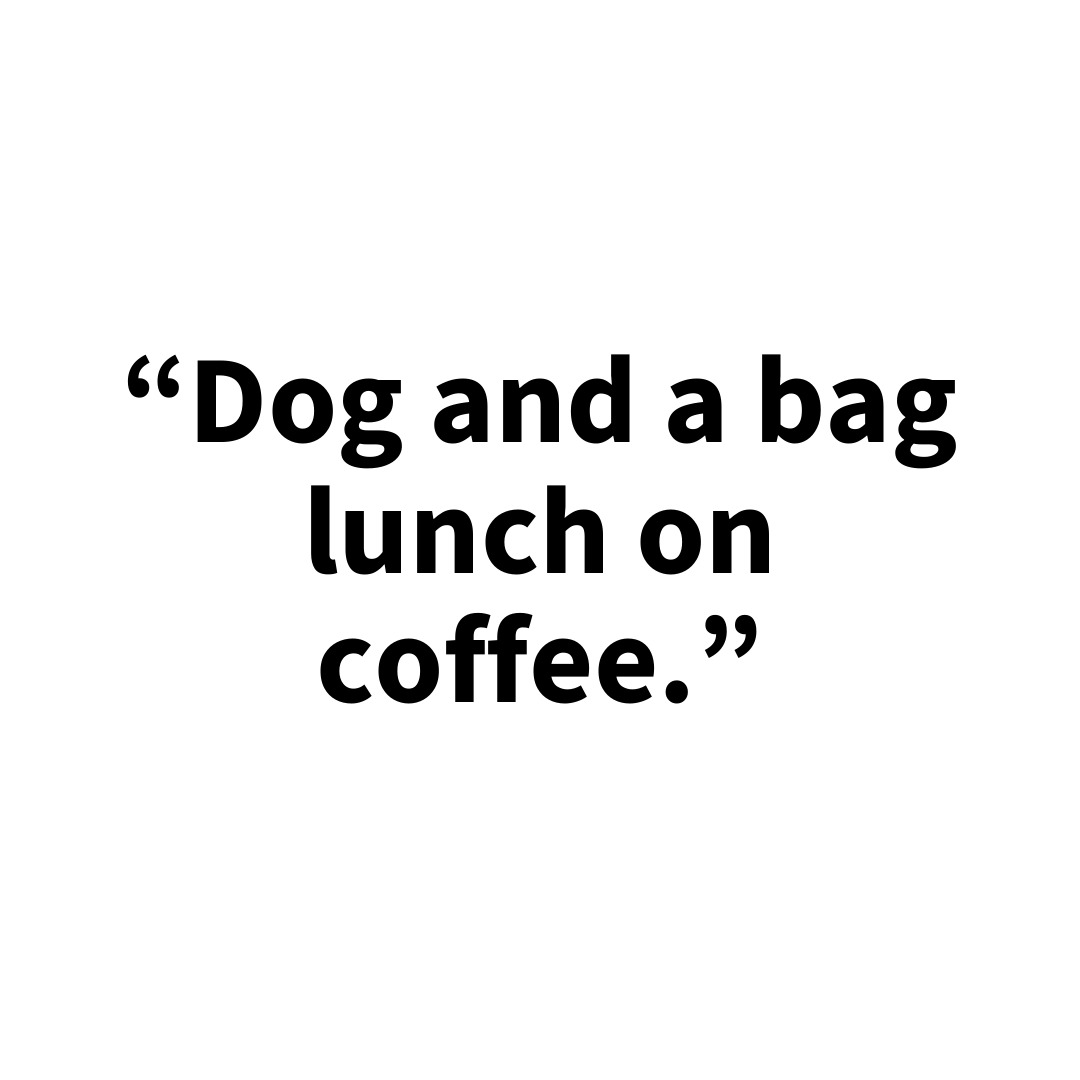} &
\includegraphics[width=0.18\textwidth]{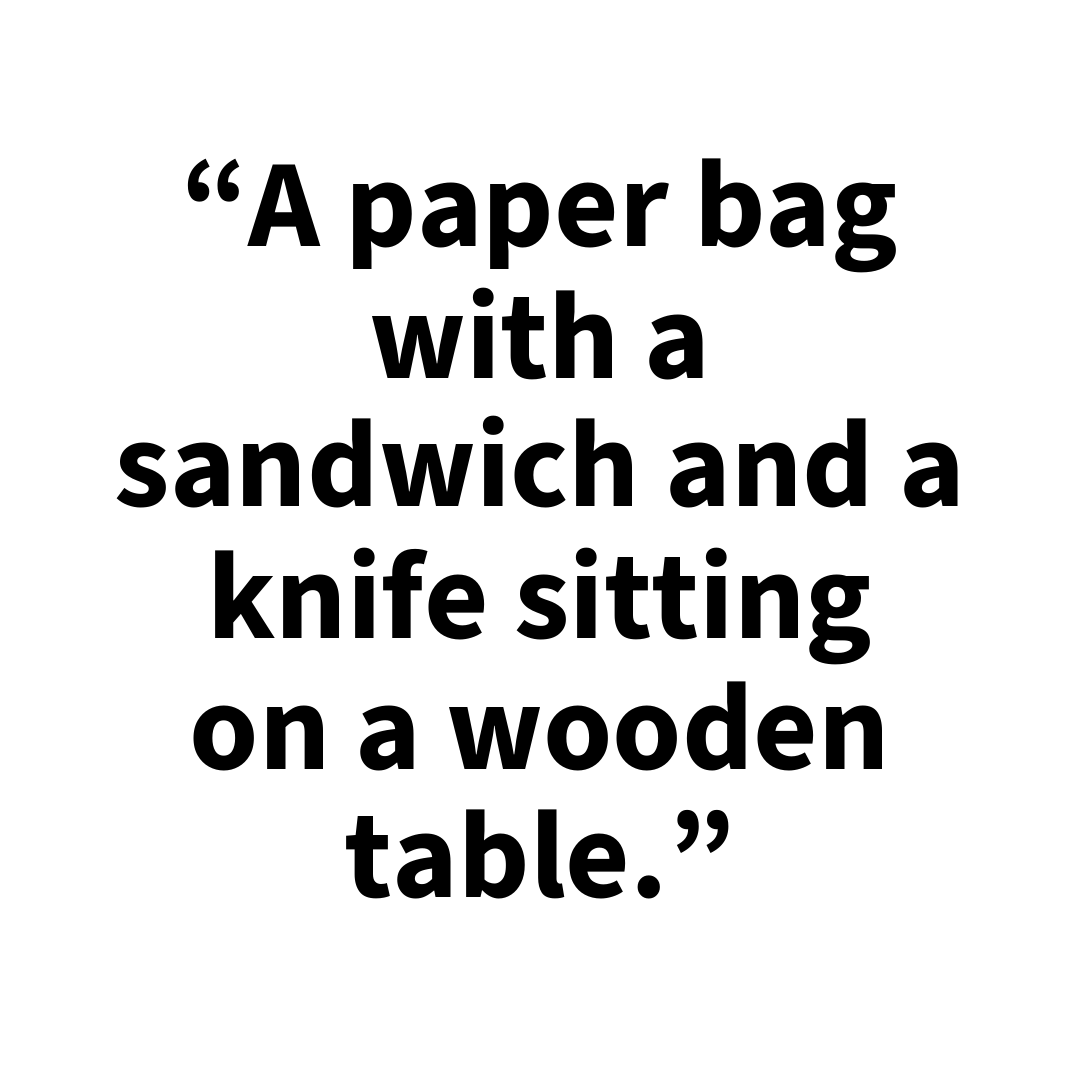} &
\includegraphics[width=0.18\textwidth]{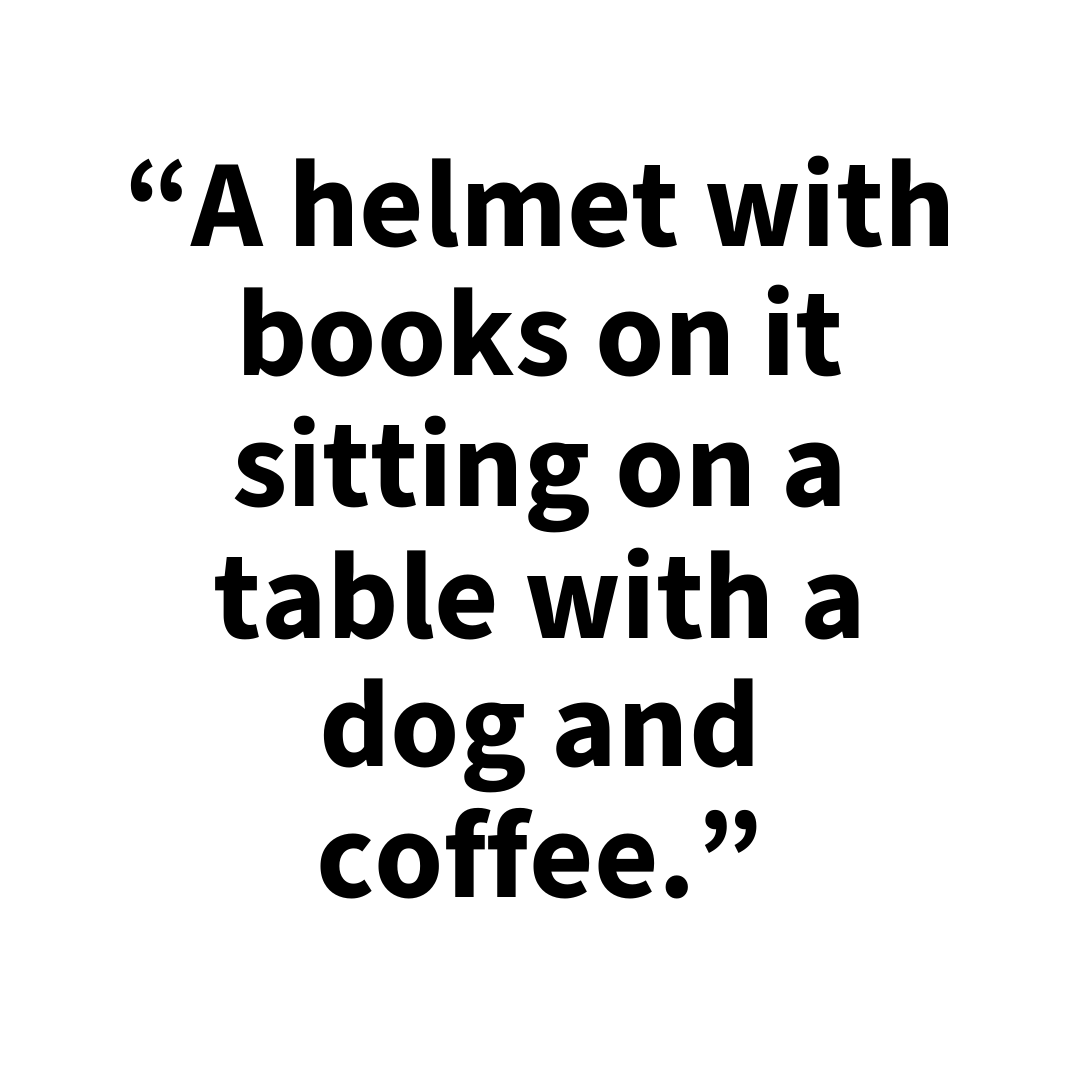} &
\includegraphics[width=0.18\textwidth]{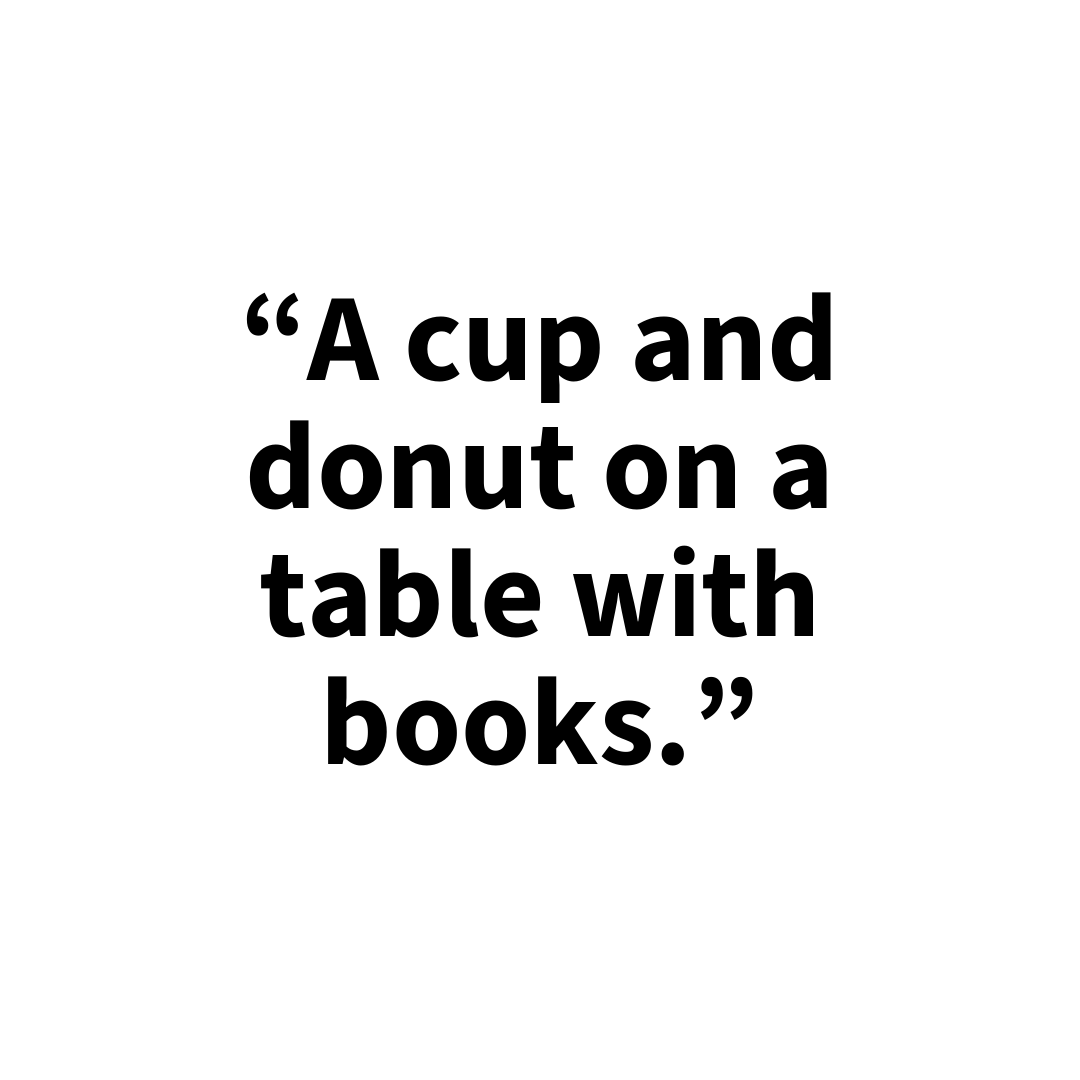} \\

\includegraphics[width=0.18\textwidth]{figures/qualitative_ita/I_to_T/065_src.png} &
\includegraphics[width=0.18\textwidth]{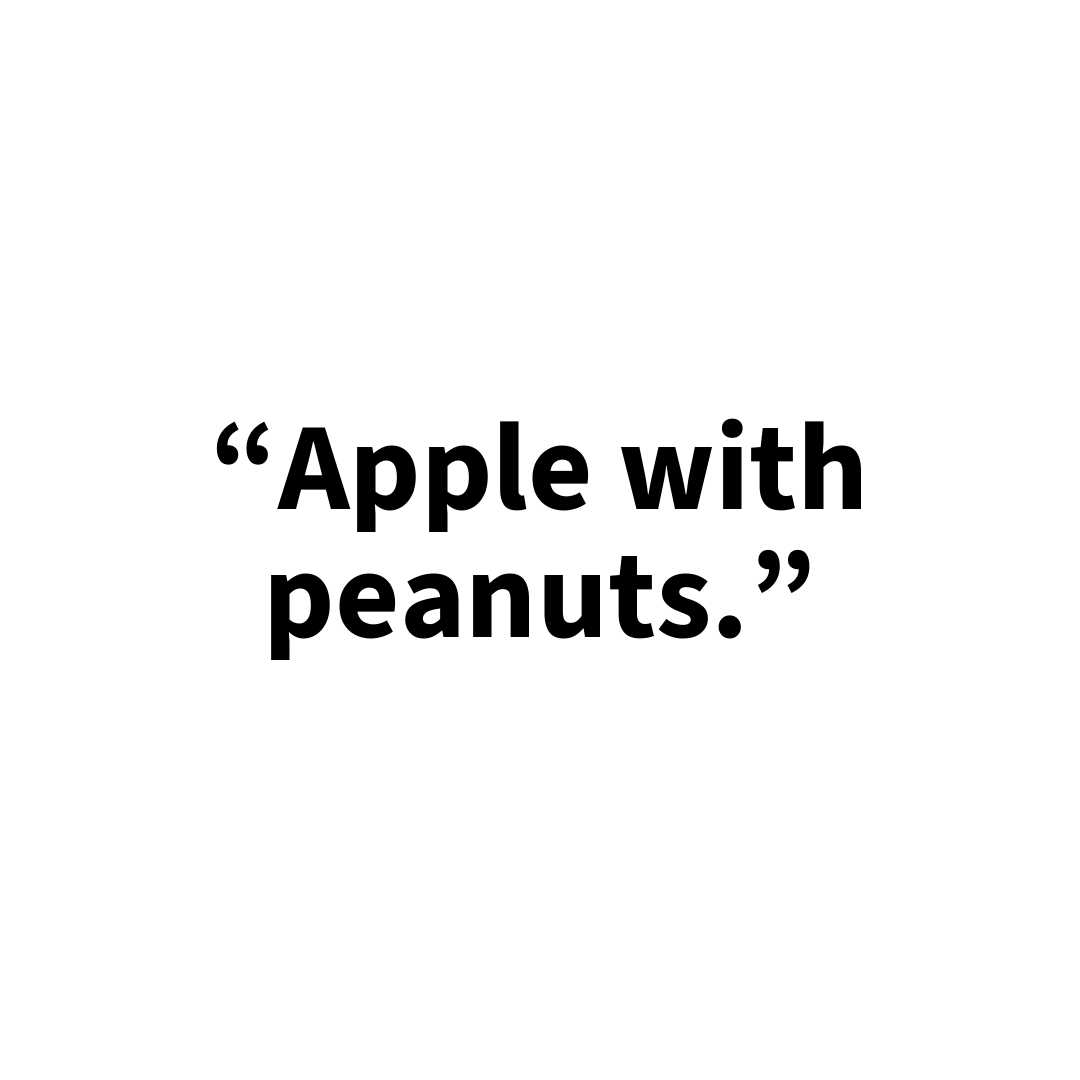} &
\includegraphics[width=0.18\textwidth]{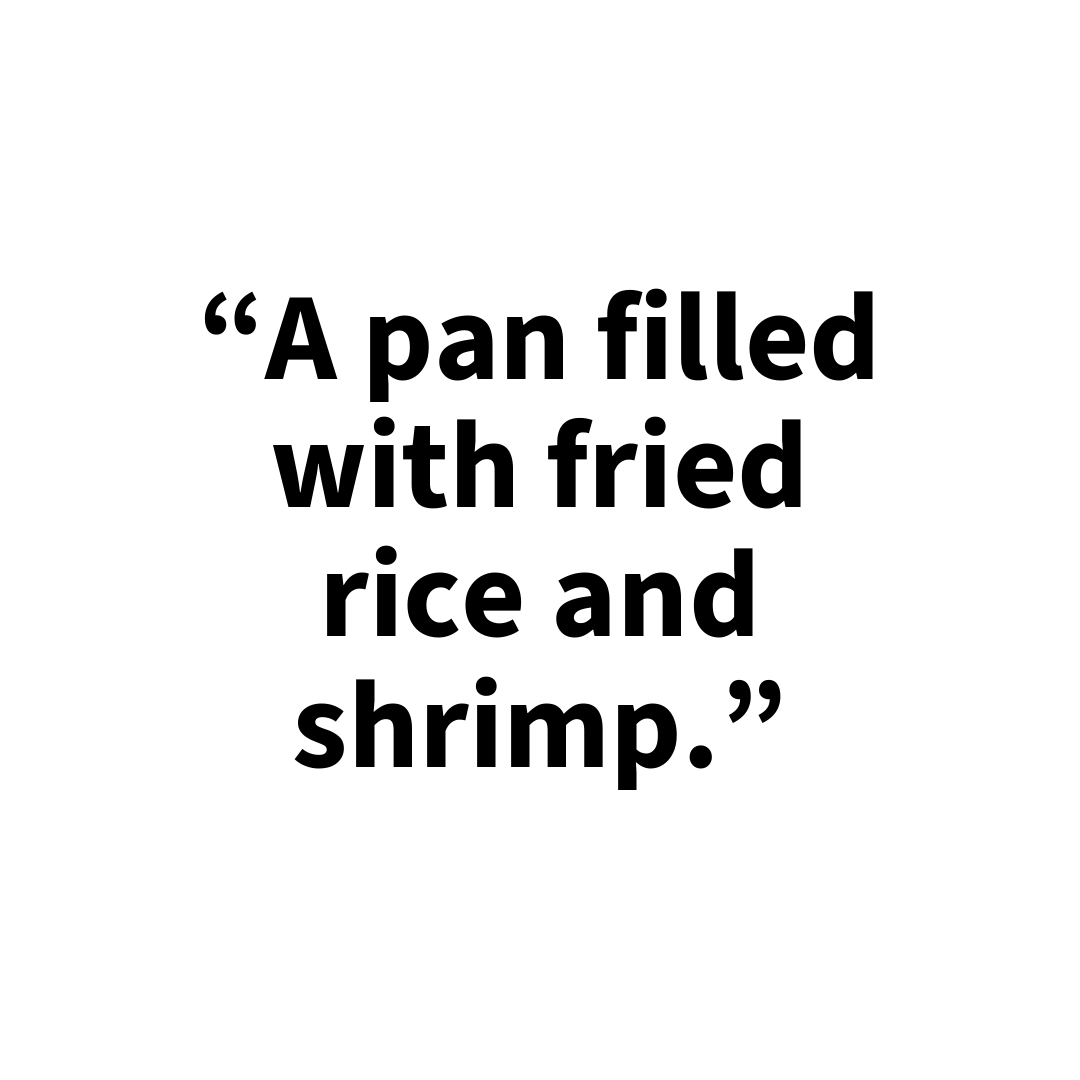} &
\includegraphics[width=0.18\textwidth]{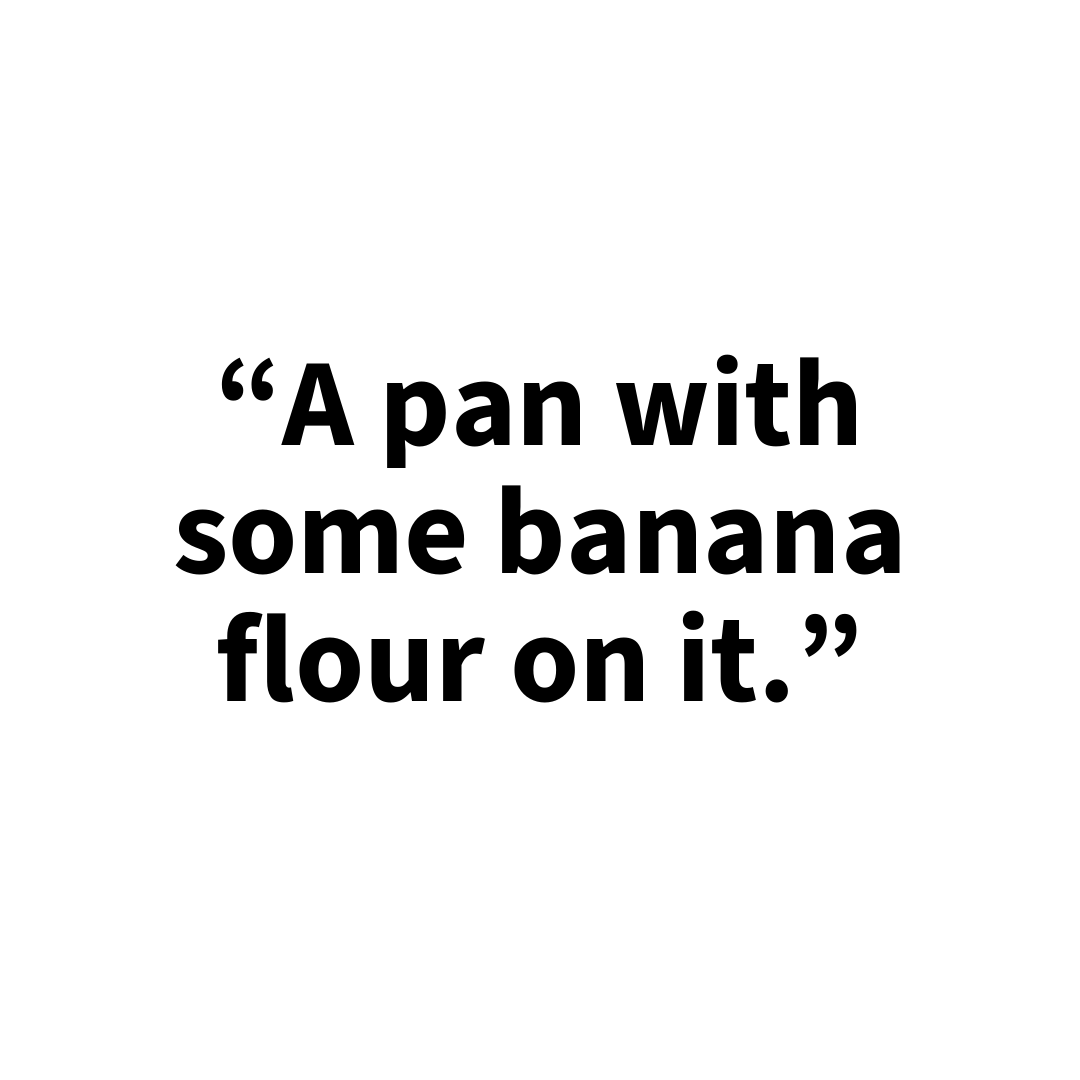} &
\includegraphics[width=0.18\textwidth]{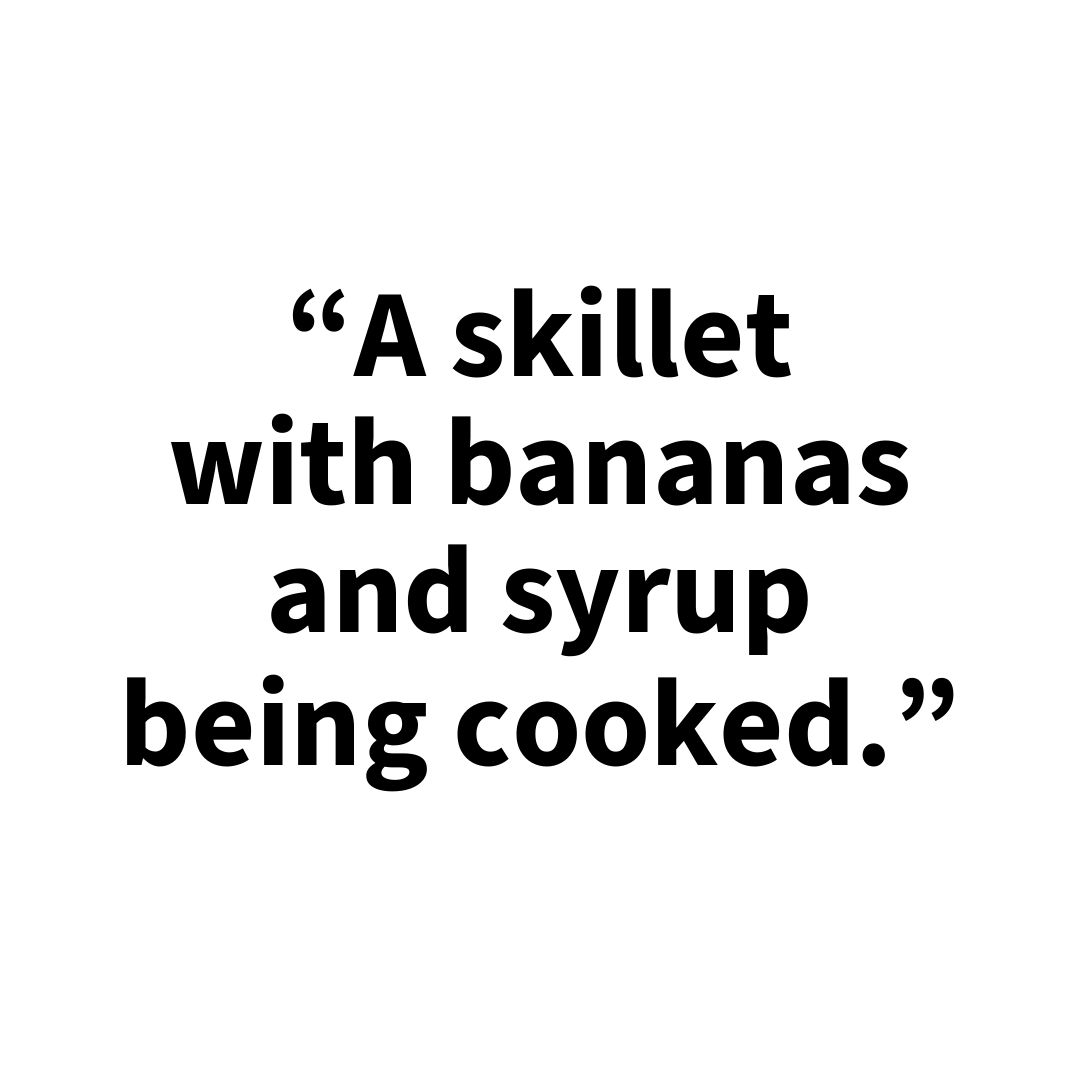} \\

\includegraphics[width=0.18\textwidth]{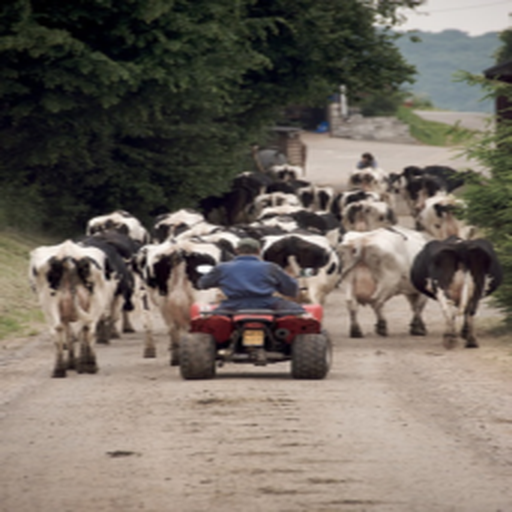} &
\includegraphics[width=0.18\textwidth]{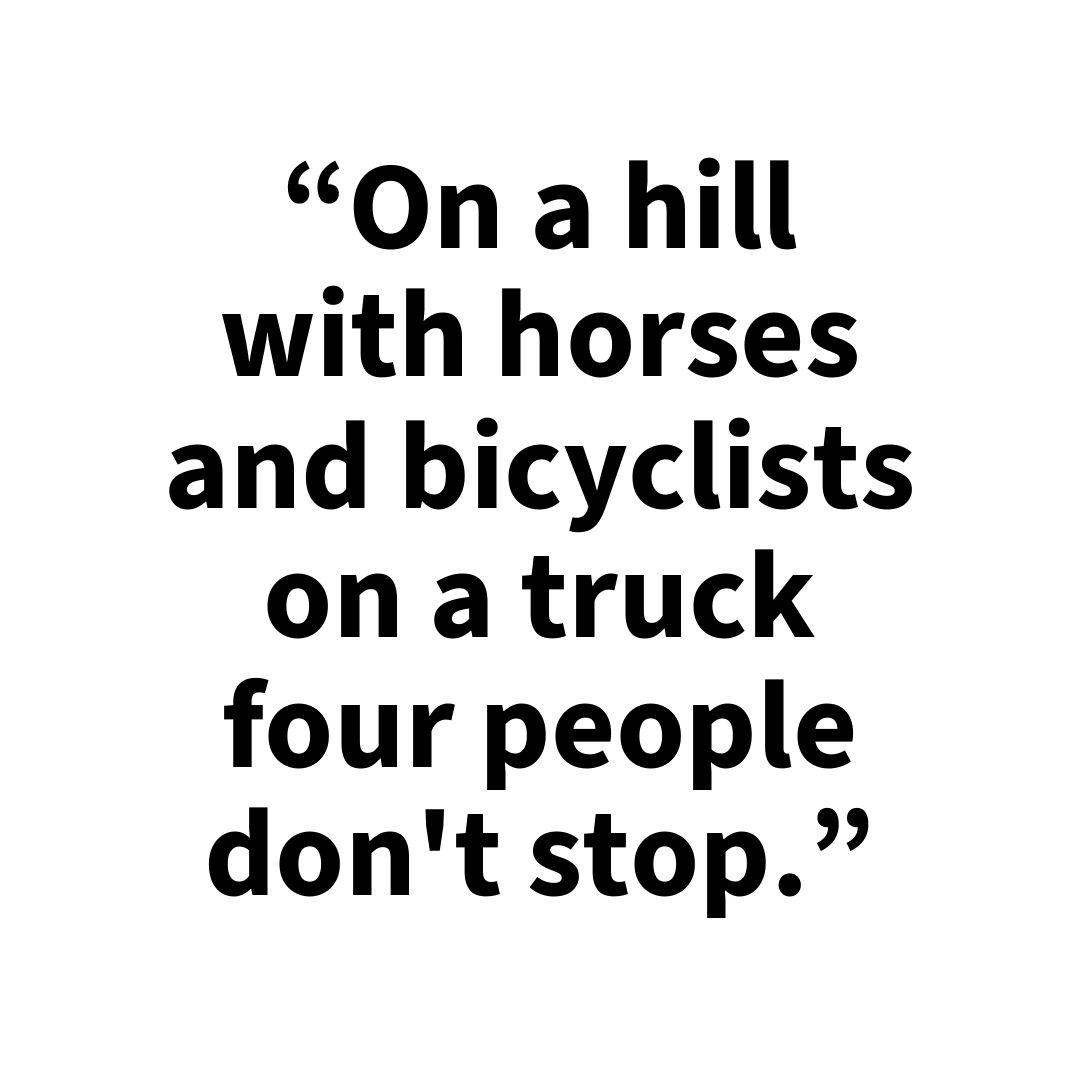} &
\includegraphics[width=0.18\textwidth]{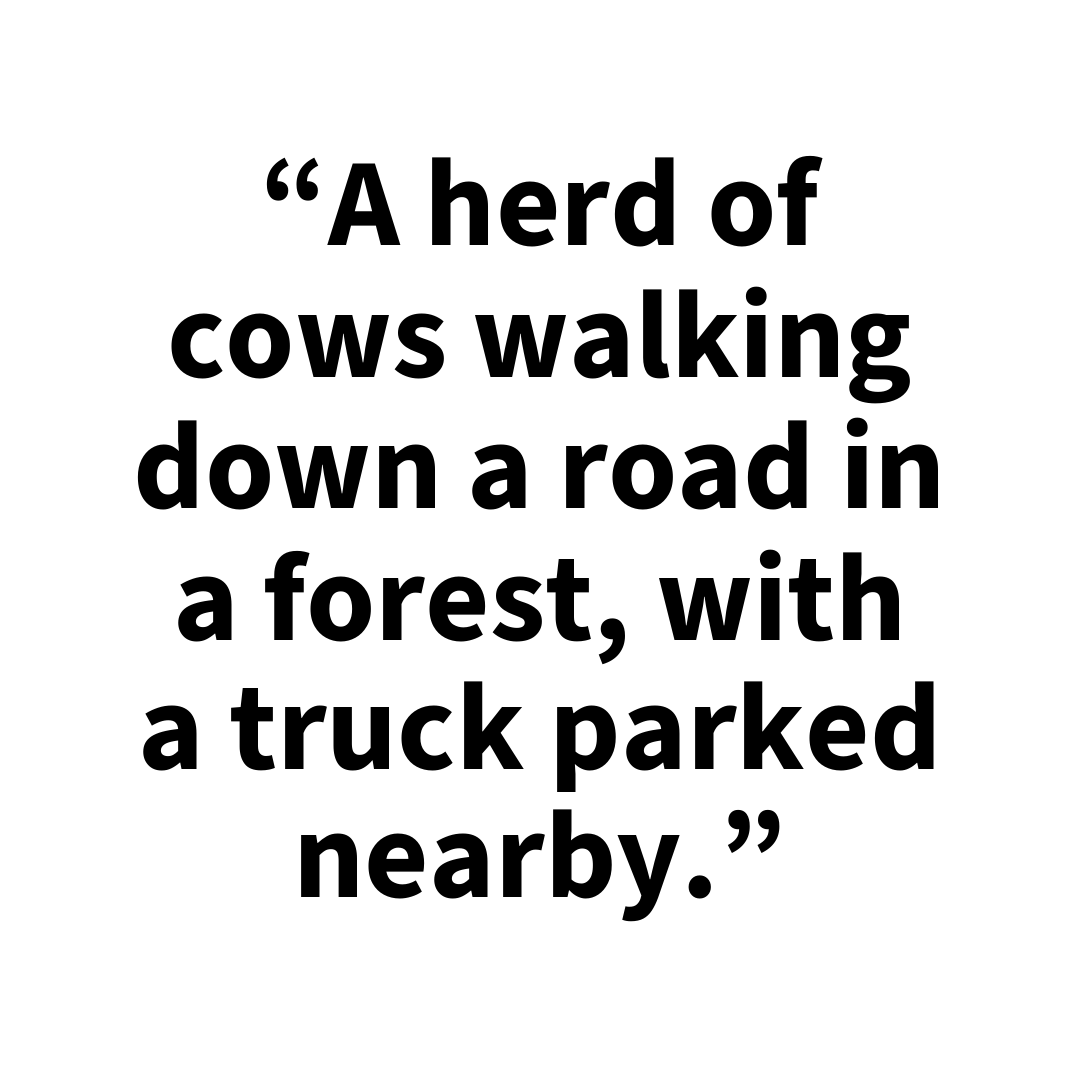} &
\includegraphics[width=0.18\textwidth]{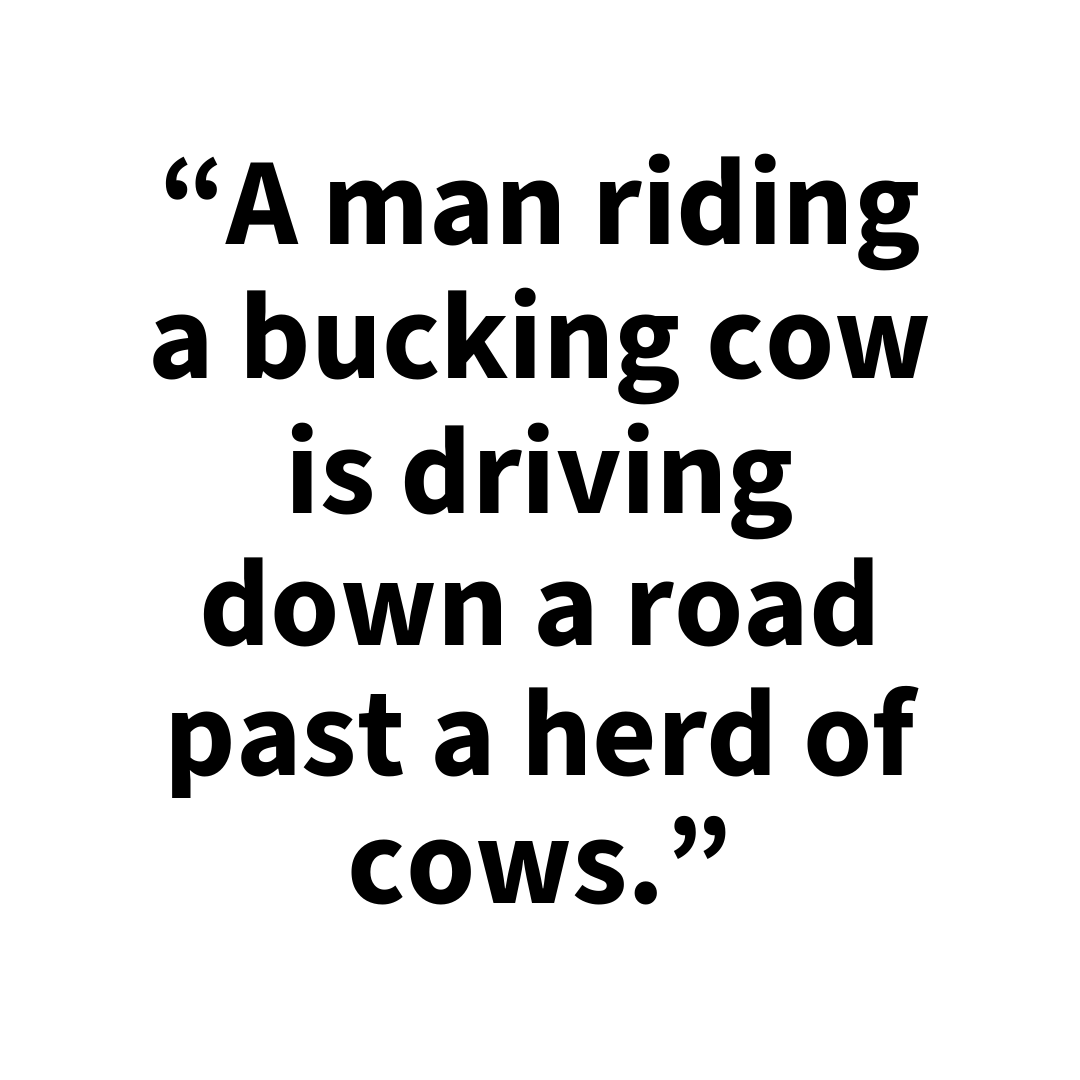} &
\includegraphics[width=0.18\textwidth]{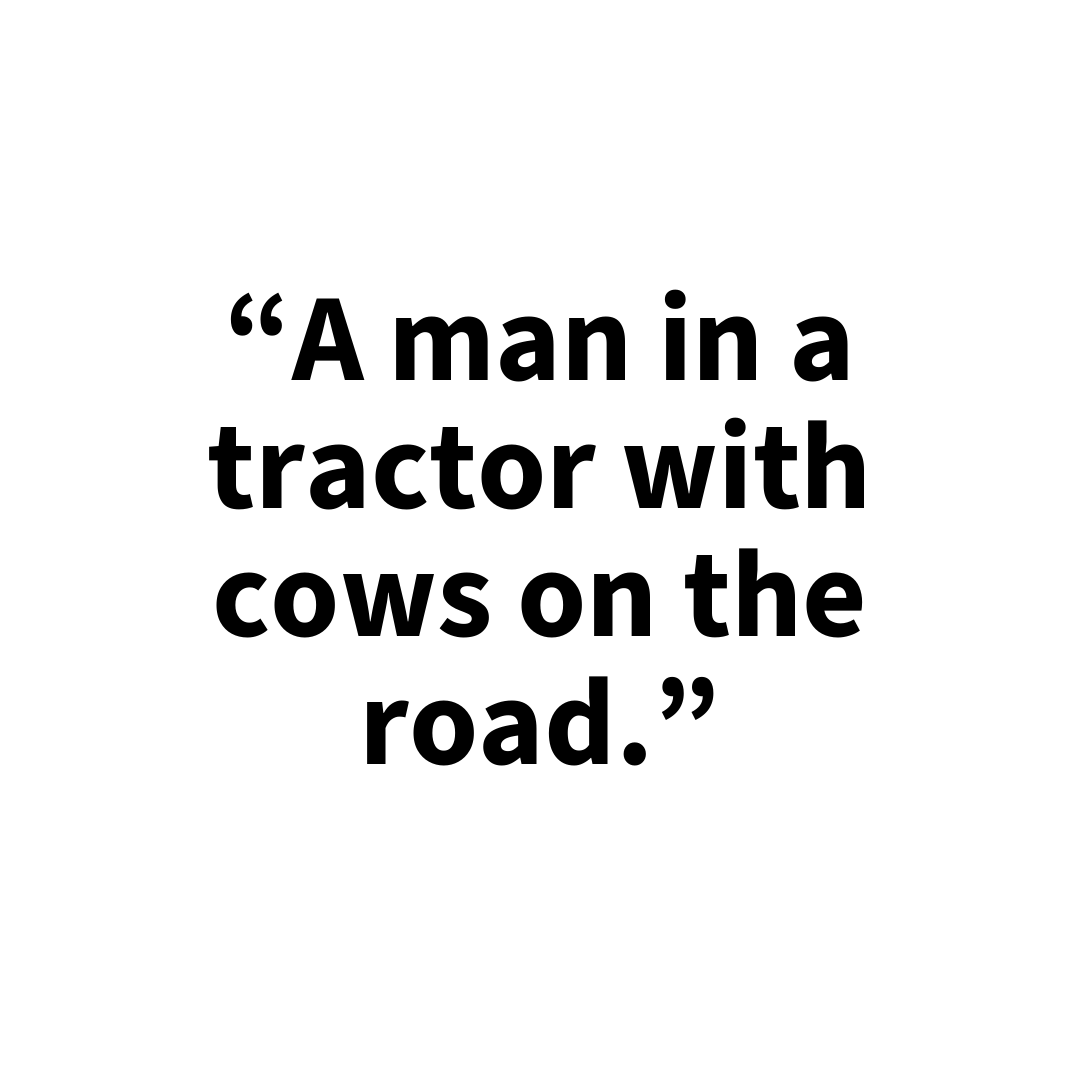} \\

\includegraphics[width=0.18\textwidth]{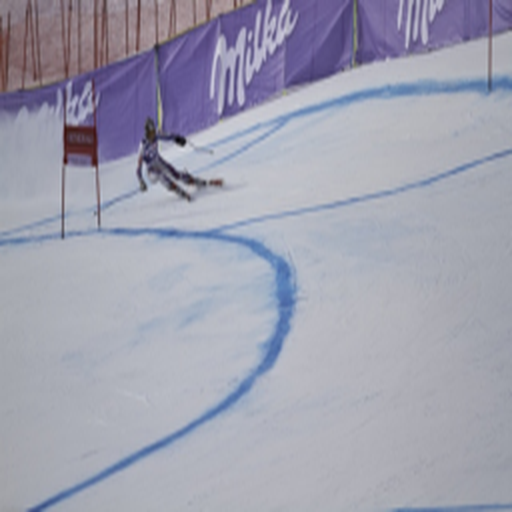} &
\includegraphics[width=0.18\textwidth]{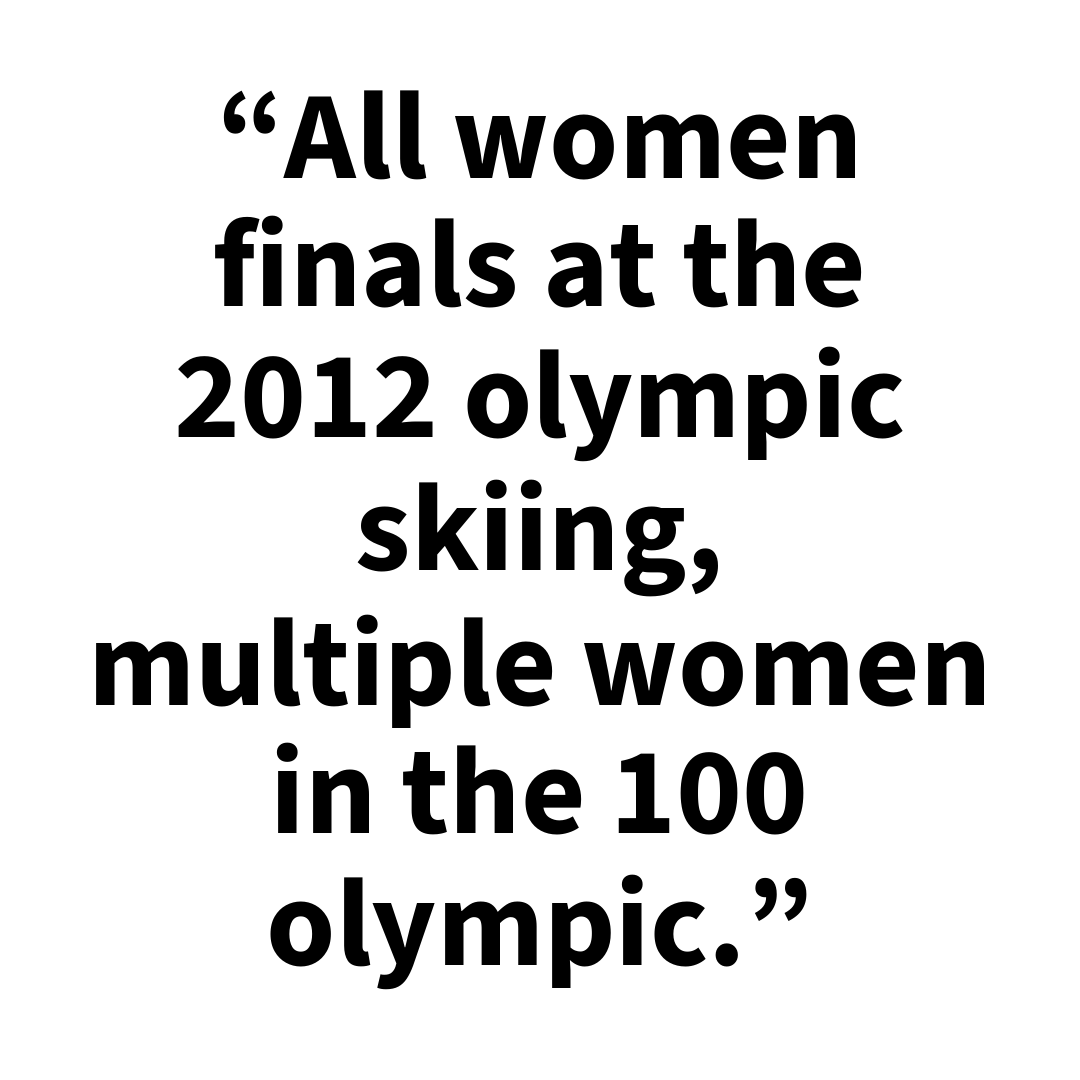} &
\includegraphics[width=0.18\textwidth]{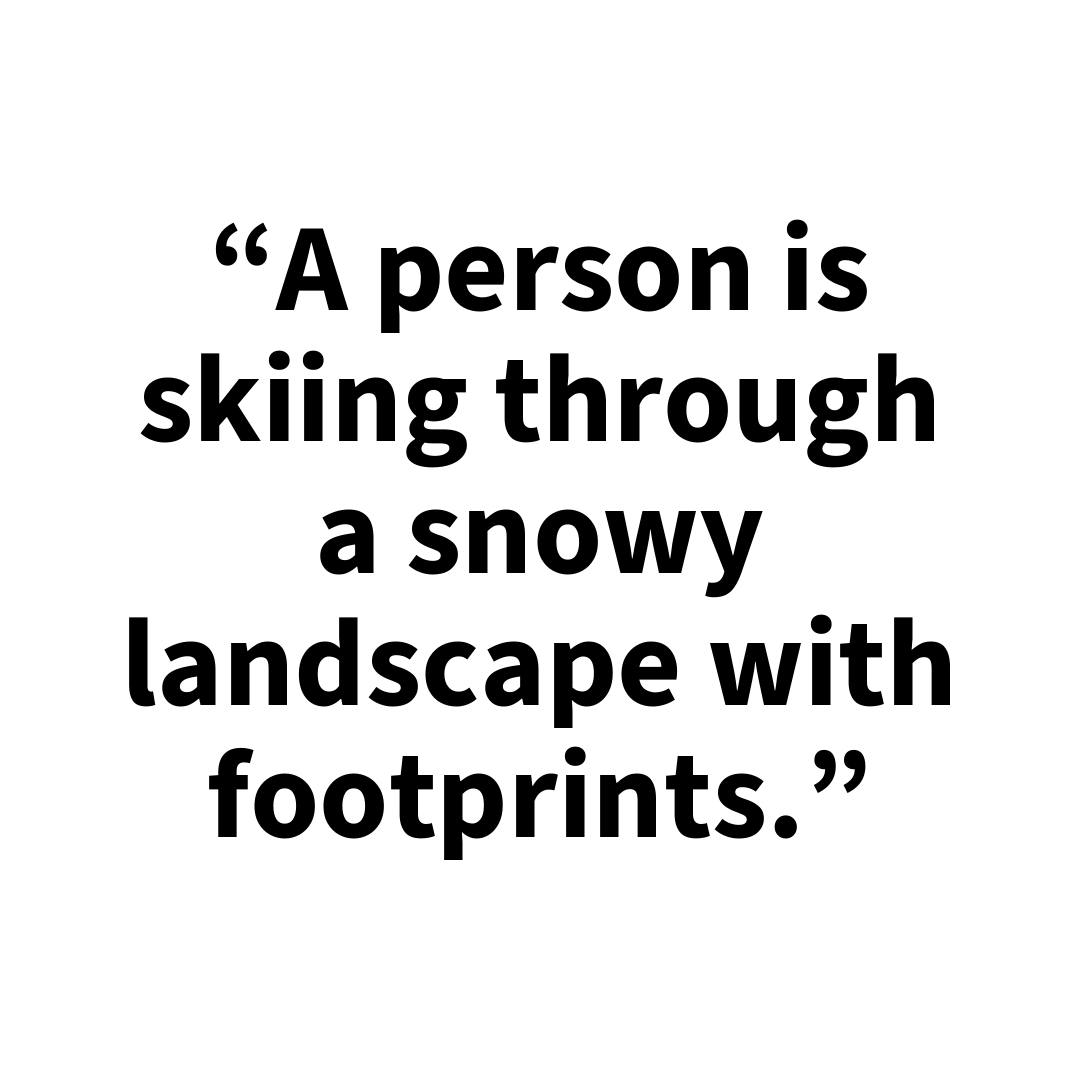} &
\includegraphics[width=0.18\textwidth]{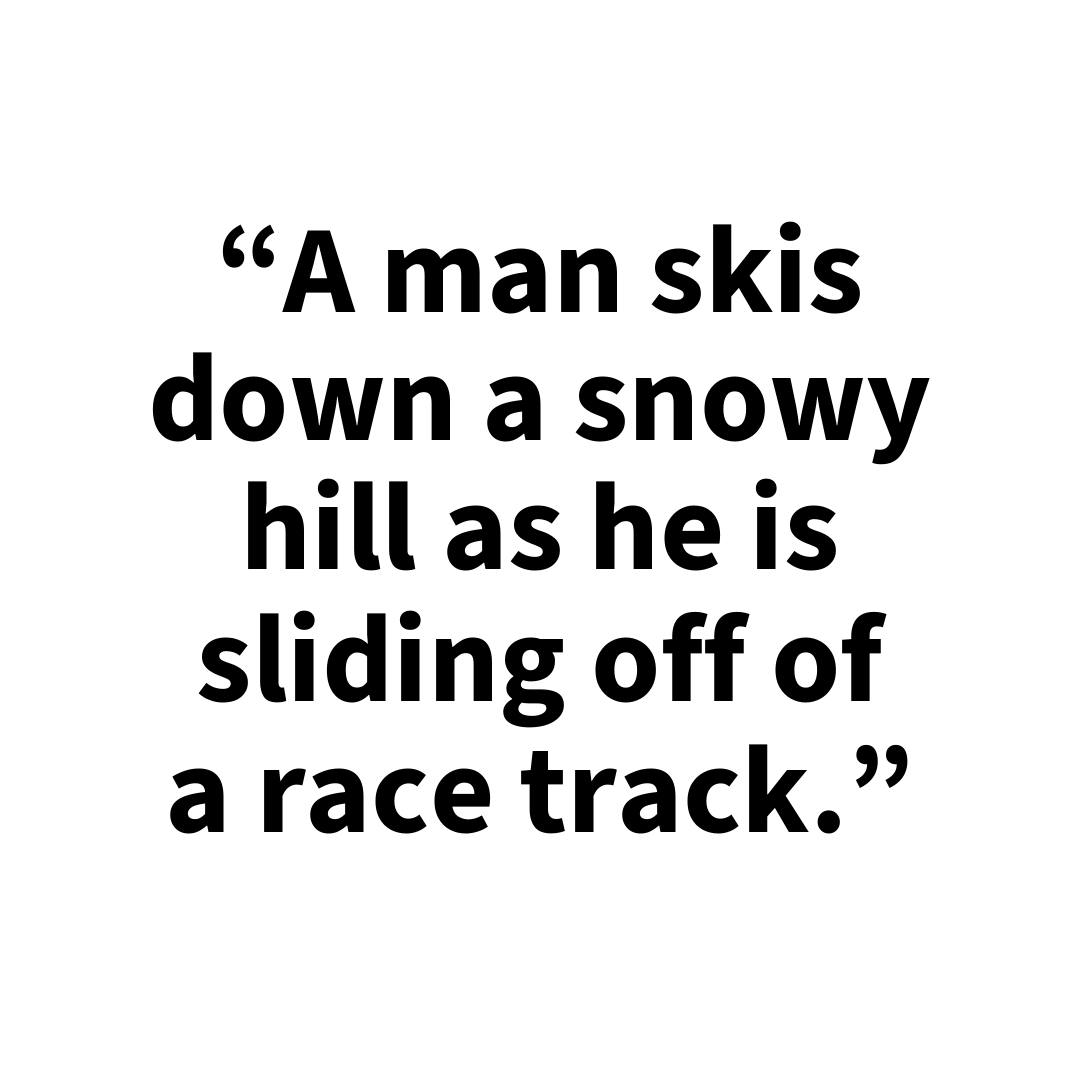} &
\includegraphics[width=0.18\textwidth]{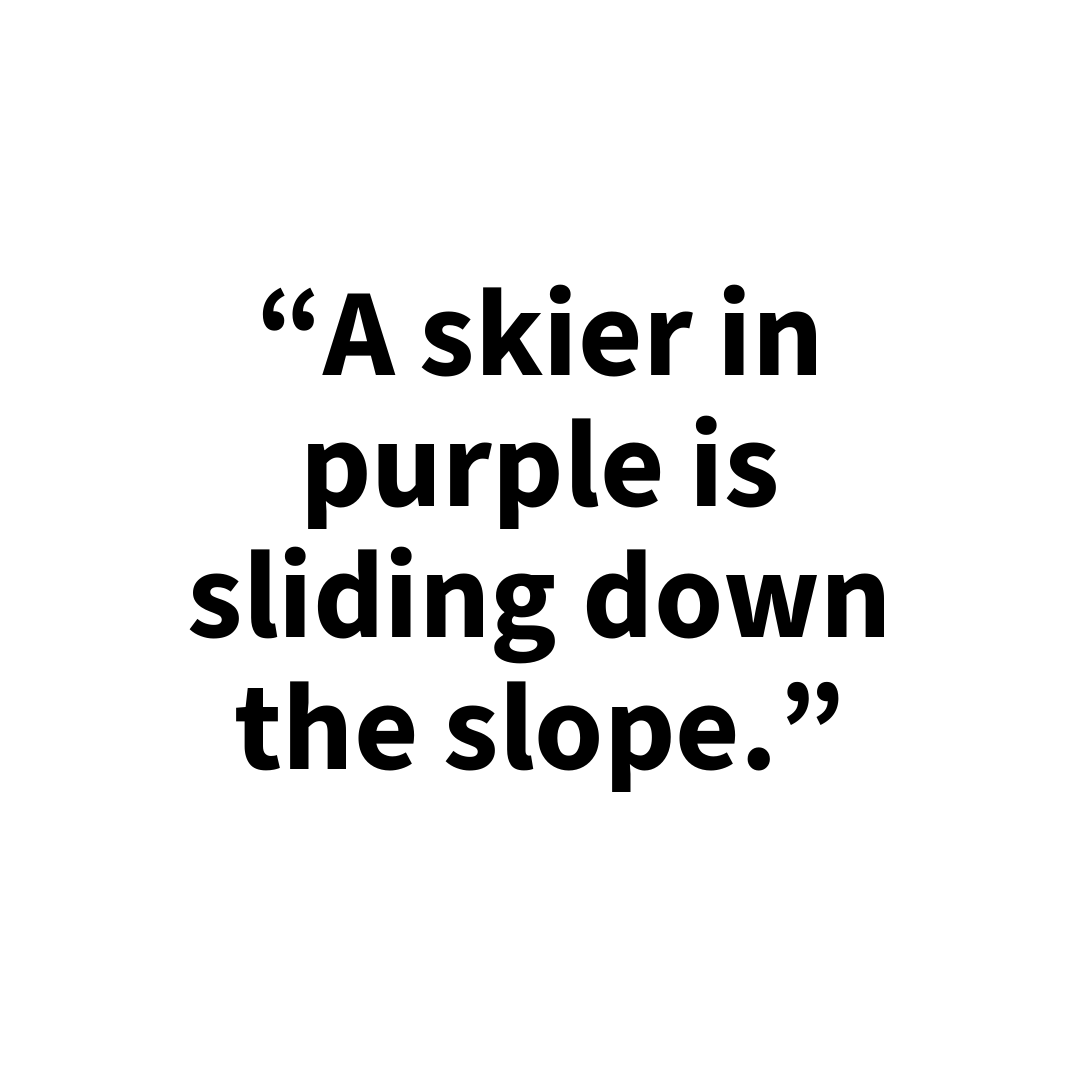} \\

\includegraphics[width=0.18\textwidth]{figures/qualitative_ita/I_to_T/080_src.png} &
\includegraphics[width=0.18\textwidth]{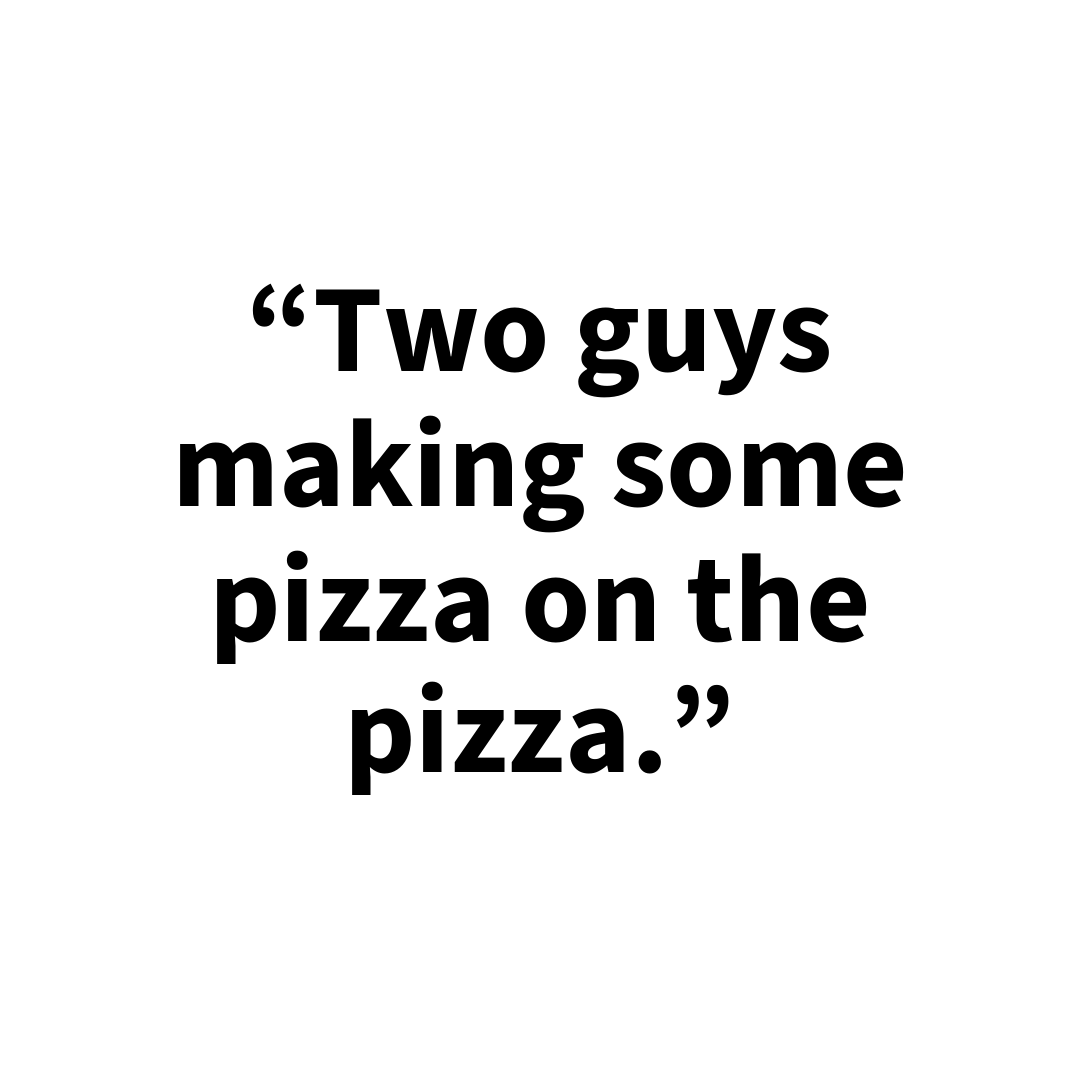} &
\includegraphics[width=0.18\textwidth]{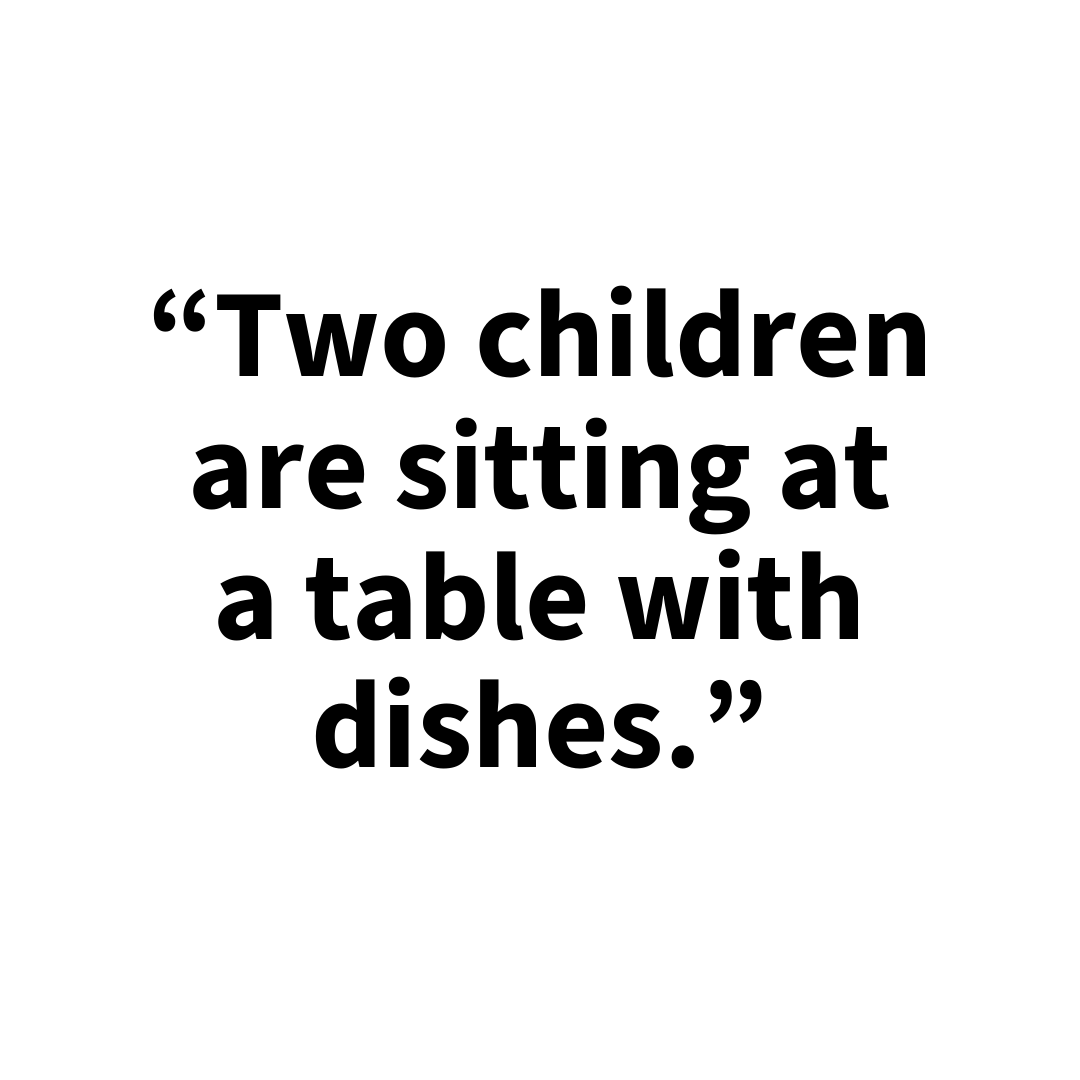} &
\includegraphics[width=0.18\textwidth]{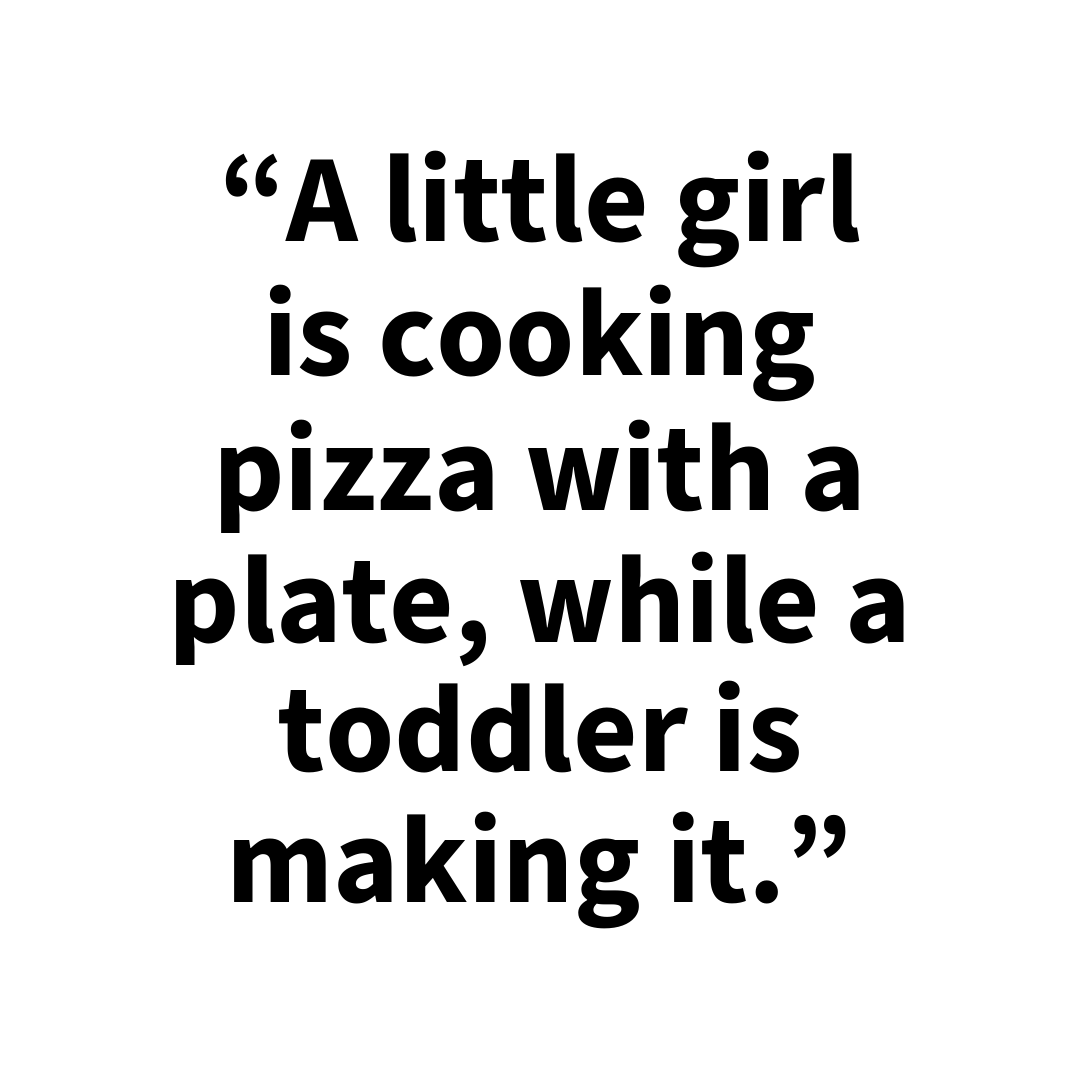} &
\includegraphics[width=0.18\textwidth]{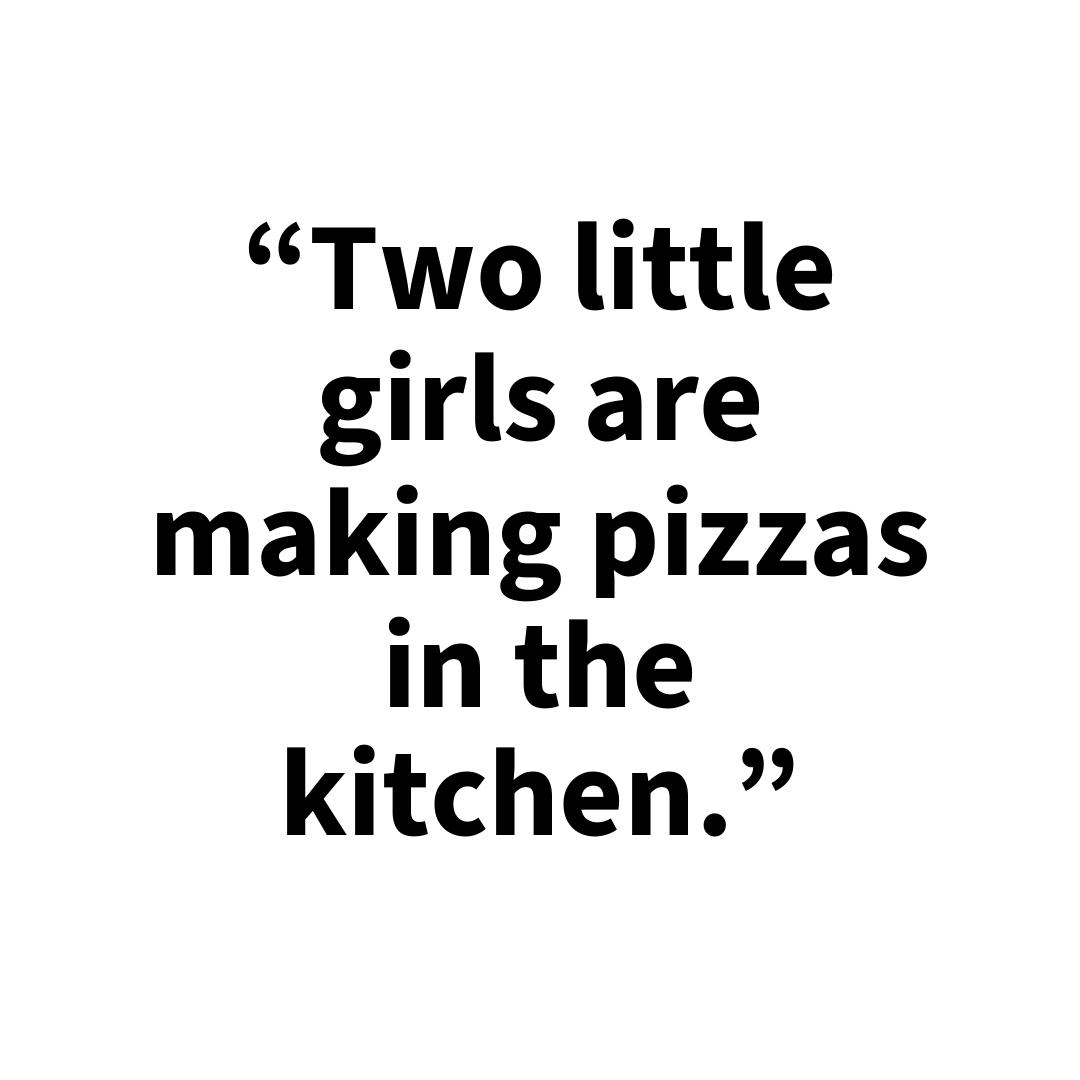} \\
\end{tabular}
\caption{\textbf{Image-to-text generation.} Given a source image (left), each method generates a caption. \Ours{}$^\dagger$ produces a caption that accurately describes the source.}
\label{fig:ita-qual-i2t}
\end{figure}

\begin{figure}[t]
\centering
\setlength{\tabcolsep}{2pt}
\renewcommand{\arraystretch}{1.0}
\begin{tabular}{c|cccc}
\footnotesize Source (A) &
\footnotesize CoDi~\cite{tang2023:codi} &
\footnotesize OmniFlow~\cite{li2025:omniflow} &
\footnotesize FlowBind~\cite{cha2026:flowbind} &
\footnotesize \textbf{\Ours{}$^\dagger$ (Ours)} \\
\midrule
\includegraphics[width=0.18\textwidth]{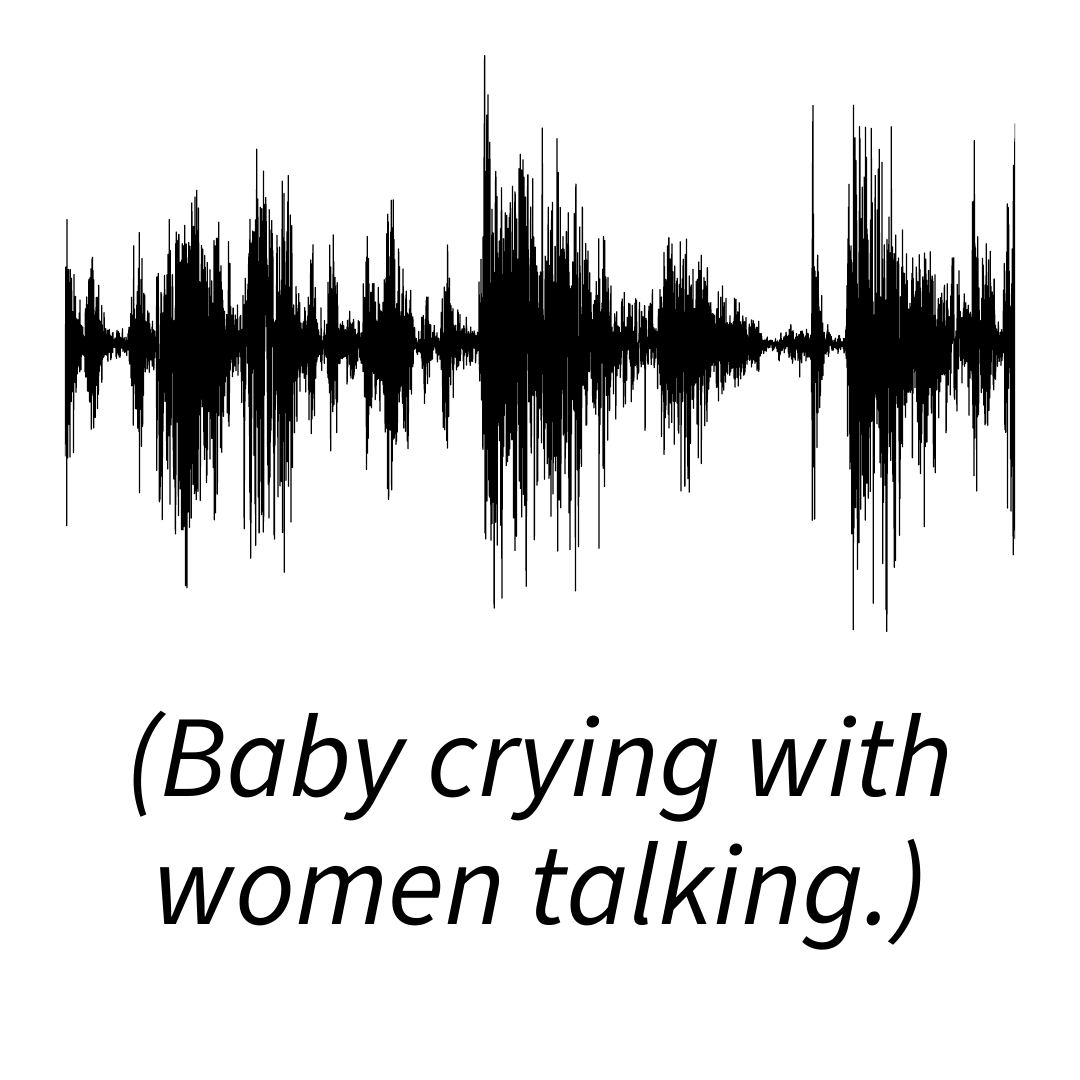} &
\includegraphics[width=0.18\textwidth]{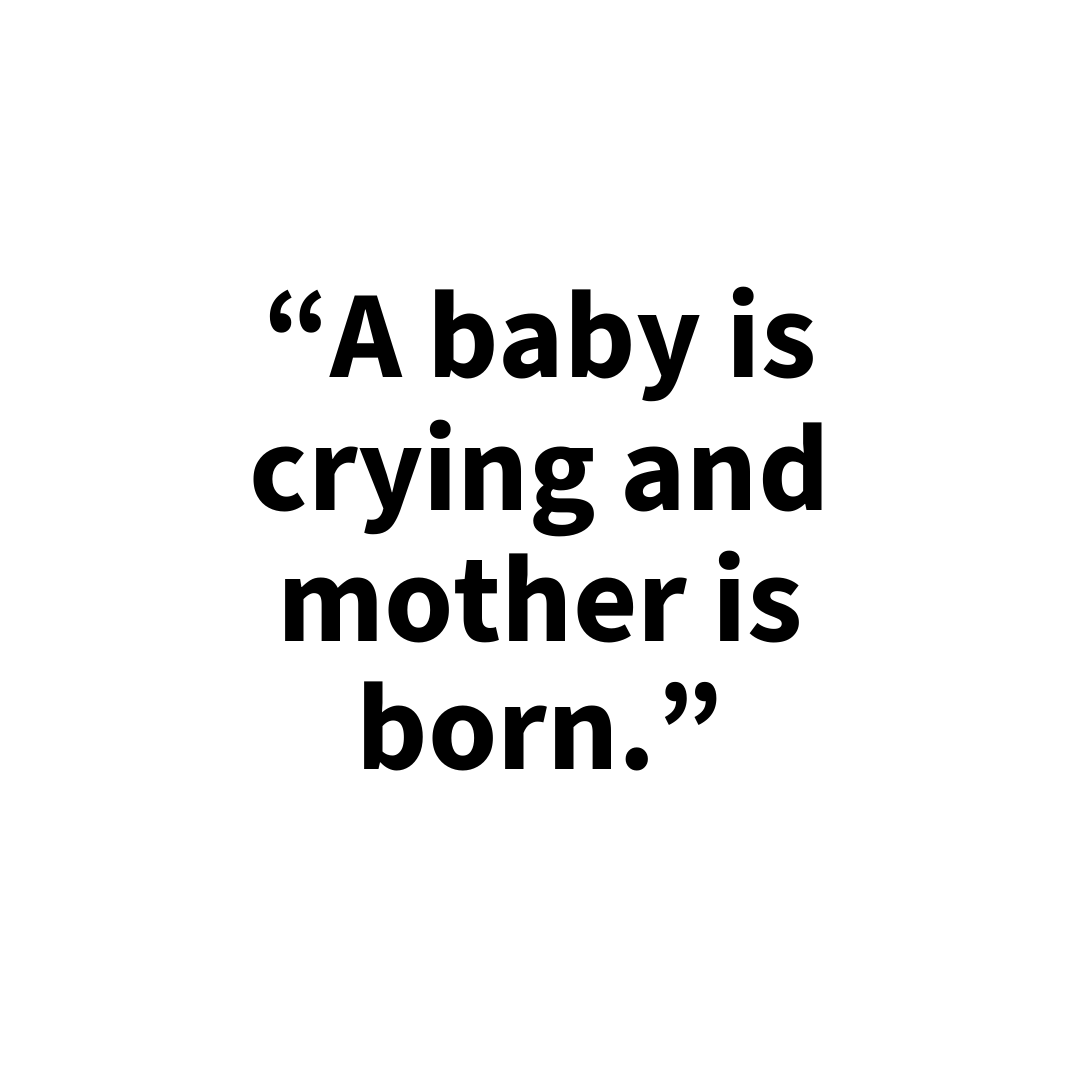} &
\includegraphics[width=0.18\textwidth]{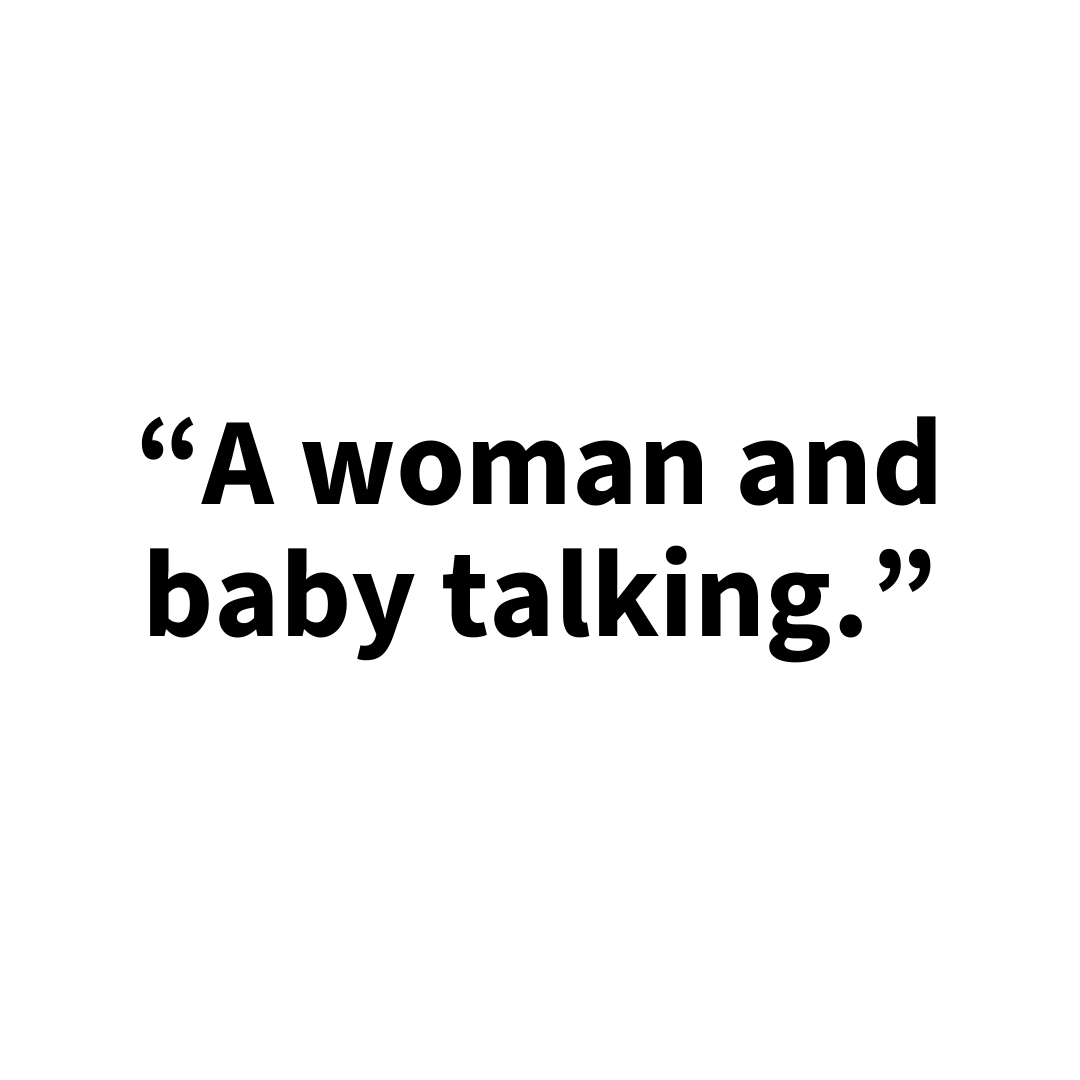} &
\includegraphics[width=0.18\textwidth]{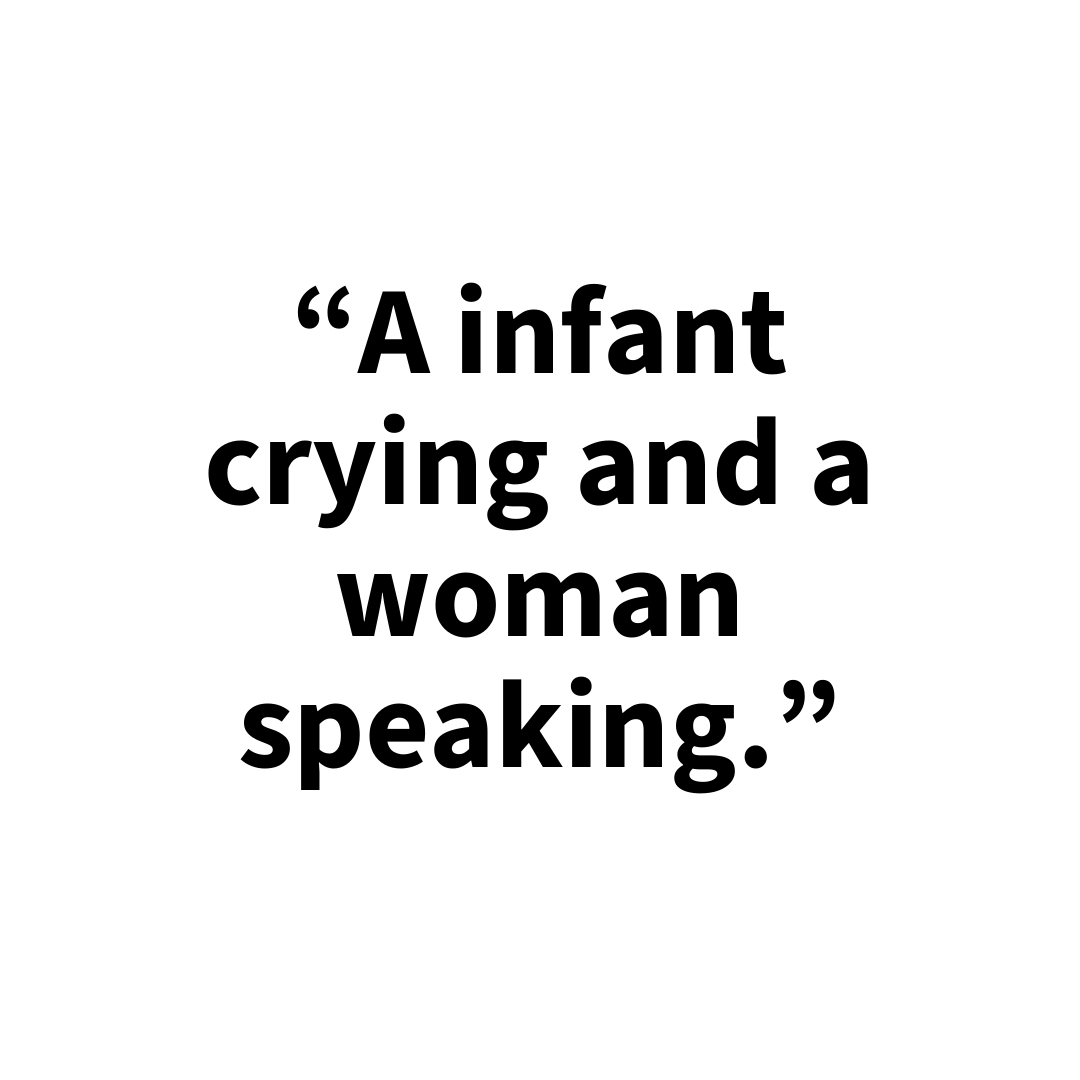} &
\includegraphics[width=0.18\textwidth]{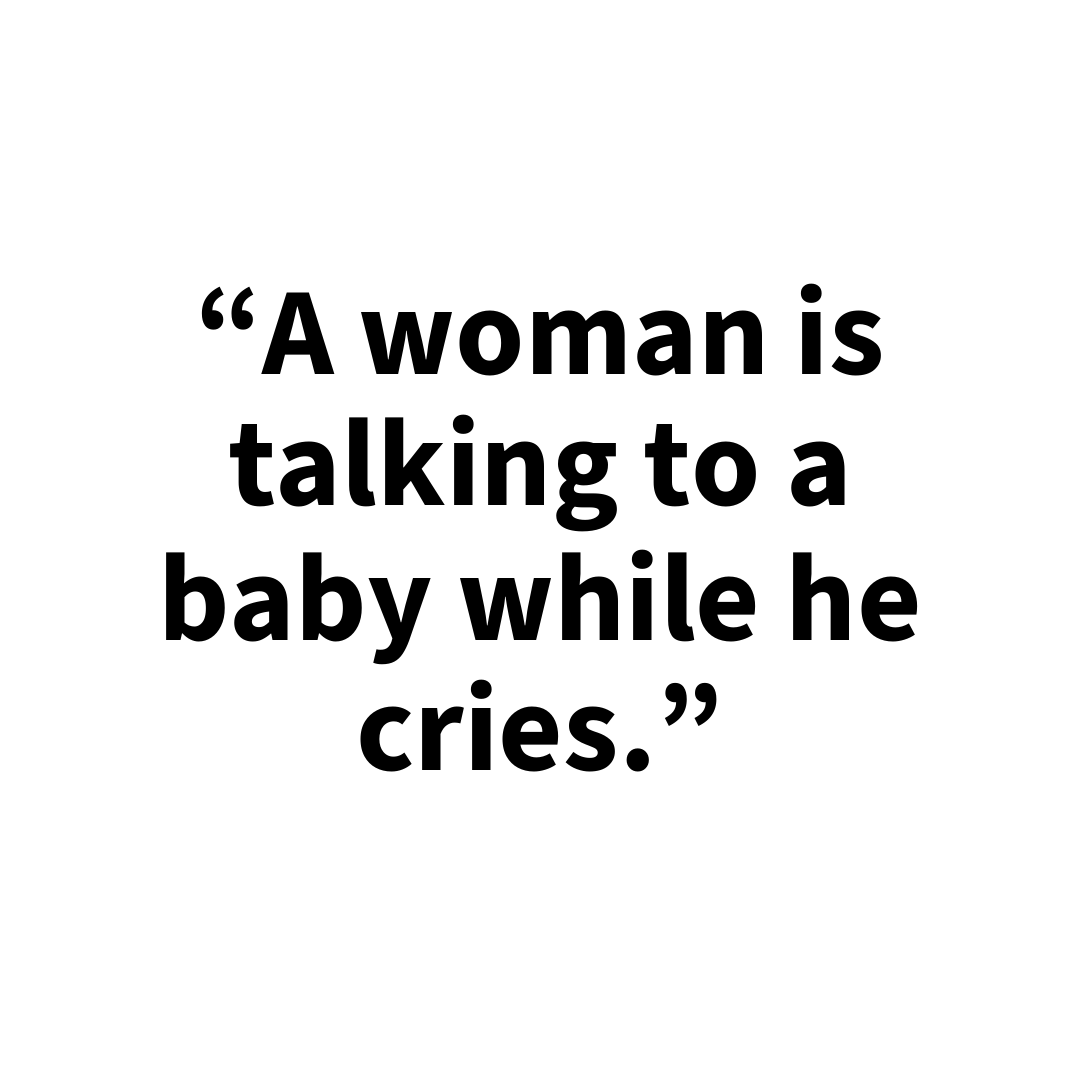} \\

\includegraphics[width=0.18\textwidth]{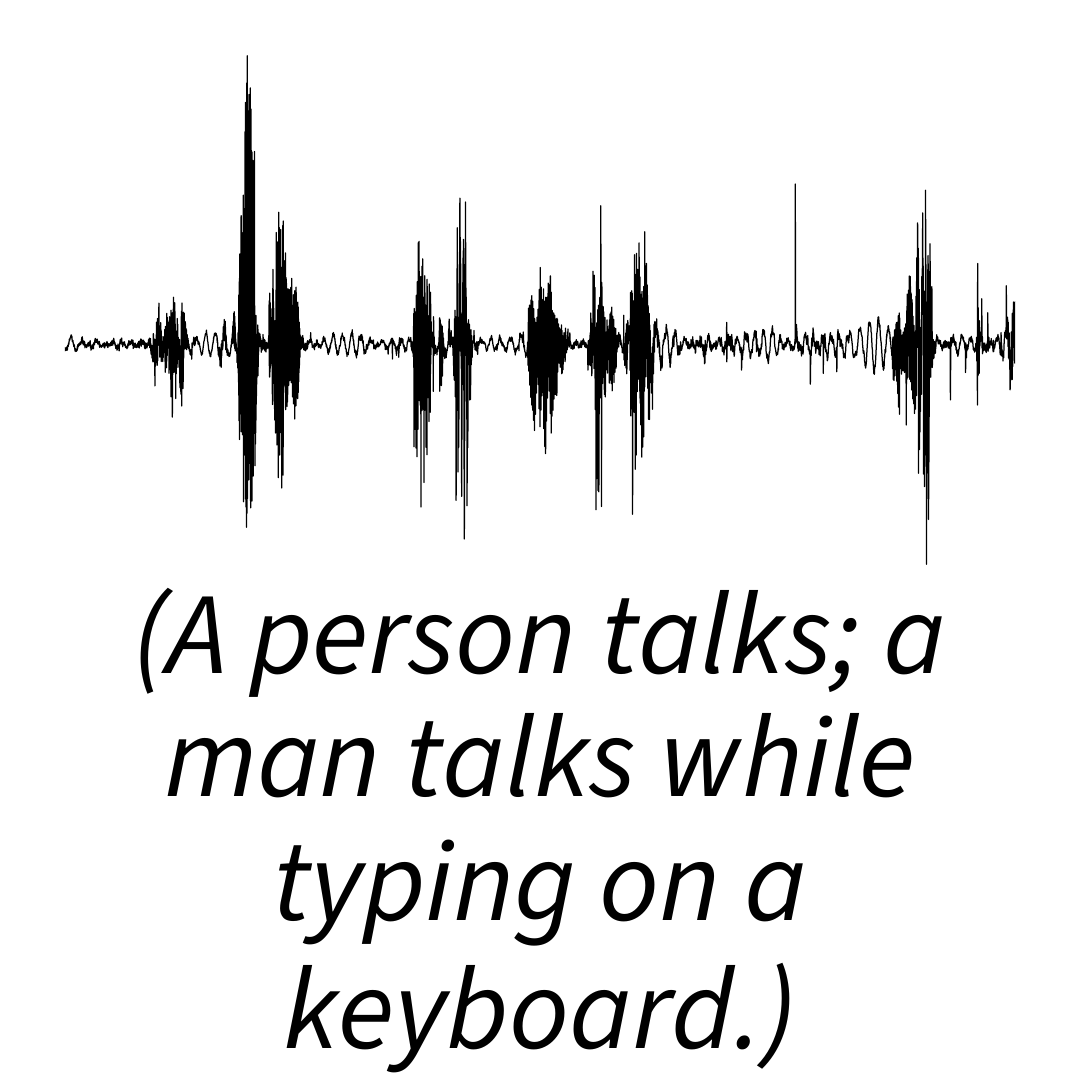} &
\includegraphics[width=0.18\textwidth]{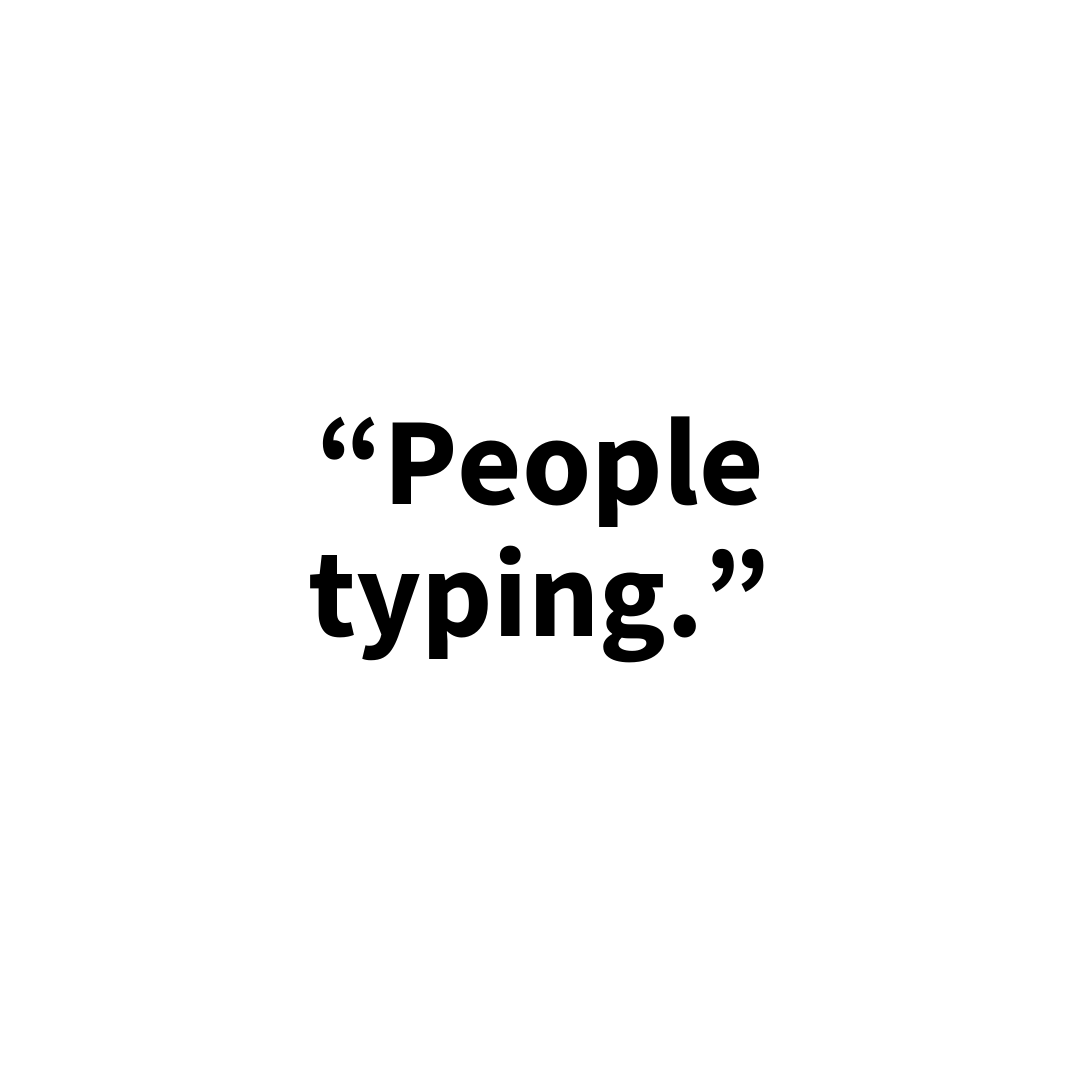} &
\includegraphics[width=0.18\textwidth]{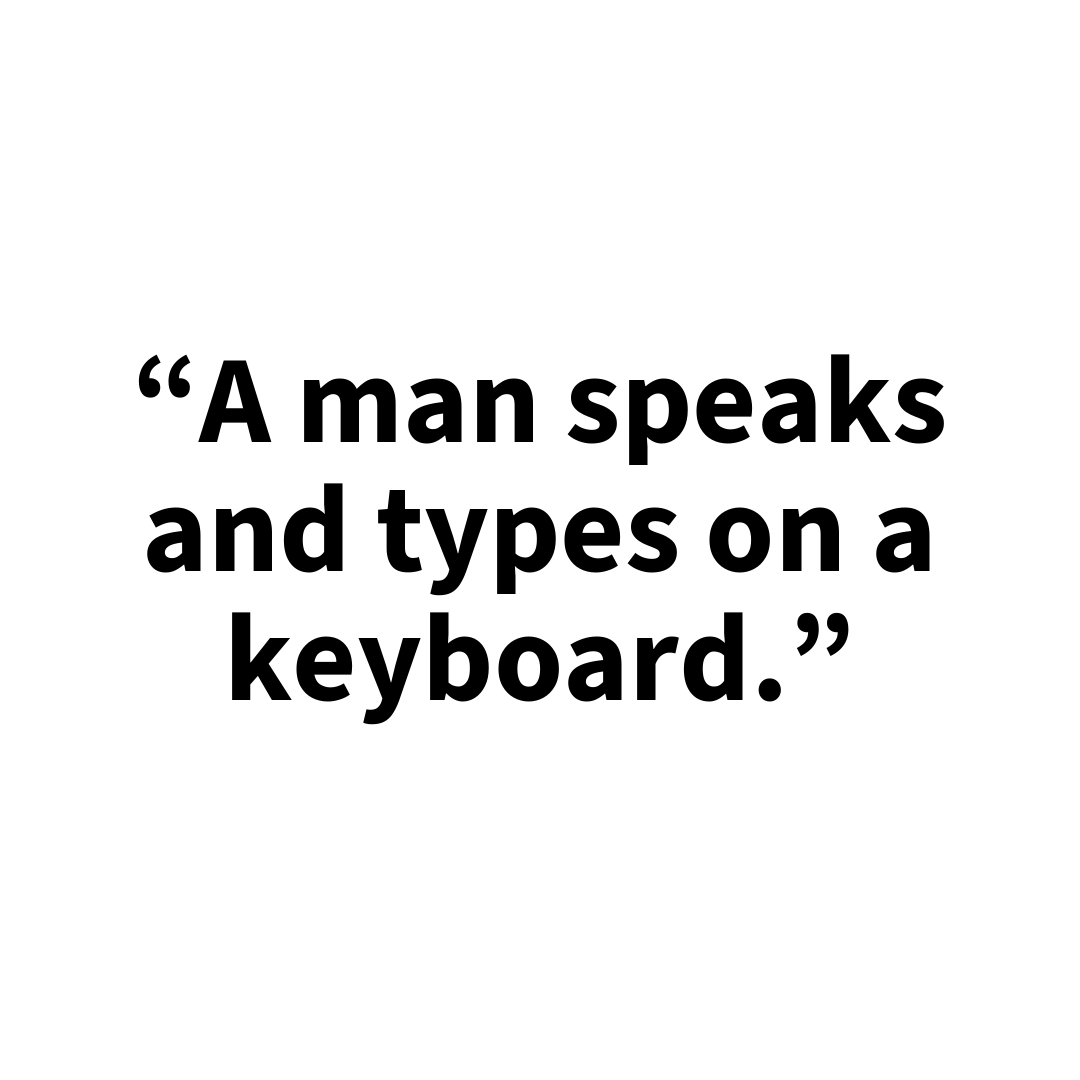} &
\includegraphics[width=0.18\textwidth]{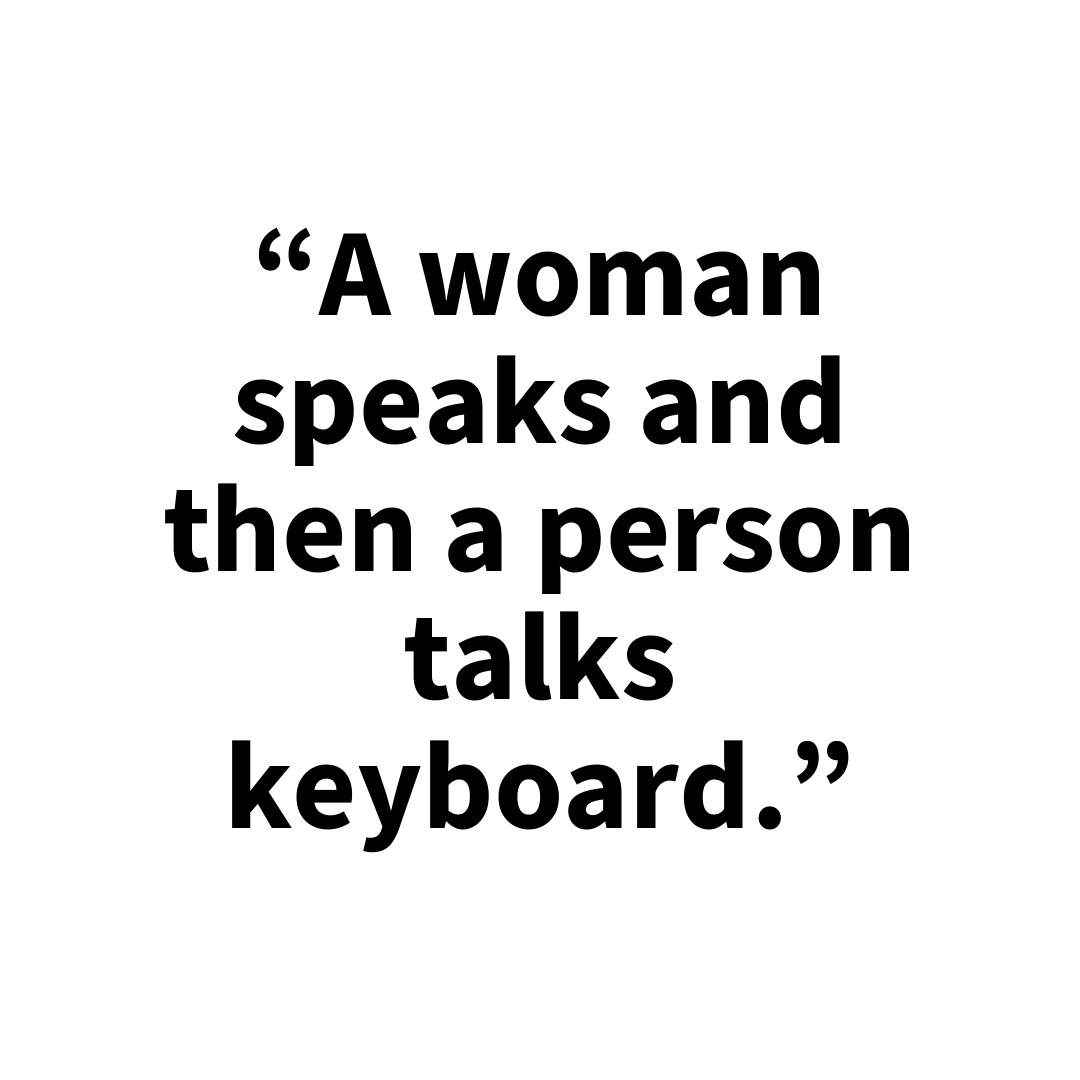} &
\includegraphics[width=0.18\textwidth]{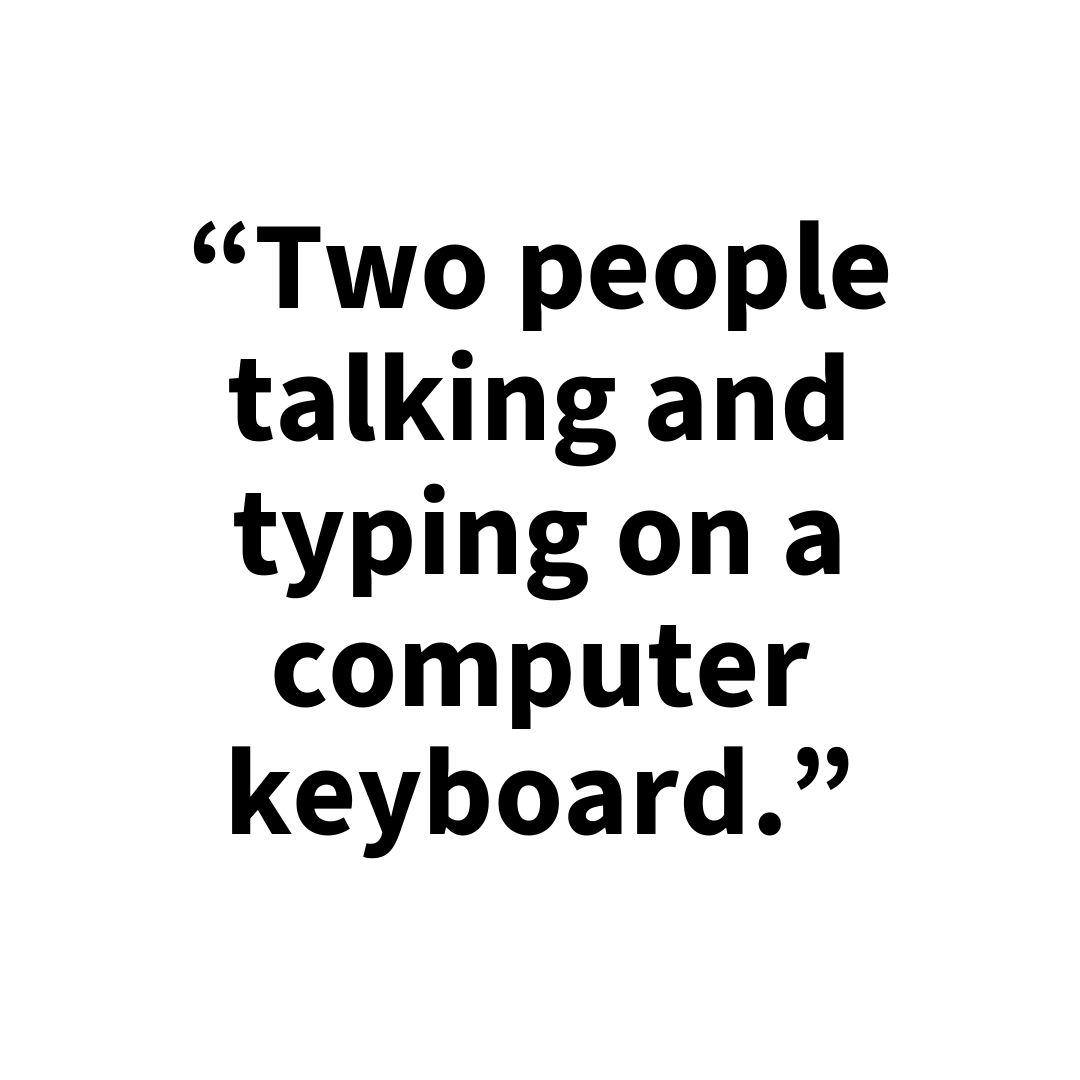} \\

\includegraphics[width=0.18\textwidth]{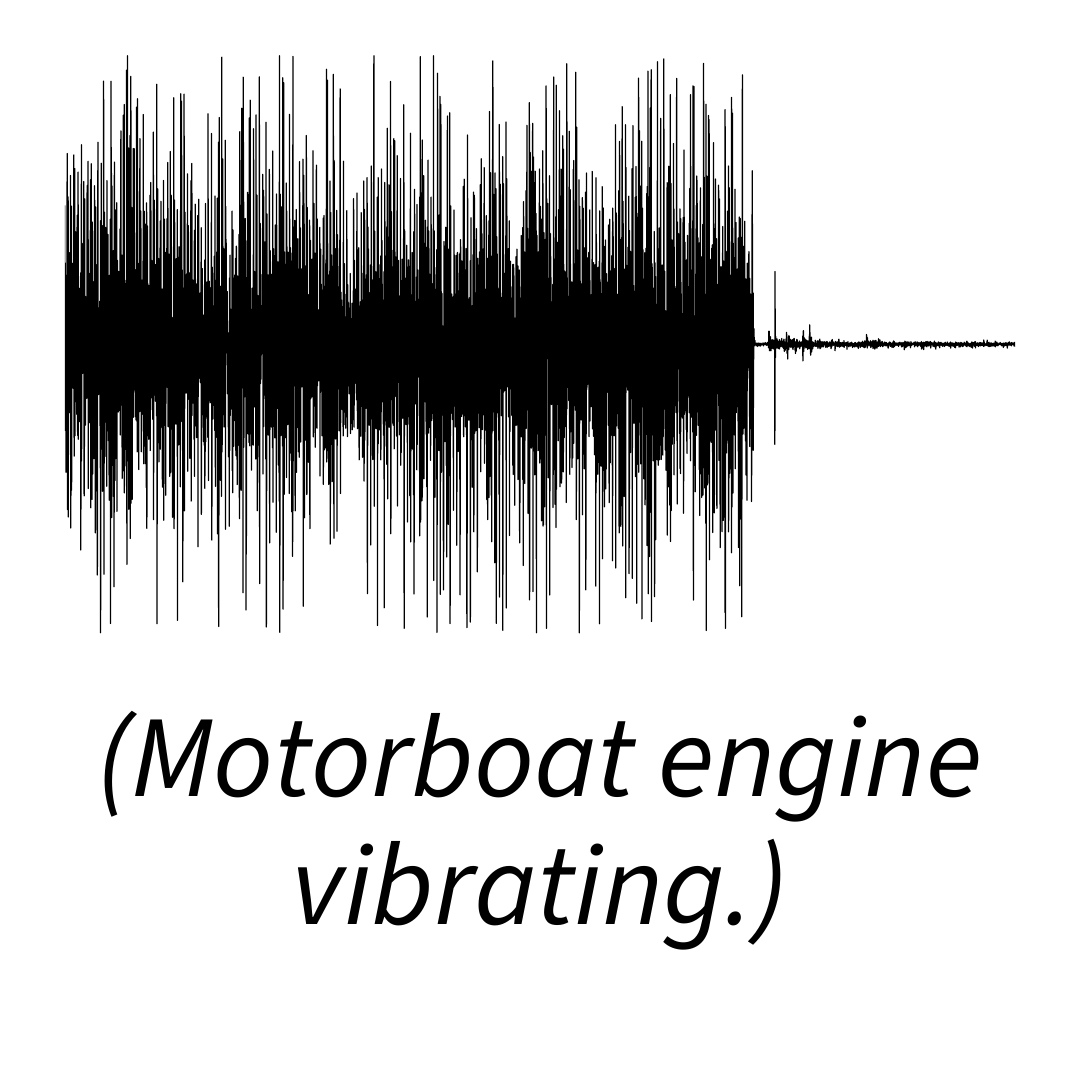} &
\includegraphics[width=0.18\textwidth]{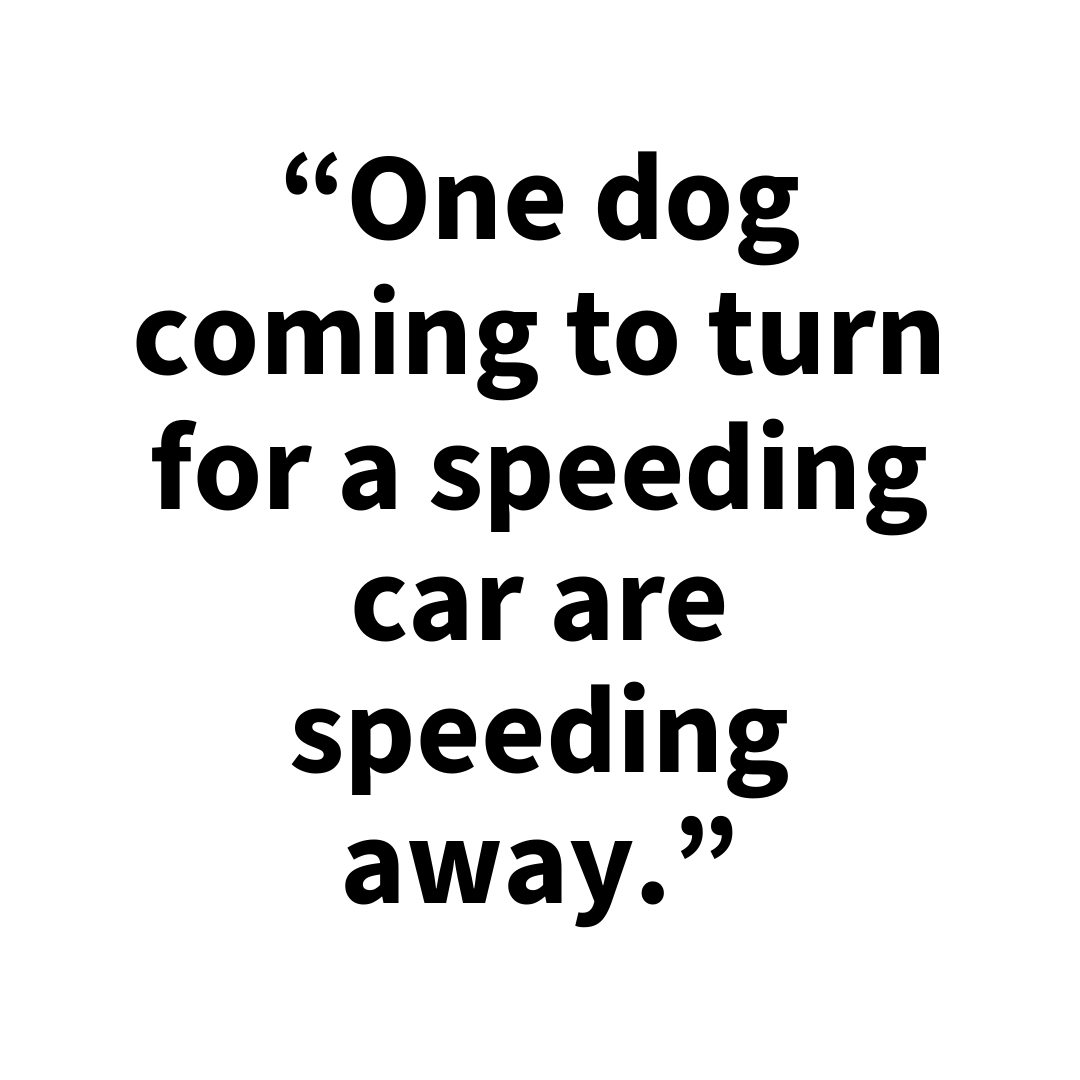} &
\includegraphics[width=0.18\textwidth]{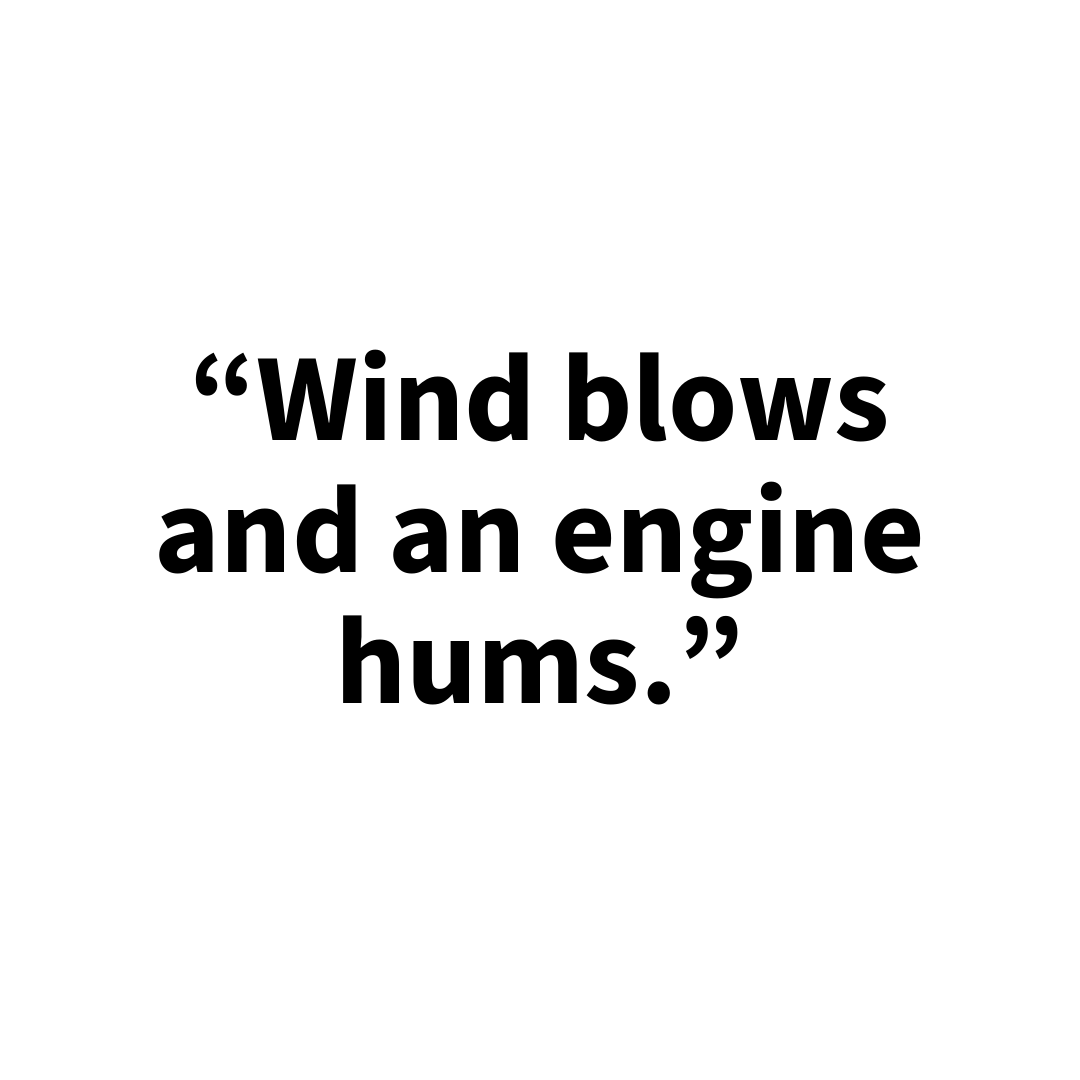} &
\includegraphics[width=0.18\textwidth]{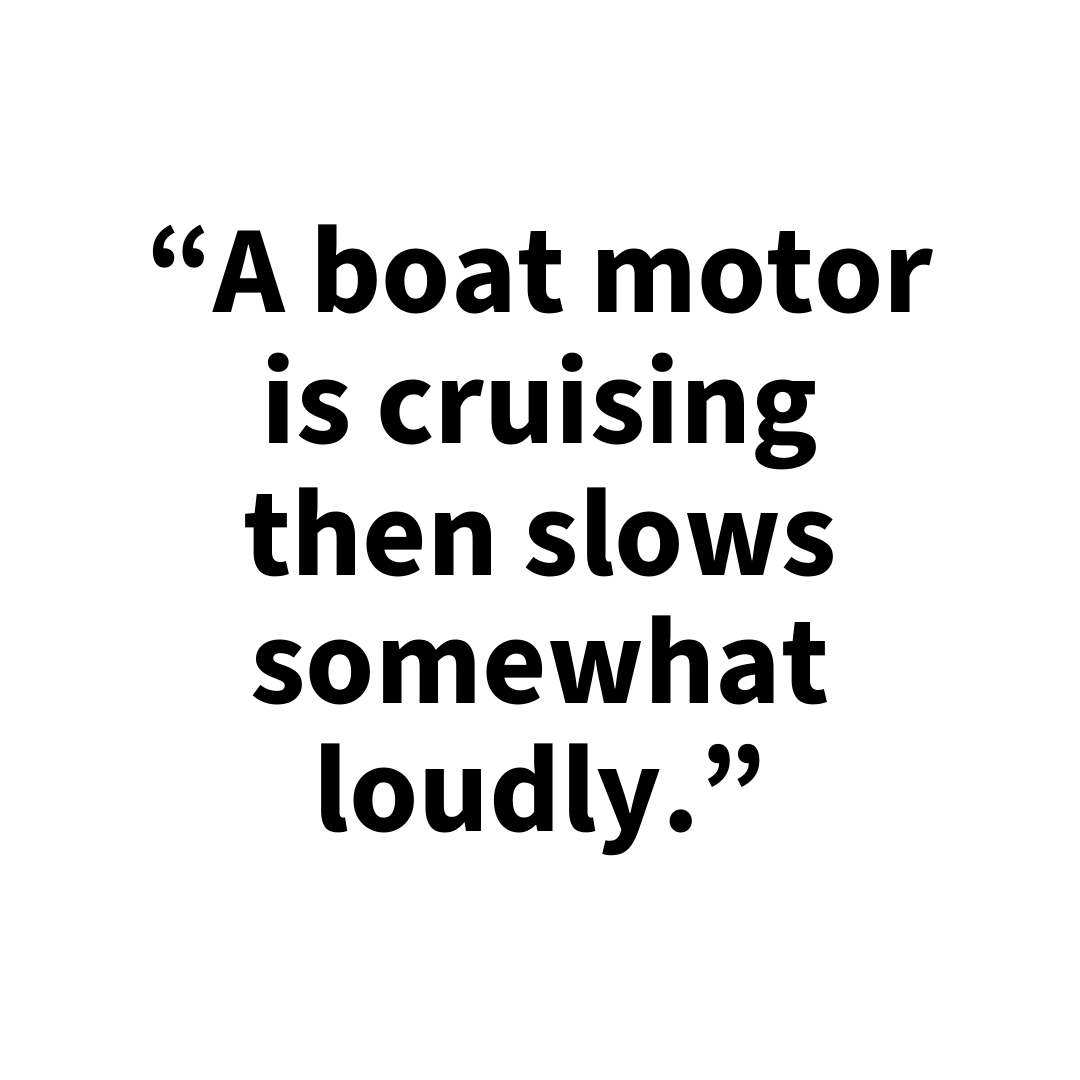} &
\includegraphics[width=0.18\textwidth]{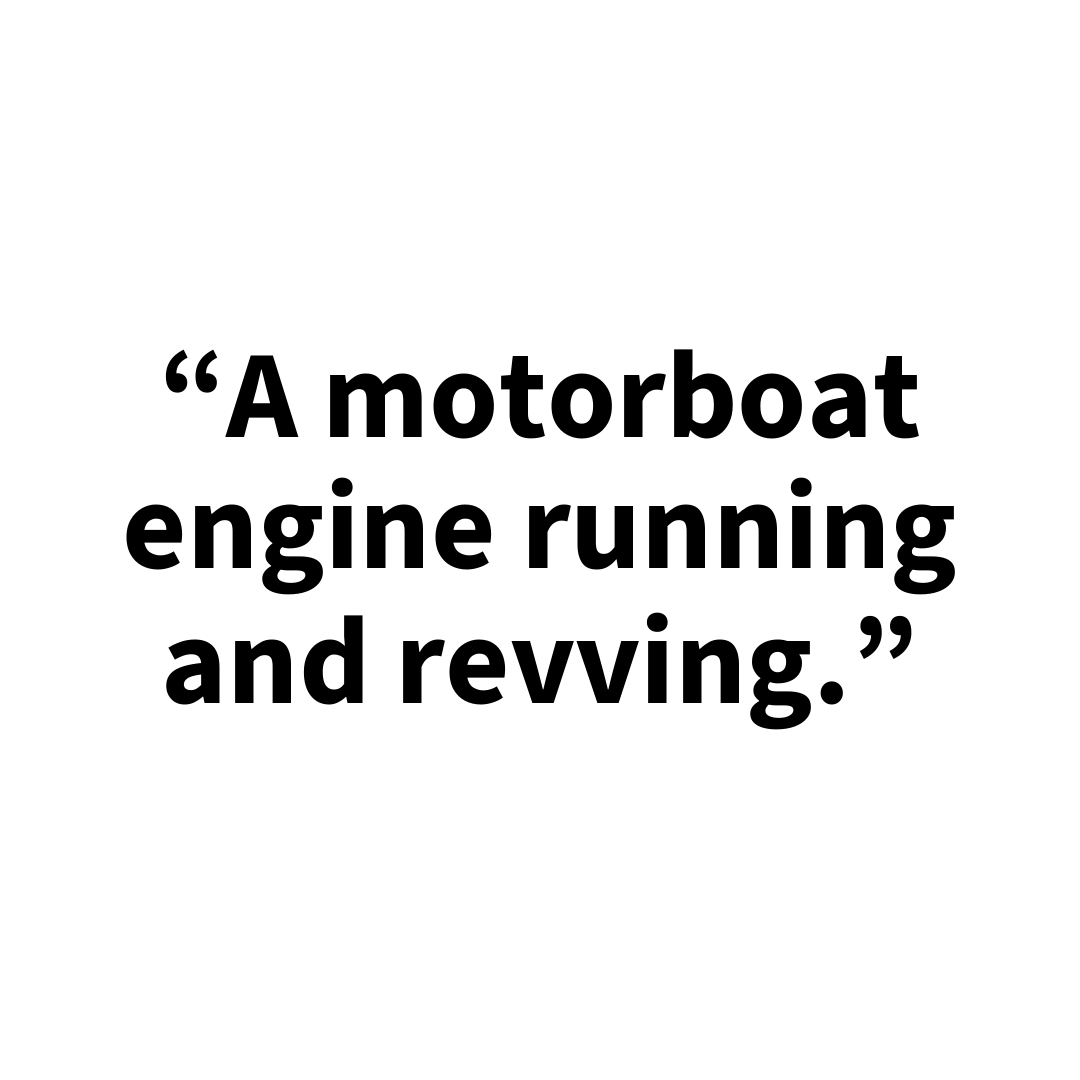} \\

\includegraphics[width=0.18\textwidth]{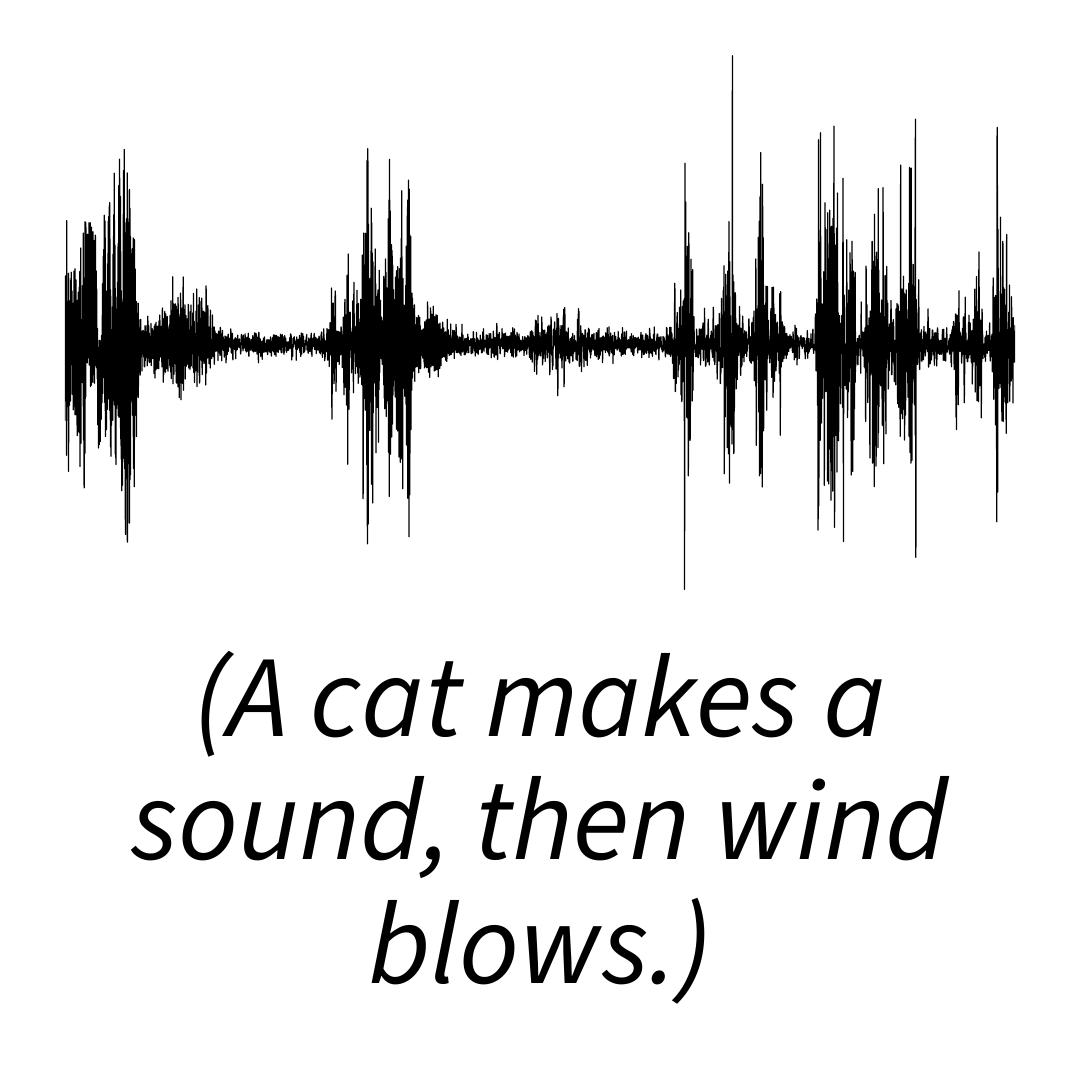} &
\includegraphics[width=0.18\textwidth]{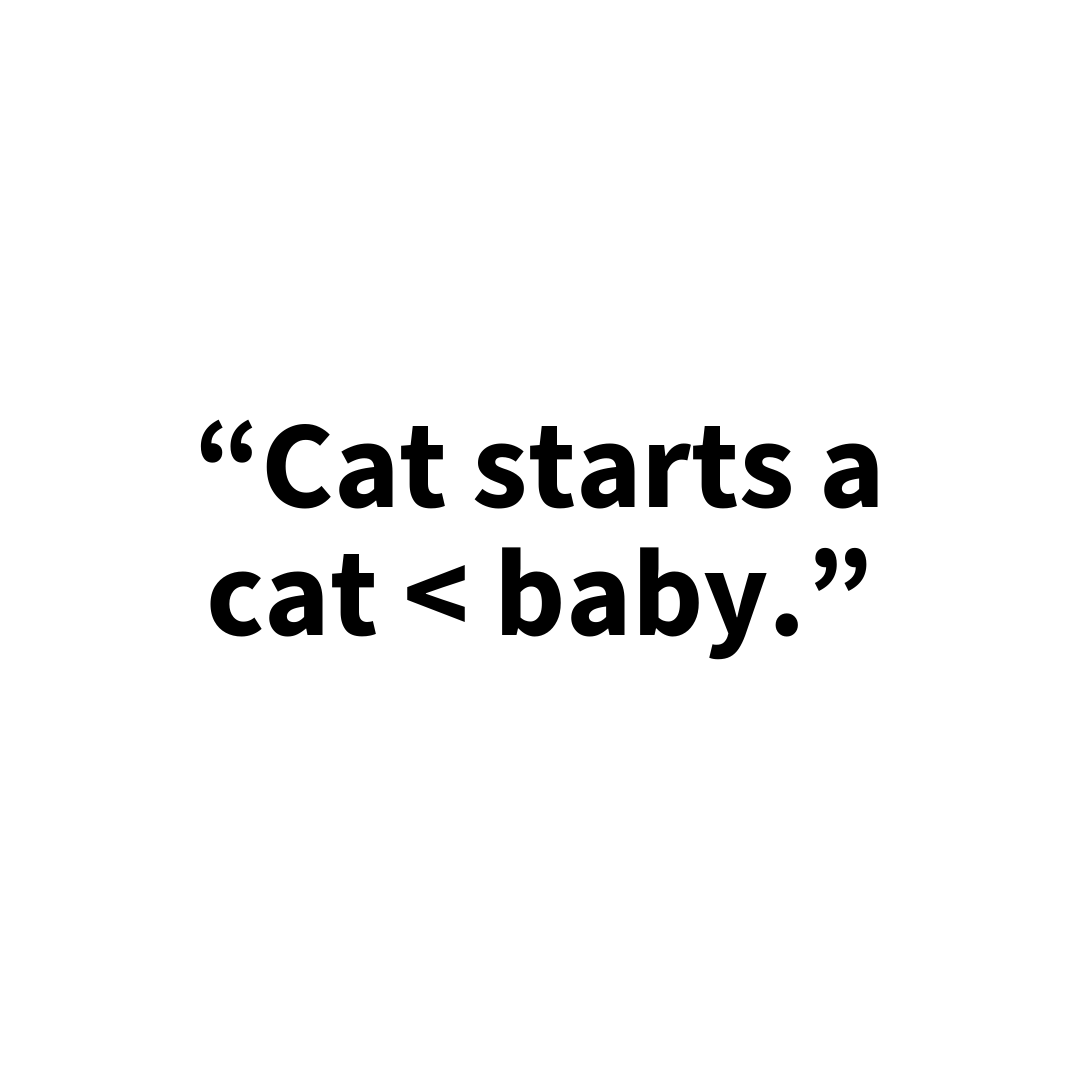} &
\includegraphics[width=0.18\textwidth]{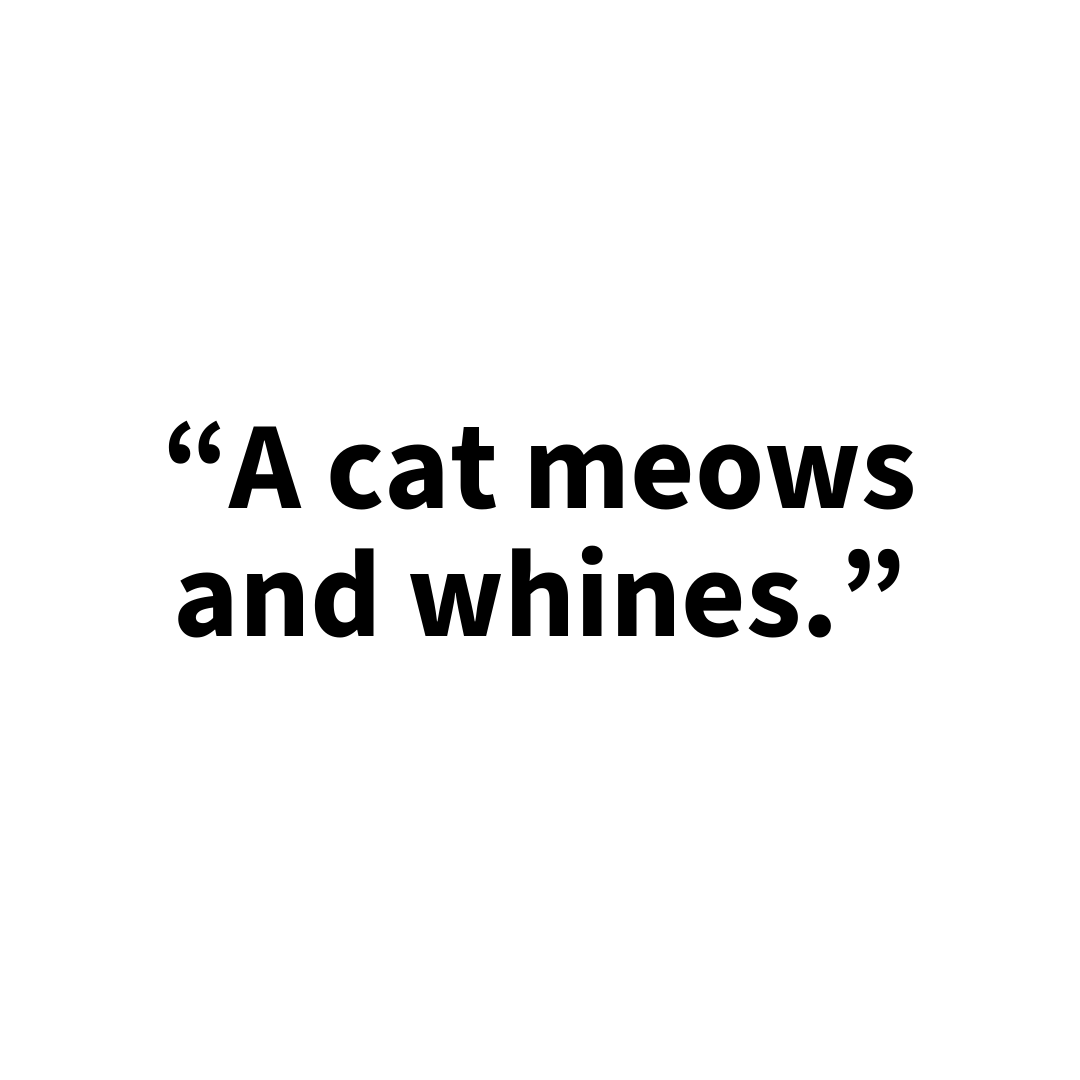} &
\includegraphics[width=0.18\textwidth]{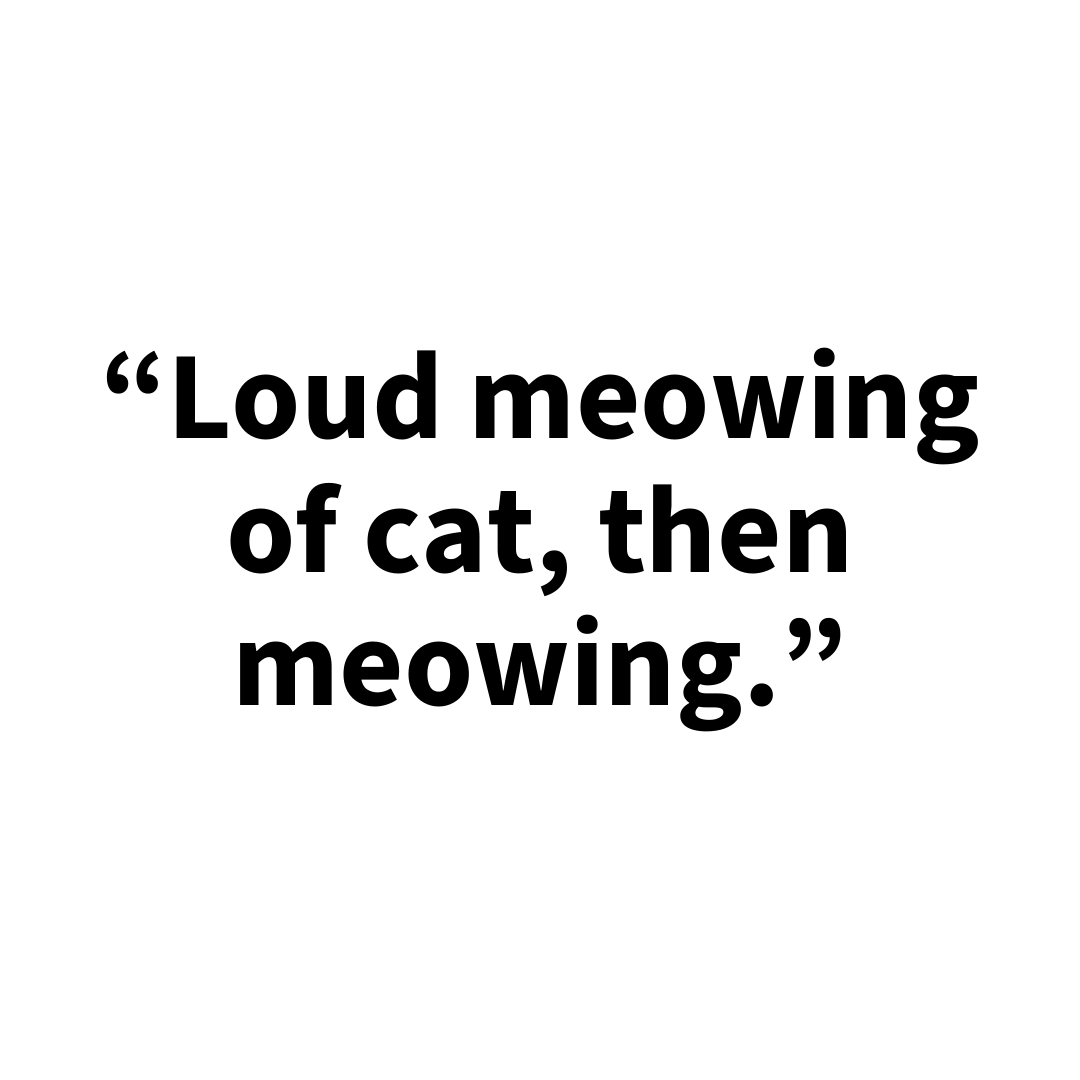} &
\includegraphics[width=0.18\textwidth]{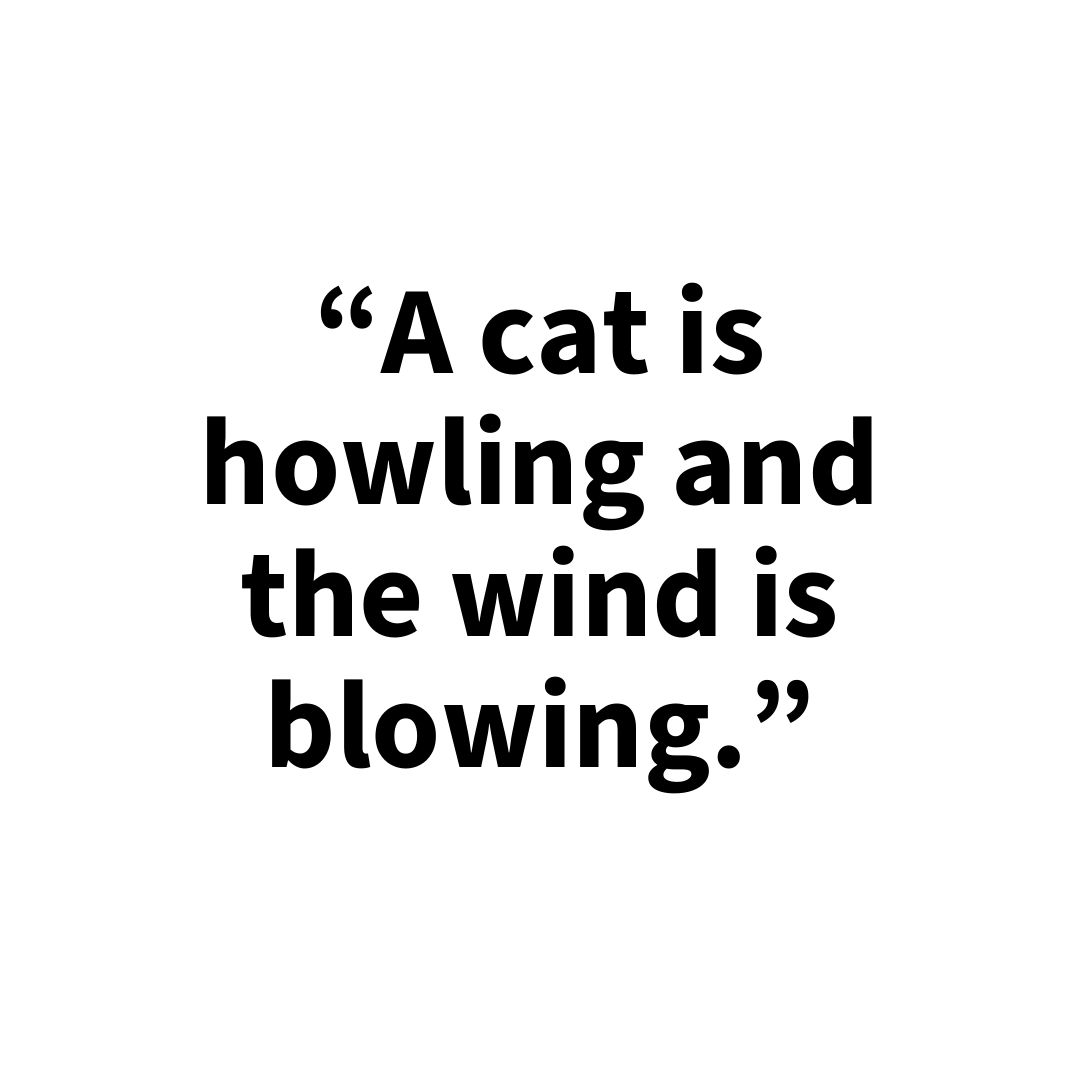} \\

\includegraphics[width=0.18\textwidth]{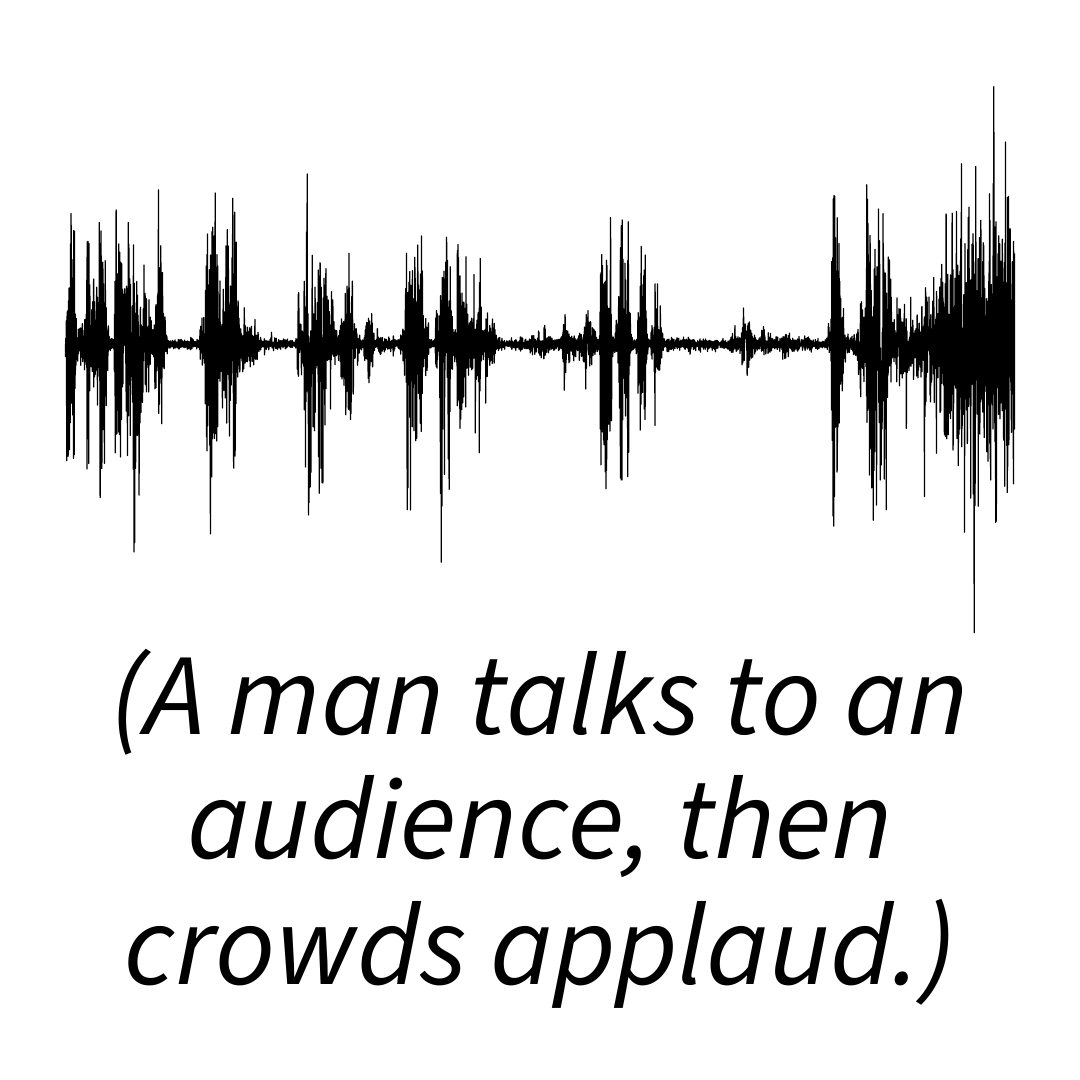} &
\includegraphics[width=0.18\textwidth]{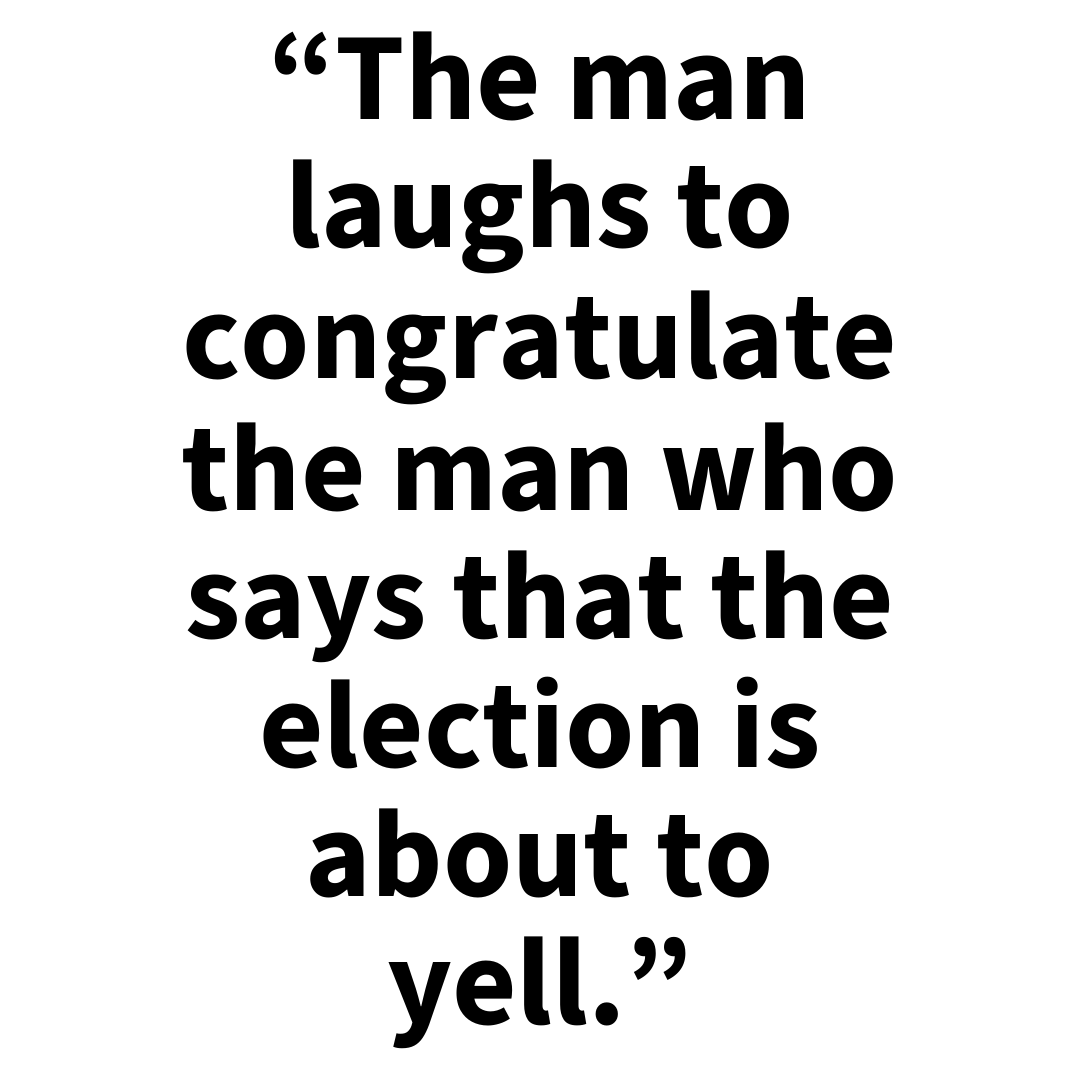} &
\includegraphics[width=0.18\textwidth]{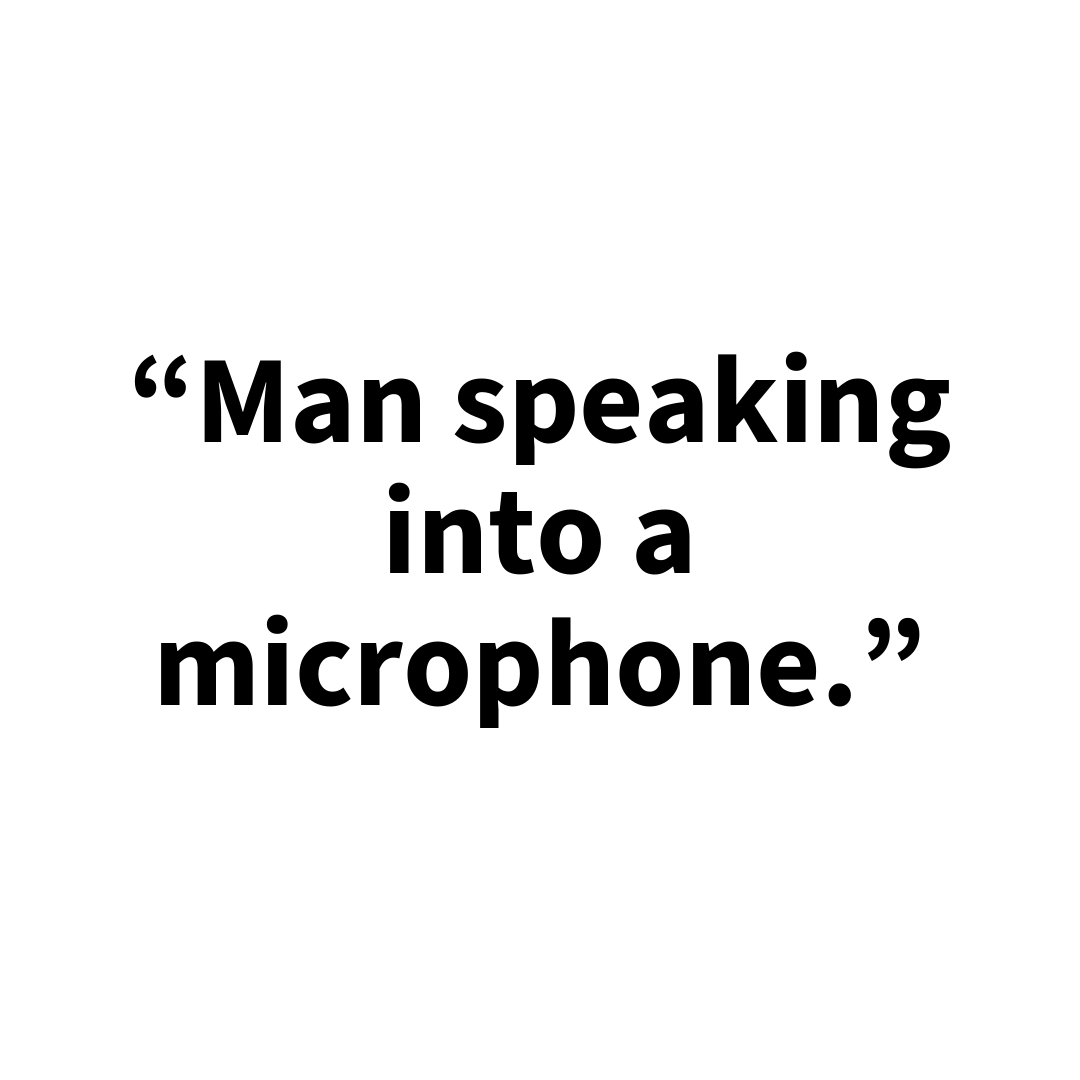} &
\includegraphics[width=0.18\textwidth]{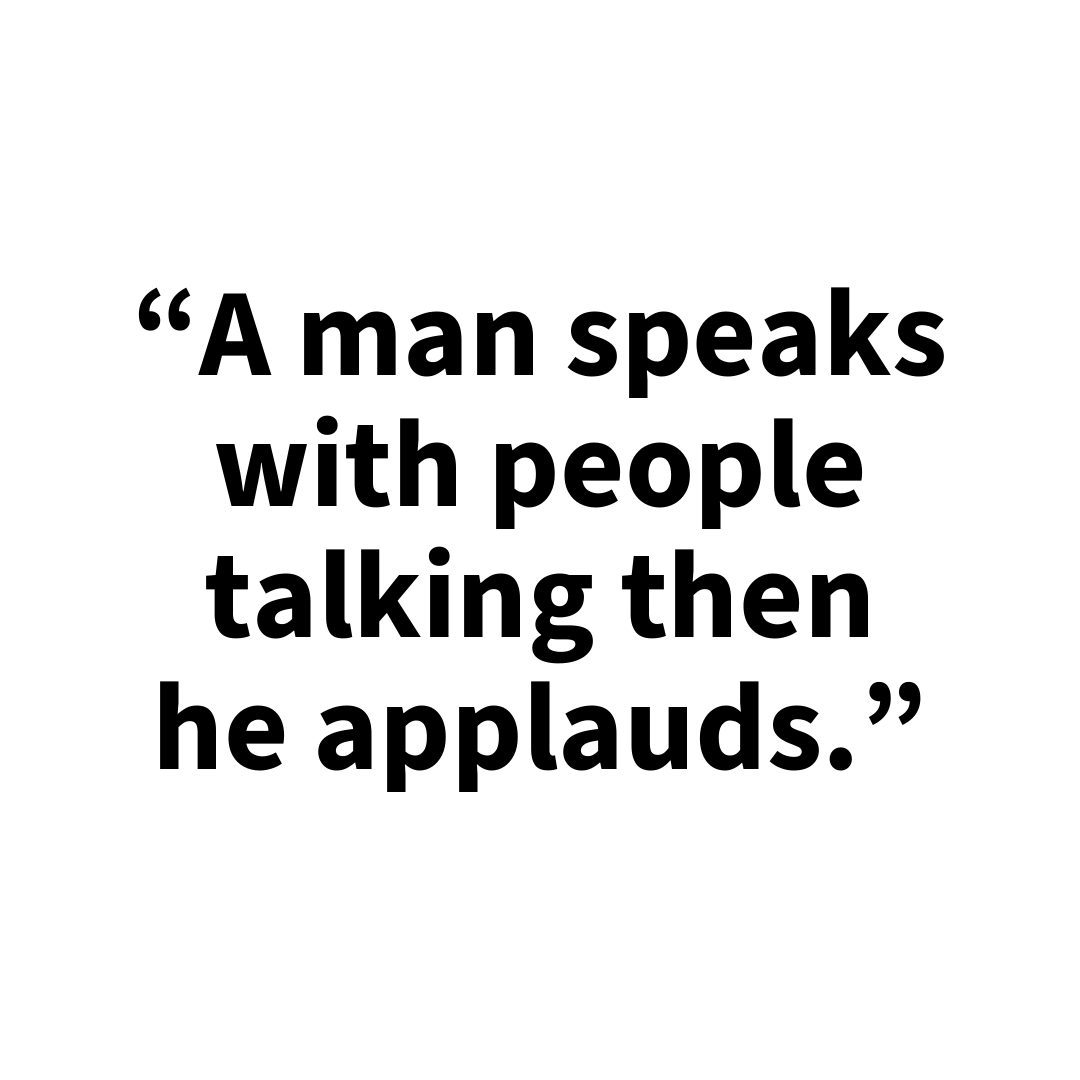} &
\includegraphics[width=0.18\textwidth]{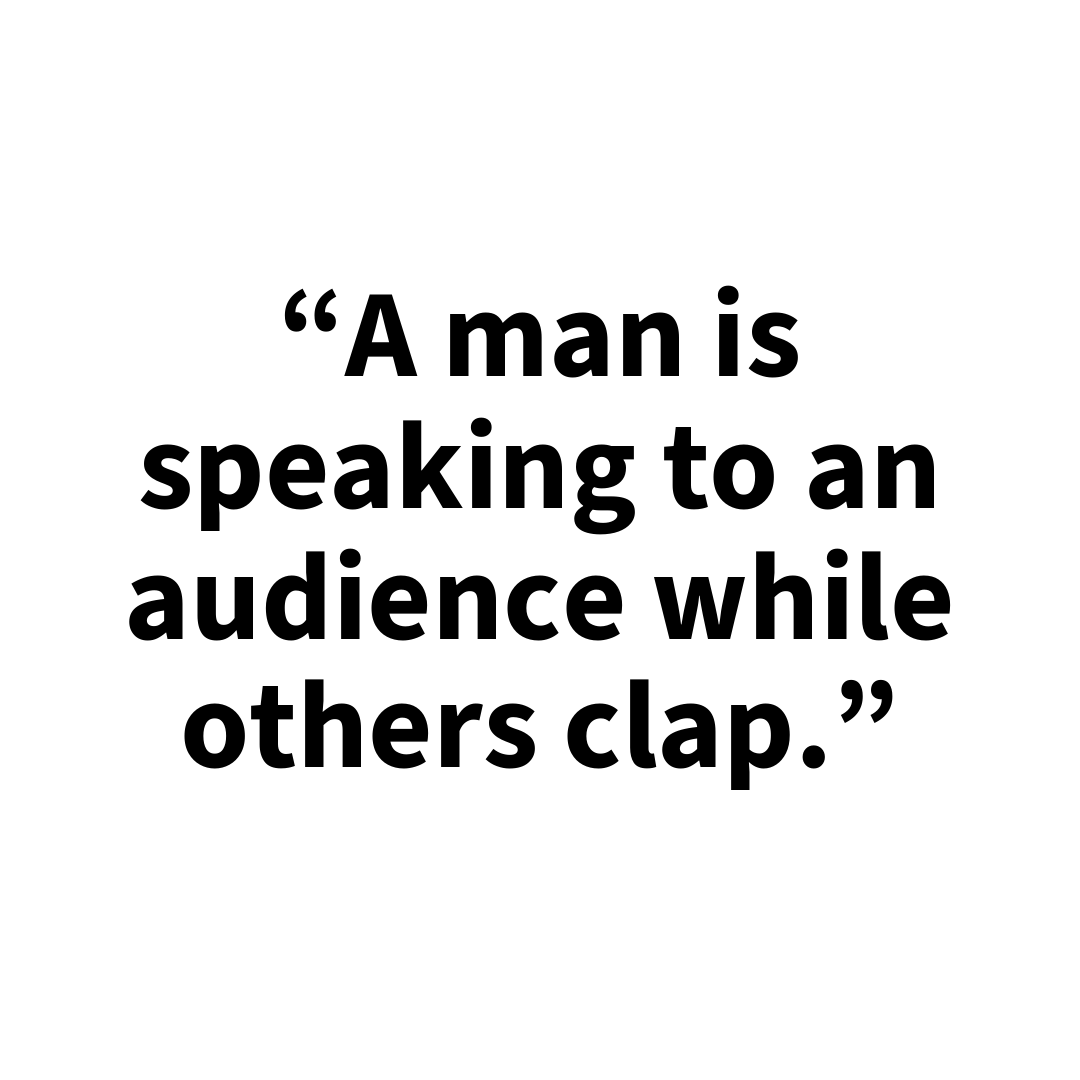} \\

\includegraphics[width=0.18\textwidth]{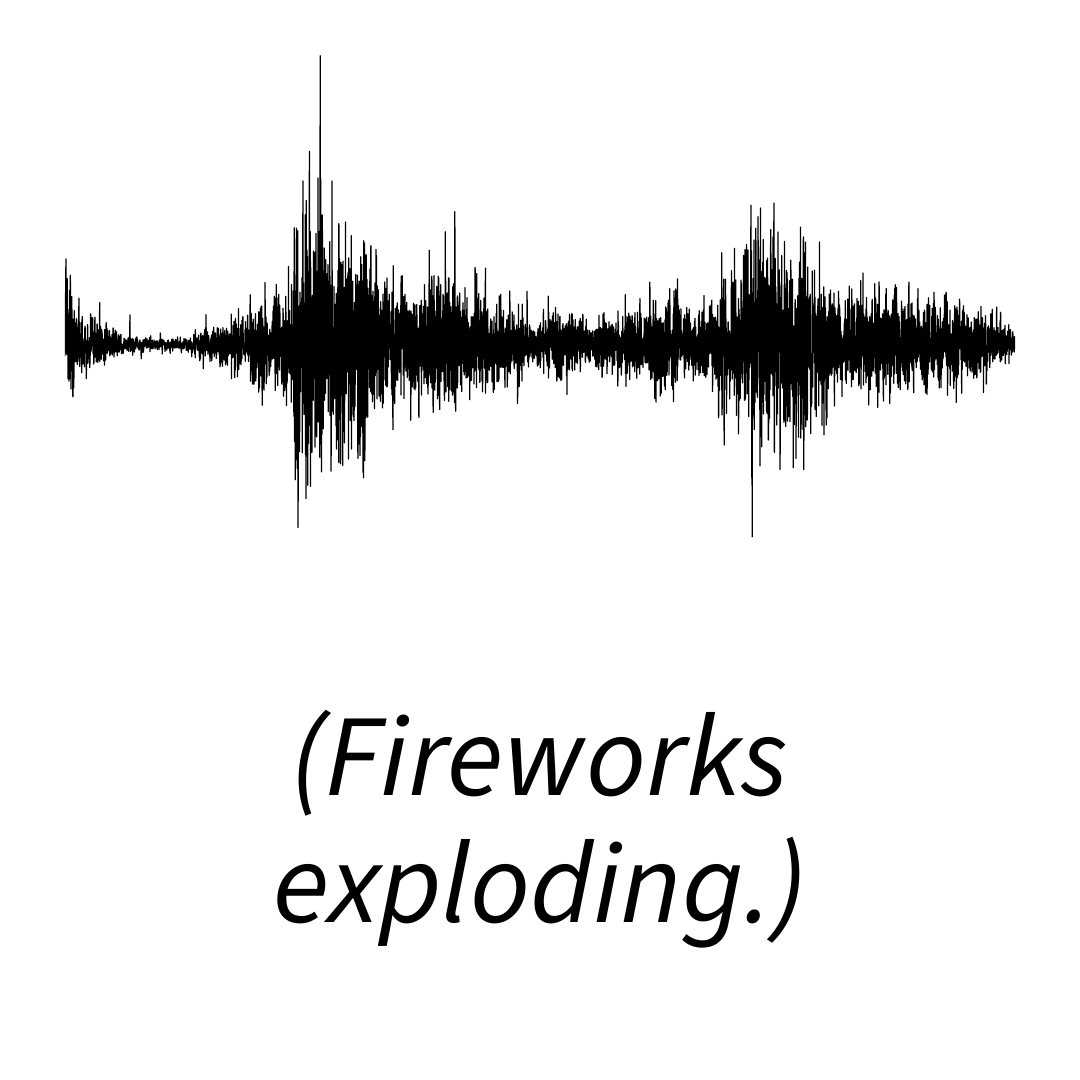} &
\includegraphics[width=0.18\textwidth]{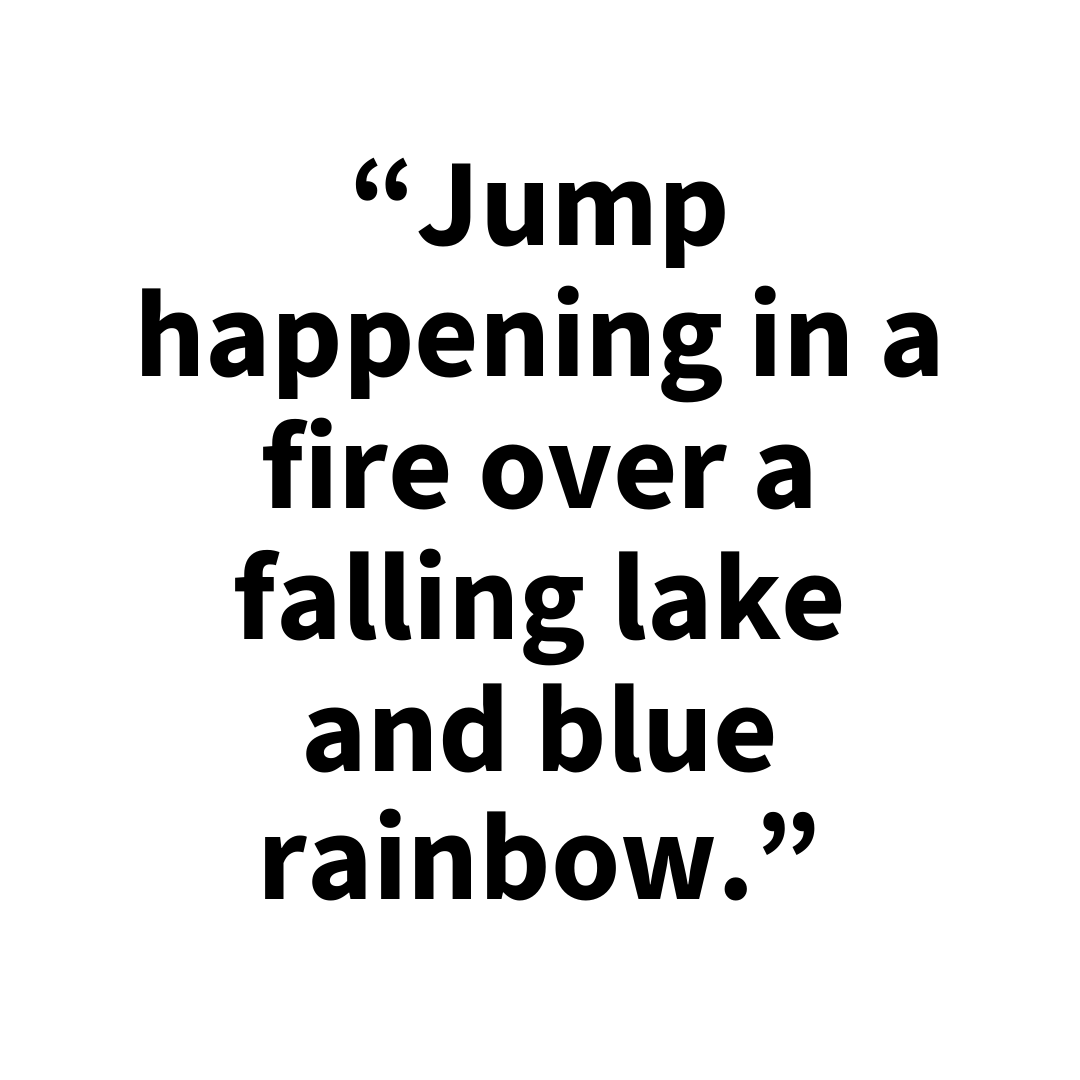} &
\includegraphics[width=0.18\textwidth]{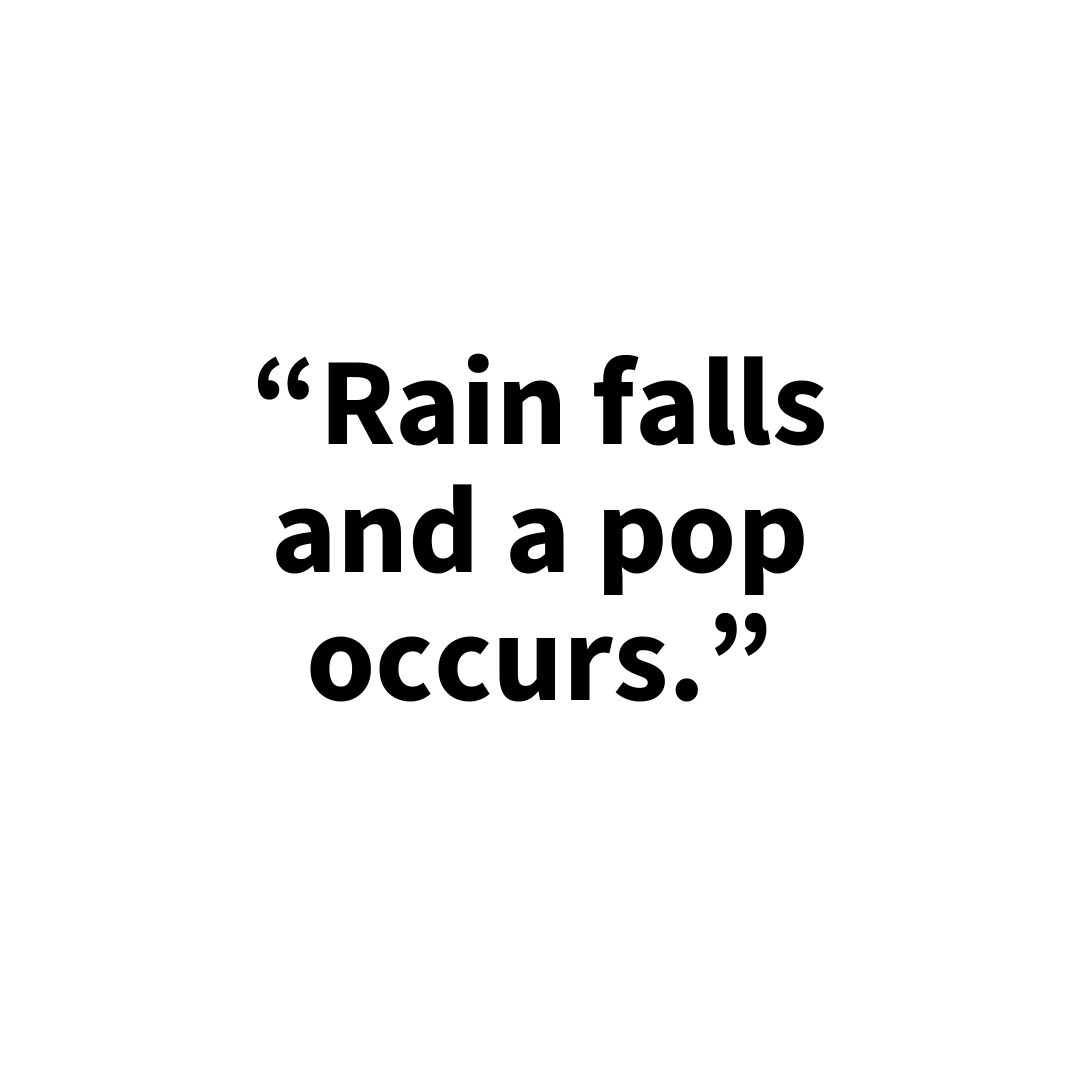} &
\includegraphics[width=0.18\textwidth]{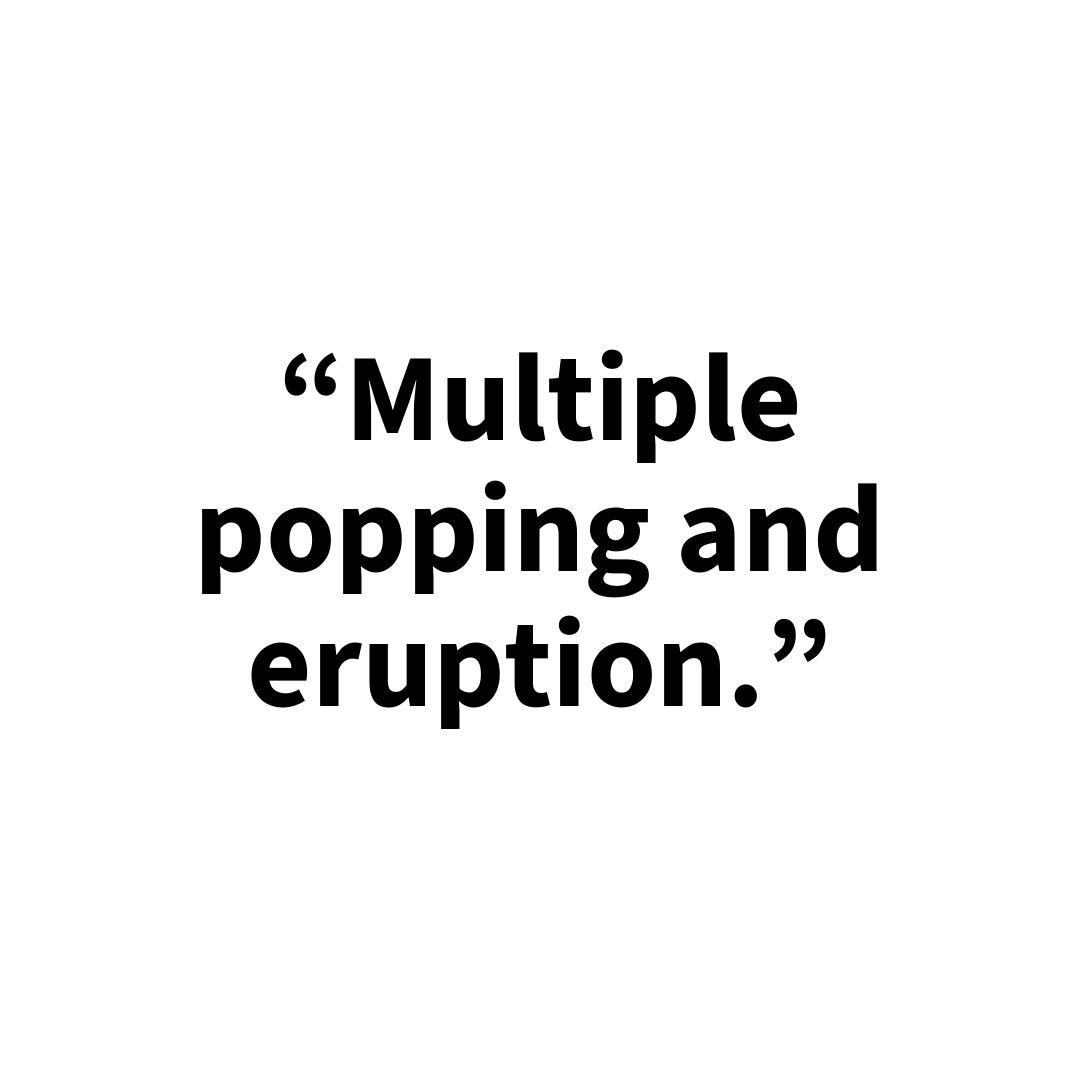} &
\includegraphics[width=0.18\textwidth]{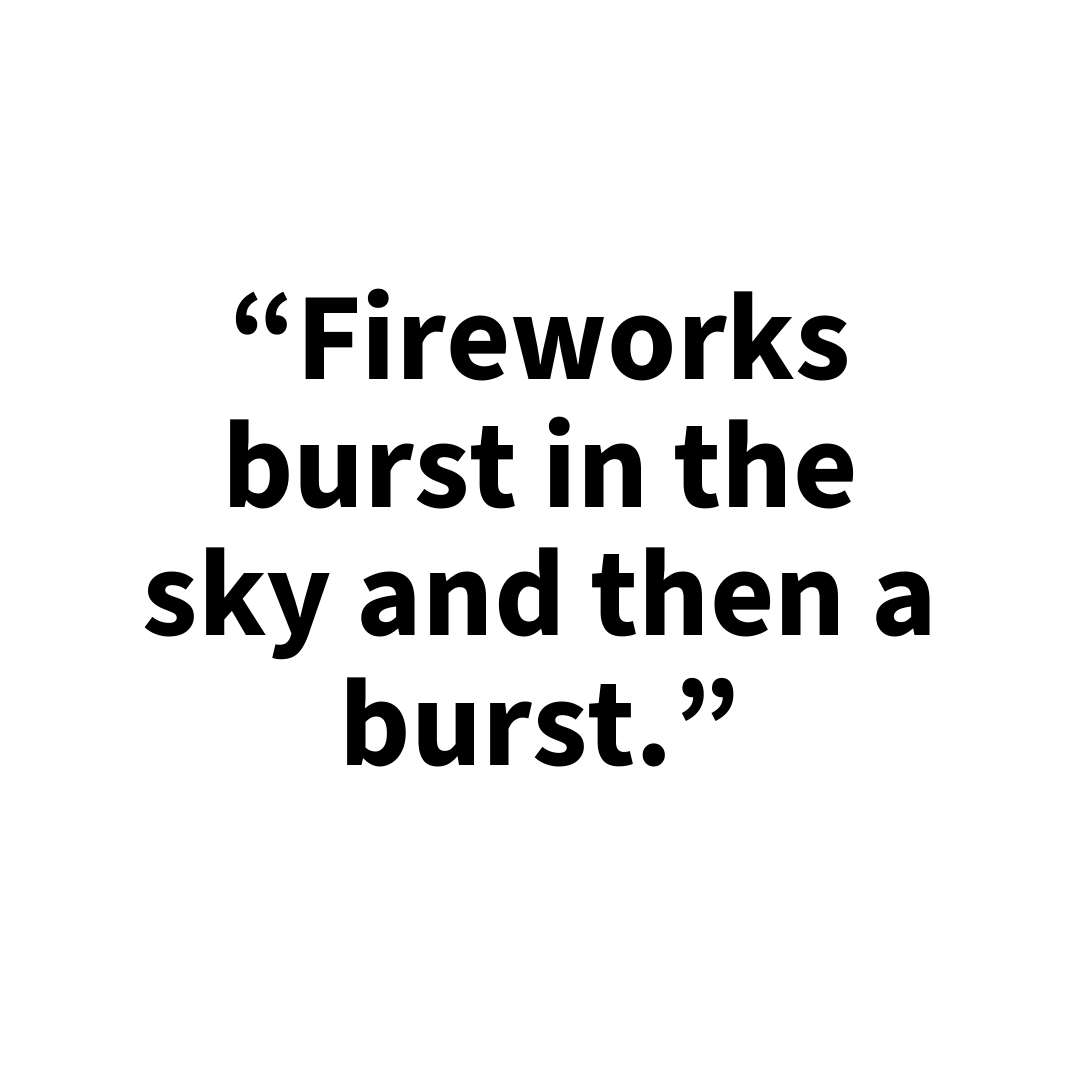} \\
\end{tabular}
\caption{\textbf{Audio-to-text generation.} Given a source audio (left), each method generates a caption.}
\label{fig:ita-qual-a2t}
\end{figure}

\begin{figure}[t]
\centering
\setlength{\tabcolsep}{2pt}
\renewcommand{\arraystretch}{1.0}
\begin{tabular}{c|cccc}
\footnotesize Source (A) &
\footnotesize CoDi~\cite{tang2023:codi} &
\footnotesize OmniFlow~\cite{li2025:omniflow} &
\footnotesize FlowBind~\cite{cha2026:flowbind} &
\footnotesize \textbf{\Ours{}$^\dagger$ (Ours)} \\
\midrule
\includegraphics[width=0.18\textwidth]{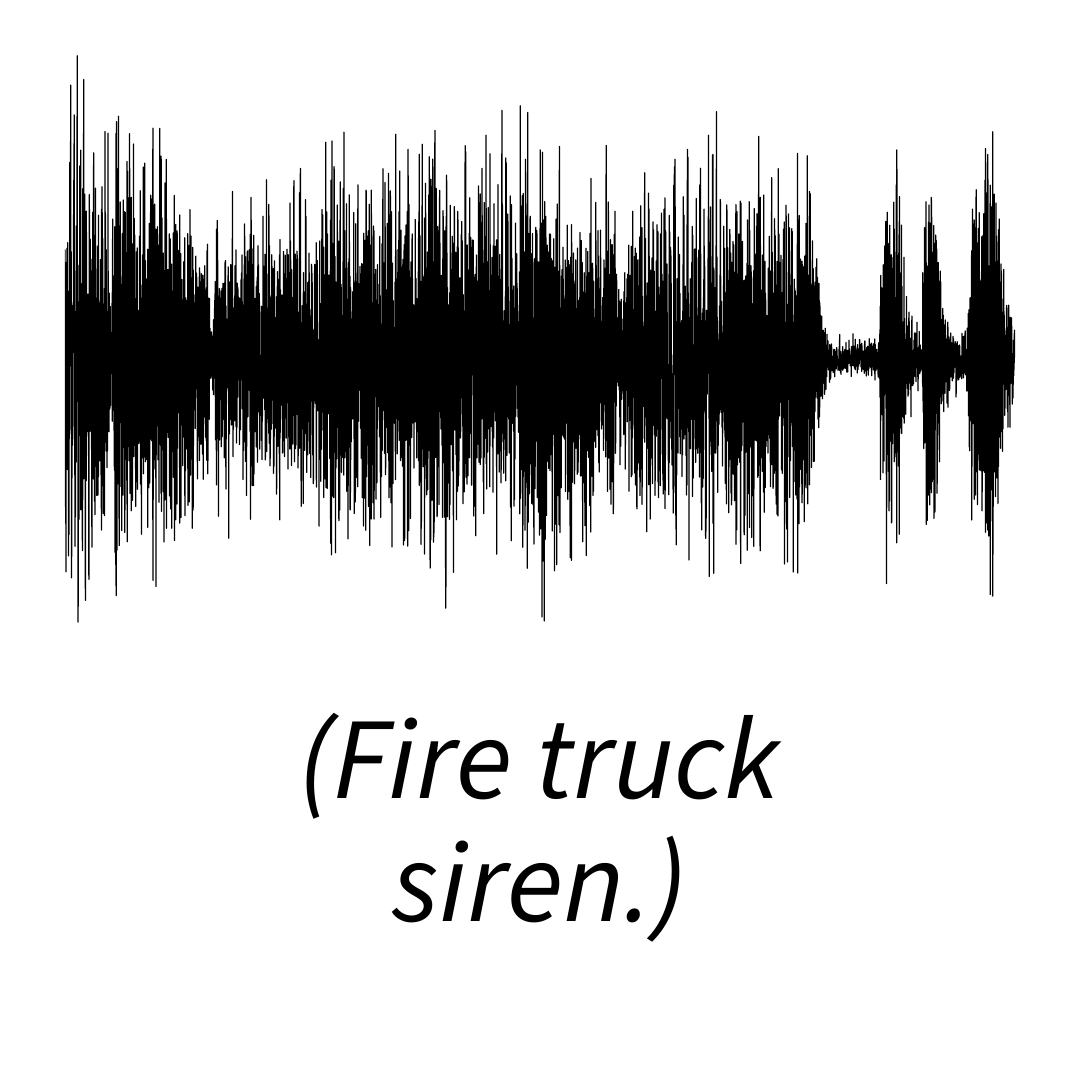} &
\includegraphics[width=0.18\textwidth]{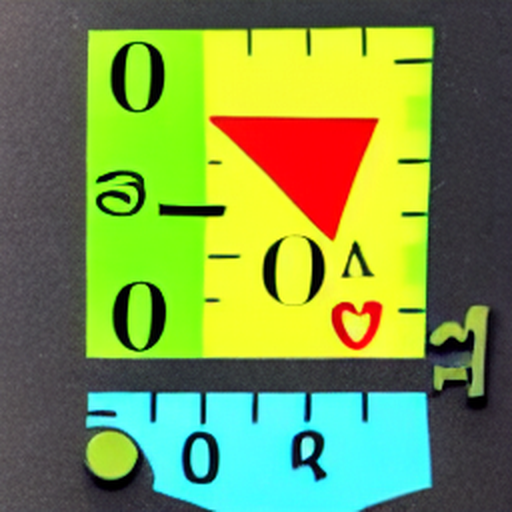} &
\includegraphics[width=0.18\textwidth]{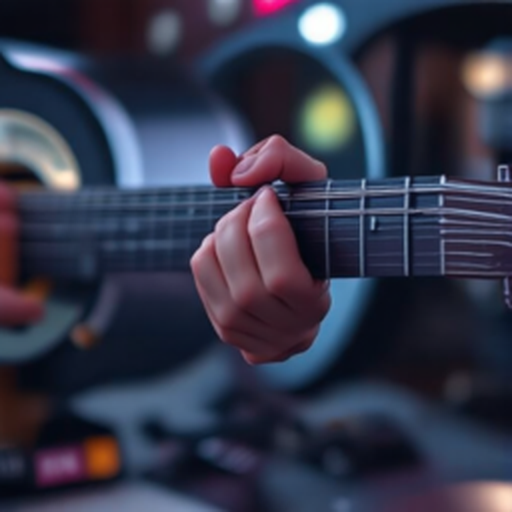} &
\includegraphics[width=0.18\textwidth]{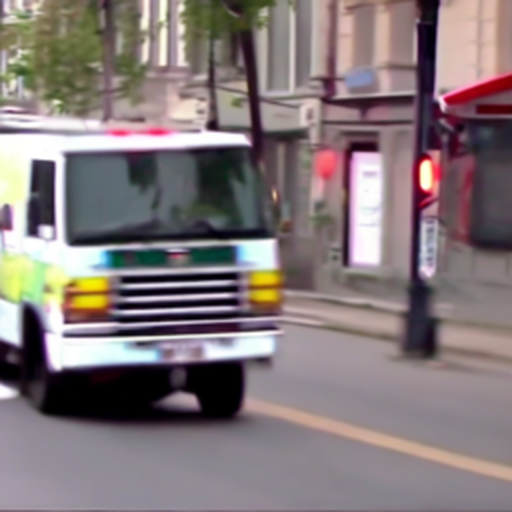} &
\includegraphics[width=0.18\textwidth]{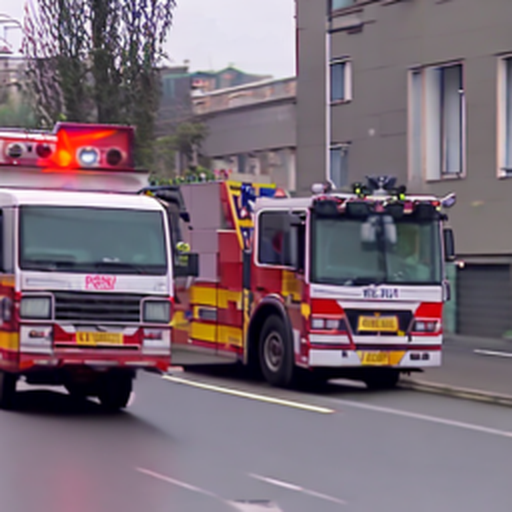} \\

\includegraphics[width=0.18\textwidth]{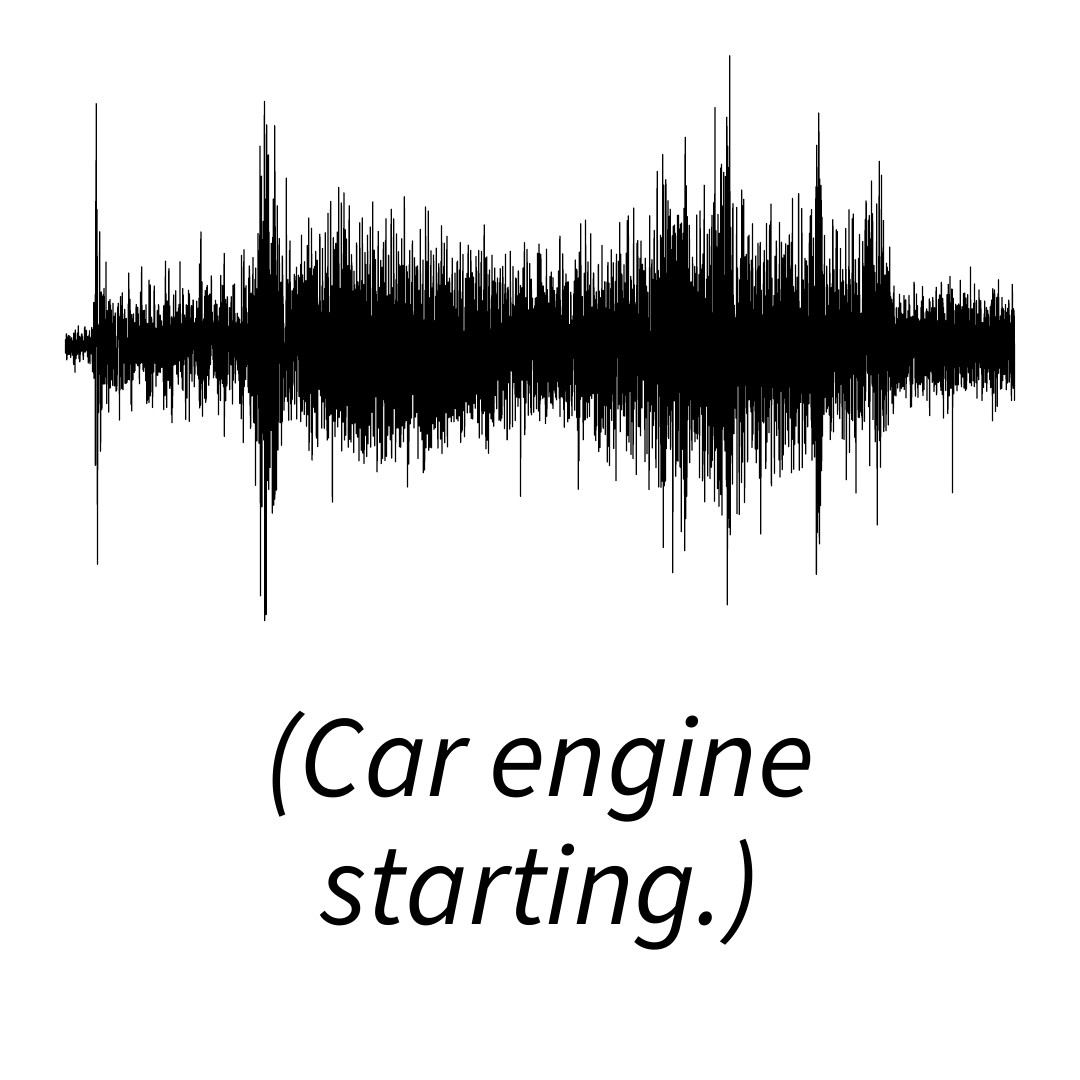} &
\includegraphics[width=0.18\textwidth]{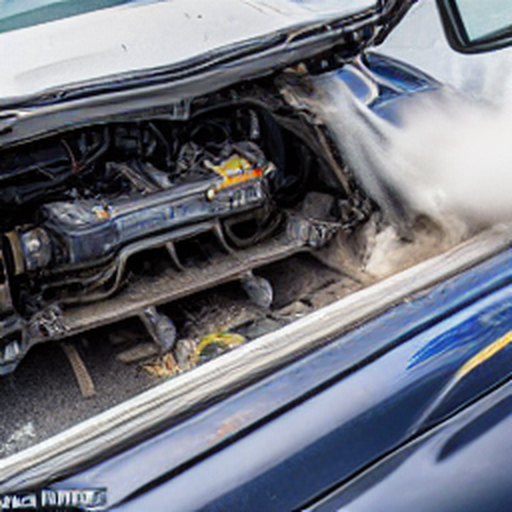} &
\includegraphics[width=0.18\textwidth]{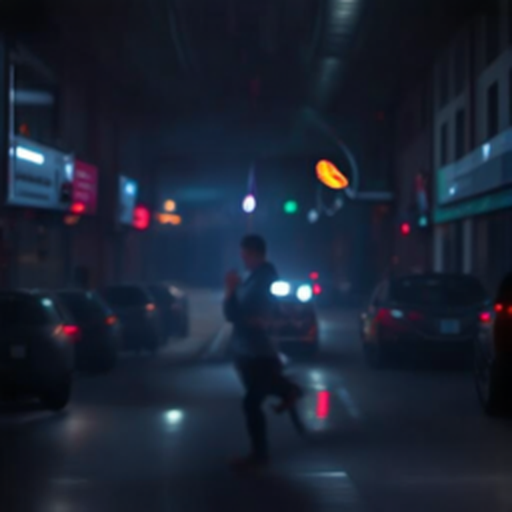} &
\includegraphics[width=0.18\textwidth]{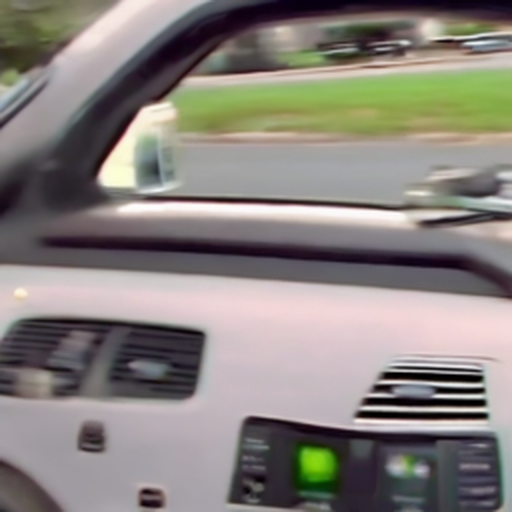} &
\includegraphics[width=0.18\textwidth]{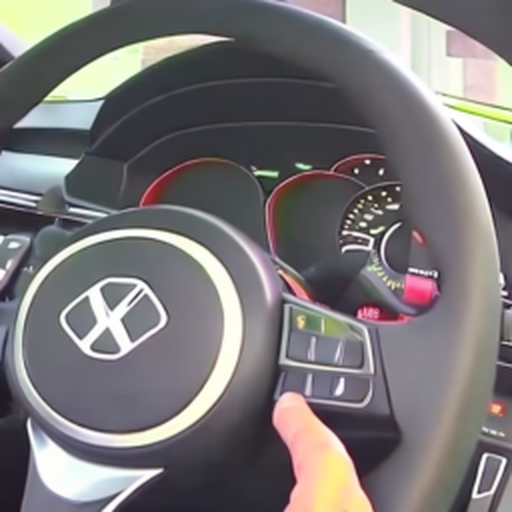} \\

\includegraphics[width=0.18\textwidth]{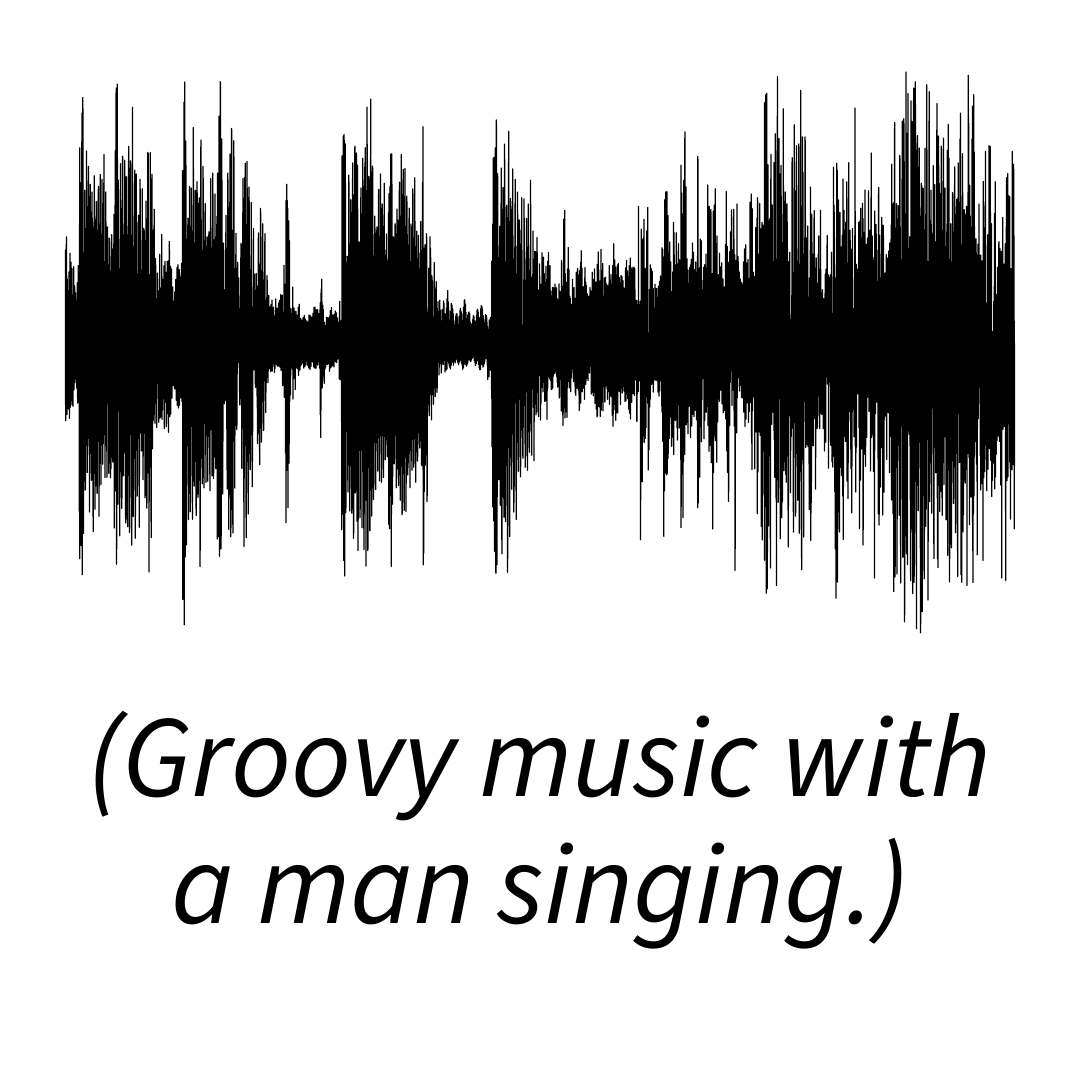} &
\includegraphics[width=0.18\textwidth]{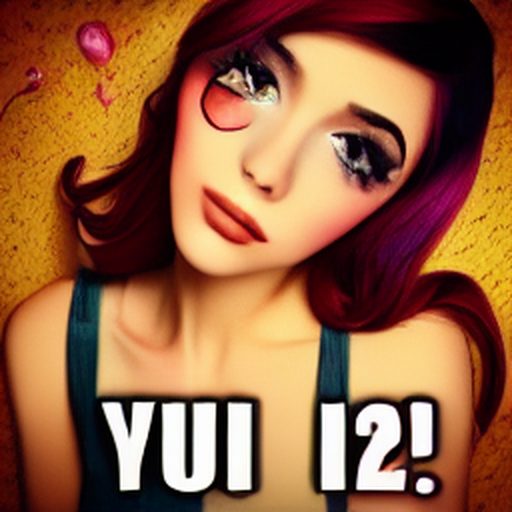} &
\includegraphics[width=0.18\textwidth]{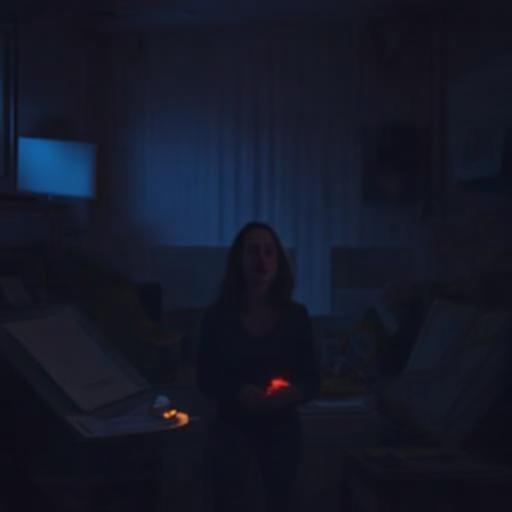} &
\includegraphics[width=0.18\textwidth]{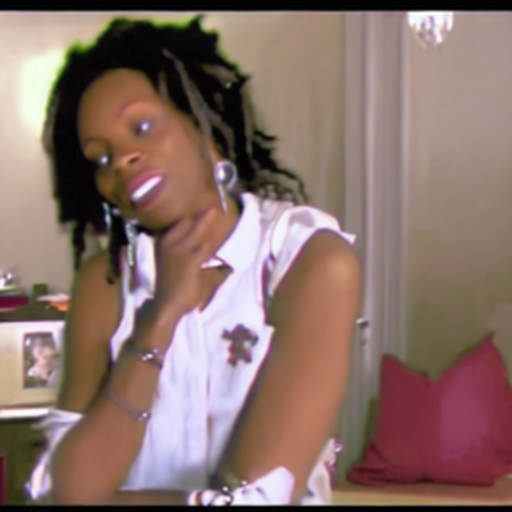} &
\includegraphics[width=0.18\textwidth]{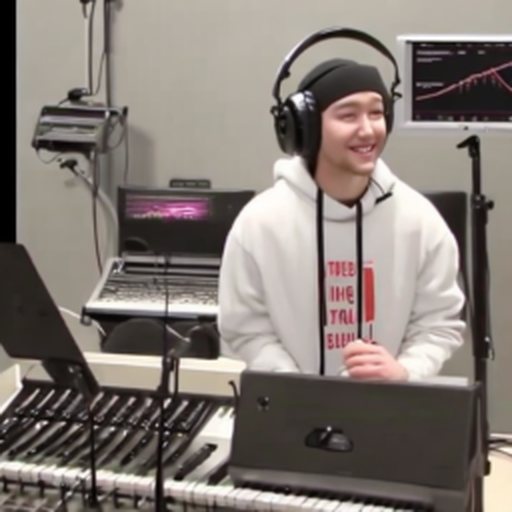} \\

\includegraphics[width=0.18\textwidth]{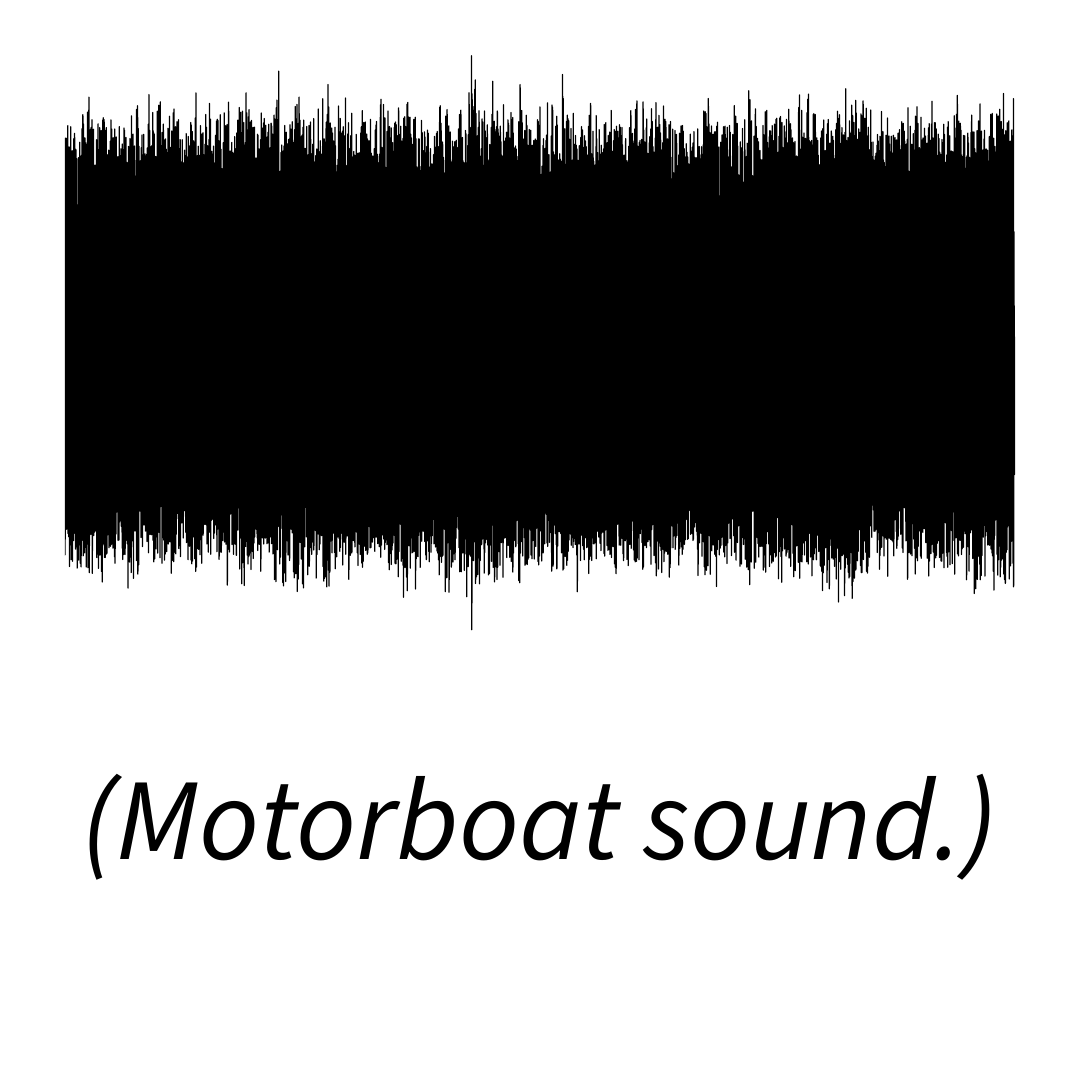} &
\includegraphics[width=0.18\textwidth]{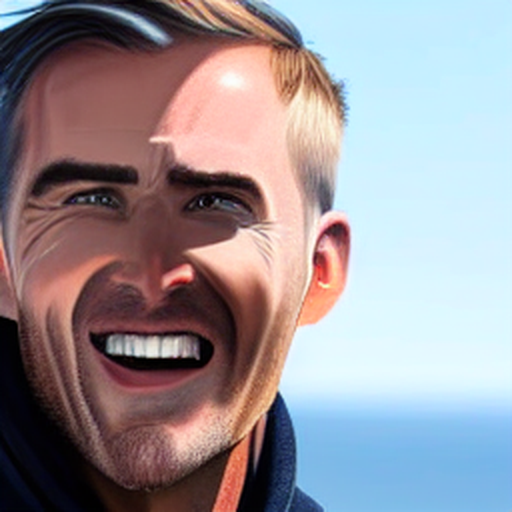} &
\includegraphics[width=0.18\textwidth]{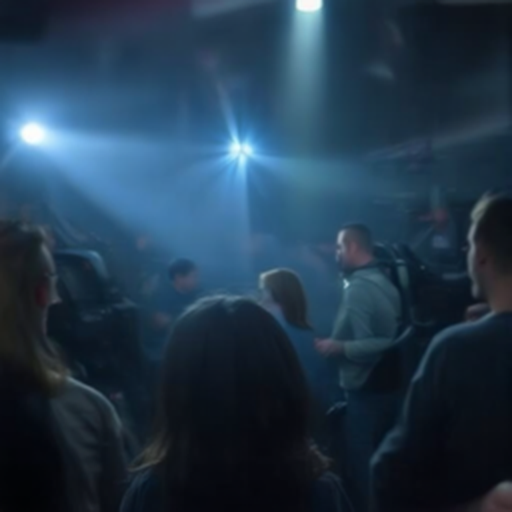} &
\includegraphics[width=0.18\textwidth]{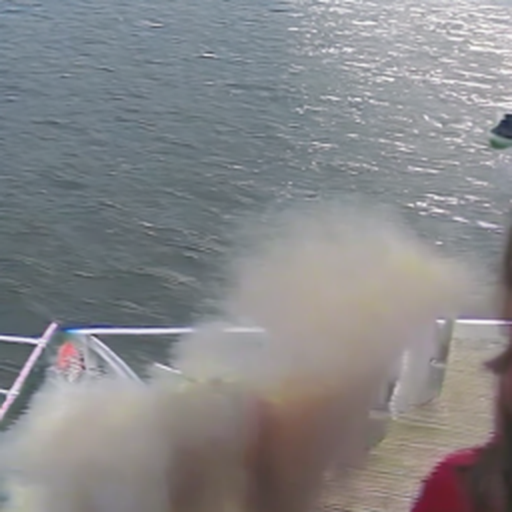} &
\includegraphics[width=0.18\textwidth]{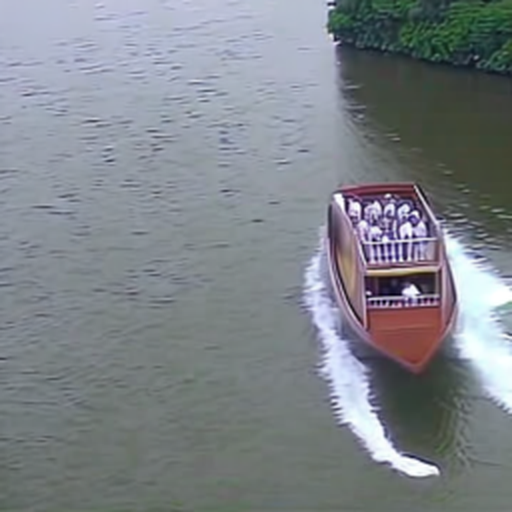} \\

\includegraphics[width=0.18\textwidth]{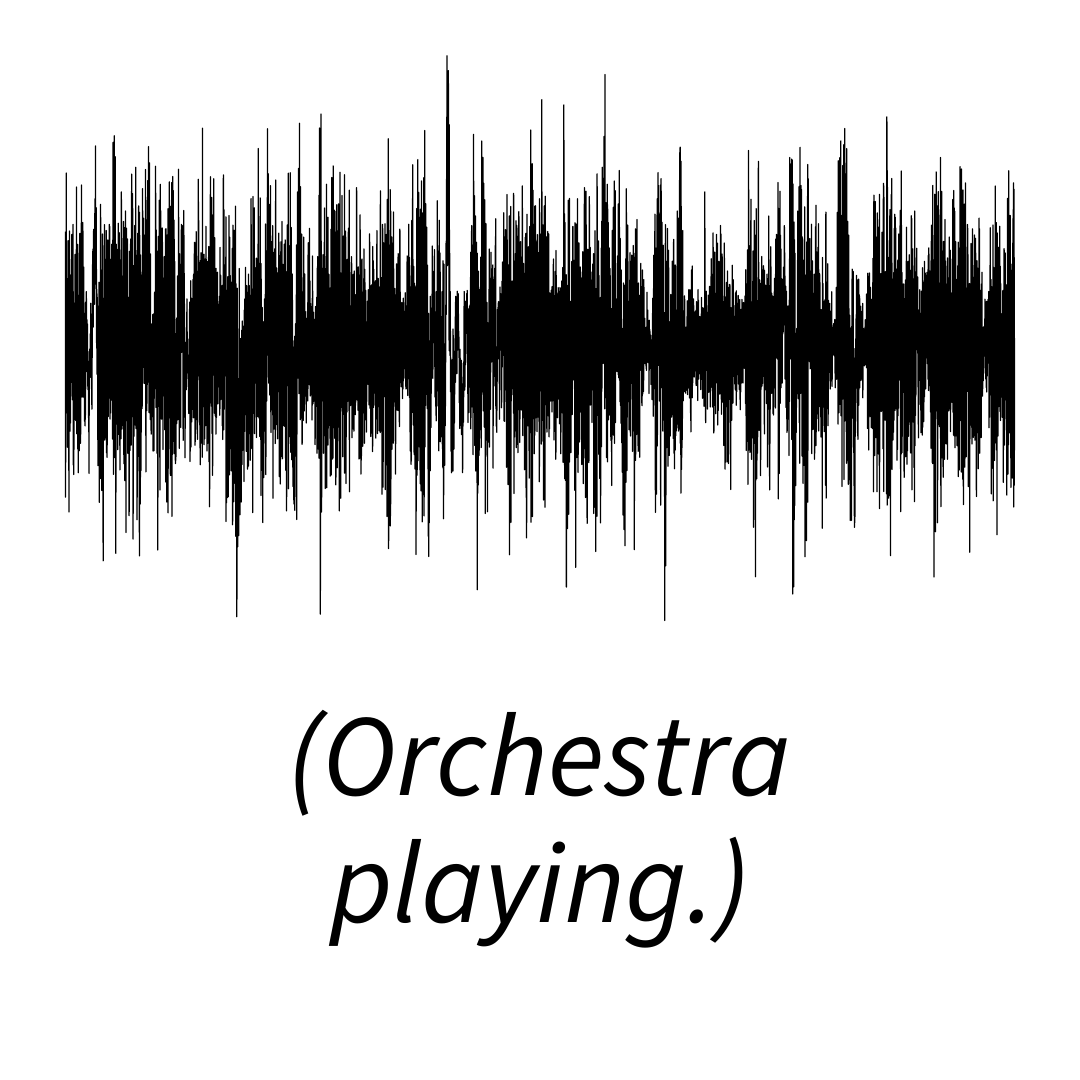} &
\includegraphics[width=0.18\textwidth]{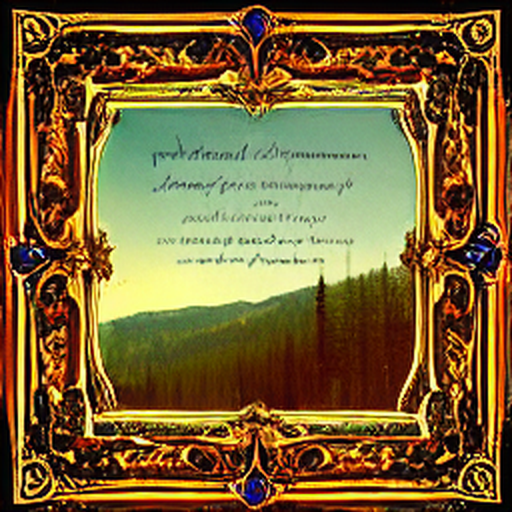} &
\includegraphics[width=0.18\textwidth]{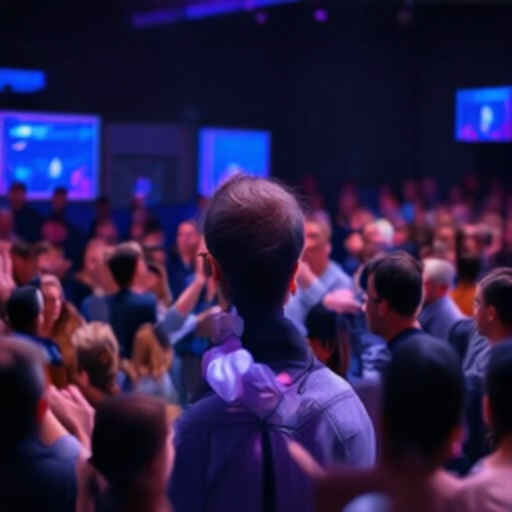} &
\includegraphics[width=0.18\textwidth]{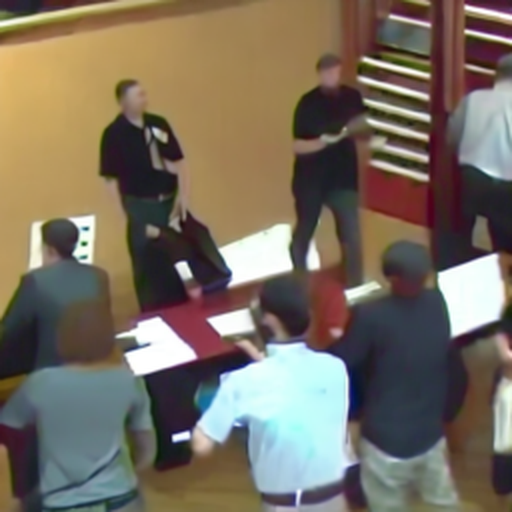} &
\includegraphics[width=0.18\textwidth]{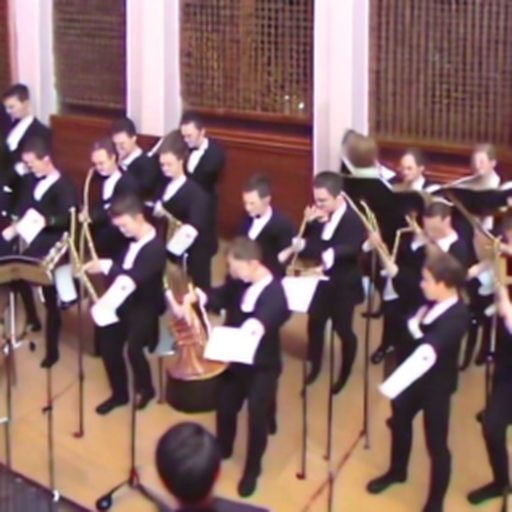} \\

\includegraphics[width=0.18\textwidth]{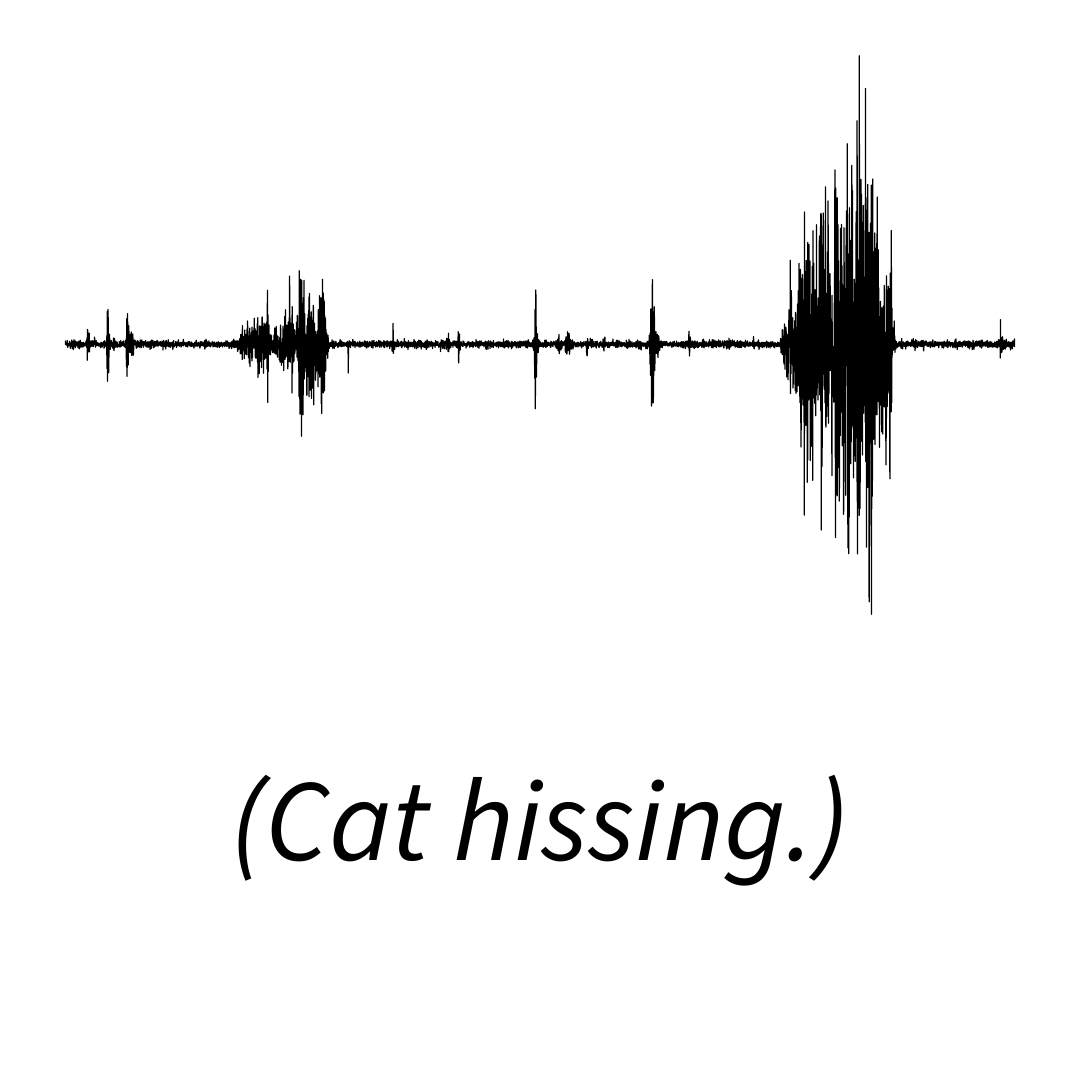} &
\includegraphics[width=0.18\textwidth]{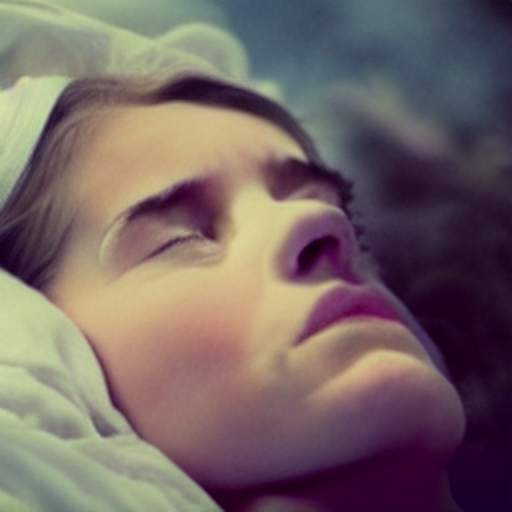} &
\includegraphics[width=0.18\textwidth]{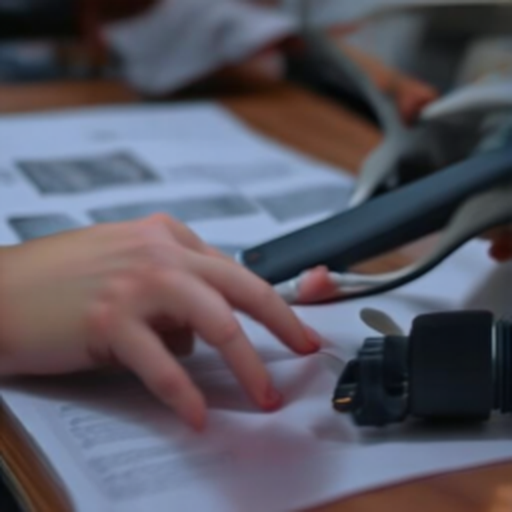} &
\includegraphics[width=0.18\textwidth]{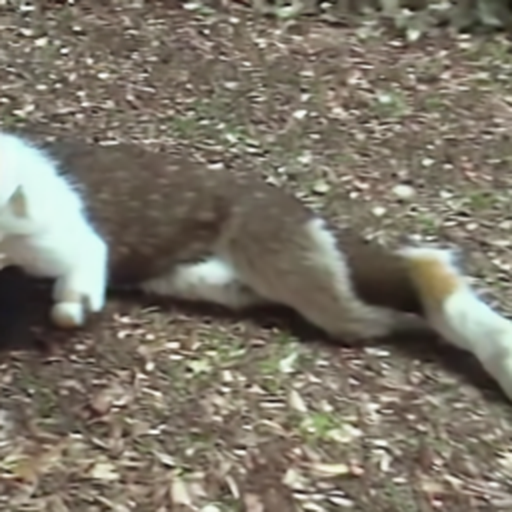} &
\includegraphics[width=0.18\textwidth]{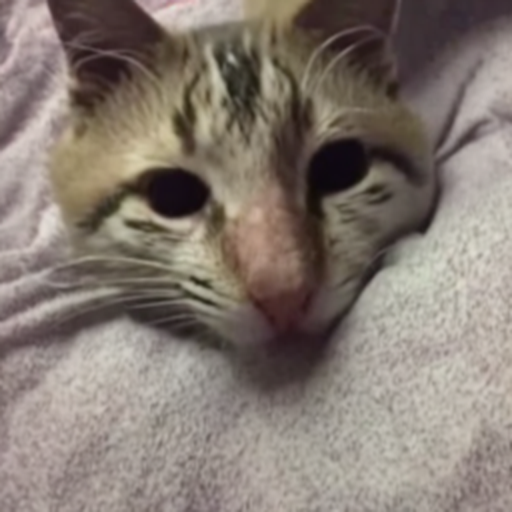} \\
\end{tabular}
\caption{\textbf{Audio-to-image generation.} Given a source audio (left), each method generates an image.}
\label{fig:ita-qual-a2i}
\end{figure}

\begin{figure}[t]
\centering
\setlength{\tabcolsep}{2pt}
\renewcommand{\arraystretch}{1.0}
\begin{tabular}{cc|cccc}
\footnotesize Source (I) &
\footnotesize Source (A) &
\footnotesize CoDi~\cite{tang2023:codi} &
\footnotesize OmniFlow~\cite{li2025:omniflow} &
\footnotesize FlowBind~\cite{cha2026:flowbind} &
\footnotesize \textbf{\Ours{}$^\dagger$ (Ours)} \\
\midrule
\includegraphics[width=0.15\textwidth]{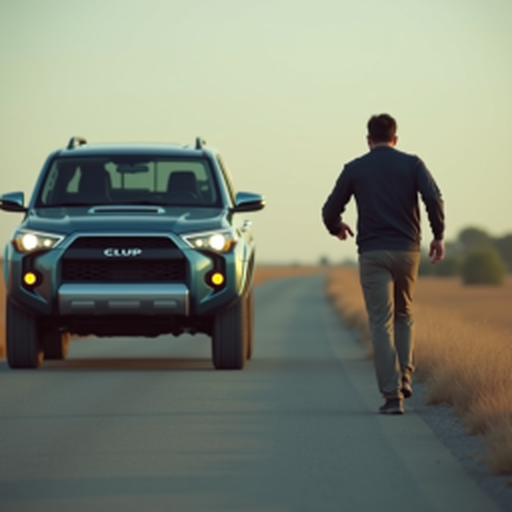} &
\includegraphics[width=0.15\textwidth]{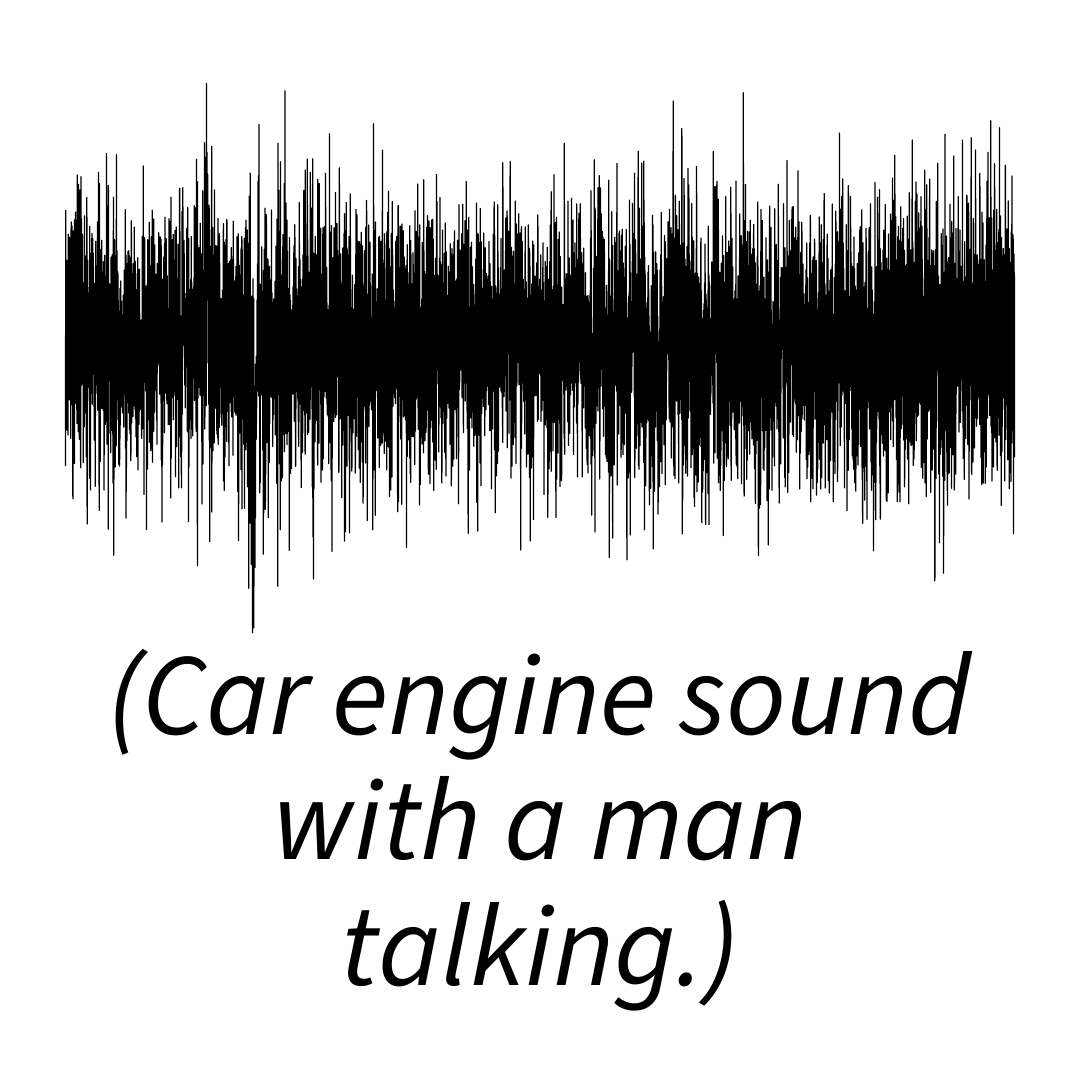} &
\includegraphics[width=0.15\textwidth]{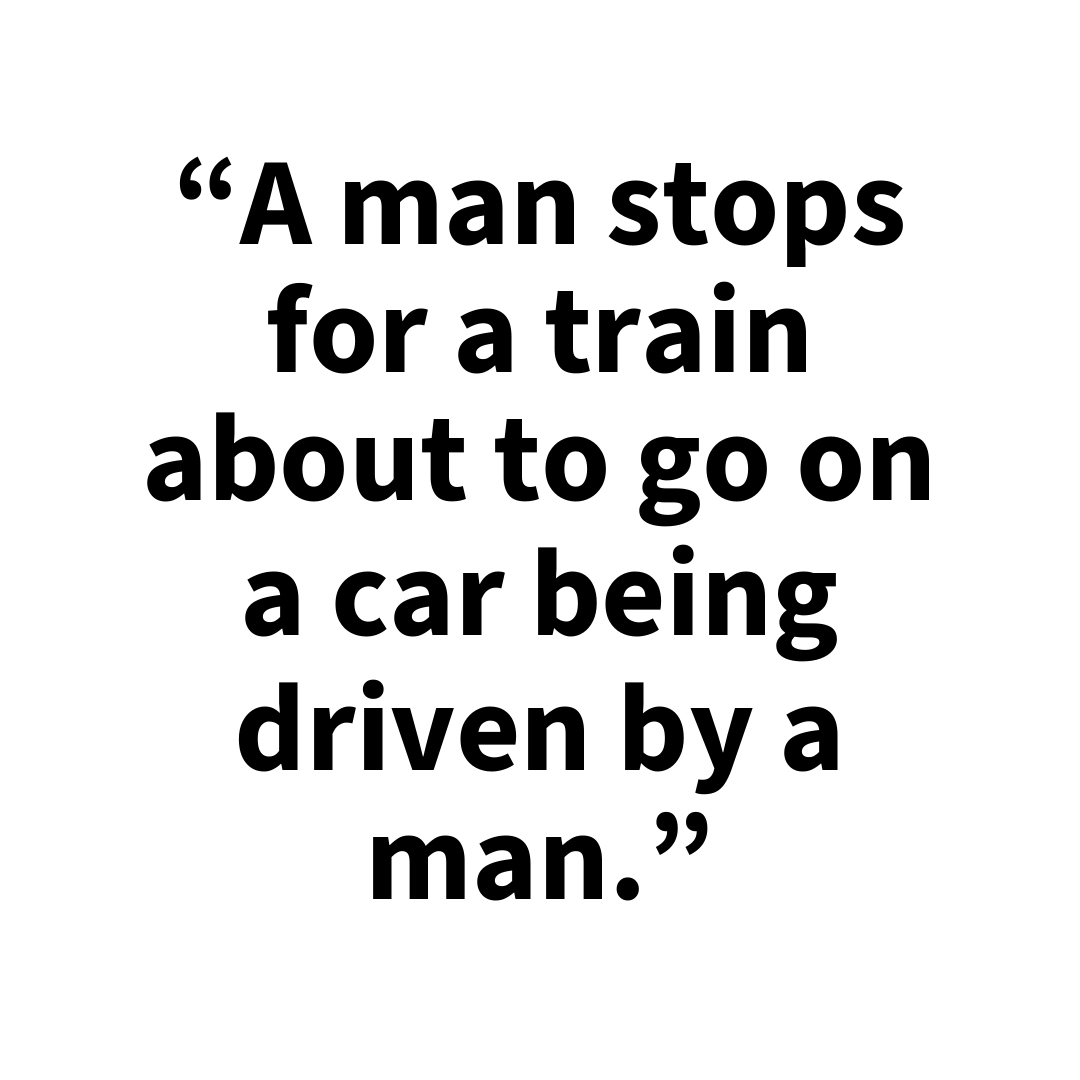} &
\includegraphics[width=0.15\textwidth]{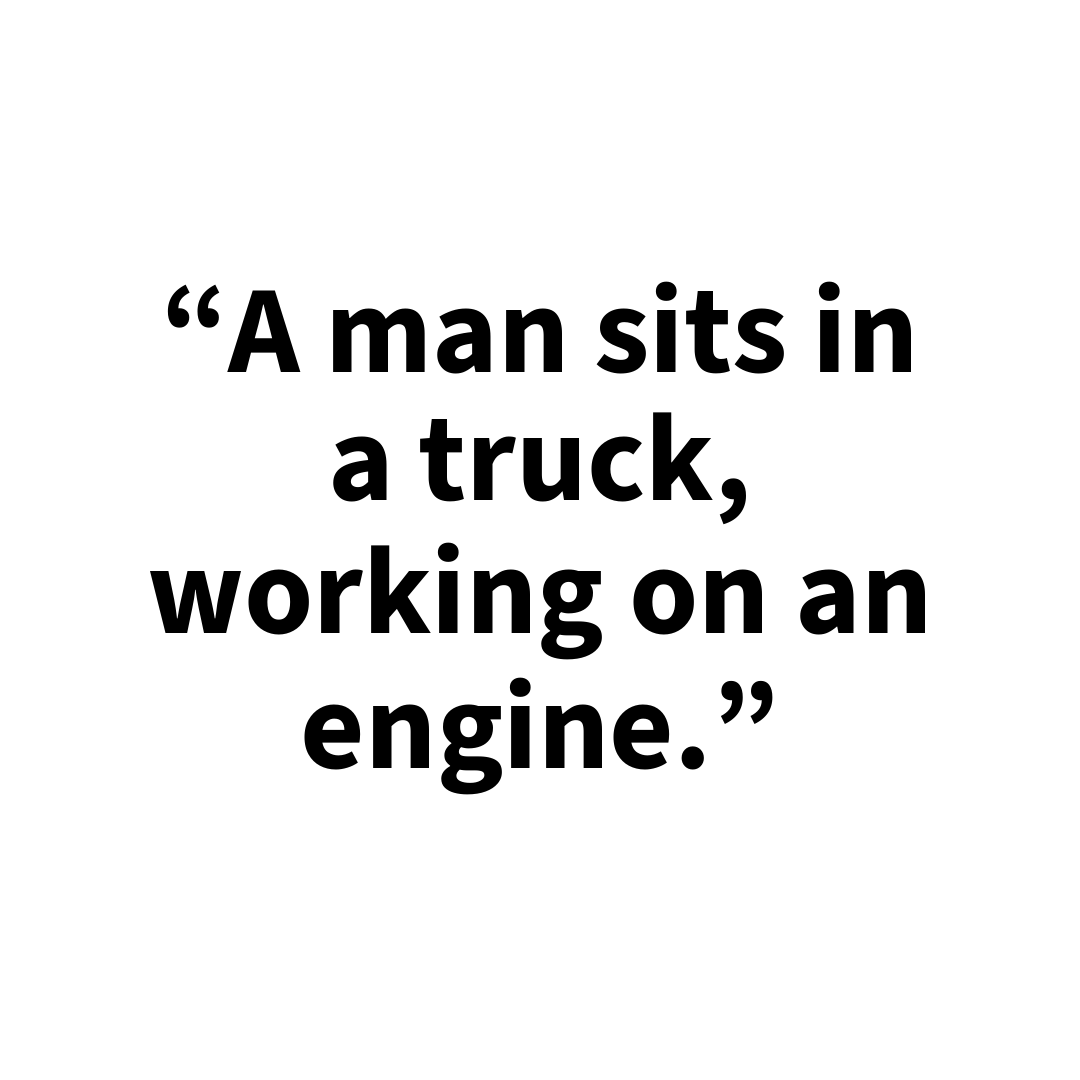} &
\includegraphics[width=0.15\textwidth]{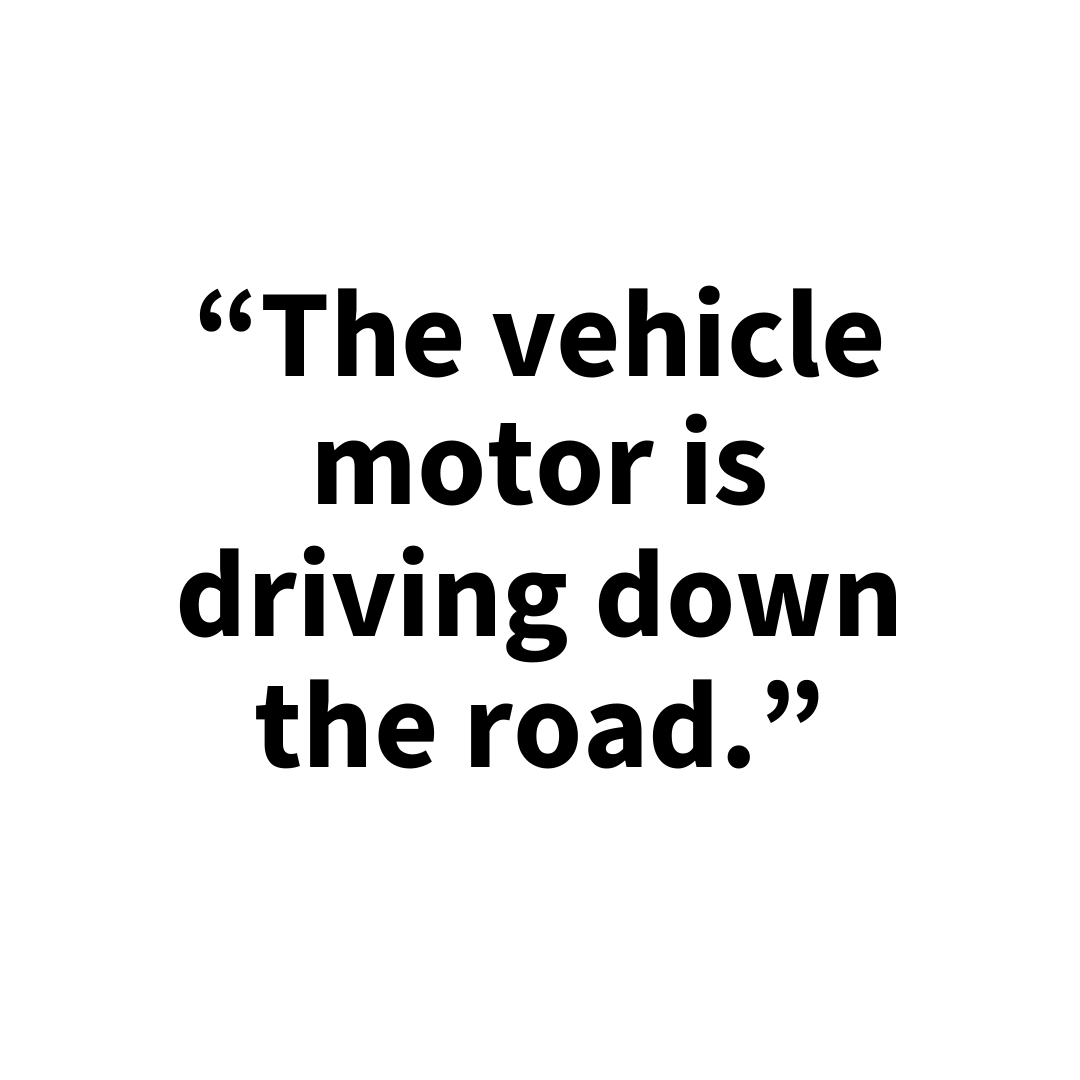} &
\includegraphics[width=0.15\textwidth]{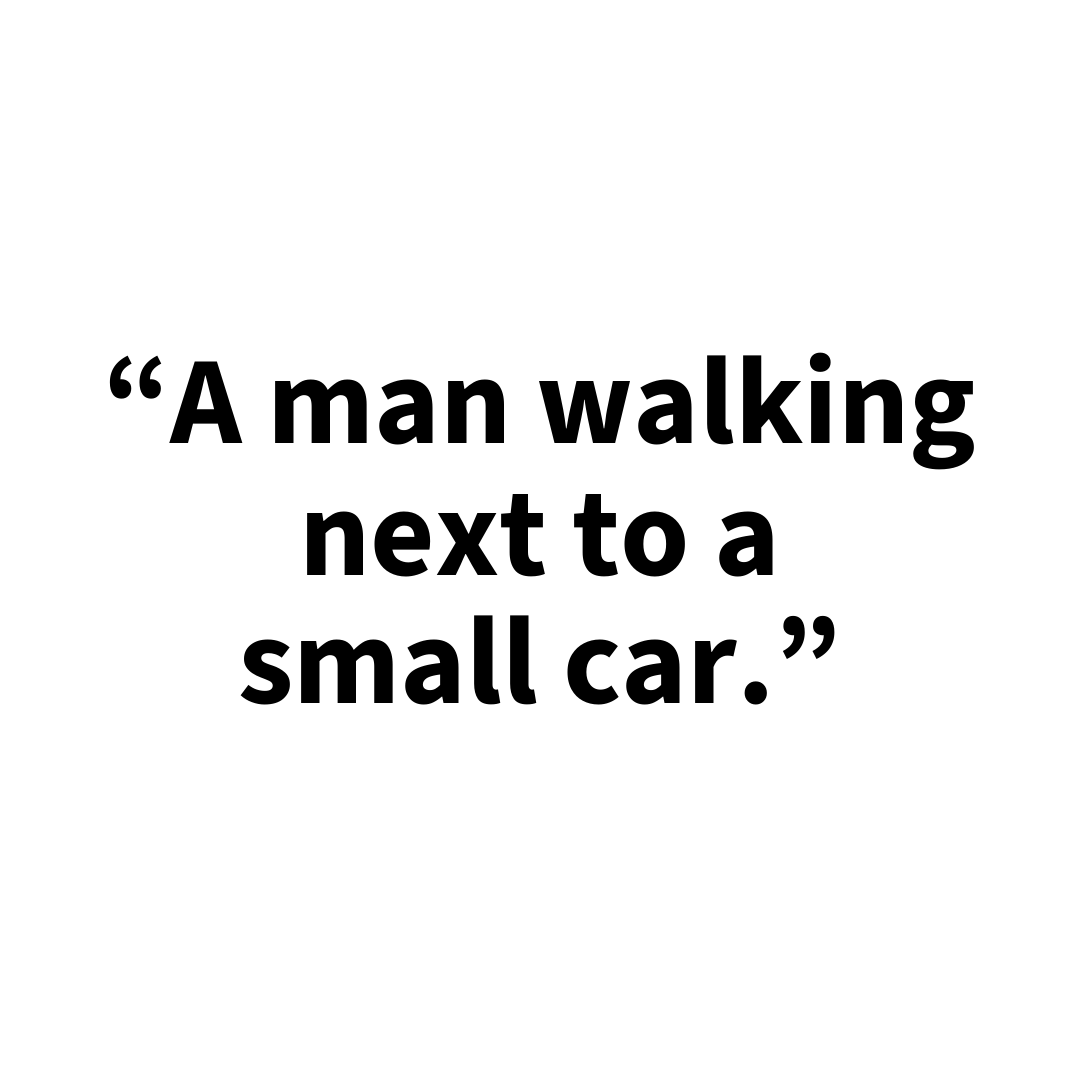} \\

\includegraphics[width=0.15\textwidth]{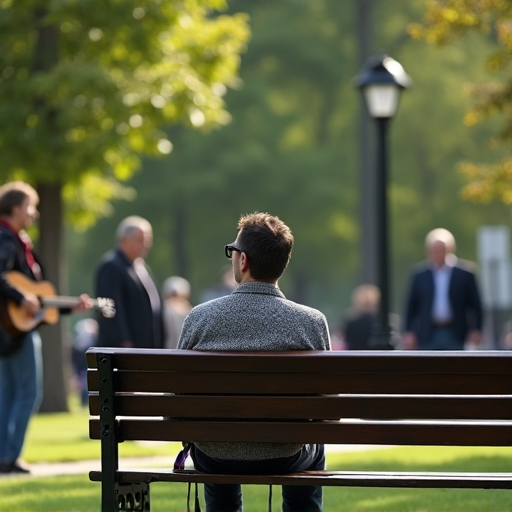} &
\includegraphics[width=0.15\textwidth]{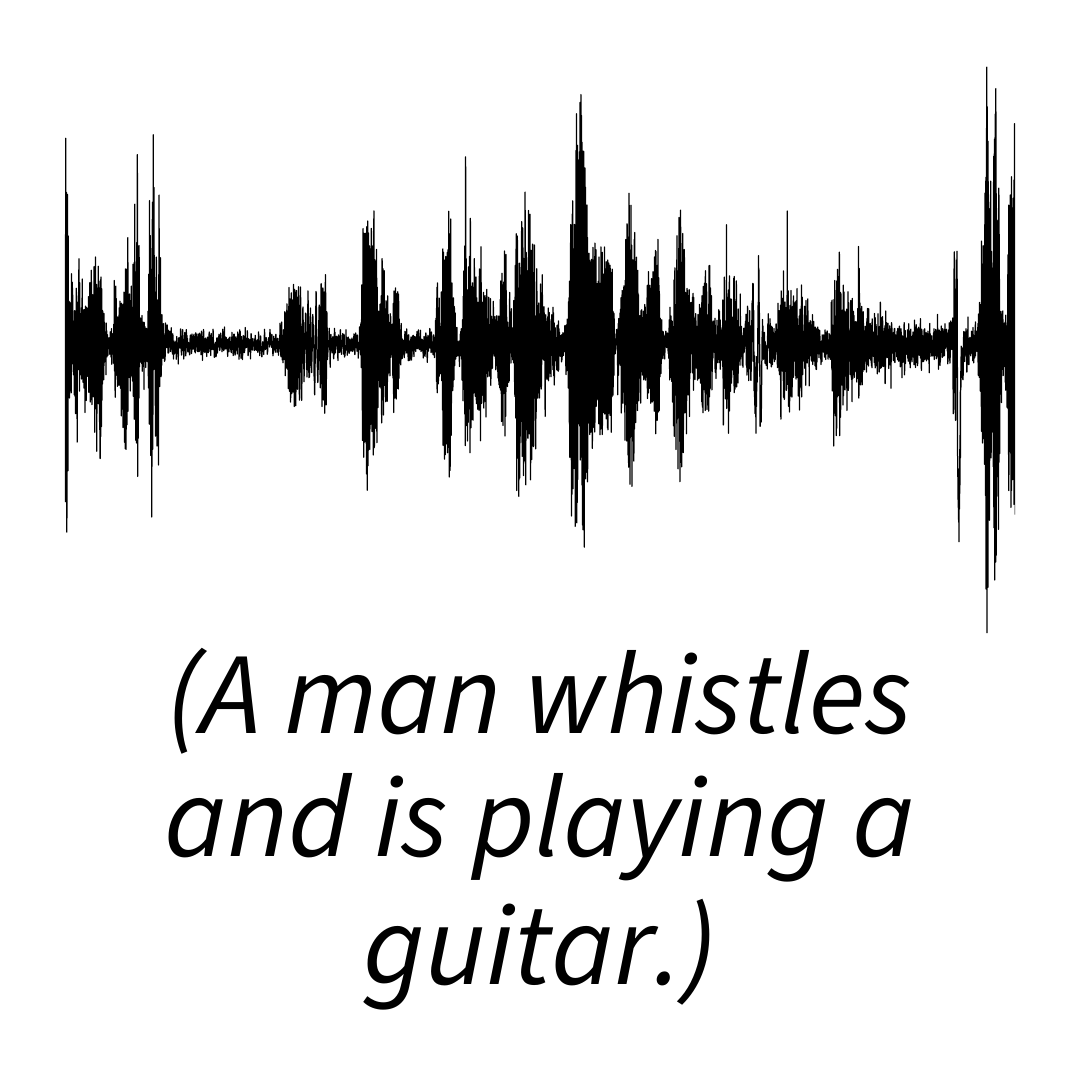} &
\includegraphics[width=0.15\textwidth]{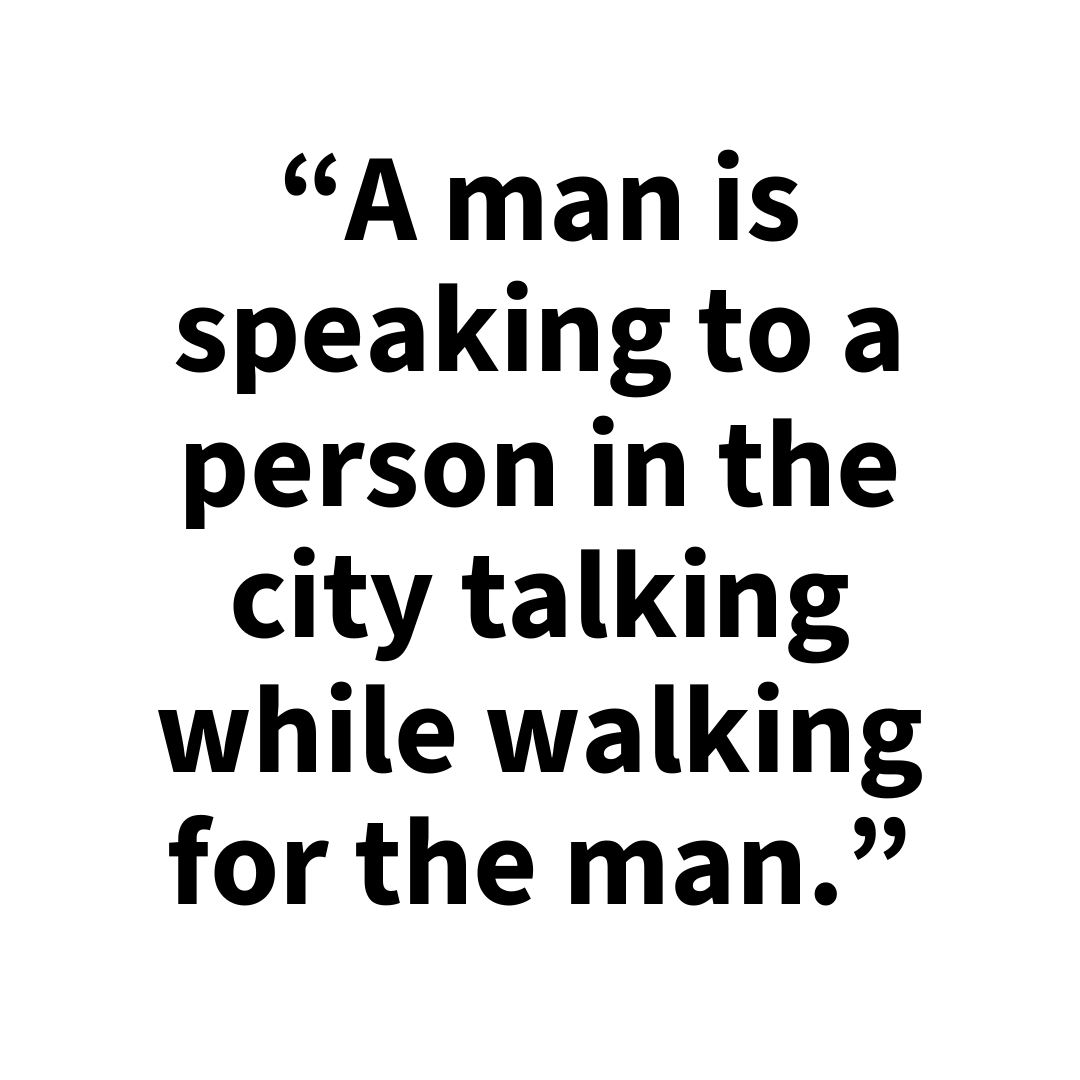} &
\includegraphics[width=0.15\textwidth]{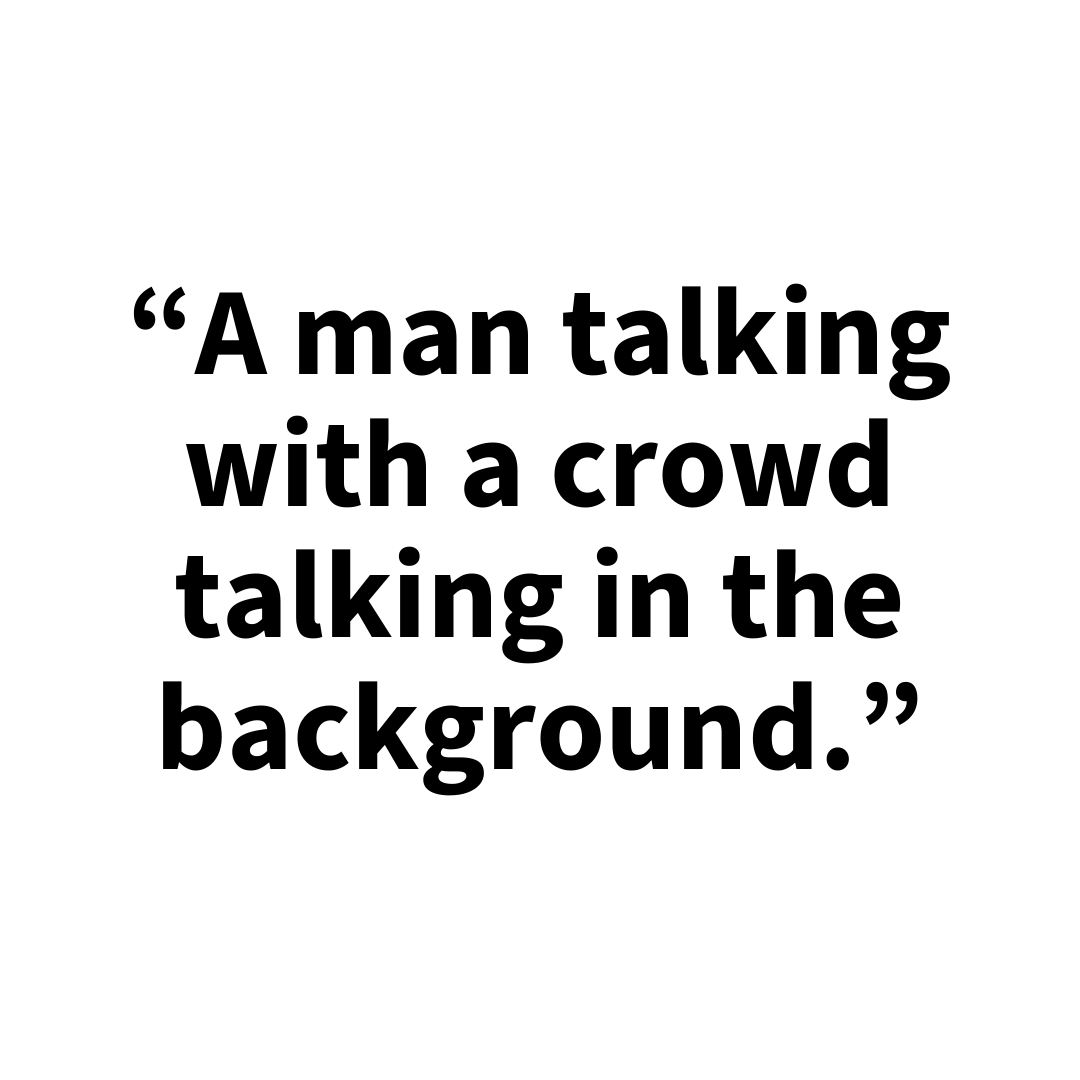} &
\includegraphics[width=0.15\textwidth]{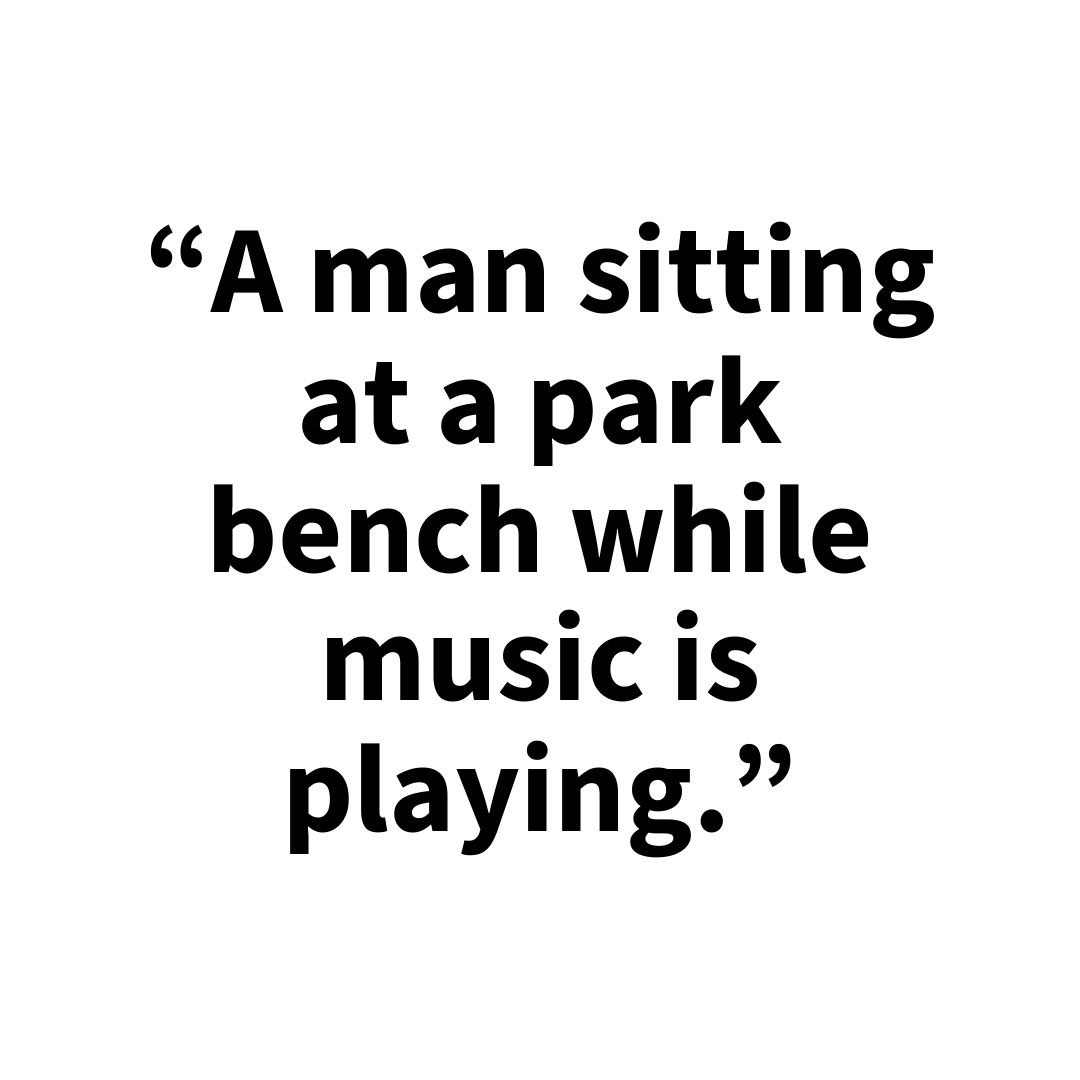} &
\includegraphics[width=0.15\textwidth]{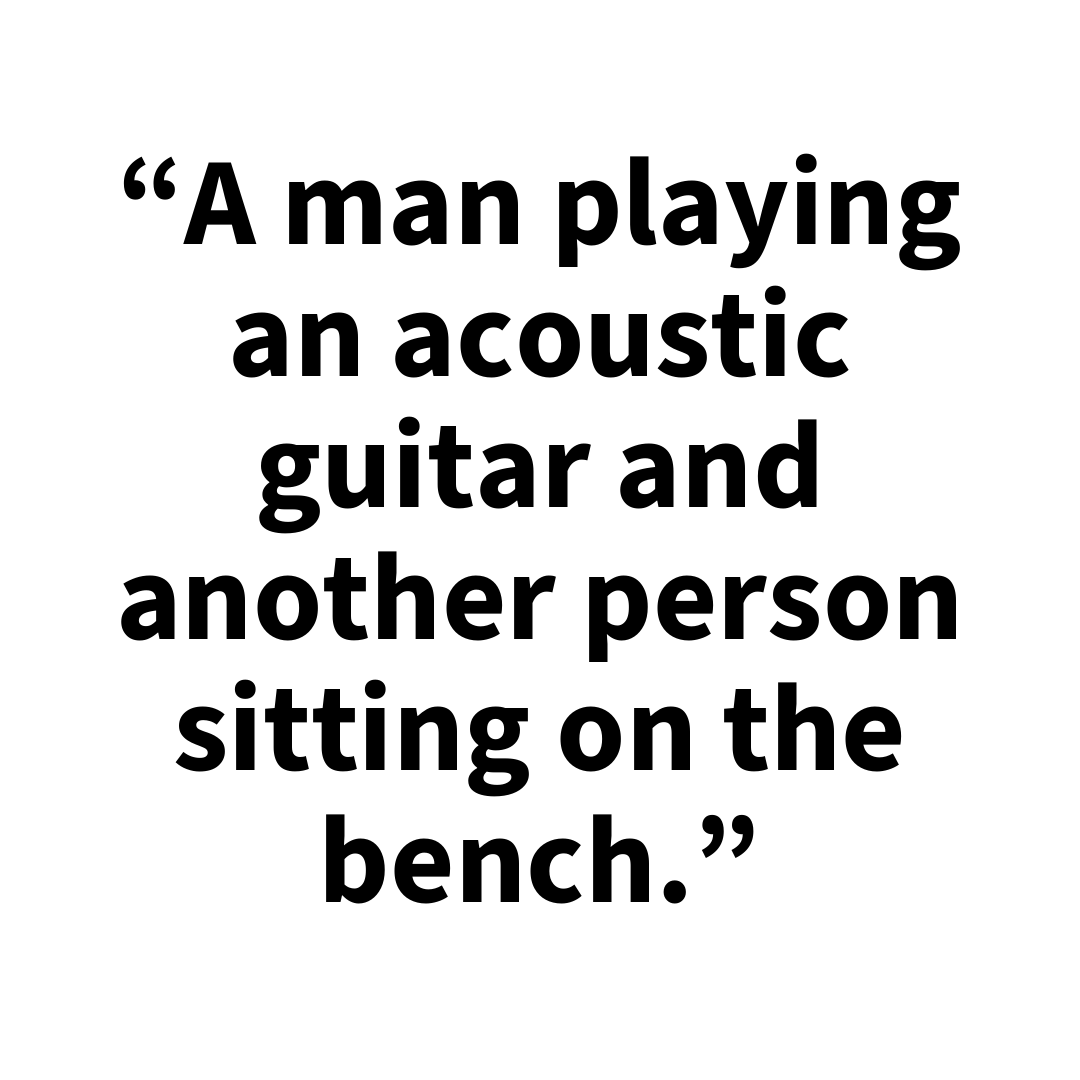} \\

\includegraphics[width=0.15\textwidth]{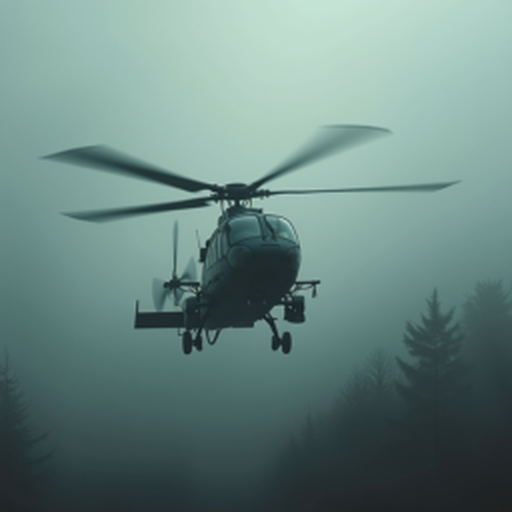} &
\includegraphics[width=0.15\textwidth]{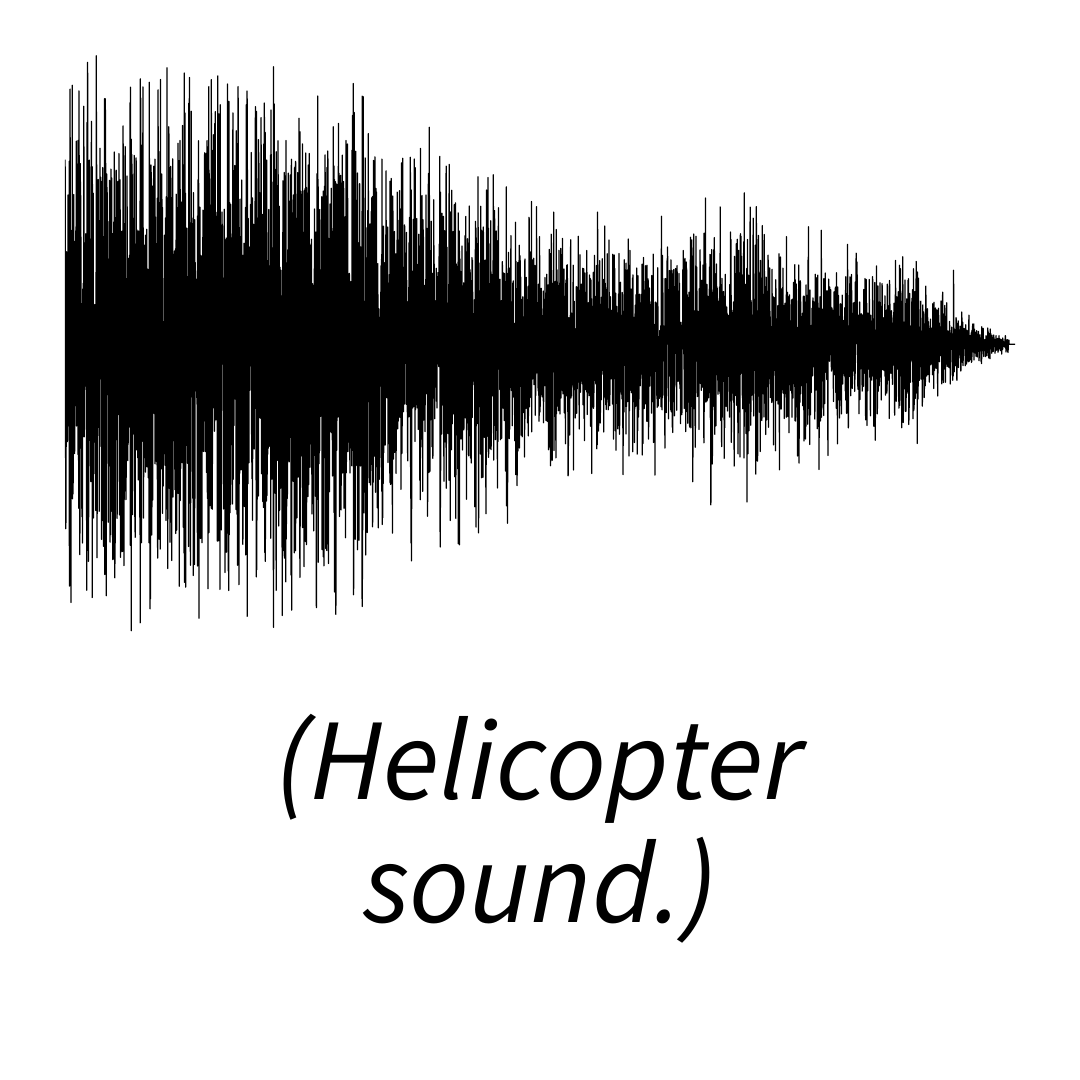} &
\includegraphics[width=0.15\textwidth]{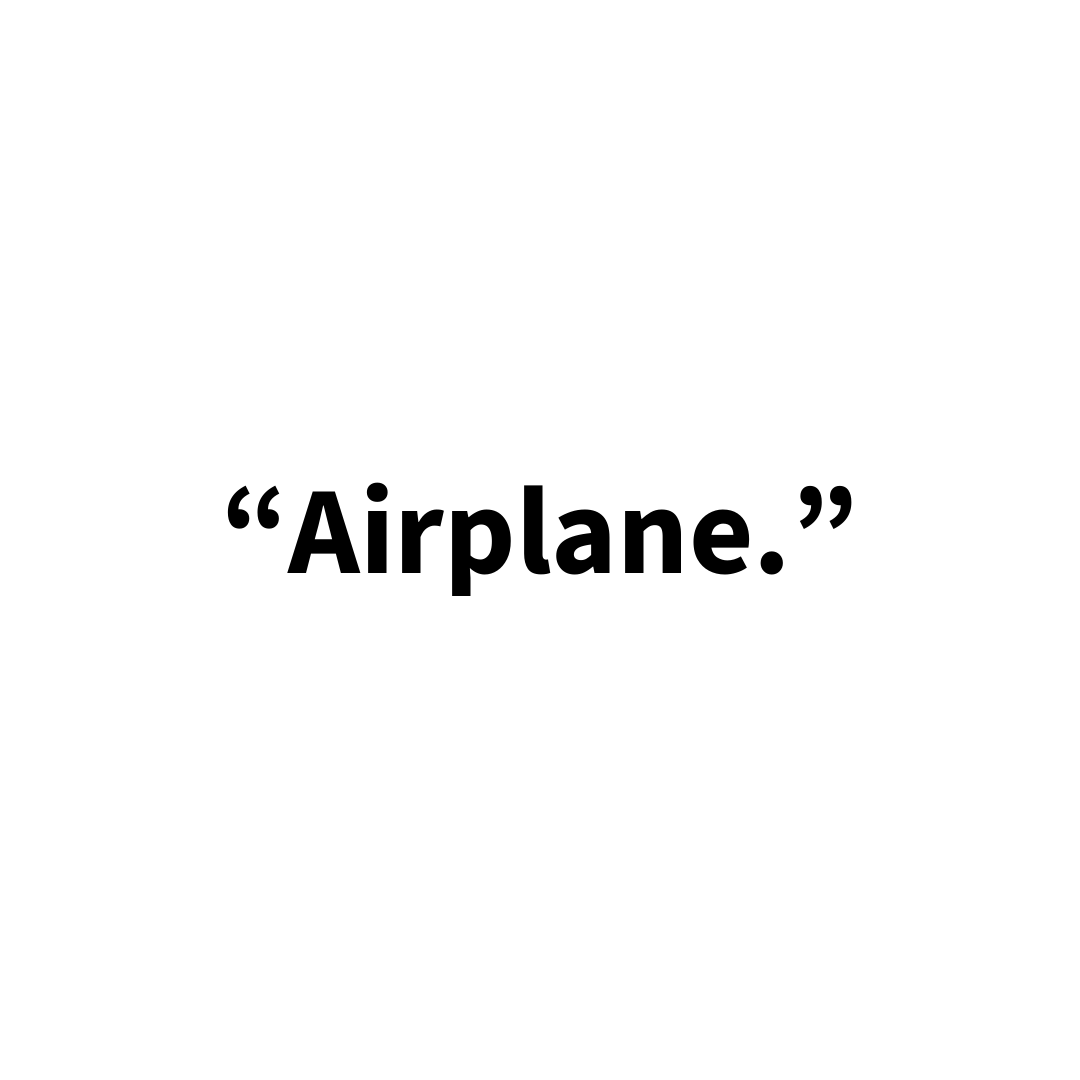} &
\includegraphics[width=0.15\textwidth]{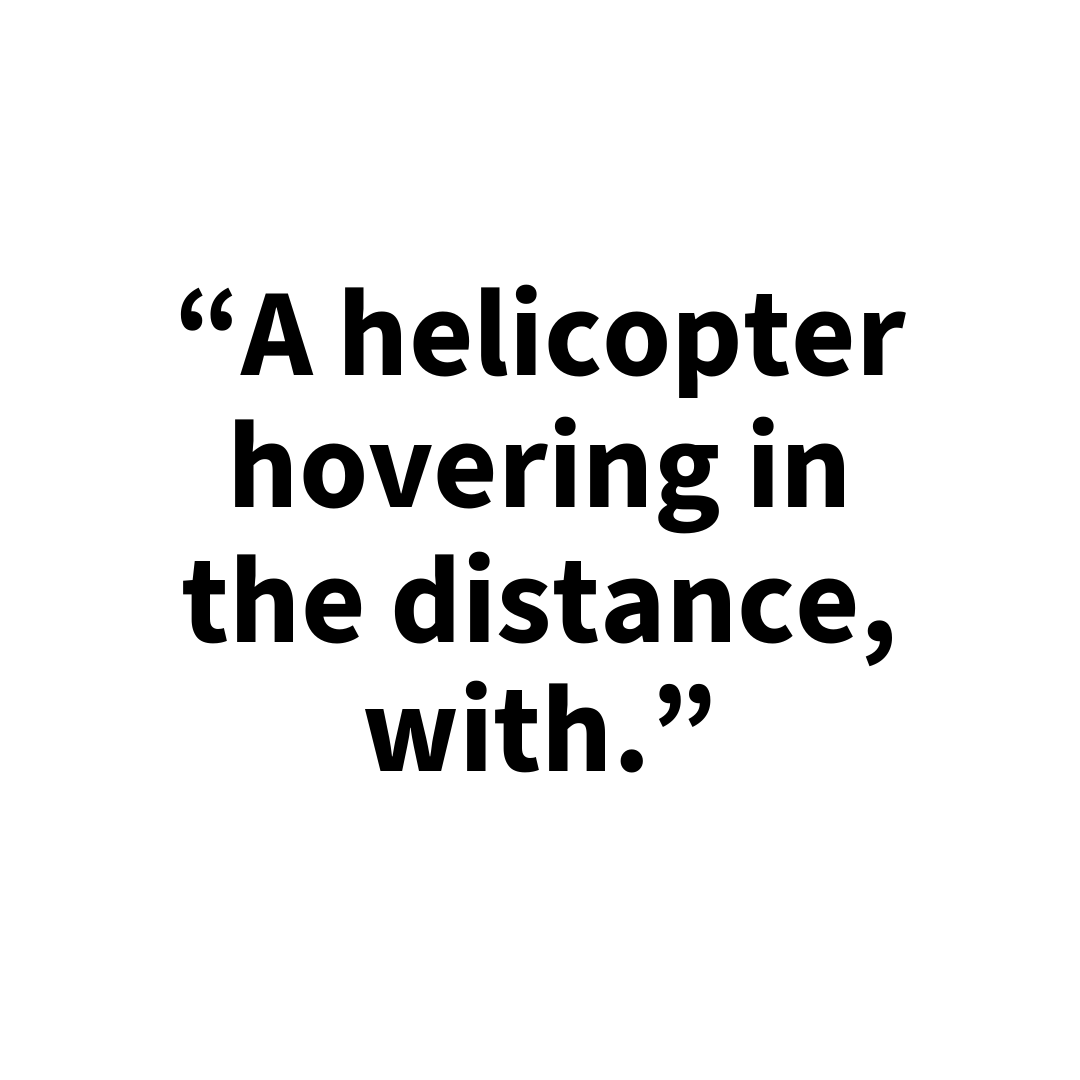} &
\includegraphics[width=0.15\textwidth]{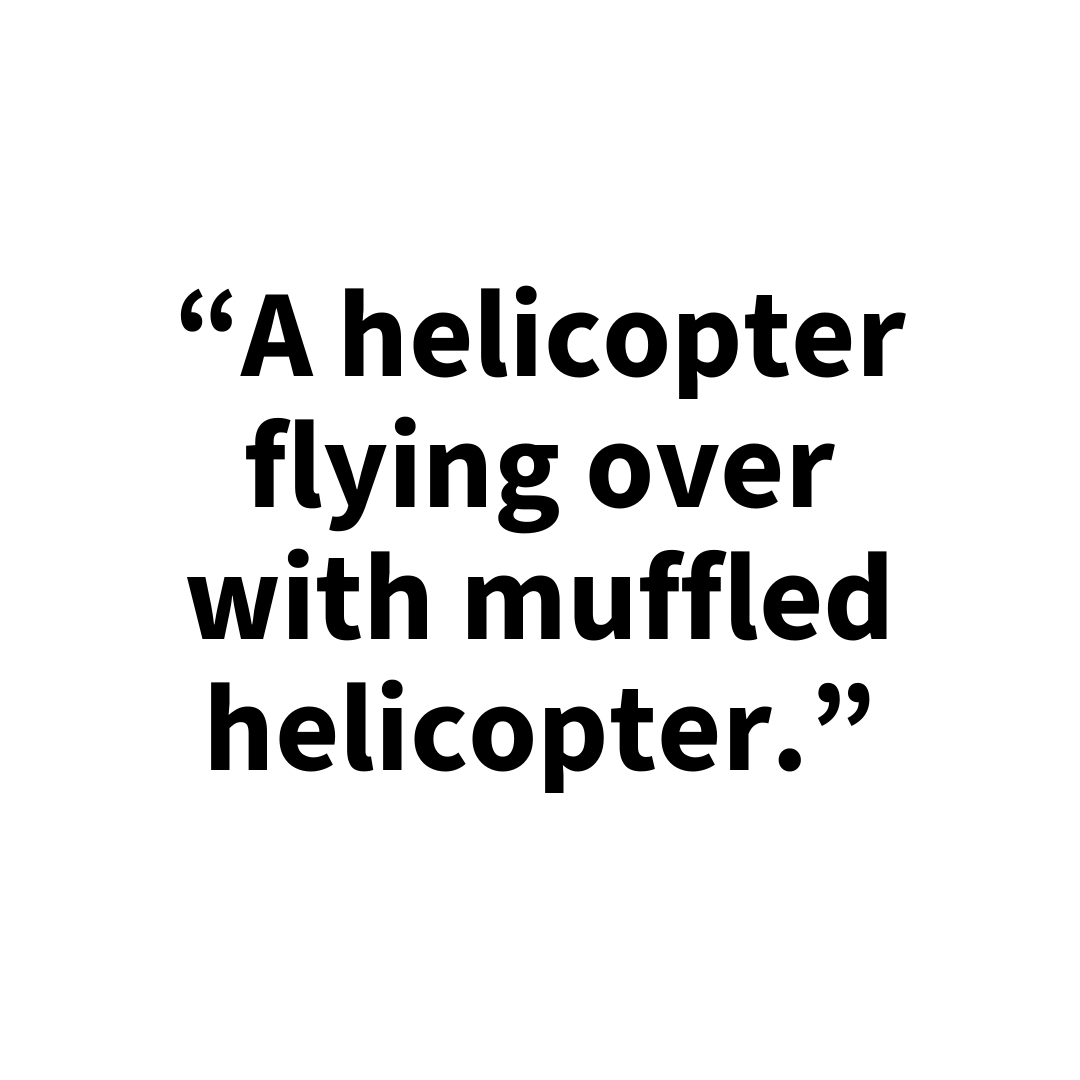} &
\includegraphics[width=0.15\textwidth]{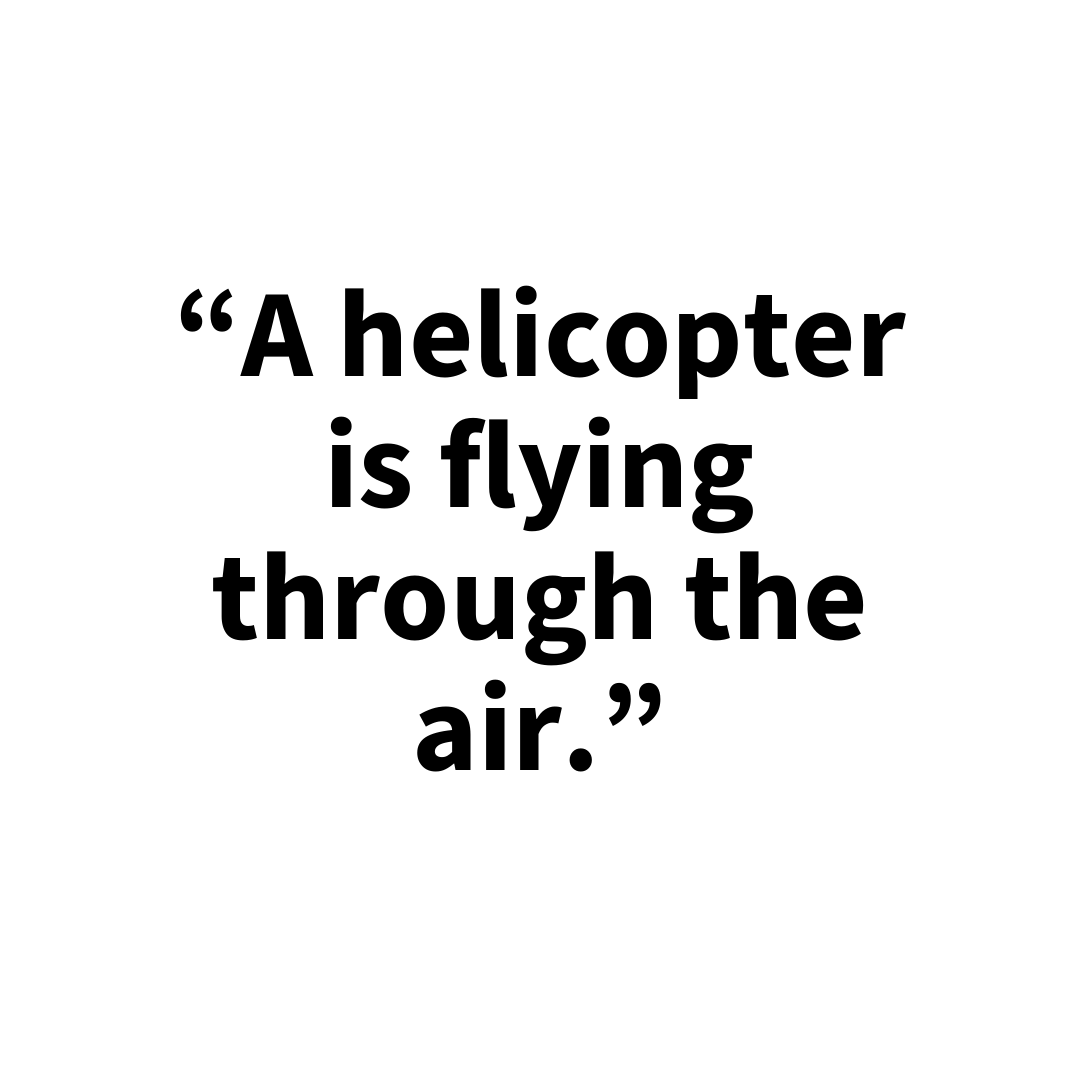} \\

\includegraphics[width=0.15\textwidth]{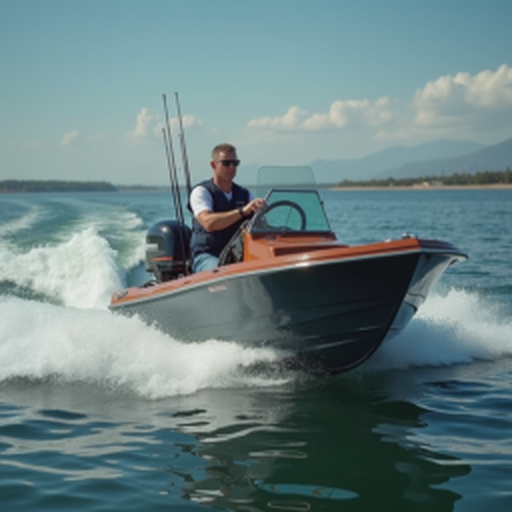} &
\includegraphics[width=0.15\textwidth]{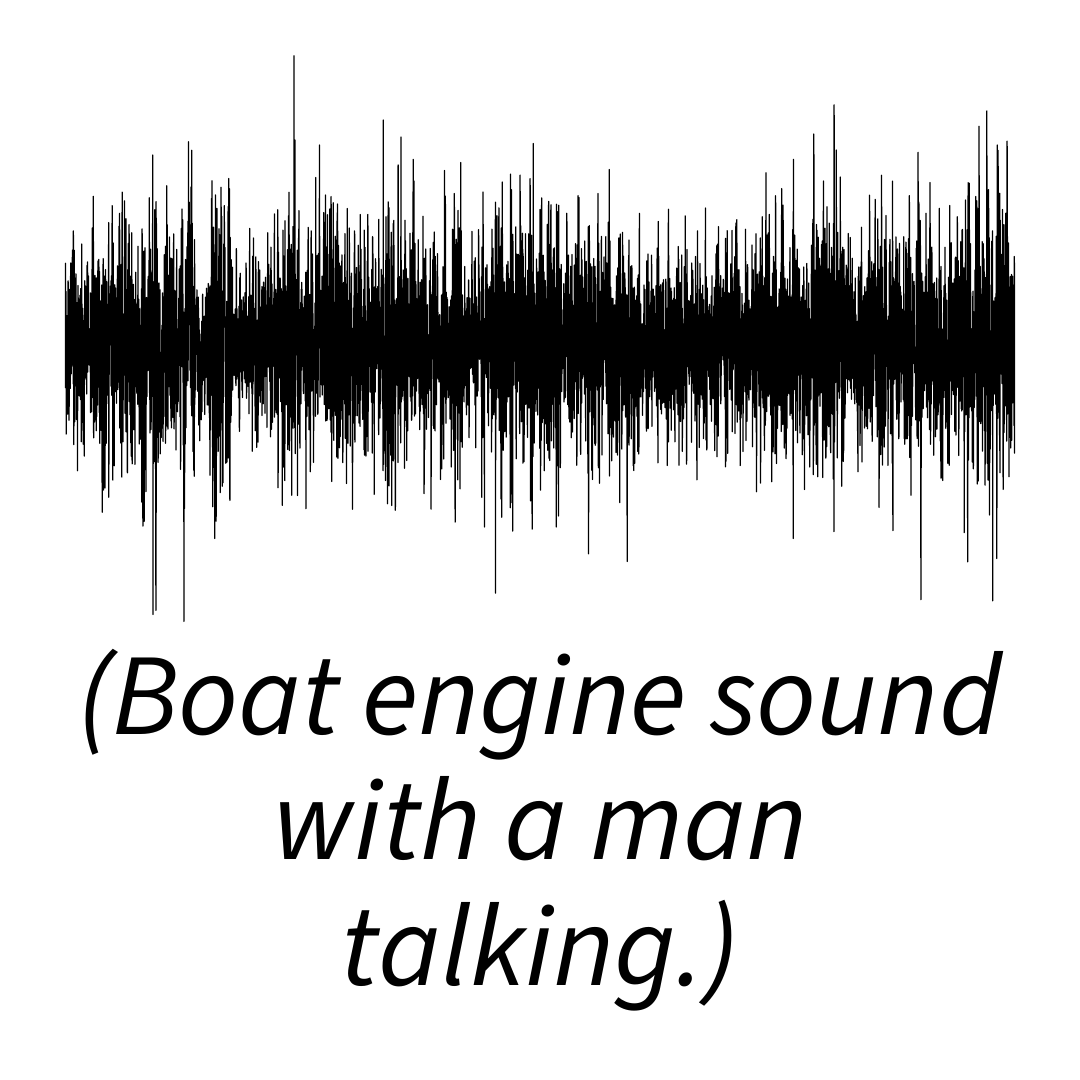} &
\includegraphics[width=0.15\textwidth]{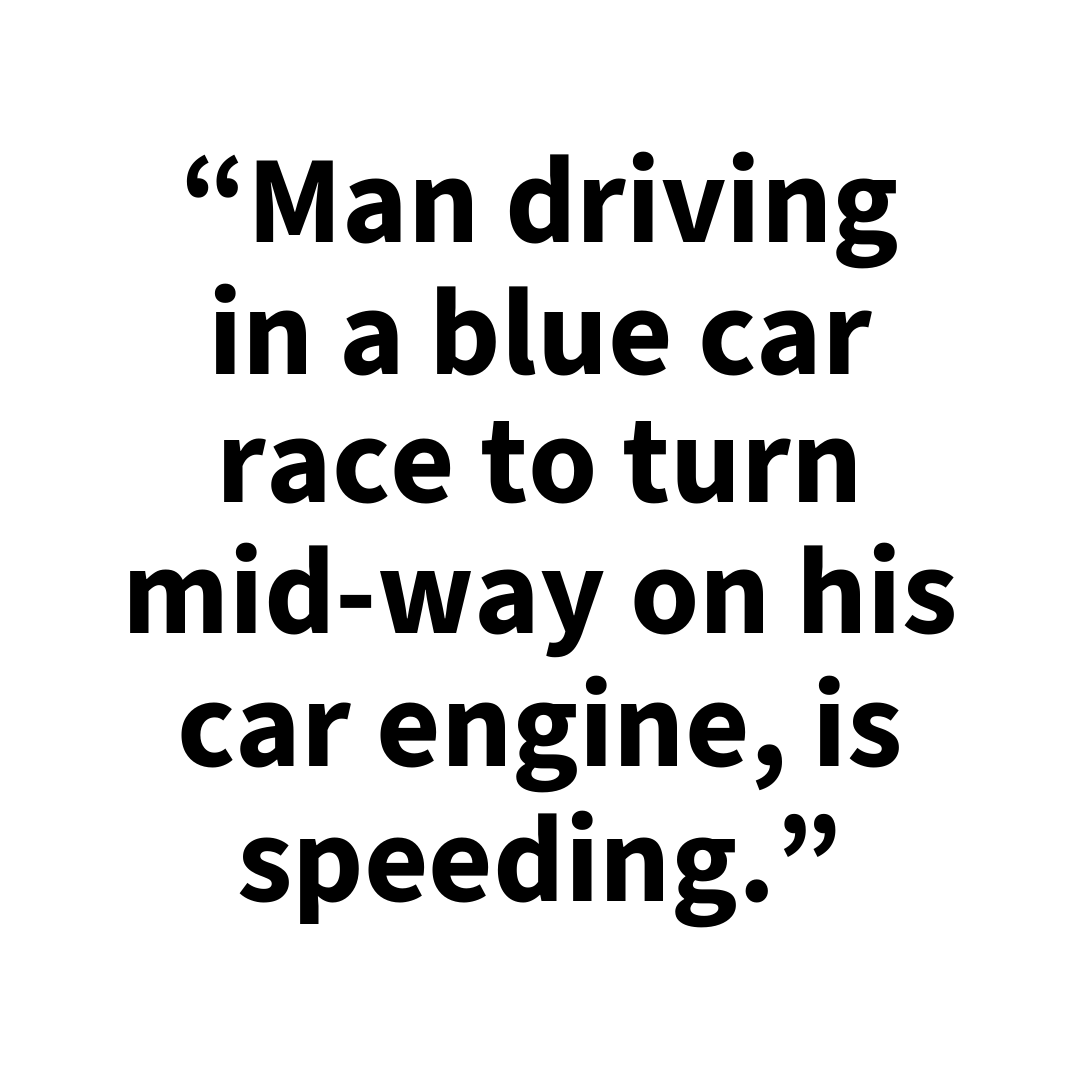} &
\includegraphics[width=0.15\textwidth]{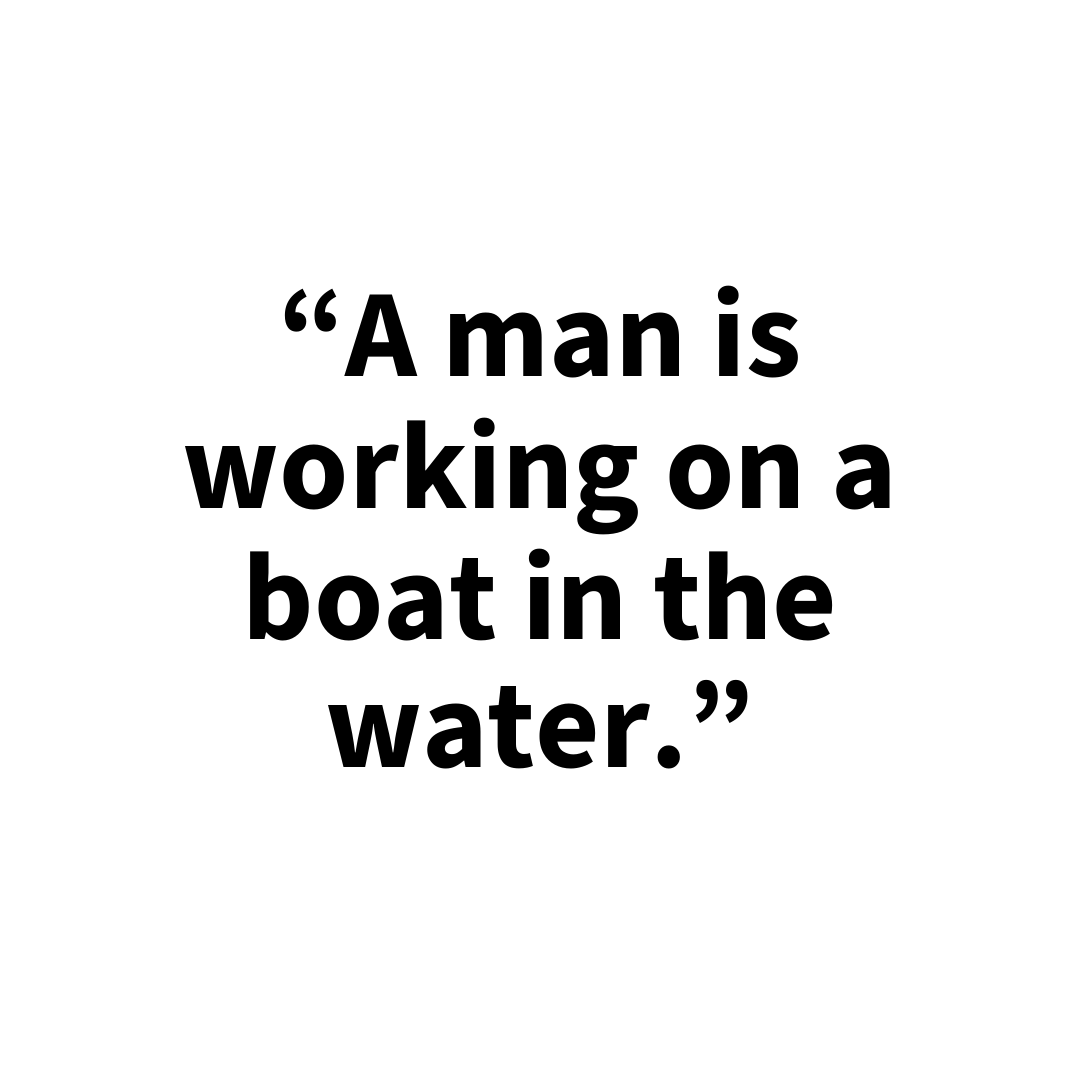} &
\includegraphics[width=0.15\textwidth]{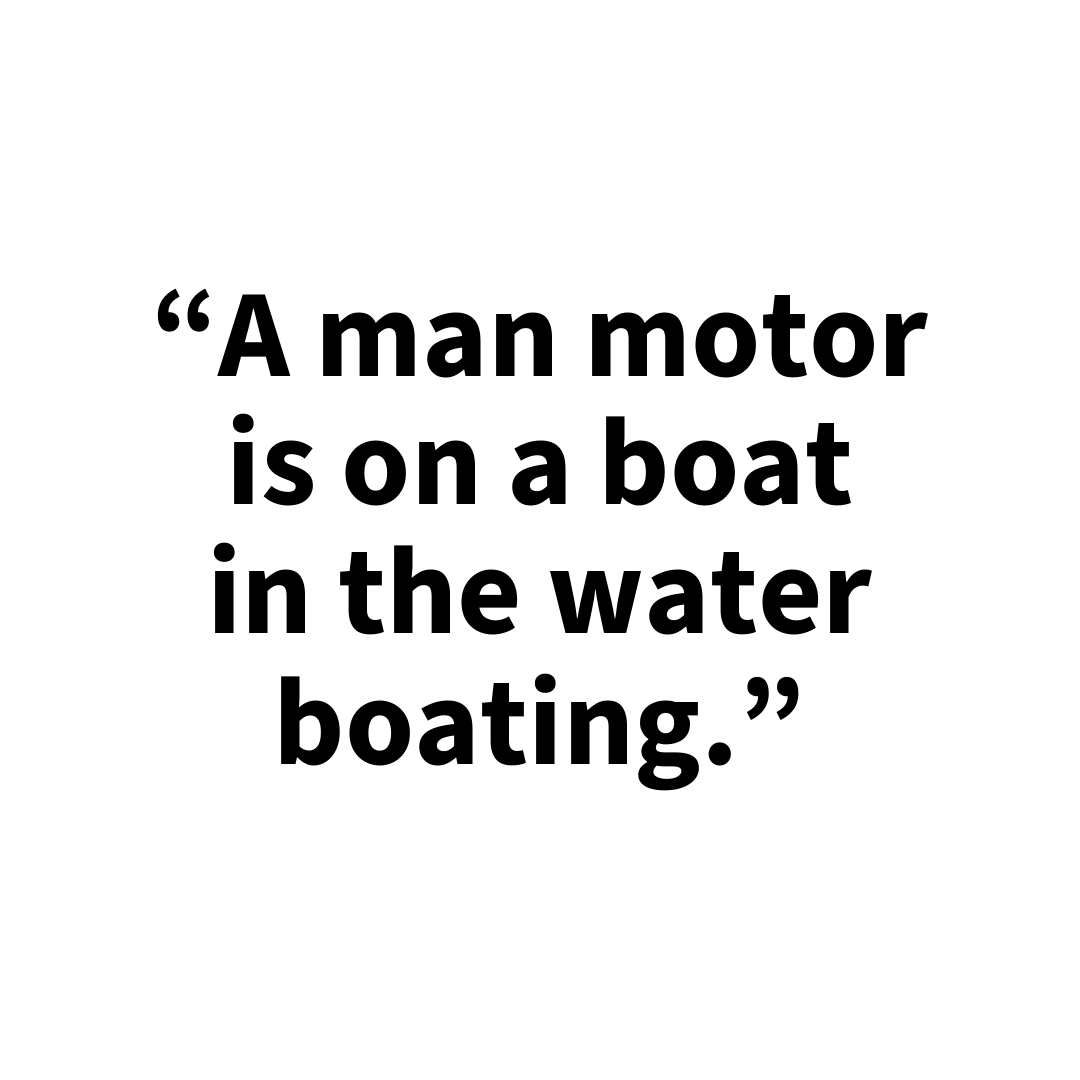} &
\includegraphics[width=0.15\textwidth]{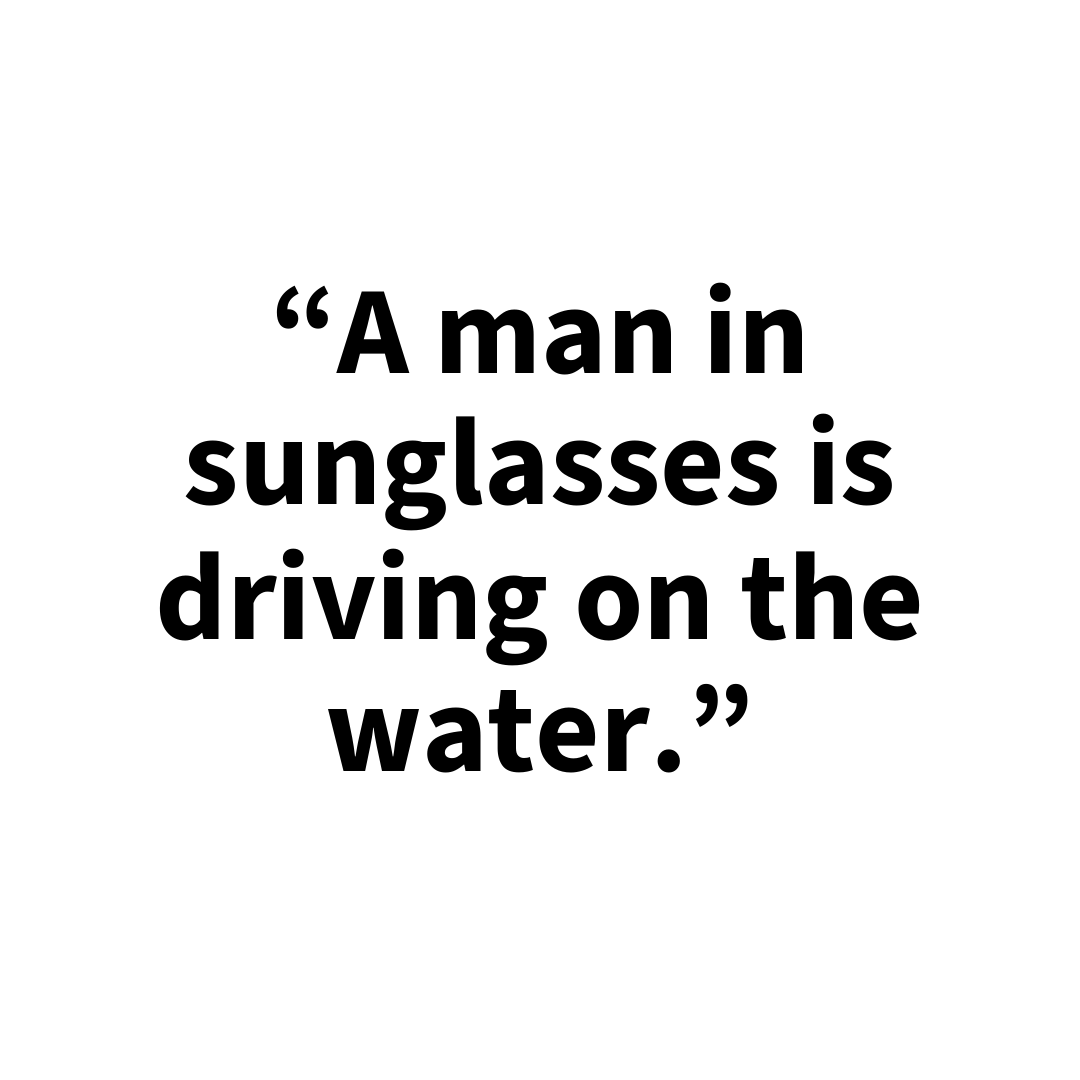} \\

\includegraphics[width=0.15\textwidth]{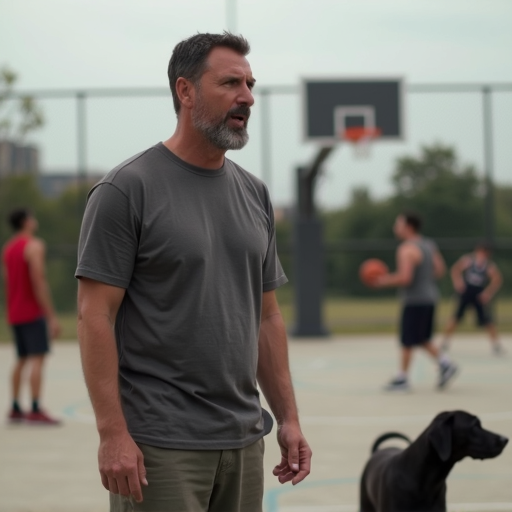} &
\includegraphics[width=0.15\textwidth]{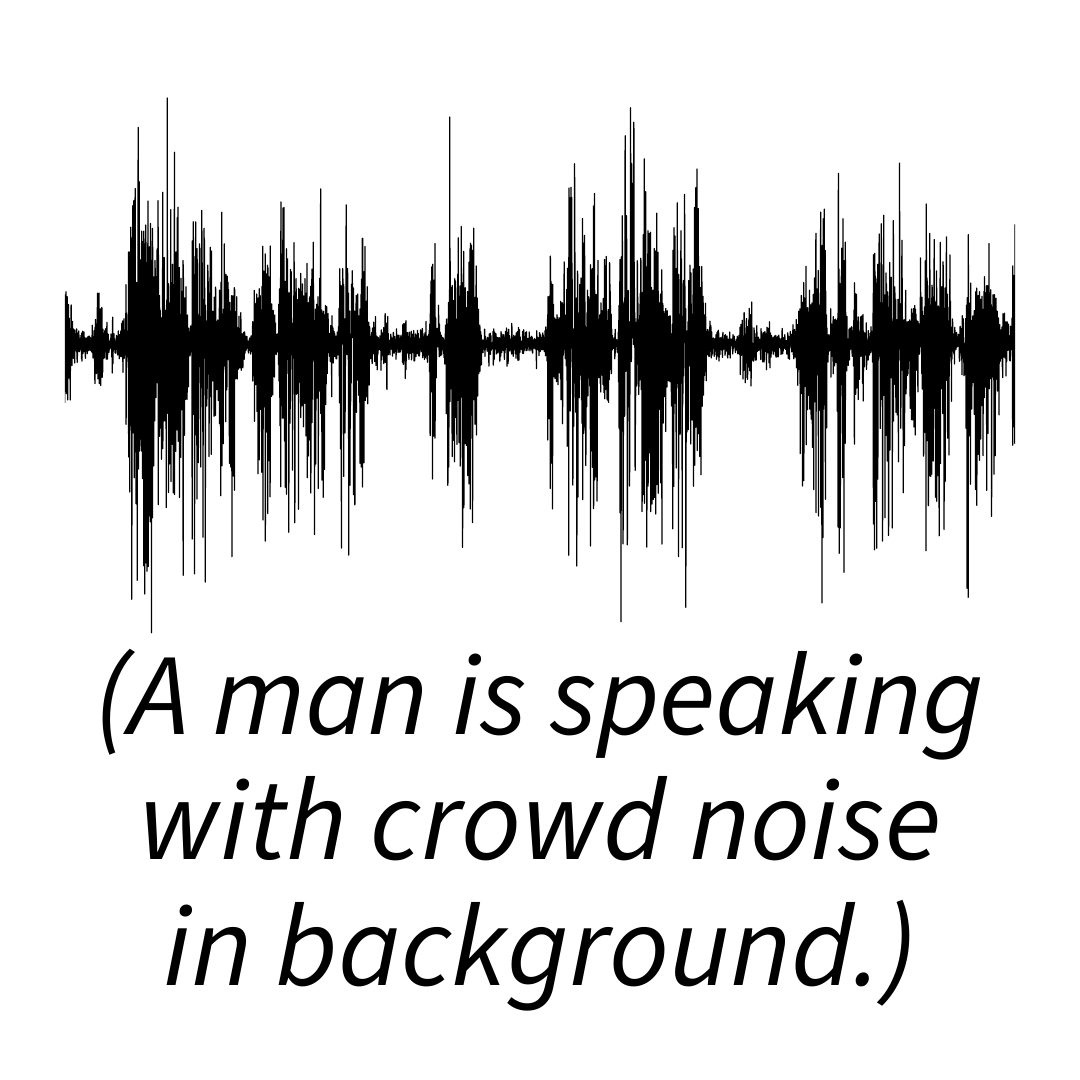} &
\includegraphics[width=0.15\textwidth]{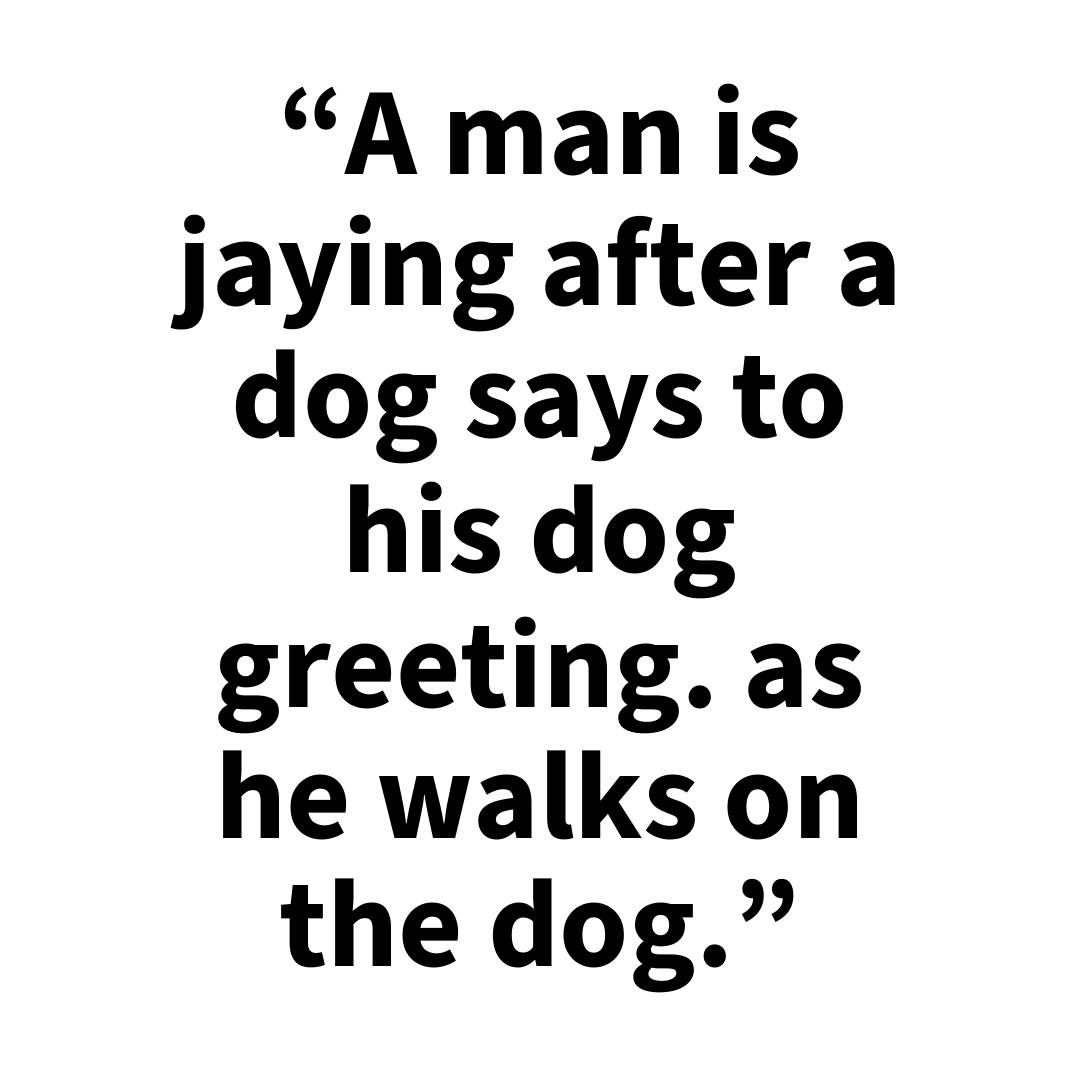} &
\includegraphics[width=0.15\textwidth]{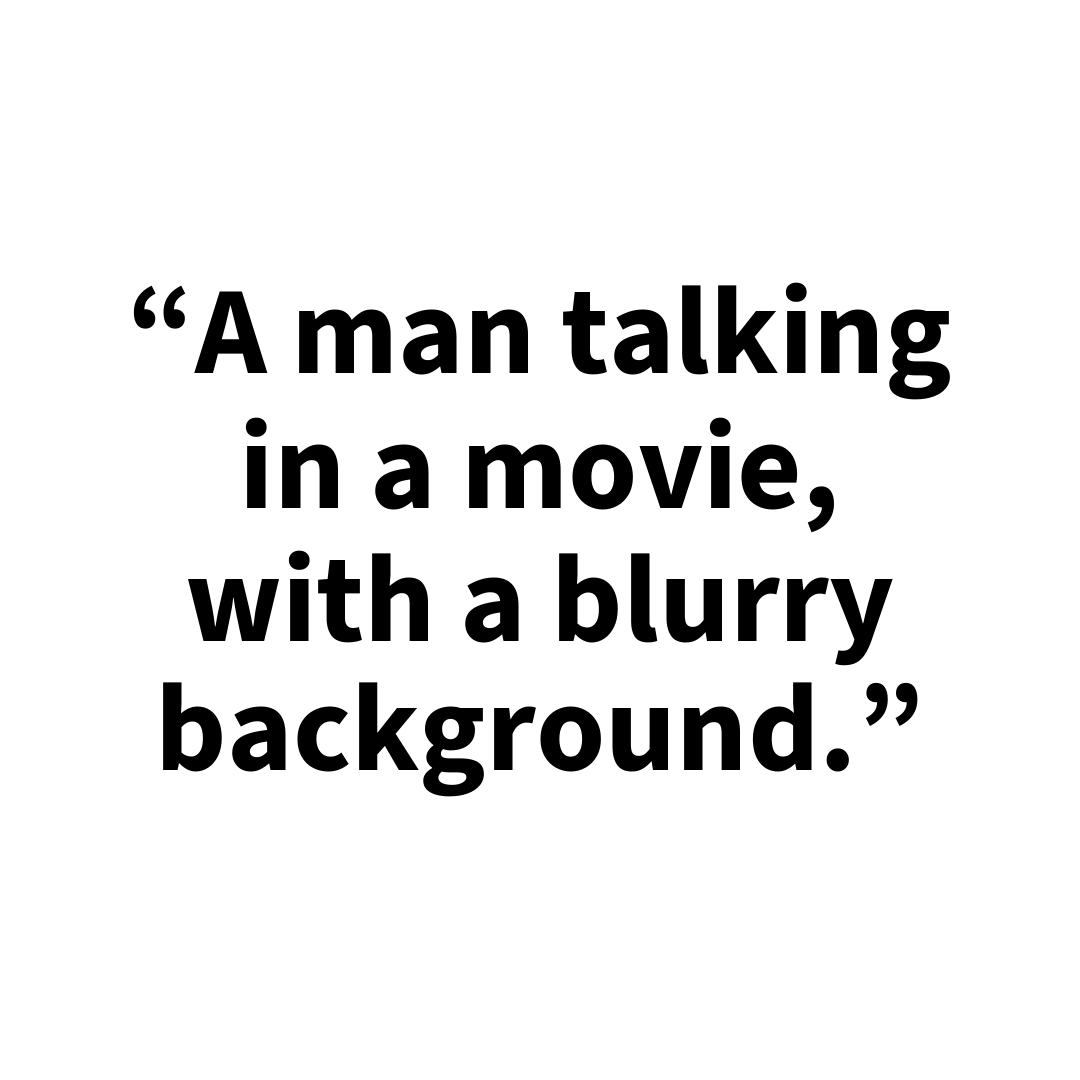} &
\includegraphics[width=0.15\textwidth]{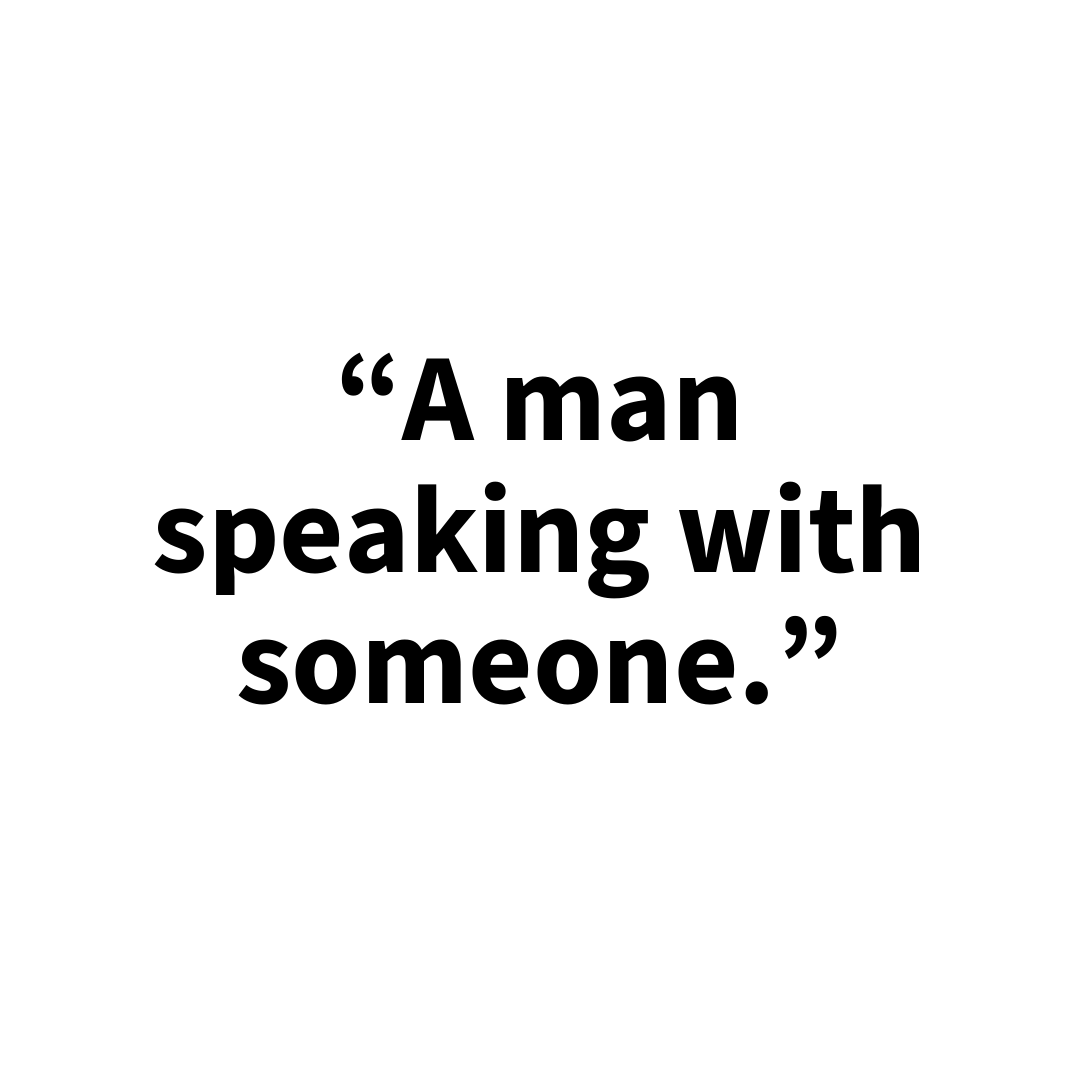} &
\includegraphics[width=0.15\textwidth]{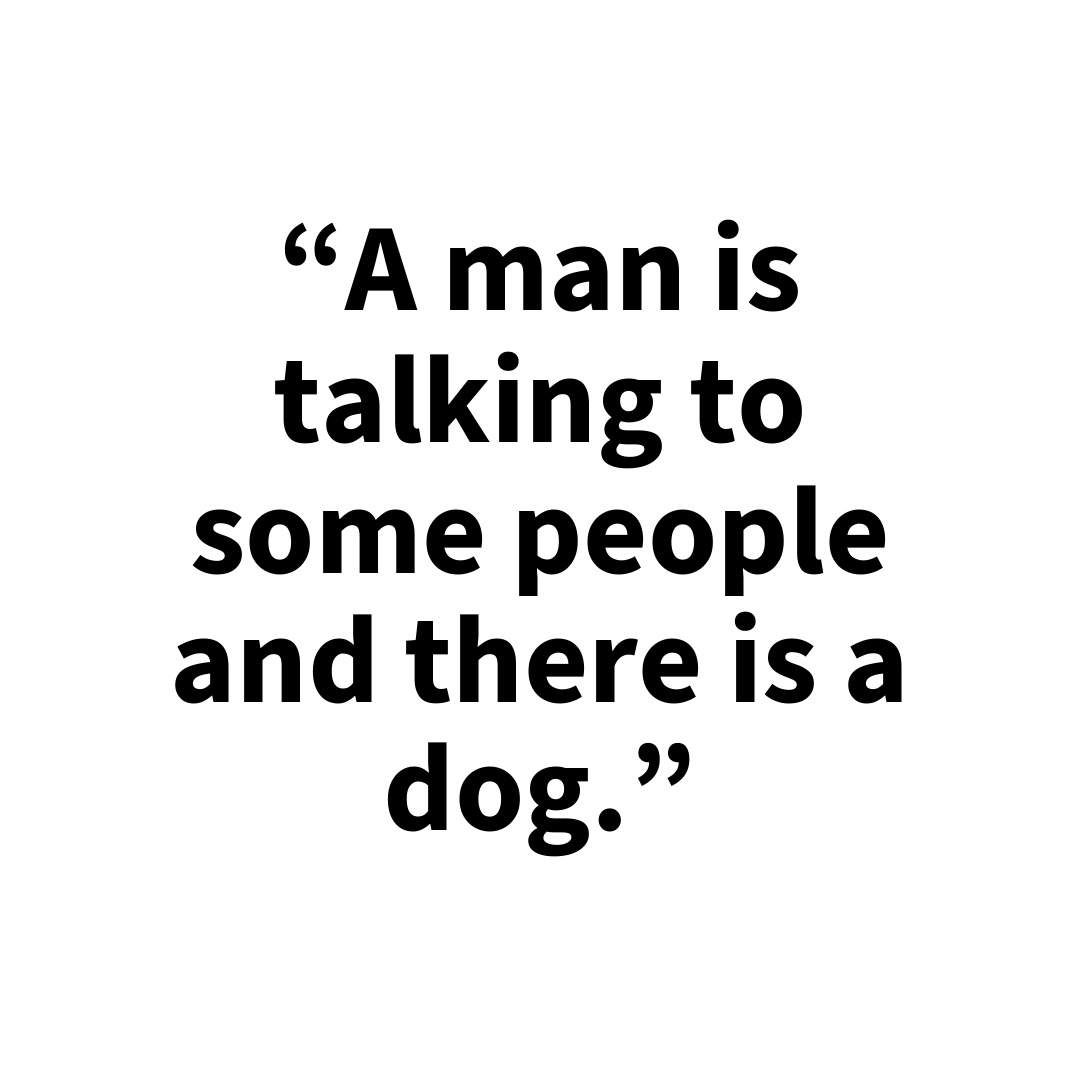} \\

\includegraphics[width=0.15\textwidth]{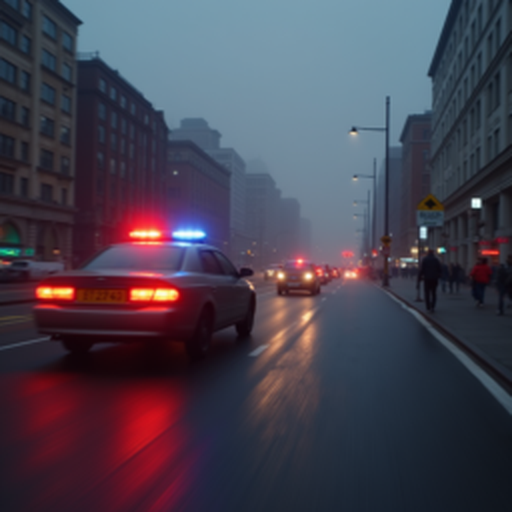} &
\includegraphics[width=0.15\textwidth]{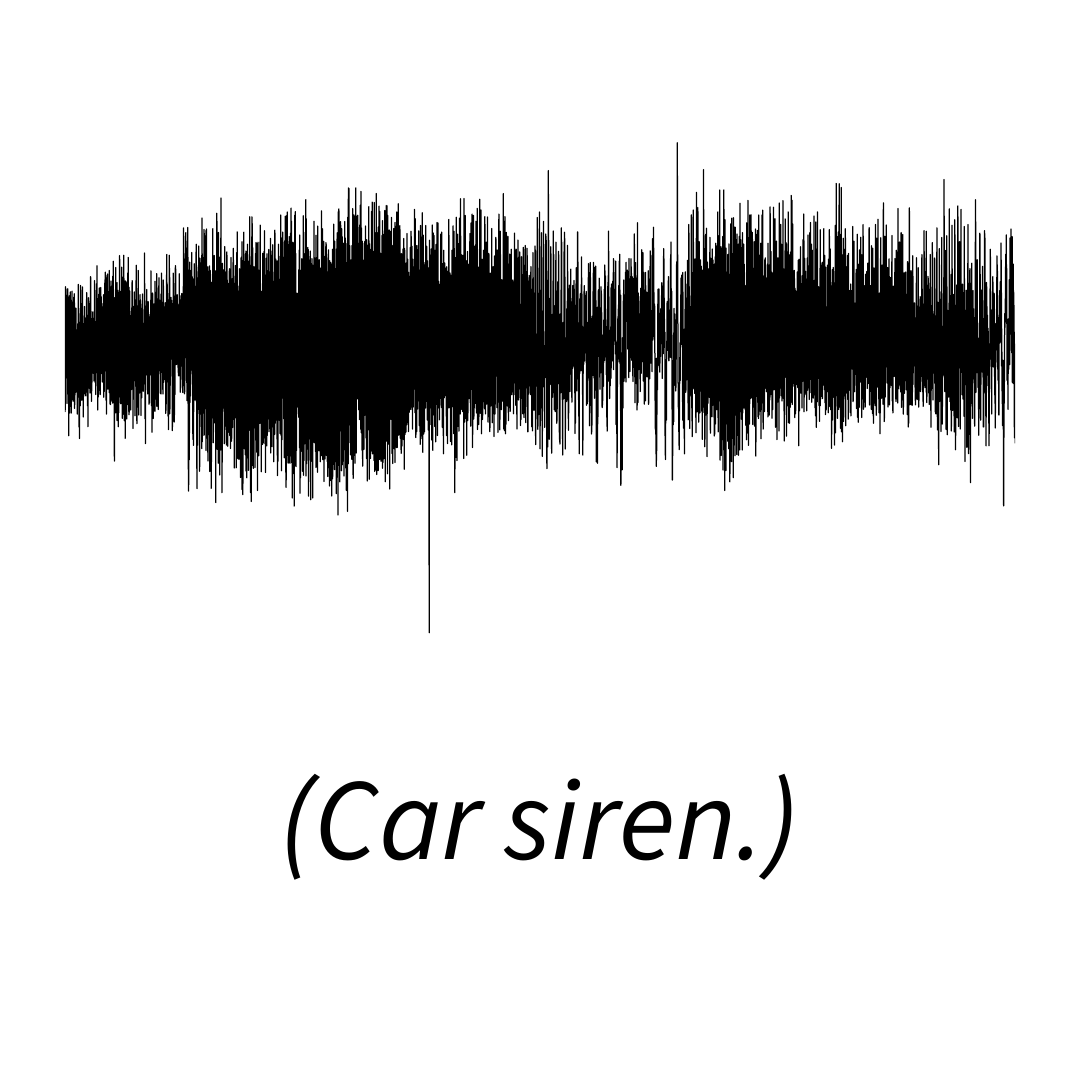} &
\includegraphics[width=0.15\textwidth]{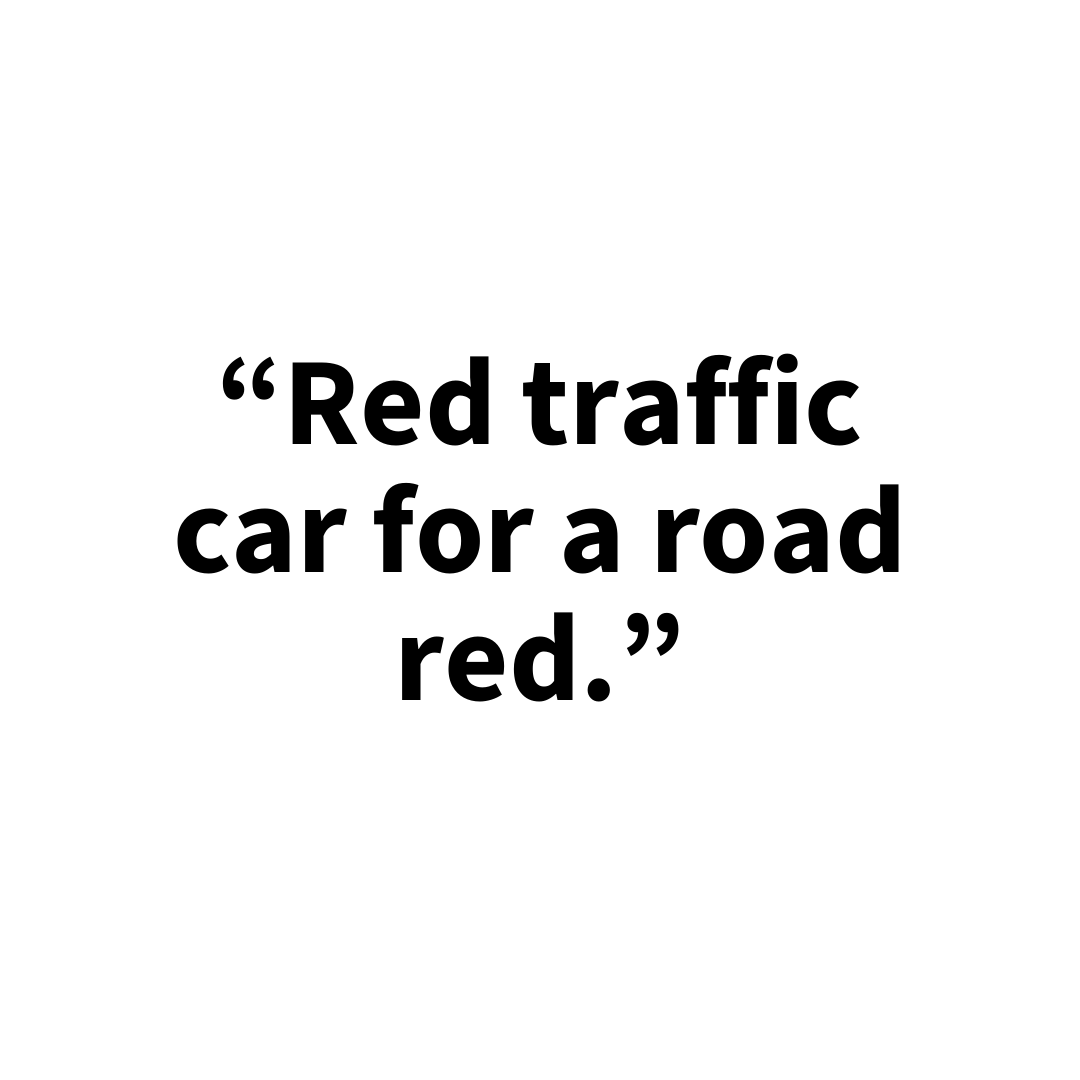} &
\includegraphics[width=0.15\textwidth]{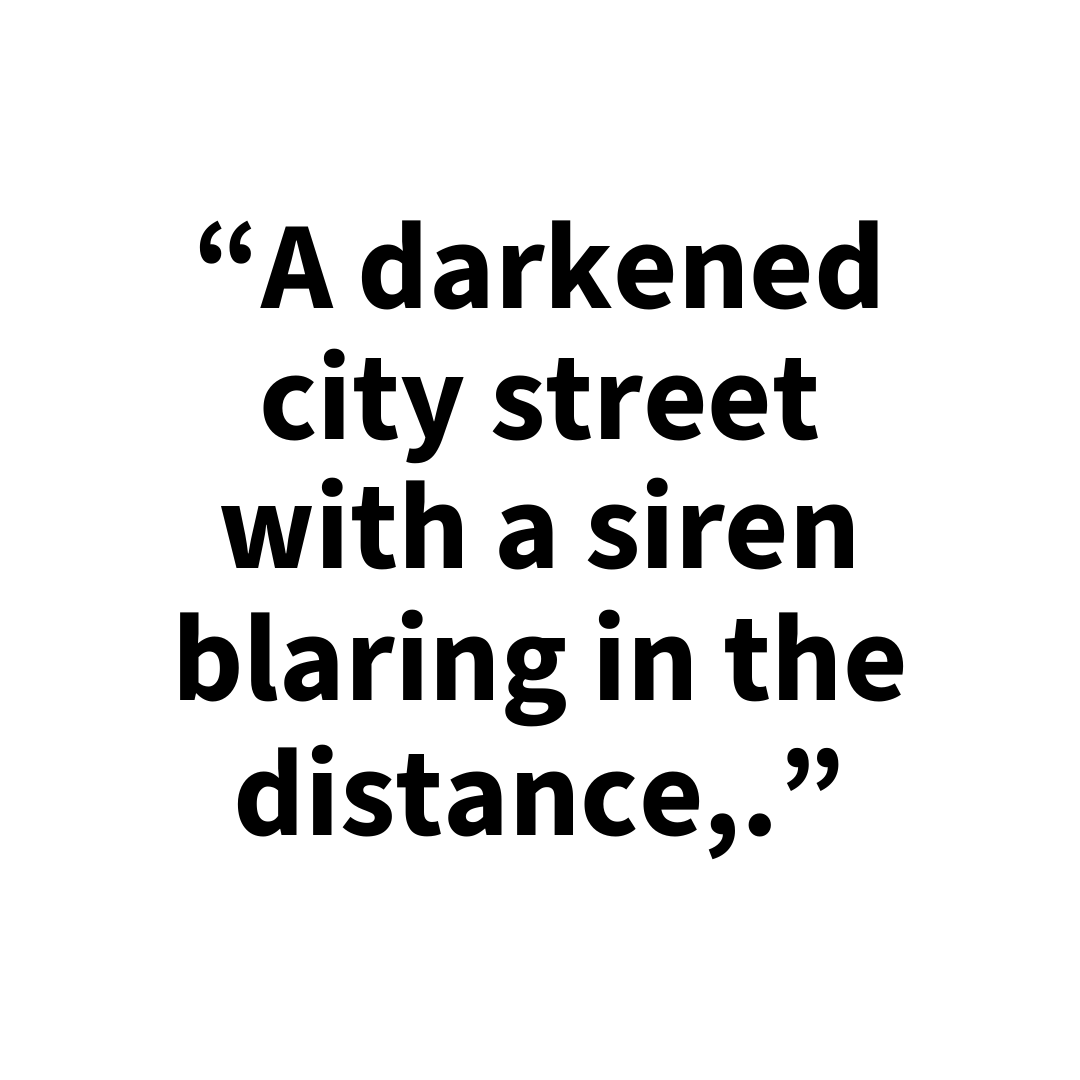} &
\includegraphics[width=0.15\textwidth]{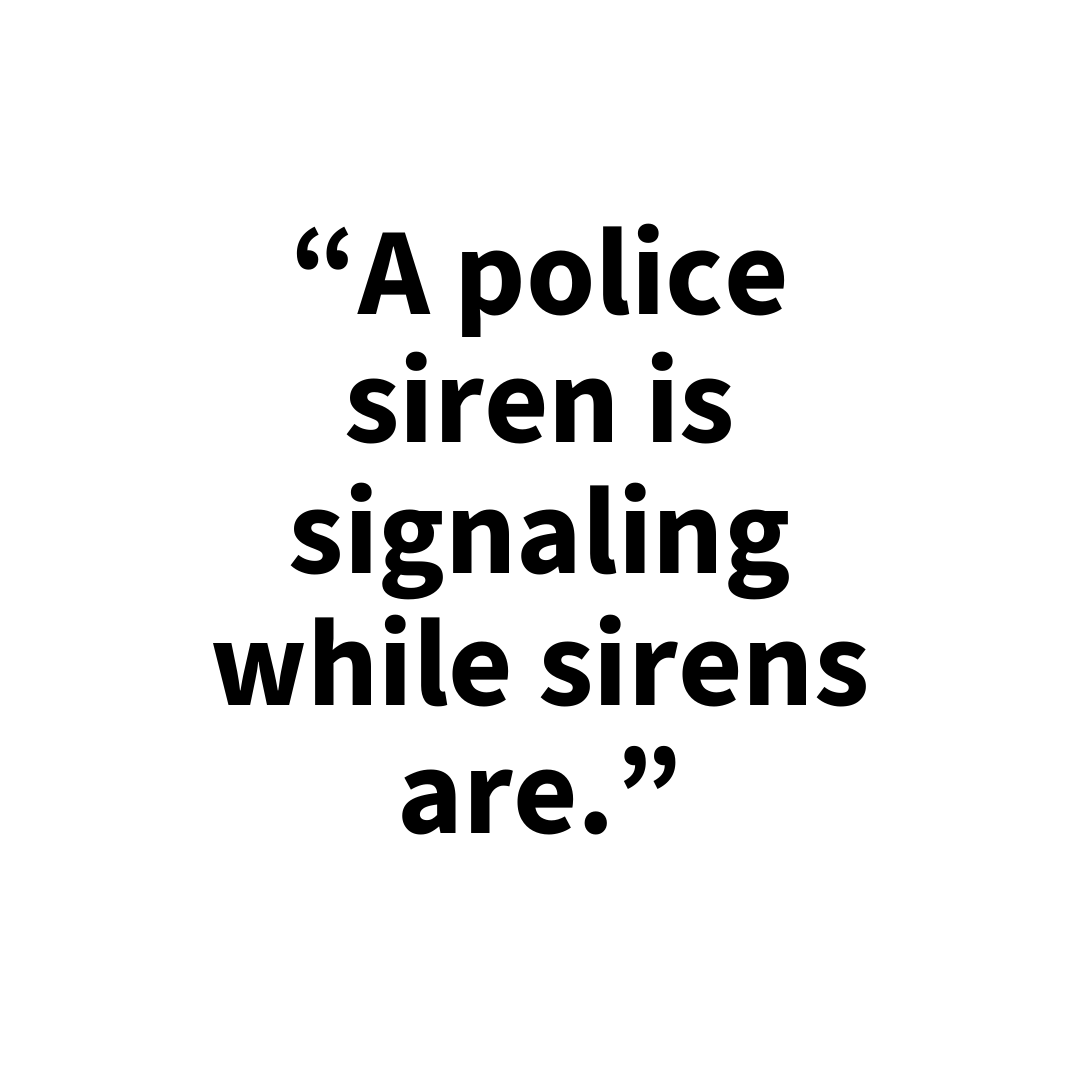} &
\includegraphics[width=0.15\textwidth]{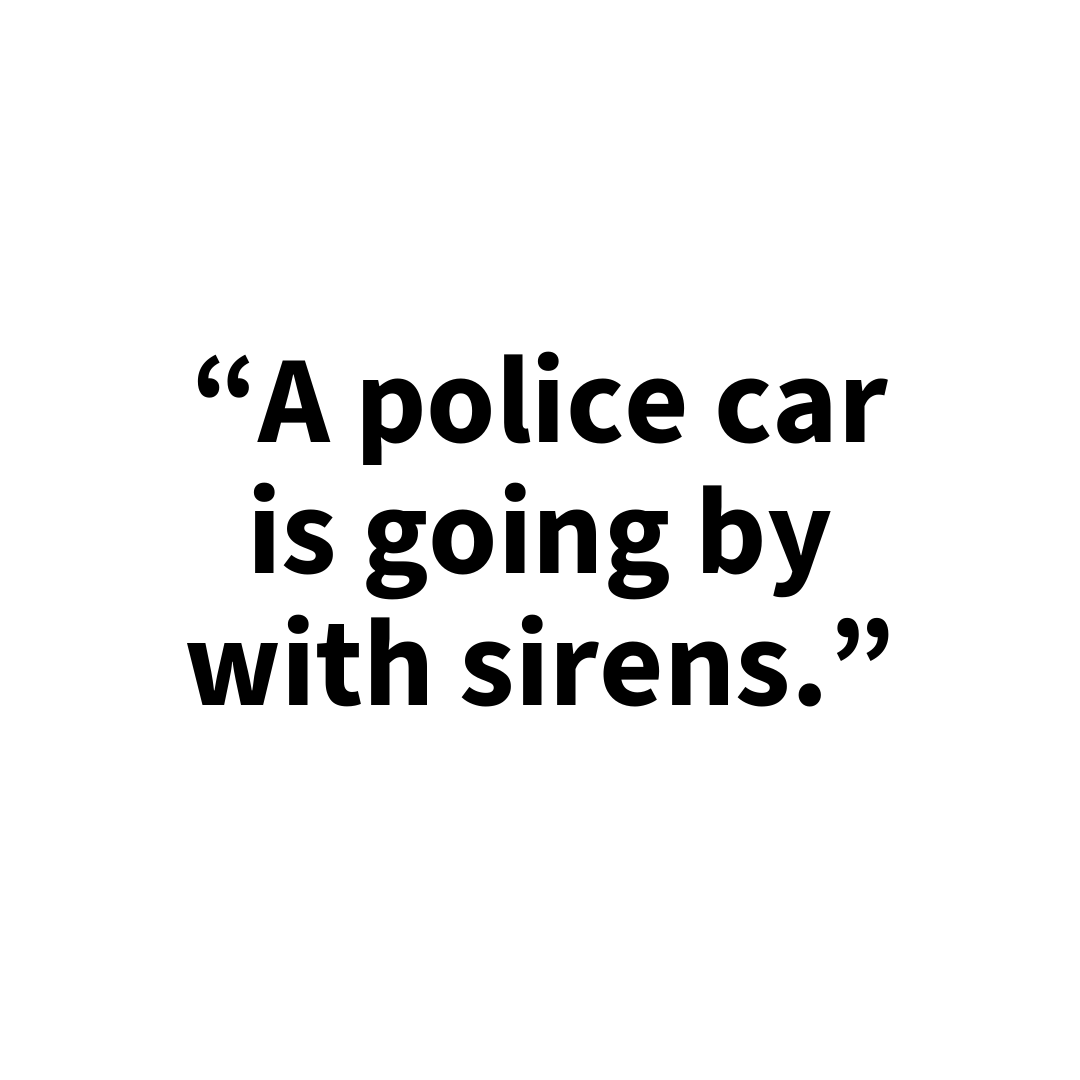} \\
\end{tabular}
\caption{\textbf{(Image $+$ Audio) $\to$ Text generation.} Given an image and an audio clip as joint sources (left), each method generates a caption.}
\label{fig:ita-qual-ia2t}
\end{figure}

\begin{figure}[t]
\centering
\setlength{\tabcolsep}{2pt}
\renewcommand{\arraystretch}{1.0}
\begin{tabular}{cc|cccc}
\footnotesize Source (T) &
\footnotesize Source (A) &
\footnotesize CoDi~\cite{tang2023:codi} &
\footnotesize OmniFlow~\cite{li2025:omniflow} &
\footnotesize FlowBind~\cite{cha2026:flowbind} &
\footnotesize \textbf{\Ours{}$^\dagger$ (Ours)} \\
\midrule
\includegraphics[width=0.15\textwidth]{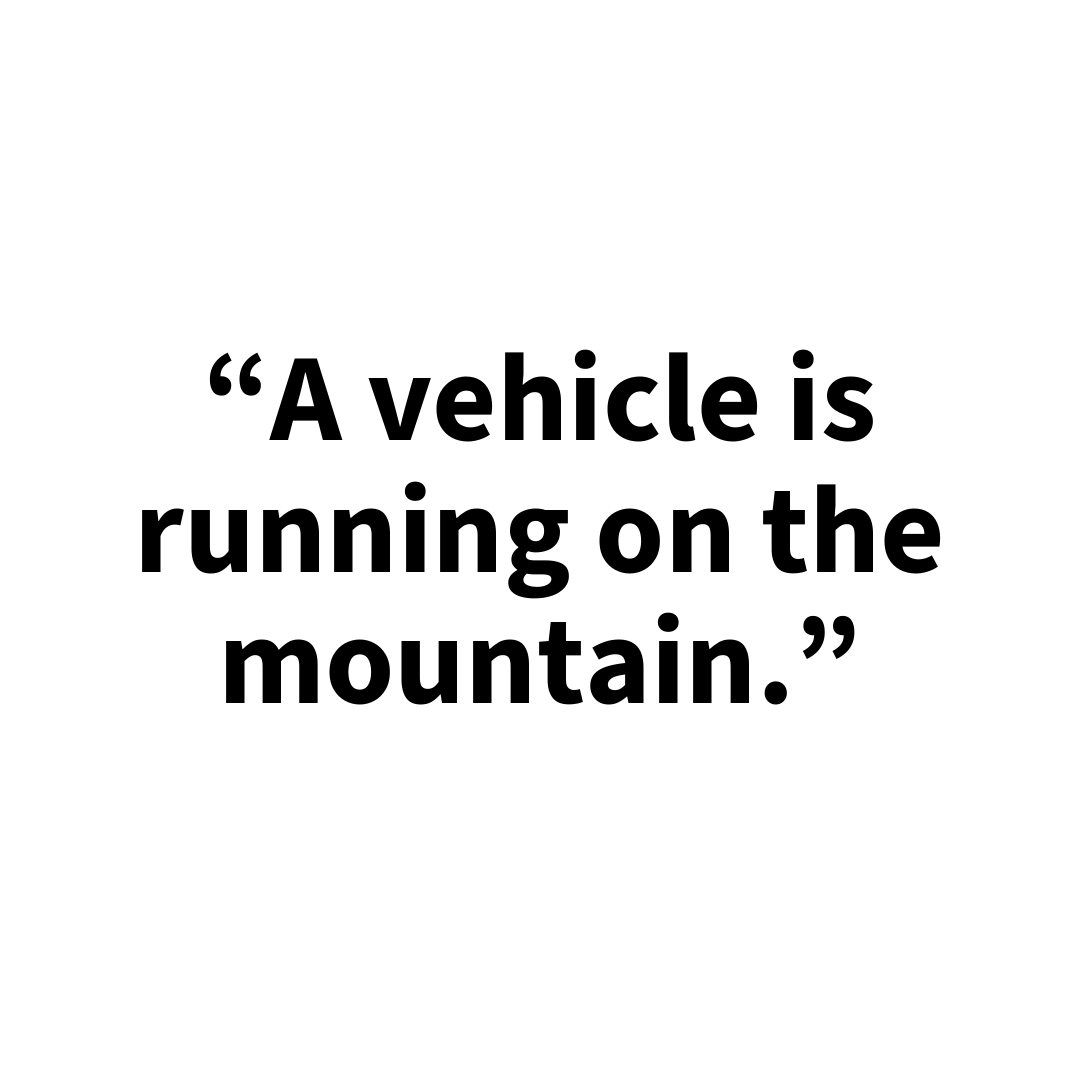} &
\includegraphics[width=0.15\textwidth]{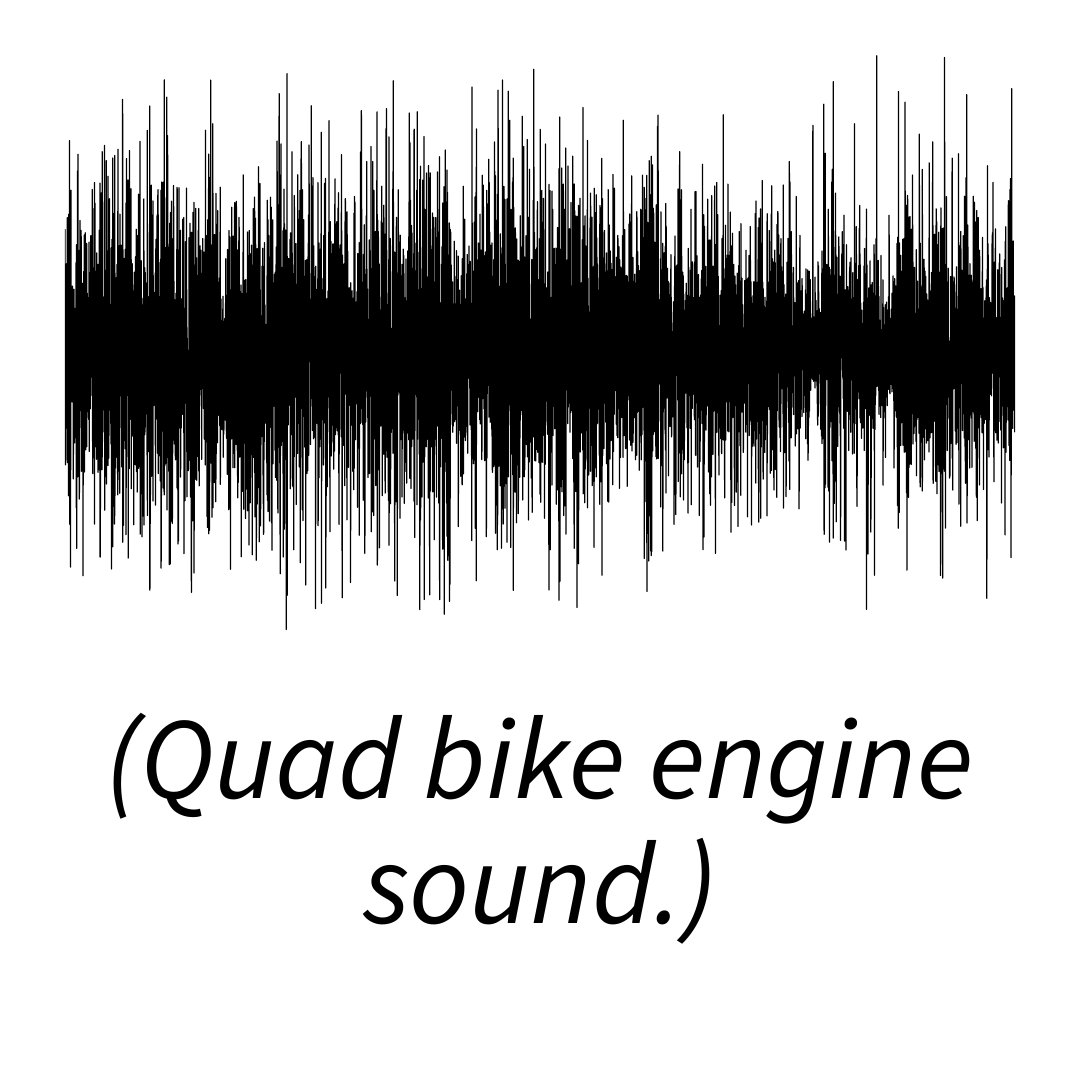} &
\includegraphics[width=0.15\textwidth]{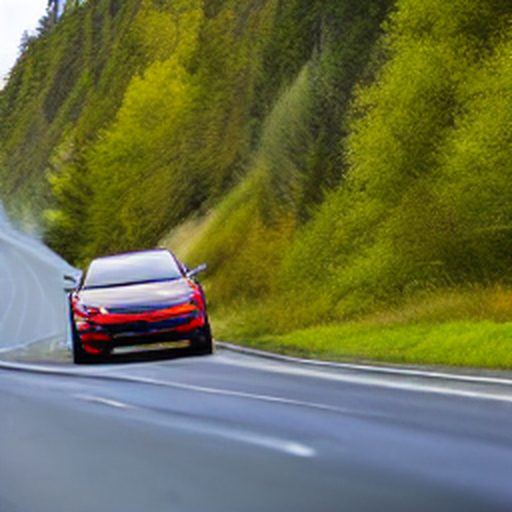} &
\includegraphics[width=0.15\textwidth]{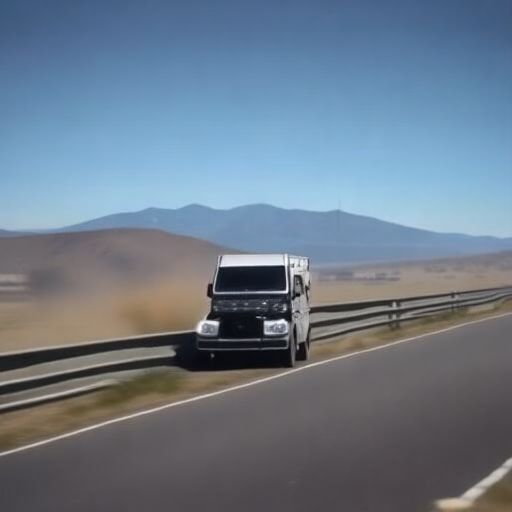} &
\includegraphics[width=0.15\textwidth]{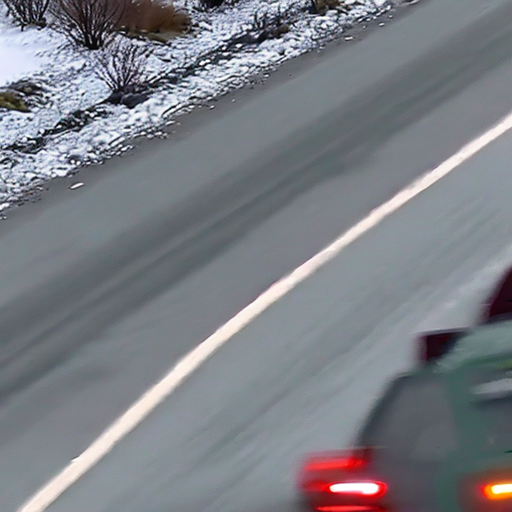} &
\includegraphics[width=0.15\textwidth]{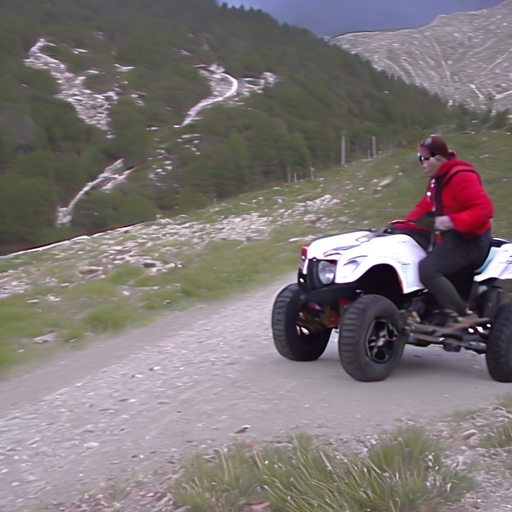} \\

\includegraphics[width=0.15\textwidth]{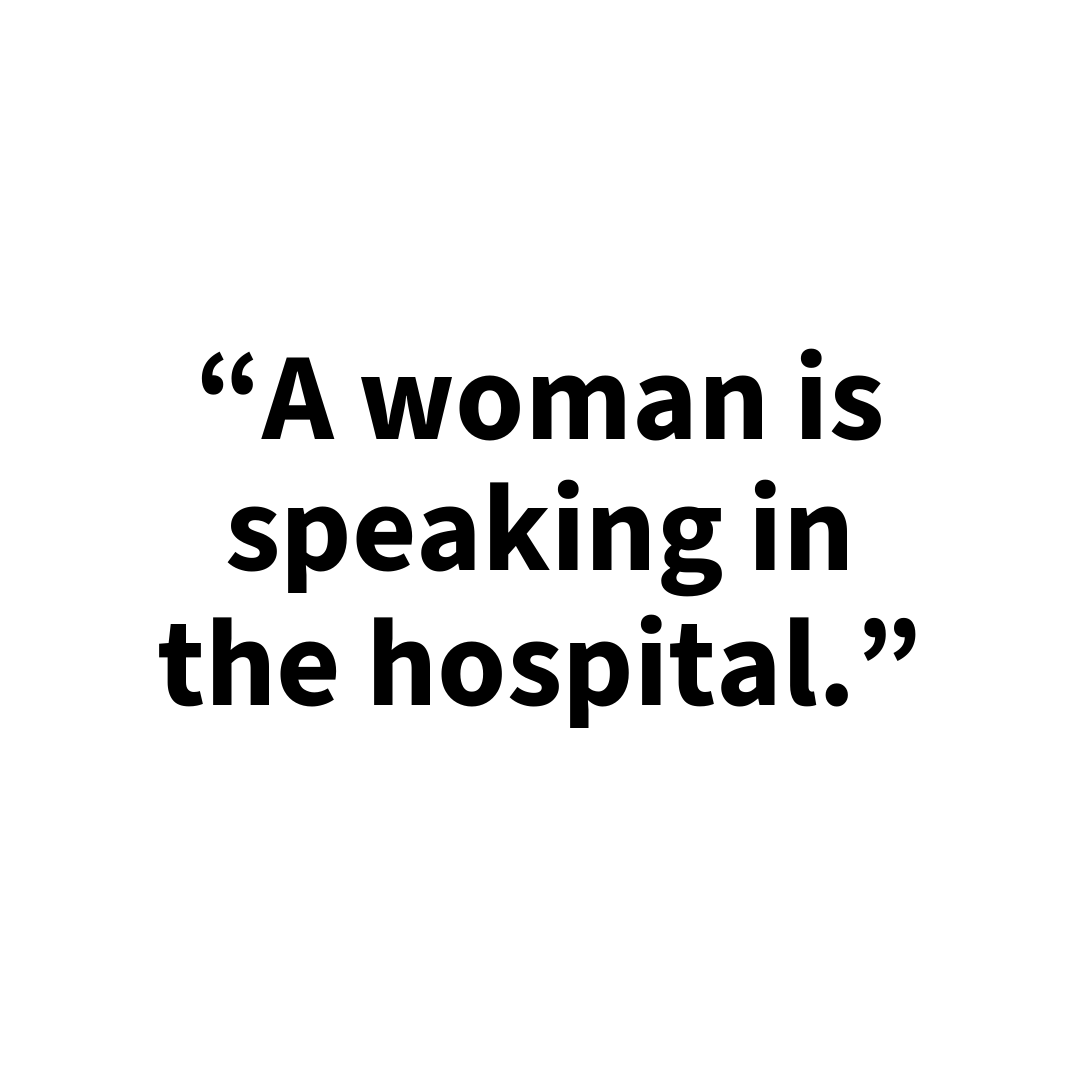} &
\includegraphics[width=0.15\textwidth]{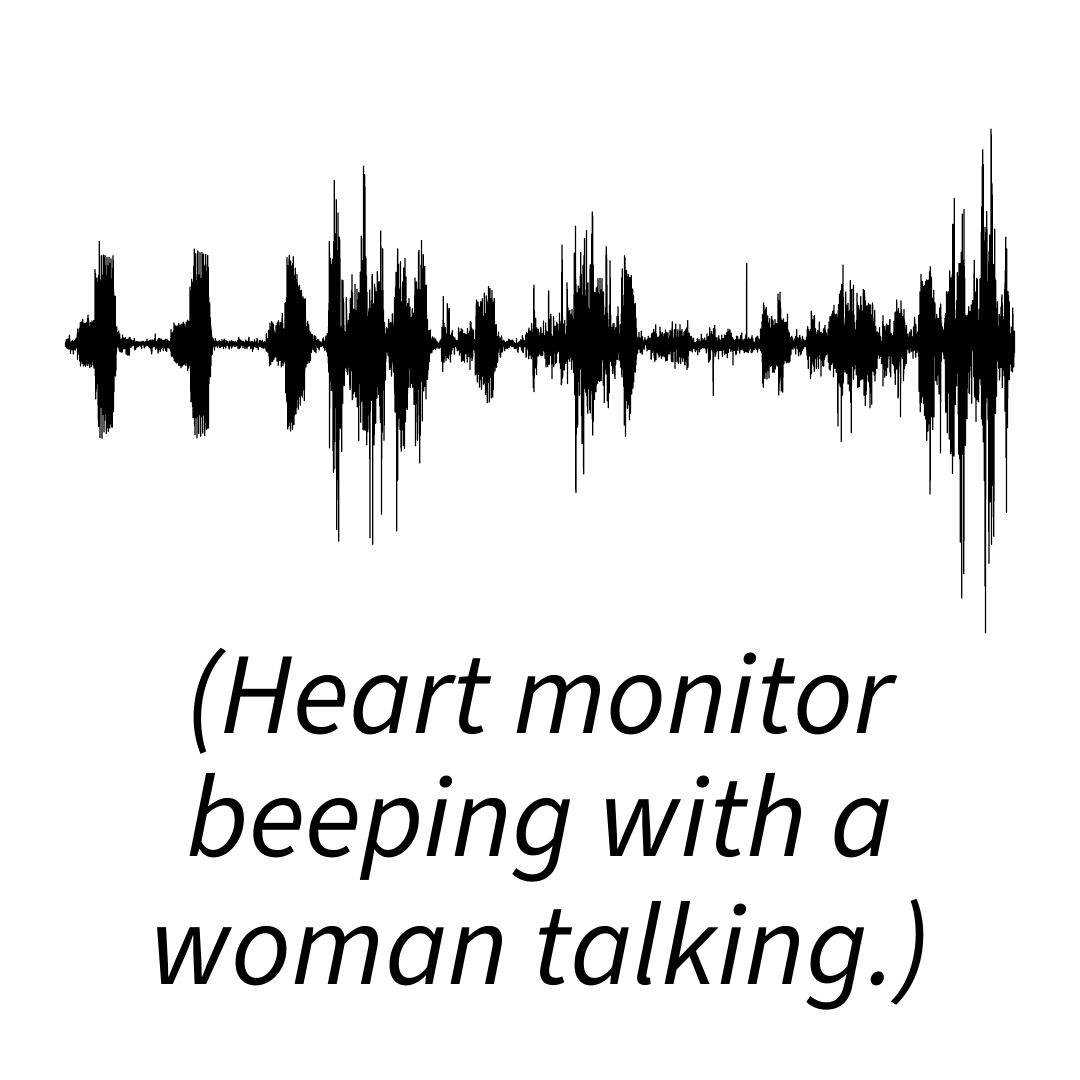} &
\includegraphics[width=0.15\textwidth]{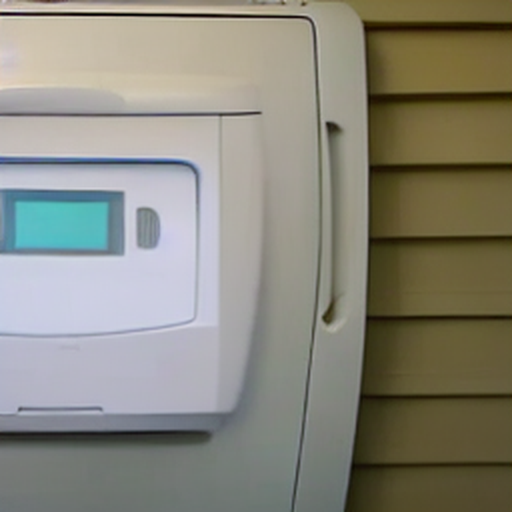} &
\includegraphics[width=0.15\textwidth]{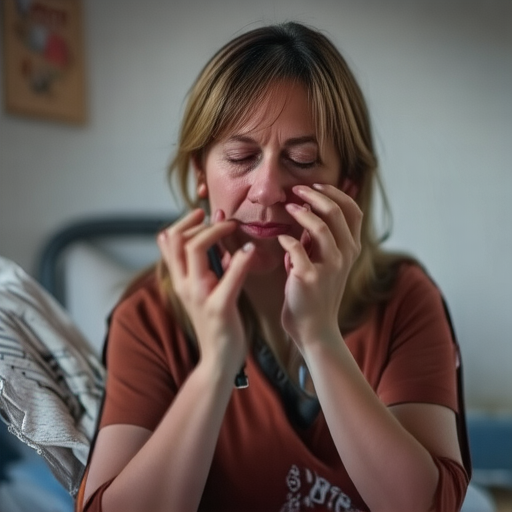} &
\includegraphics[width=0.15\textwidth]{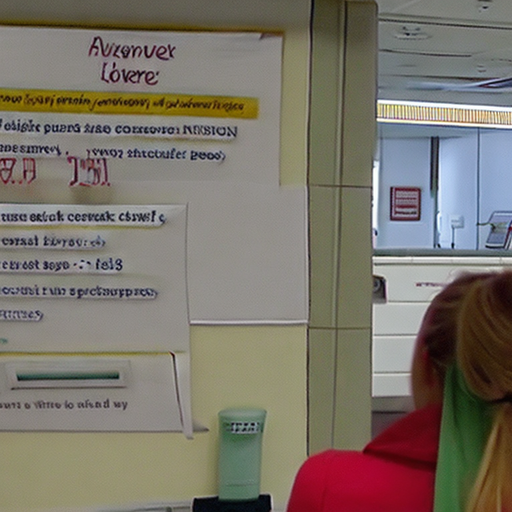} &
\includegraphics[width=0.15\textwidth]{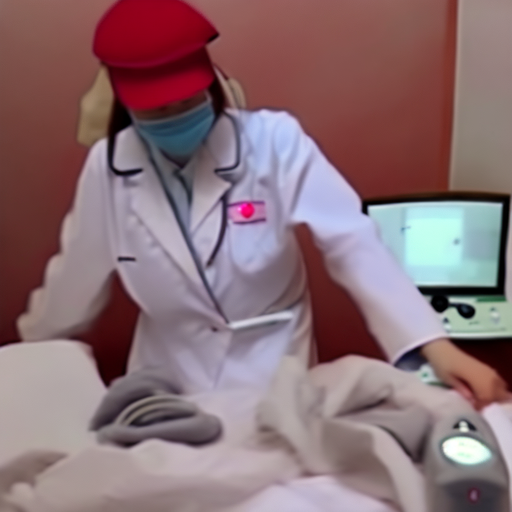} \\

\includegraphics[width=0.15\textwidth]{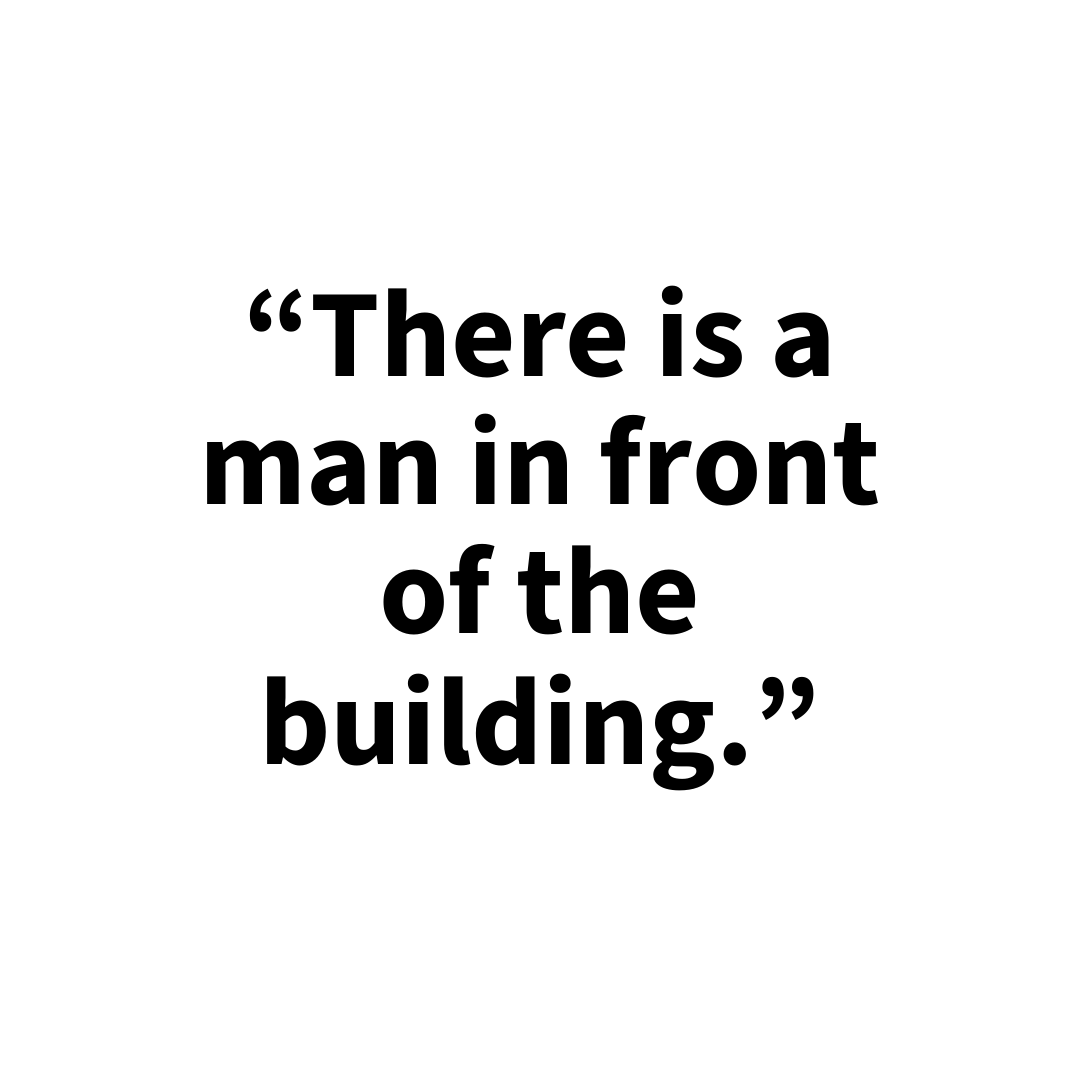} &
\includegraphics[width=0.15\textwidth]{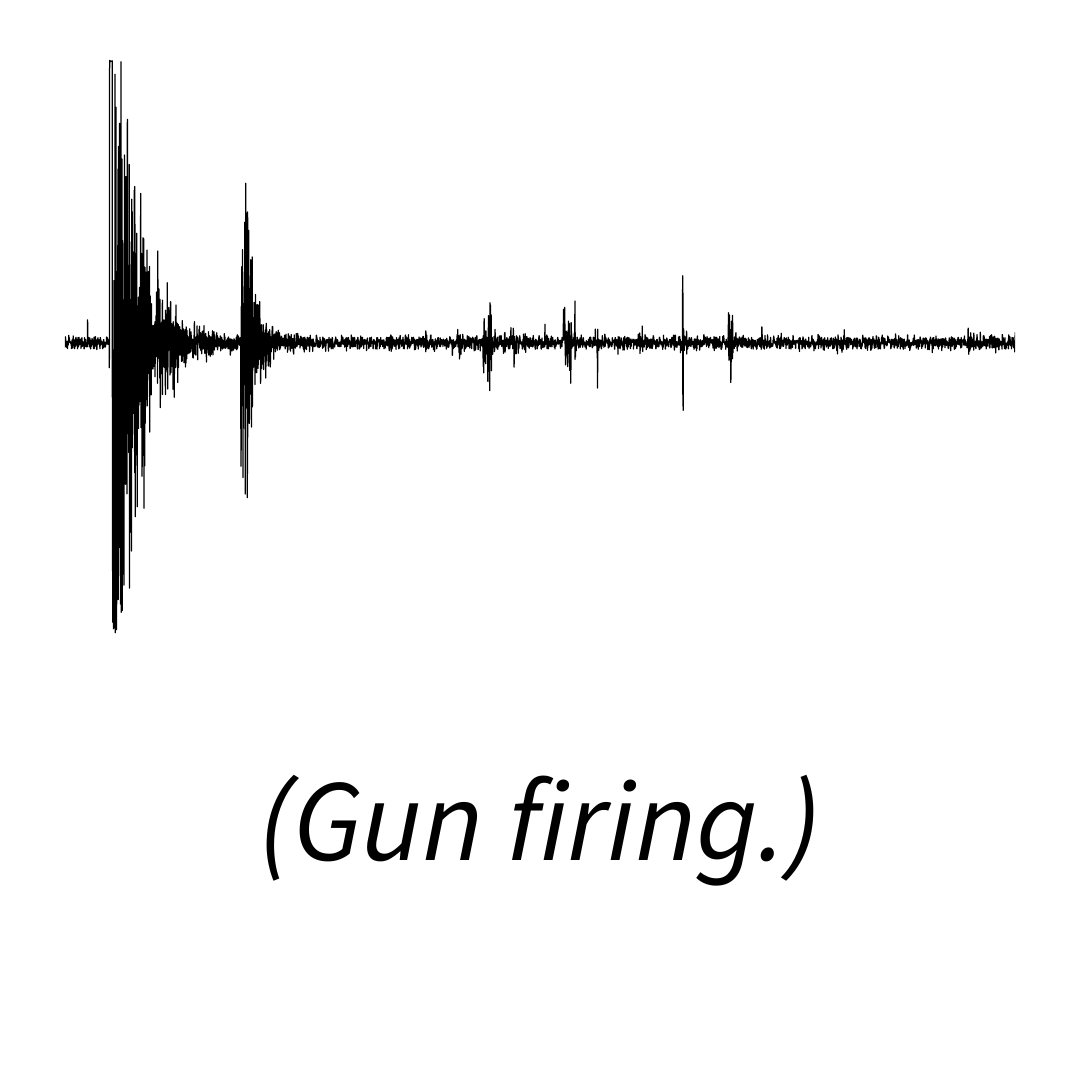} &
\includegraphics[width=0.15\textwidth]{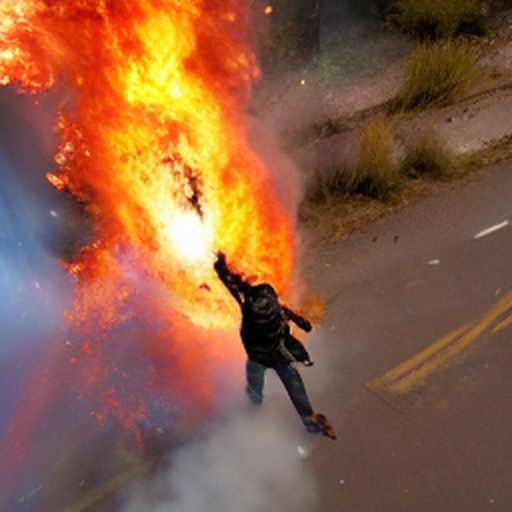} &
\includegraphics[width=0.15\textwidth]{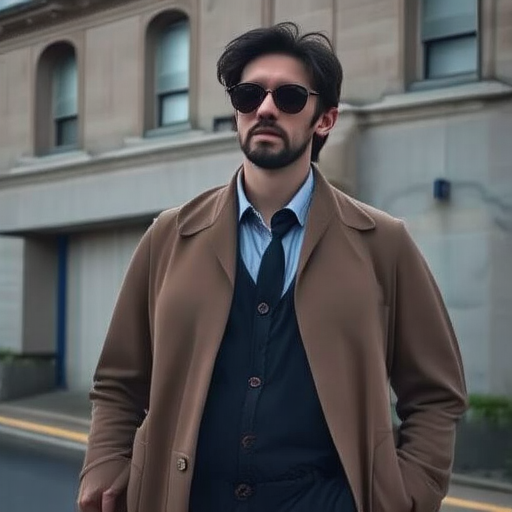} &
\includegraphics[width=0.15\textwidth]{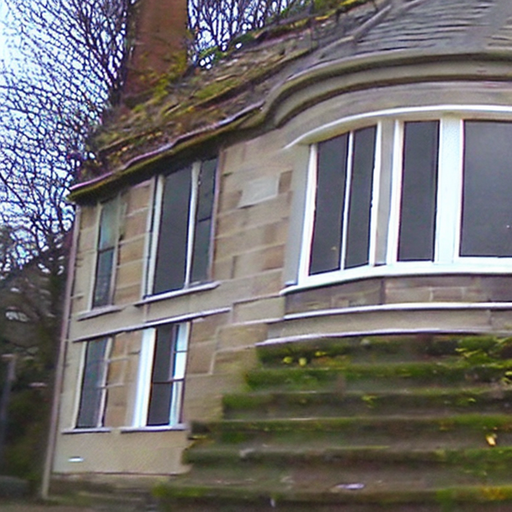} &
\includegraphics[width=0.15\textwidth]{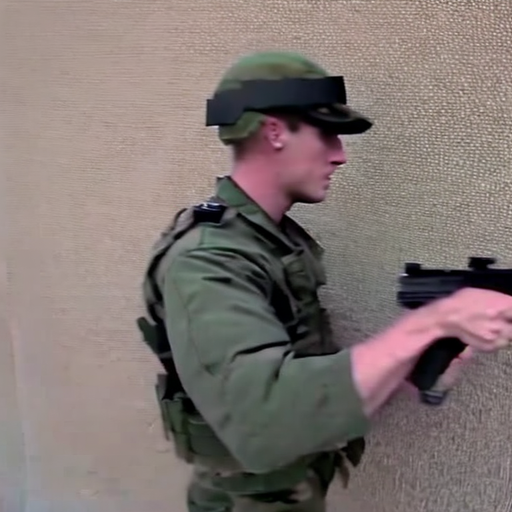} \\

\includegraphics[width=0.15\textwidth]{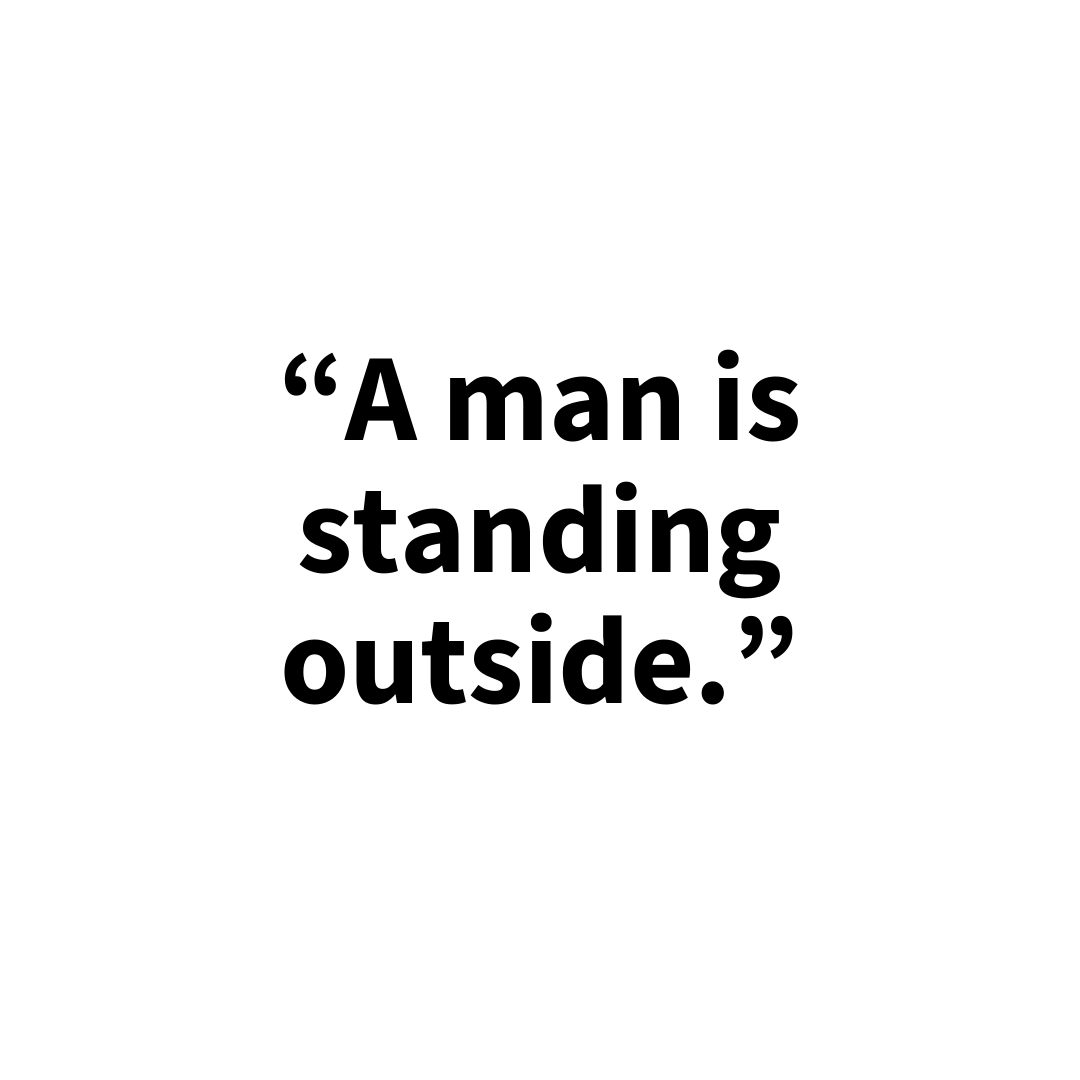} &
\includegraphics[width=0.15\textwidth]{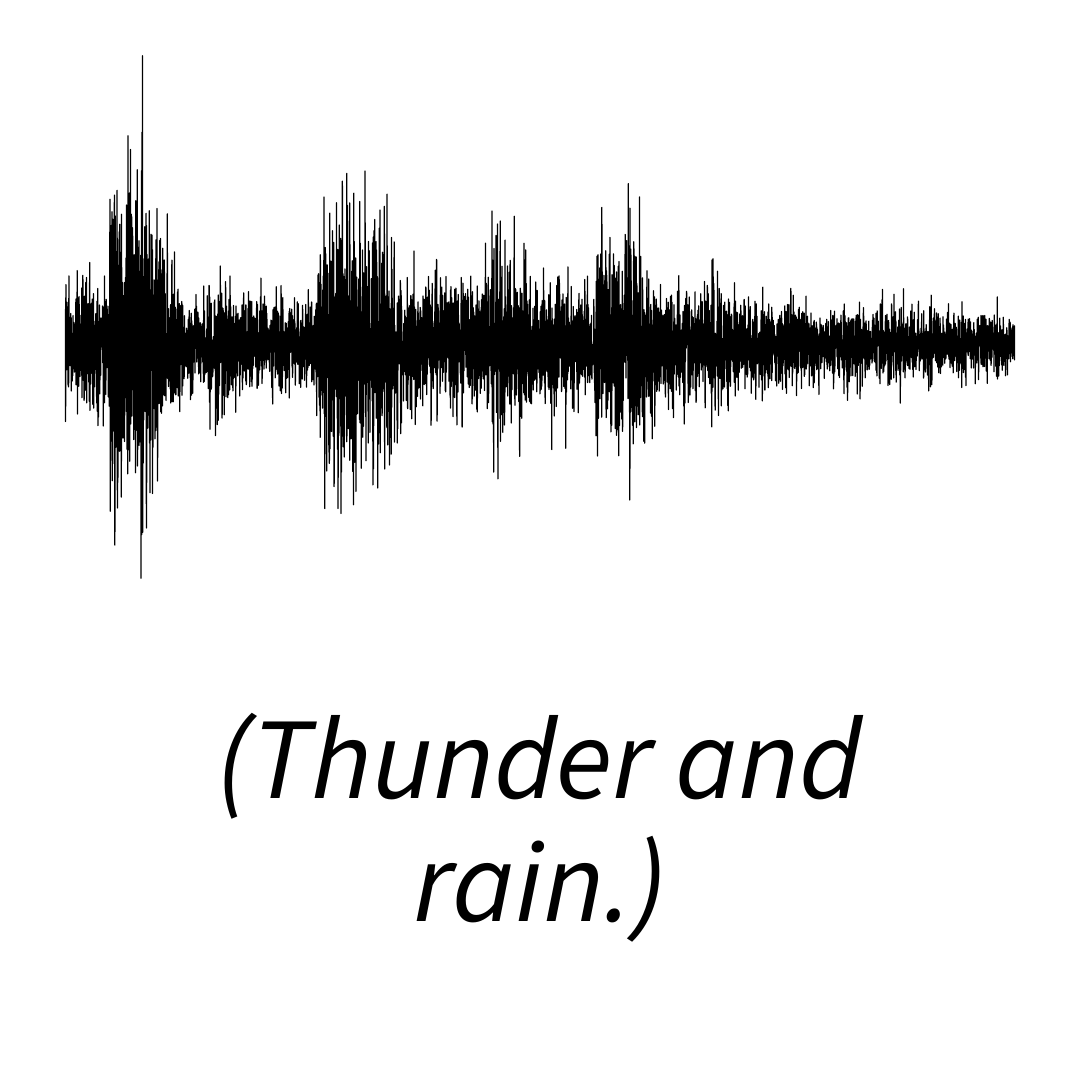} &
\includegraphics[width=0.15\textwidth]{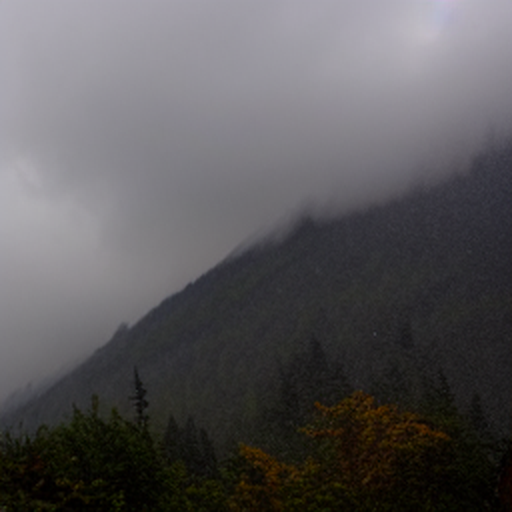} &
\includegraphics[width=0.15\textwidth]{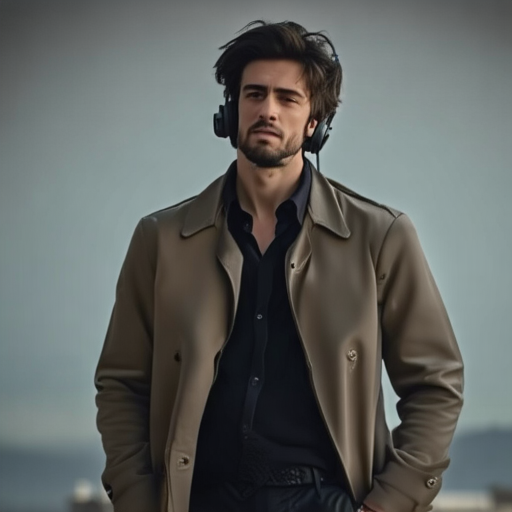} &
\includegraphics[width=0.15\textwidth]{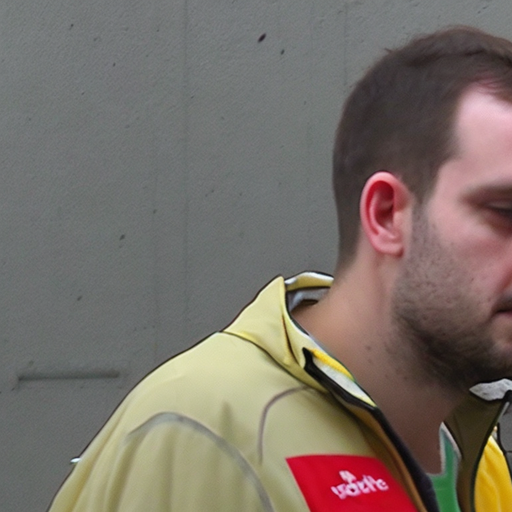} &
\includegraphics[width=0.15\textwidth]{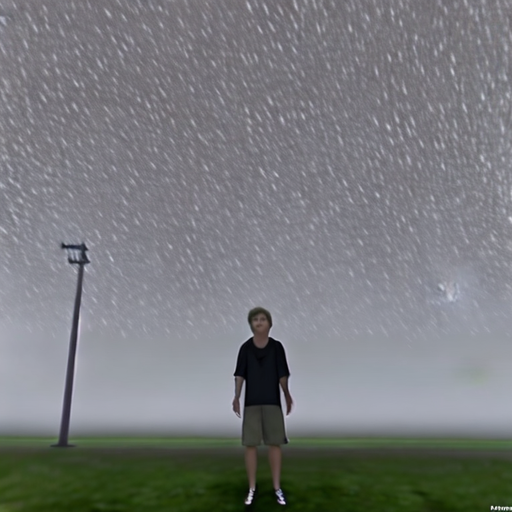} \\

\includegraphics[width=0.15\textwidth]{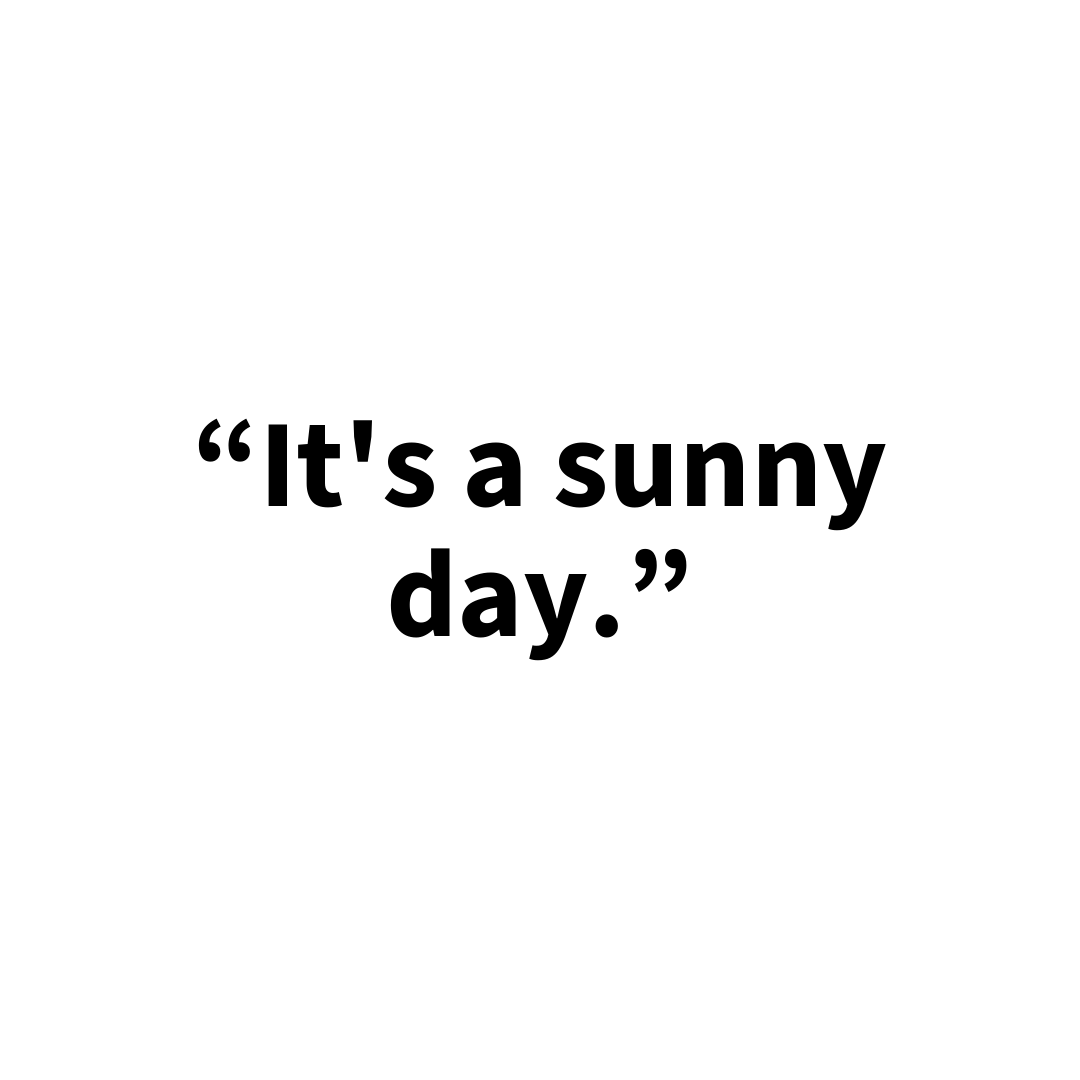} &
\includegraphics[width=0.15\textwidth]{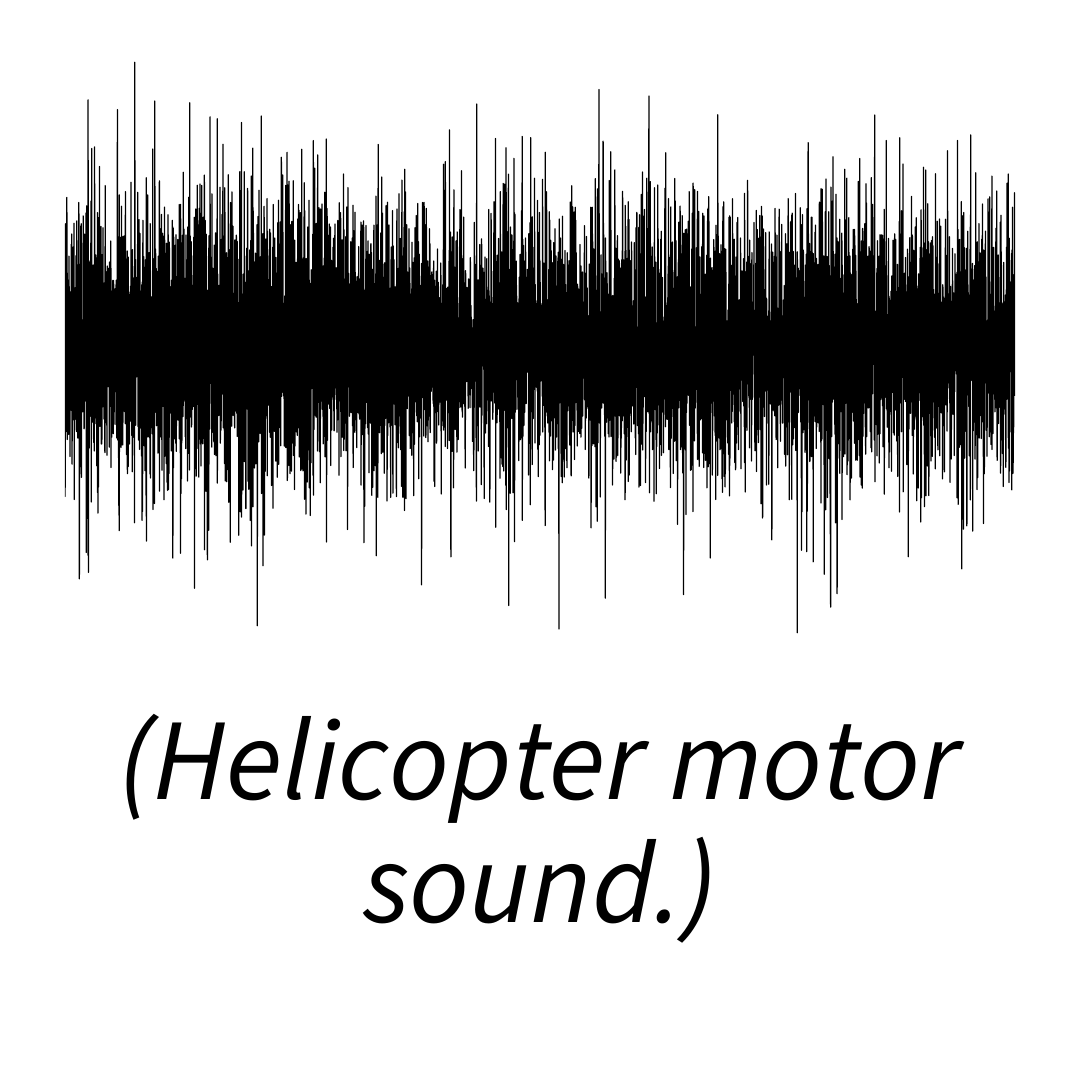} &
\includegraphics[width=0.15\textwidth]{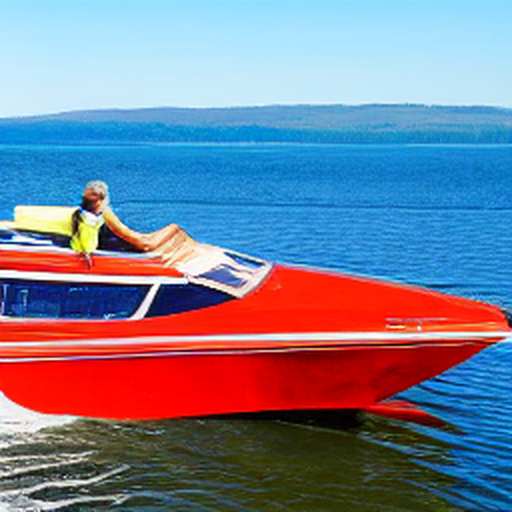} &
\includegraphics[width=0.15\textwidth]{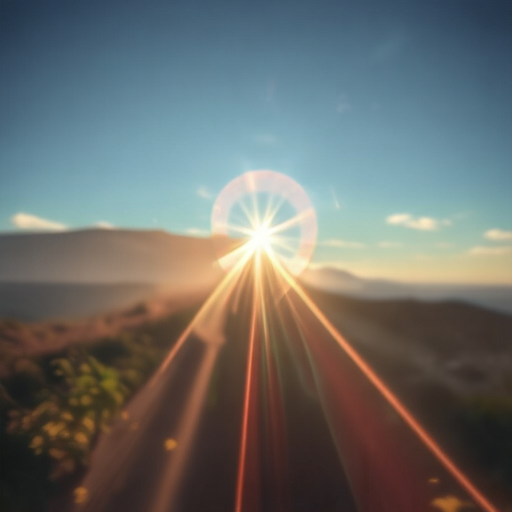} &
\includegraphics[width=0.15\textwidth]{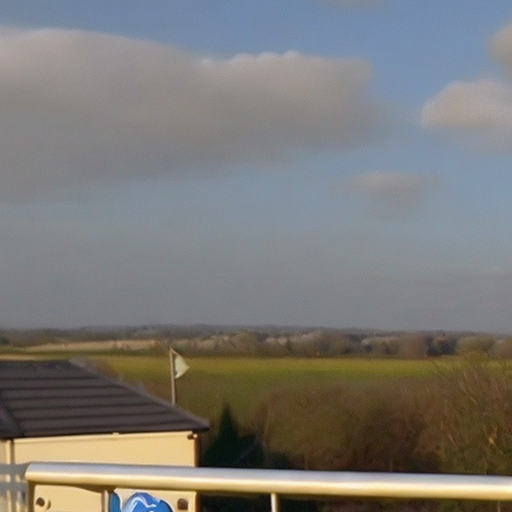} &
\includegraphics[width=0.15\textwidth]{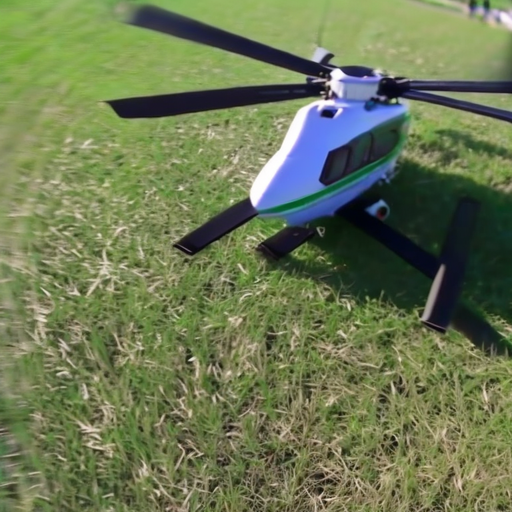} \\

\includegraphics[width=0.15\textwidth]{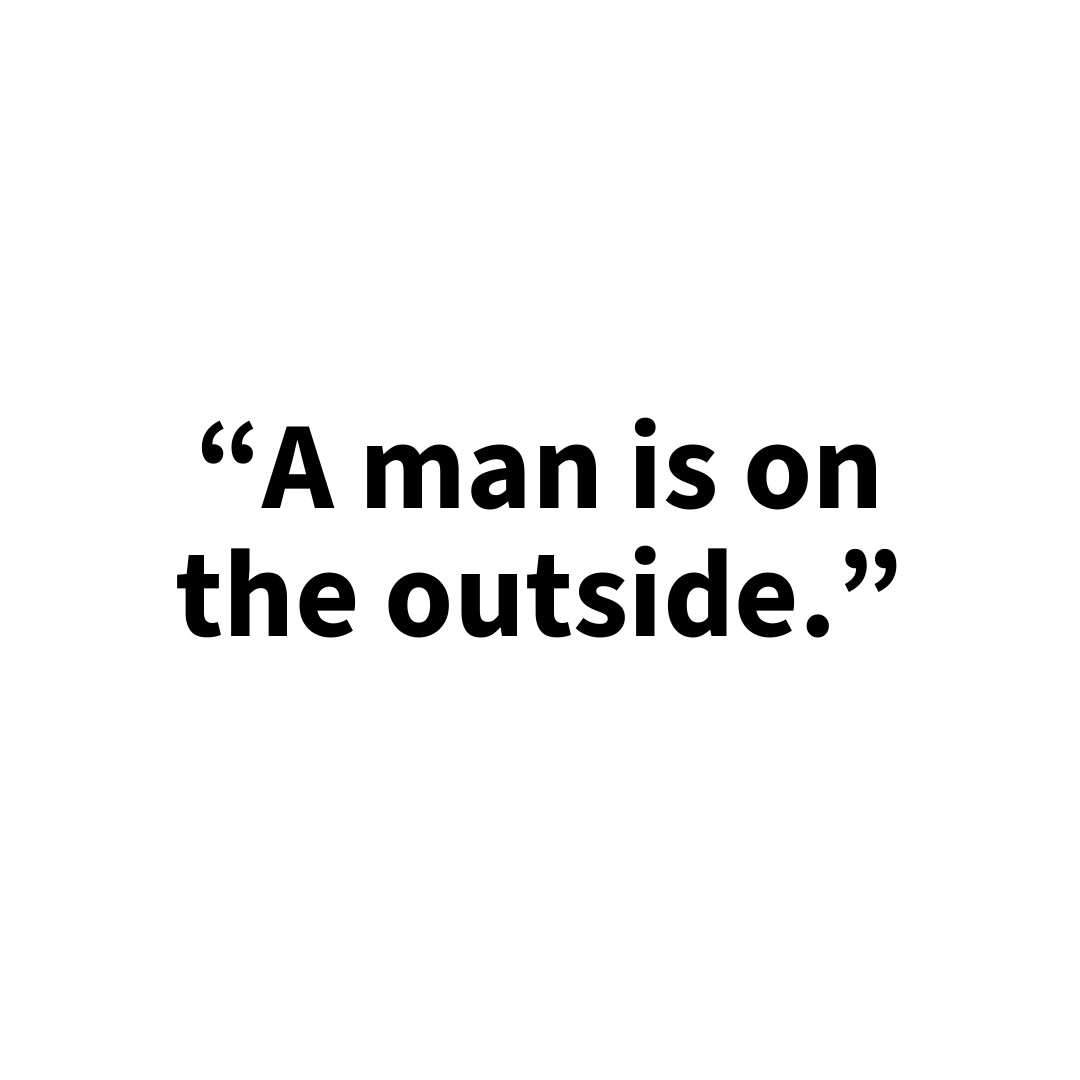} &
\includegraphics[width=0.15\textwidth]{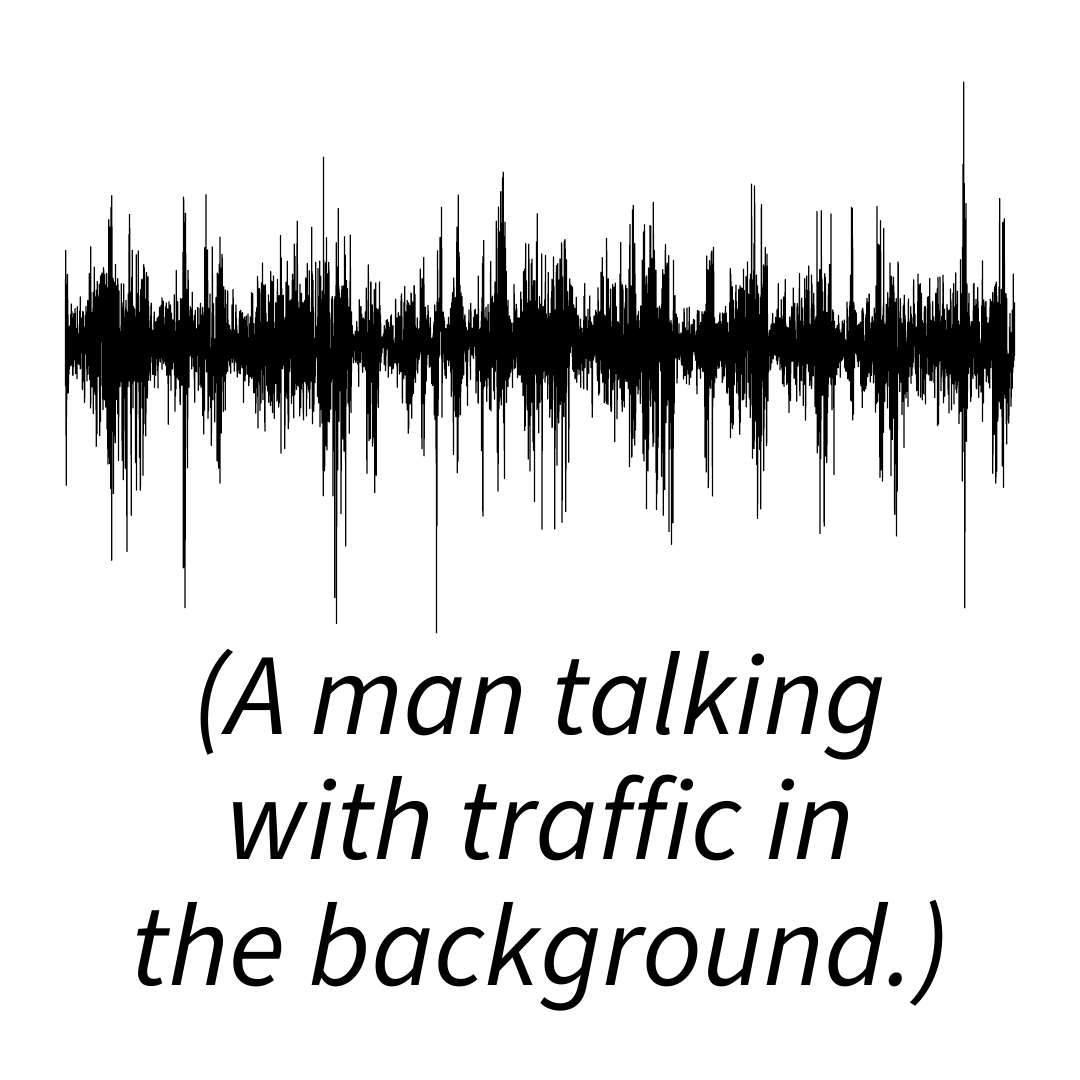} &
\includegraphics[width=0.15\textwidth]{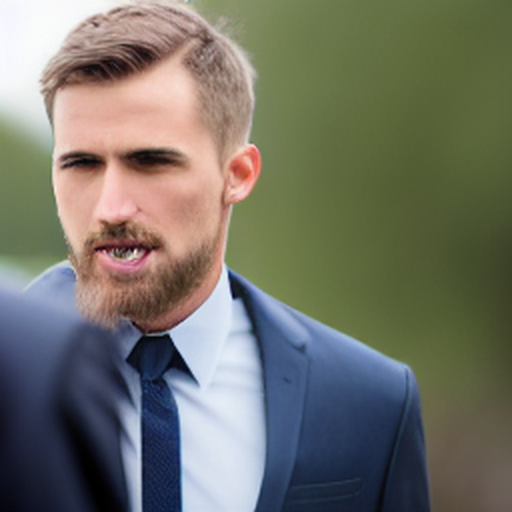} &
\includegraphics[width=0.15\textwidth]{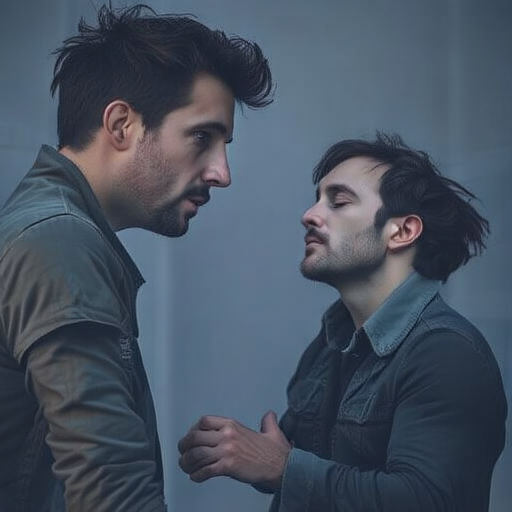} &
\includegraphics[width=0.15\textwidth]{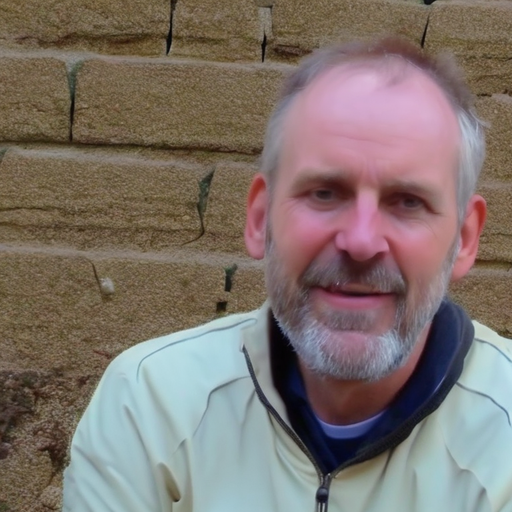} &
\includegraphics[width=0.15\textwidth]{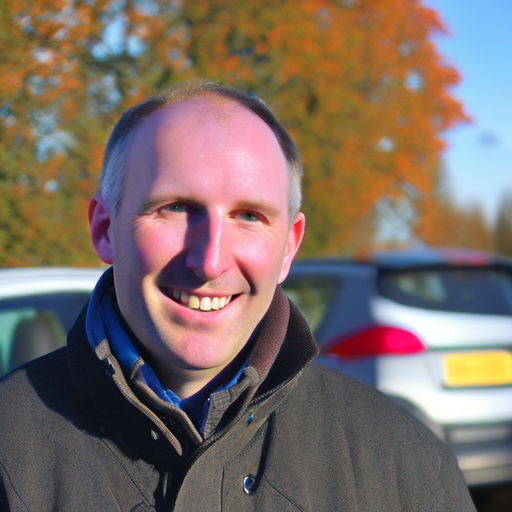} \\
\end{tabular}
\caption{\textbf{(Text $+$ Audio) $\to$ Image generation.} Given a text and an audio clip as joint sources (left), each method generates an image.}
\label{fig:ita-qual-ta2i}
\end{figure}

We provide additional qualitative results for image-text-audio any-to-any generation in Figs.~\ref{fig:ita-qual-t2i}--\ref{fig:ita-qual-uncond-multimodal_vae}. 
\Ours{}$^\dagger$ shows strong source alignment in one-to-one and many-to-one generation, and produces more coherent multimodal triples in unconditional co-generation. 
For qualitative examples involving audio, we include the corresponding raw audio files in the supplementary material. 
For tasks where audio is the target modality, i.e., text-to-audio, image-to-audio, and (text+image)-to-audio, we provide the full qualitative outputs, including the associated text and image results, in the supplementary material.


\begin{table}[t]
\small
\centering
\caption{\textbf{Full comparison of one-to-one generation fidelity on image-text-audio.} Specialists are shown as reference rows, and multimodal VAE baselines use the same backbone and compute as \Ours{}.
\textcolor{softgray}{Gray} marks UnifiedIO2-L results, which do not support audio--image translation.
\textbf{Bold} and \underline{underline} mark the best and second-best generalist results.}
\label{tab:ita-fidelity}
\begin{tabular}{>{\centering\arraybackslash}p{1.6cm}!{\color{lightgray}\vrule}l!{\color{midgray}\vrule}cccccc}
\toprule
\multirow{2}{*}{Category} & \multicolumn{1}{c!{\color{midgray}\vrule}}{\multirow{2}{*}{Method}}
& T$\rightarrow$I & I$\rightarrow$T & T$\rightarrow$A & A$\rightarrow$T & I$\rightarrow$A & A$\rightarrow$I \\
& \multicolumn{1}{c!{\color{midgray}\vrule}}{} & FID $\downarrow$ & CIDEr $\uparrow$ & FAD $\downarrow$ & CIDEr $\uparrow$ & FAD $\downarrow$ & FID $\downarrow$ \\
\midrule
\multirow{6}{*}{\emph{Specialists}}
& FLUX.1~\cite{flux2024}        & 24.81 & --     & --   & --    & --   & -- \\
& LLaVA-NeXT~\cite{liu2024llavanext}    & --    & 44.60 & --   & --    & --   & -- \\
& TangoFlux~\cite{hung2025:tangoflux}     & --    & --     & 2.47 & --    & --   & -- \\
& Qwen2-Audio~\cite{chu2024:qwen2audio}   & --    & --     & --   & 21.39  & --   & -- \\
& Seeing \& Hearing~\cite{xing2024:seeinghearing}   & --    & --     & --   & --  & 7.47   & -- \\
& Sound2Vision~\cite{sungbin2024:sound2vision}   & --    & --     & --   & --  & --   & 25.59 \\
\midrule
\multirow{6}{*}{\emph{Generalists}}
& \textcolor{softgray}{UnifiedIO2-L}~\cite{lu2023:unifiedio2}  
& \textcolor{softgray}{22.03} 
& \textcolor{softgray}{91.36} 
& \textcolor{softgray}{7.13} 
& \textcolor{softgray}{15.18} 
& \textcolor{softgray}{--} 
& \textcolor{softgray}{--} \\
& CoDi~\cite{tang2023:codi}          & 24.97            & 11.97           & 8.67            & 7.70             & 13.92           & 53.60 \\
& OmniFlow~\cite{li2025:omniflow}      & 20.52            & 26.60           & \underline{4.08}& 31.42            & 4.87            & 97.56 \\
& FlowBind~\cite{cha2026:flowbind}      & 20.13 & 42.57 & 4.64           & \textbf{54.41}   & 2.12 & 28.47 \\
& \textbf{\Ours{} (Ours)}      & \underline{14.73}   & \underline{57.72}  & \textbf{3.52}   & 43.72 & \underline{1.98}   & \underline{22.61} \\
& \textbf{\Ours{}$^\dagger$ (Ours)}      & \textbf{14.40}   & \textbf{58.31}  & \textbf{3.52}   & \underline{47.38}& \textbf{1.90}   & \textbf{22.97} \\
\midrule
\multirow{2}{*}{\makecell{\emph{Multimodal} \\ \emph{VAEs}}}
& MMVAE~\cite{shi2019:mmvae}           & 14.94 & 59.34 & 3.55 & 47.43 & 1.92 & 23.09 \\
& MoPoE~\cite{sutter2021:mopoe}         & 14.60 & 59.74 & 3.46 & 50.34 & 2.04 & 23.57 \\
\bottomrule
\end{tabular}
\end{table}

\begin{table}[t]
\small
\centering
\caption{\textbf{Full comparison of one-to-one cross-modal alignment on image-text-audio.}
Specialists are shown as reference rows, and multimodal VAE baselines use the same backbone and compute as \Ours{}.
\textcolor{softgray}{Gray} marks UnifiedIO2-L results, which do not support audio--image translation.
\textbf{Bold} and \underline{underline} mark the best and second-best generalist results.}
\label{tab:ita-alignment}
\begin{tabular}{>{\centering\arraybackslash}p{1.6cm}!{\color{lightgray}\vrule}l!{\color{midgray}\vrule}cccccc}
\toprule
\multirow{2}{*}{Category} & \multicolumn{1}{c!{\color{midgray}\vrule}}{\multirow{2}{*}{Method}}
& T$\rightarrow$I & I$\rightarrow$T & T$\rightarrow$A & A$\rightarrow$T & I$\rightarrow$A & A$\rightarrow$I \\
& \multicolumn{1}{c!{\color{midgray}\vrule}}{} & CLIP $\uparrow$ & CLIP $\uparrow$ & CLAP $\uparrow$ & CLAP $\uparrow$ & AIS $\uparrow$ & AIS $\uparrow$ \\
\midrule
\multirow{6}{*}{\emph{Specialists}}
& FLUX.1~\cite{flux2024}        & 30.36 & --     & --   & --    & --   & -- \\
& LLaVA-NeXT~\cite{liu2024llavanext}    & --    & 29.47 & --   & --    & --   & -- \\
& TangoFlux~\cite{hung2025:tangoflux}     & --    & --     & 46.96 & --    & --   & -- \\
& Qwen2-Audio~\cite{chu2024:qwen2audio}   & --    & --     & --   & 45.45  & --   & -- \\
& Seeing \& Hearing~\cite{xing2024:seeinghearing}   & --    & --     & --   & --  & 51.96   & -- \\
& Sound2Vision~\cite{sungbin2024:sound2vision}   & --    & --     & --   & --  & --   & 97.40 \\
\midrule
\multirow{6}{*}{\emph{Generalists}}
& \textcolor{softgray}{UnifiedIO2-L}~\cite{lu2023:unifiedio2}
& \textcolor{softgray}{29.73}
& \textcolor{softgray}{\textbf{29.97}}
& \textcolor{softgray}{15.22}
& \textcolor{softgray}{35.48}
& \textcolor{softgray}{--}
& \textcolor{softgray}{--} \\
& CoDi~\cite{tang2023:codi}          & 29.70            & 26.05           & 16.72            & 32.72            & 59.45           & 84.47 \\
& OmniFlow~\cite{li2025:omniflow}      & \textbf{30.76}   & 27.48           & 30.07 & \textbf{44.74}   & 76.52           & 64.22 \\
& FlowBind~\cite{cha2026:flowbind}      & 28.07            & 29.14 & 27.13           & \underline{41.80}& 89.23 & 93.94 \\
& \textbf{\Ours{} (Ours)}      & \underline{29.94}& 29.05           & \underline{31.45}   & 39.38            & \underline{89.92}  & \underline{95.59} \\
& \textbf{\Ours{}$^\dagger$ (Ours)}      & 29.92 & \underline{29.18}           & \textbf{31.81}   & 39.45            & \textbf{90.11}  & \textbf{95.61} \\
\midrule
\multirow{2}{*}{\makecell{\emph{Multimodal} \\ \emph{VAEs}}}
& MMVAE~\cite{shi2019:mmvae}           & 29.68 & 29.14 & 30.09 & 38.30 & 90.05 & 95.11 \\
& MoPoE~\cite{sutter2021:mopoe}         & 29.59 & 29.05 & 31.43 & 38.98 & 90.82 & 94.20 \\
\bottomrule
\end{tabular}
\end{table}


\newcommand{\softline}{\arrayrulecolor{gray!30}\hline\arrayrulecolor{black}}

\newcommand{\unconCell}[2]{%
  \begin{tabular}{@{}c@{\hspace{2pt}}c@{\hspace{2pt}}c@{}}%
    \includegraphics[width=0.13\textwidth]{figures/qualitative_ita/uncond/generalists/#1_#2_caption_text.pdf} &
    \includegraphics[width=0.13\textwidth]{figures/qualitative_ita/uncond/generalists/#1_#2_audio.pdf} &
    \includegraphics[width=0.13\textwidth]{figures/qualitative_ita/uncond/generalists/#1_#2.png}%
  \end{tabular}%
}

\begin{figure}[t]
\centering
\setlength{\tabcolsep}{4pt}
\renewcommand{\arraystretch}{1.2}
\begin{tabular}{lcc}
\toprule
\footnotesize CoDi~\cite{tang2023:codi} &
\multicolumn{2}{c}{\raisebox{-.45\height}{\includegraphics[width=0.82\textwidth]{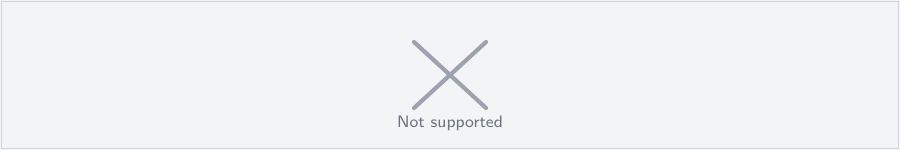}}} \\
\addlinespace[4pt]
\footnotesize OmniFlow~\cite{li2025:omniflow} &
\unconCell{483}{omniflow} &
\unconCell{484}{omniflow} \\
\addlinespace[2pt]
\footnotesize FlowBind~\cite{cha2026:flowbind} &
\multicolumn{2}{c}{\raisebox{-.45\height}{\includegraphics[width=0.82\textwidth]{figures/placeholder_not_supported.pdf}}} \\
\addlinespace[4pt]
\footnotesize \textbf{\Ours{}$^\dagger$ (Ours)} &
\unconCell{483}{ours_hell_200} &
\unconCell{484}{ours_hell_200} \\
\bottomrule
\end{tabular}
\caption{\textbf{Unconditional co-generation among generalists.} Each row shows two jointly sampled (text, audio, image) triples from a method's prior. CoDi~\cite{tang2023:codi} and FlowBind~\cite{cha2026:flowbind} do not support fully unconditional joint sampling.}
\label{fig:ita-qual-uncond-generalists}
\end{figure}

\newcommand{\unconVCell}[2]{%
  \begin{tabular}{@{}c@{\hspace{2pt}}c@{\hspace{2pt}}c@{}}%
    \includegraphics[width=0.13\textwidth]{figures/qualitative_ita/uncond/mvaes/#1_#2_caption_text.pdf} &
    \includegraphics[width=0.13\textwidth]{figures/qualitative_ita/uncond/mvaes/#1_#2_audio.pdf} &
    \includegraphics[width=0.13\textwidth]{figures/qualitative_ita/uncond/mvaes/#1_#2.png}%
  \end{tabular}%
}
\begin{figure}[t]
\centering
\setlength{\tabcolsep}{4pt}
\renewcommand{\arraystretch}{1.2}
\begin{tabular}{lcc}
\toprule
\footnotesize MMVAE~\cite{shi2019:mmvae} &
\unconVCell{451}{moe_200} &
\unconVCell{241}{moe_200} \\
\softline
\addlinespace[3pt]
\footnotesize MoPoE~\cite{sutter2021:mopoe} &
\unconVCell{451}{mopoe_200} &
\unconVCell{241}{mopoe_200} \\
\softline
\addlinespace[3pt]
\footnotesize \textbf{\Ours{}$^\dagger$ (Ours)} &
\unconVCell{451}{ours_hell_200} &
\unconVCell{241}{ours_hell_200} \\
\bottomrule
\end{tabular}
\caption{\textbf{Unconditional co-generation against multimodal VAE baselines.} Each row shows two jointly sampled (text, audio, image) triples. The multimodal VAE baselines share the same backbone and compute as \Ours{}$^\dagger$.}
\label{fig:ita-qual-uncond-multimodal_vae}
\end{figure}

\end{document}